\documentclass[runningheads]{llncs}

\usepackage{eccv}

\usepackage{eccvabbrv}

\usepackage{graphicx}
\usepackage{caption} %
\usepackage{subcaption}
\usepackage{booktabs}
\usepackage{amsmath}
\usepackage{multirow}
\usepackage{array} %
\usepackage{tikz}
\usetikzlibrary{spy,calc}
\usepackage[table]{xcolor}
\definecolor{color3}{RGB}{232,245,233} %
\usepackage{placeins} %
\usepackage{dblfloatfix} %

\usepackage[accsupp]{axessibility}  %
\usepackage{tikz}
\usetikzlibrary{spy,calc} %
\usepackage{pgf}          %
\usepackage{wrapfig}

\usepackage{hyperref}

\usepackage{orcidlink}

\newcommand{\myname}[0]{UniFixer} %

\begin{document}

\title{UniFixer: A Universal Reference-Guided Fixer for Diffusion-Based View Synthesis}

\titlerunning{UniFixer: A Universal Reference-Guided Fixer}

\author{Sihan Chen\inst{1}\orcidlink{0009-0001-7042-148X} \and
Xiang Zhang\textsuperscript{\textdagger}\inst{1, 2}\orcidlink{0000-0003-2004-5794} \and
Yang Zhang\inst{2}\orcidlink{0000-0002-2381-6067} \and
Tunc Aydin\inst{2}\orcidlink{0009-0002-0415-1415} \and
Christopher Schroers\inst{2}\orcidlink{0000-0003-1473-1878}}

\authorrunning{S. Chen, X. Zhang, Y. Zhang, T. Aydin, C. Schroers}

\institute{ETH Zürich, Switzerland \\ \email{\{sihchen, xianzhang\}@ethz.ch}\\ \and
DisneyResearch|Studios, Switzerland \\
\email{\{yang.zhang, tunc.aydin, christopher.schroers\}@disneyresearch.com}\\
}
\maketitle
\begingroup
\renewcommand{\thefootnote}{}
\footnotetext{\textsuperscript{\textdagger}Corresponding author}
\endgroup

\begin{abstract}
With the recent surge of generative models, diffusion-based approaches have become mainstream for view synthesis tasks, either in an explicit depth-warp-inpaint or in an implicit end-to-end manner. Despite their success, both paradigms often suffer from noticeable quality degradation, \eg, blurred details and distorted structures, caused by pixel-to-latent compression and diffusion hallucination. In this paper, we investigate diffusion degradation from three key dimensions (\ie, spatial, temporal, and backbone-related) and propose \textbf{\myname}, a universal reference-guided framework that fixes diverse degradation artifacts via a coarse-to-fine strategy. Specifically, a reference pre-alignment module is first designed to perform coarse alignment between the reference view and the degraded novel view. A global structure anchoring mechanism then rectifies geometric distortions to ensure structural fidelity, followed by a local detail injection module that recovers fine-grained texture details for high-quality view synthesis. 
Our \myname\ serves as a plug-and-play refiner that achieves zero-shot fixing across different types of diffusion degradation, and extensive experiments verify our state-of-the-art performance on novel view synthesis and stereo conversion.


\end{abstract}

\begin{figure*}[t]
  \centering
  \includegraphics[width=\textwidth]{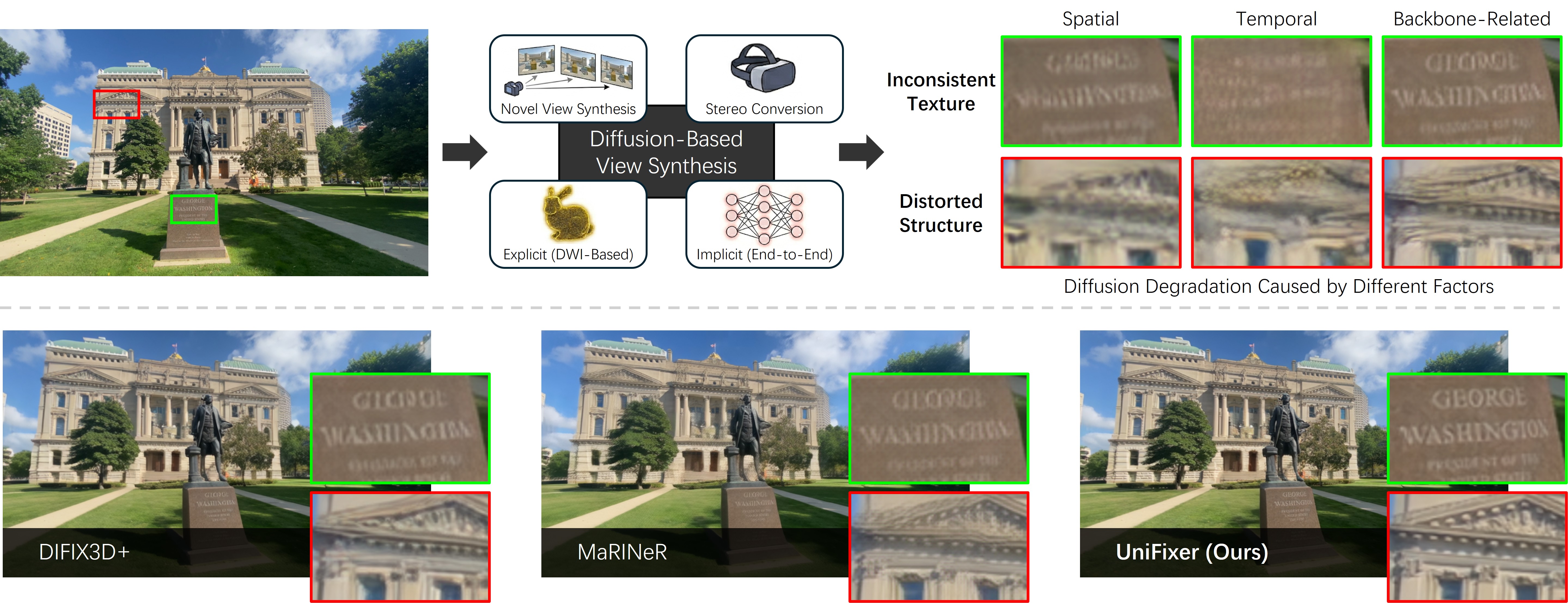}
  \caption{Existing diffusion-based view synthesis, including explicit/implicit novel view synthesis and stereo conversion approaches, often suffer from diffusion degradations (\eg, inconsistent textures and distorted structures) due to pixel-to-latent compression and diffusion hallucination. Moreover, diffusion degradation varies with different spatial resolutions, temporal dynamics, and diffusion backbones, posing significant challenges for high-fidelity view synthesis. Compared with previous novel view fixers, our \myname\ shows state-of-the-art performance and generalizes to different types of degradations for plug-and-play enhancement.}
  \label{fig:teaser}
  \vspace{-0.8em}
\end{figure*}

\section{Introduction}
\label{sec:intro}
View synthesis aims to generate high-fidelity images at novel camera poses, facilitating immersive scene perception beyond the originally captured viewpoints. Two pivotal tasks have emerged within this domain: Novel View Synthesis (NVS), which renders a scene from previously unseen camera poses~\cite{yu2024viewcrafter,bai2025recammaster,zhang2026hairguard,fan2025omniview}, and Stereo Conversion (SC), a specialized form of NVS that lifts monocular content to stereoscopic pairs for 3D displays~\cite{zhao2024stereocrafter,yu2025mono2stereo,shen2025stereopilot,metzger2025elastic3d}. These capabilities underpin a wide range of applications, including free-viewpoint video, immersive telepresence, AR/VR, 3D filmmaking, and scalable 3D content creation from casual captures~\cite{mehl2024stereoconv,bahmani2025lyra,liang2024wonderland}.

\par 
Recently, diffusion-based generative models have reshaped the landscape of view synthesis. Existing methods typically fall into two paradigms: (i) \textit{explicit Depth-Warp-Inpaint (DWI) methods} that estimate depth, warp observations to the target view, and finally inpaint disoccluded regions~\cite{yu2024viewcrafter,jiang2025vace,zhao2024stereocrafter,yu2025mono2stereo}, and (ii) \textit{implicit end-to-end approaches} that directly generate target views conditioned on the input observations~\cite{bai2025recammaster, shen2025stereopilot,fan2025omniview}. Despite promising results achieved, both paradigms often produce novel views with degraded structures and texture details (\cref{fig:teaser}), due to pixel-to-latent compression and diffusion-induced hallucination~\cite{zhang2025high,zhang2026hairguard,metzger2025elastic3d}. Moreover, diffusion degradation varies with multifaceted factors, \eg, spatial resolutions, temporal dynamics, and diffusion backbones, underscoring the pressing demand for a universal and robust novel view fixer.

\par 
Several fixers have been developed to improve view synthesis performance by utilizing high-quality reference views~\cite{wu2025difix3d+,bosiger2024mariner}. For instance, DIFIX3D+ leverages pre-trained diffusion priors to alleviate artifacts in the rendered novel views~\cite{wu2025difix3d+}, and MaRINeR performs deep feature matching to aggregate high-quality features from reference views~\cite{bosiger2024mariner}. However, existing designs primarily operate within low-resolution latent or feature spaces for novel view fixing, which creates an information bottleneck that hinders the transfer of fine-grained textures. Furthermore, previous approaches remain vulnerable to domain shifts and struggle to bridge large viewpoint gaps, often resulting in the loss of high-frequency local details or even the distortions of structures in cases with severe degradation (\cref{fig:teaser}).

\par 
Our goal is to develop a universal novel view fixer that performs robustly across different tasks, model backbones, and degradation patterns without additional re-training. To this end, we present \textbf{\myname}, a reference-guided framework to fix diverse diffusion degradations in a coarse-to-fine manner. In particular, our \myname\ consists of three main modules: Reference Pre-Alignment (RPA), Global Structure Anchoring (GSA), and Local Detail Injection (LDI). Firstly, we warp the reference view to align with the degraded novel view via RPA, which performs coarse alignment to alleviate the spatial search burden for subsequent feature extraction and correspondence matching. We then employ GSA to aggregate deep features from the reference views and fix structure distortions in the novel views. Finally, we design LDI with deformable convolution and gating mechanisms to adaptively compensate for geometric misalignments and filter out warping artifacts, recovering fine-grained local details for high-fidelity view synthesis. Benefiting from the proposed coarse-to-fine refinement scheme, our \myname\ shows remarkable generalization across different degradation types and achieves state-of-the-art performance when applied to NVS and SC tasks in a plug-and-play manner.

In summary, our main contributions are three-fold:
\begin{itemize}
  \item We provide a comprehensive analysis of diffusion degradation in view synthesis across three key dimensions: spatial resolution, temporal dynamics, and model architectures. The results show degradation artifacts in even state-of-the-art view synthesis approaches and reveal significant distribution gaps among different degradation types. 
  \item We propose a universal fixer (\myname) to handle diverse diffusion degradation in a coarse-to-fine manner. After aligning high-quality reference views via RPA, we employ GSA and LDI modules to progressively correct structure distortions and recover fine-grained texture details for high-quality view synthesis.
  \item  Extensive experiments verify the state-of-the-art performance of our \myname\ in NVS and SC tasks. Furthermore, our method exhibits robust zero-shot generalization to unseen degradation types and functions as a plug-and-play fixer that improves both explicit and implicit view synthesis methods. 
\end{itemize}

\section{Related Work}
\label{sec:related}

\subsection{Explicit View Synthesis}
\label{subsec:explicit}
Explicit diffusion-based view synthesis methods typically adopt a Depth-Warp-Inpaint (DWI) pipeline~\cite{wang2024stereodiffusion,yu2024viewcrafter,ren2025gen3c,zhang2025high,ren2025gen3c,zhang2026hairguard}: they first estimate depth/disparity (and camera pose if needed)~\cite{yang2024depth,ke2024repurposing,zhang2024betterdepth,lin2025depth}, warp available observations to the target view~\cite{niklaus2020softmax}, and then inpaint disoccluded regions using generative diffusion models~\cite{ju2024brushnet,jiang2025vace,yu2024viewcrafter}. Building upon this pipeline, ViewCrafter employs 3D point clouds for novel view rendering and iterative scene reconstruction~\cite{yu2024viewcrafter}. GEN3C proposes a 3D-informed video diffusion model to produce consistent novel views from spatial-temporal 3D caches~\cite{ren2025gen3c}. Considering that the depth quality is crucial for explicit methods, the recent approach HairGuard designs a depth fixer to improve novel view synthesis quality~\cite{zhang2026hairguard}. For stereo conversion, StereoCrafter proposes an auto-regressive strategy and tiled processing to handle videos with different lengths and resolutions~\cite{zhao2024stereocrafter}, and M2SVid extends the spatial attention in the Stable Video Diffusion~\cite{blattmann2023svdiff} for high-quality inpainting~\cite{shvetsova2025m2svid}. Although DWI-based approaches show advantages in controllability, geometry estimation and warping can be imperfect in practice. For instance, depth/disparity errors and imperfect visibility reasoning often introduce misalignment and distortions near occlusion boundaries and thin structures. Meanwhile, reflective/refractive regions can further make correspondence ambiguous due to multi-depth (non-Lambertian) effects~\cite{shen2025stereopilot}.

\subsection{Implicit View Synthesis}

Implicit view synthesis methods usually perform end-to-end generation without explicit warping or geometry estimation~\cite{bai2025recammaster,chu2025wanmove,zhou2025stablevirtual,fan2025omniview,gao2024cat3d}. For instance, ReCamMaster~\cite{bai2025recammaster} directly synthesizes novel views conditioned on the input observations and camera information (e.g., via Pl\"ucker camera embeddings~\cite{sitzmann2021light,he2024cameractrl}), leveraging diffusion priors for multi-view reasoning and view synthesis. 
Considering the view-dependent effects and depth ambiguity in complex scenarios, \eg, scenes with transparent objects or reflective surfaces, implicit view synthesis approaches are also employed for stereo conversion~\cite{shen2025stereopilot,behrens2025stereospace,xing2025stereoworld}. Eye2eye utilizes two diffusion models as a generator and a refiner to produce stereo pairs with realistic view-dependent effects~\cite{geyer2025eye2eye}. StereoPilot~\cite{shen2025stereopilot} leverages a pre-trained video diffusion transformer to generate stereoscopic outputs in an end-to-end manner, learning stereo priors from data to address depth ambiguity. While implicit models show promising results under complex scenes and may reduce the dependency on geometry estimators, existing methods often suffer from diffusion degradations, \eg, blurred details and distorted structures, due to pixel-to-latent compression and diffusion-induced hallucination.

\subsection{Reference-Guided Fixer}
Reference images have been widely used for enhancement in computer vision tasks like super-resolution~\cite{lu2021masa,cao2022reference,bernasconi2025rebair} and style transfer~\cite{an2021artflow,yoo2019photorealistic,kolkin2022neural}. In view synthesis tasks, a desirable fixer should preserve the overall layout and color distribution while correcting structure distortions and recovering fine-grained details in the novel views. The previous method DIFIX3D+ employs pre-trained diffusion models to aggregates reference features in the latent space for novel view enhancement~\cite{wu2025difix3d+}. However, latent-space fusion can limit the transfer of fine-grained details from the reference image and becomes brittle when large spatial gaps exist between the reference and the degraded view. MaRINeR leverages deep feature matching to estimate the texture correspondence for guided refinement, but the feature matching is often performed on low-resolution feature maps due to computational efficiency, making it difficult to recover fine-grained textures~\cite{bosiger2024mariner}. In contrast, our \myname\ designs a coarse-to-fine pipeline to improve both global structures and local texture details for high-fidelity view synthesis.

\section{Degradation Analysis}
\label{sec:degradation}
Due to pixel-to-latent compression and diffusion hallucination, existing diffusion-based view synthesis methods often suffer from degradations like distorted structures and inconsistent textures, as illustrated in \cref{fig:teaser}. To facilitate the analysis of diffusion degradations, we first introduce a feature visualization approach using t-SNE~\cite{van2008visualizing}. Given a degraded image $I_{deg}$, \ie, outputs from diffusion-based view synthesis methods, and its corresponding ground truth $I_{gt}$, we extract per-image features $F_{deg}$ and $F_{gt}$ using a frozen DINOv3~\cite{simeoni2025dinov3} encoder $\Phi(\cdot)$, \ie,
\def\imgWidth{0.325\linewidth} %
\def\cropWidth{0.16\linewidth} %

\def\rebigone{(1.95, -0.6)} %
\def\refour{(-0.17,-0.08)} %

\def\sizewidth{1.2cm} %
\def\sizeheight{0.8cm} %
\def\ssmag{4}

\begin{figure*}[!t]
  \centering
  \tikzstyle{img} = [rectangle, minimum width=\imgWidth,draw=black]
  \vspace{1mm}
  \includegraphics[width=\linewidth]{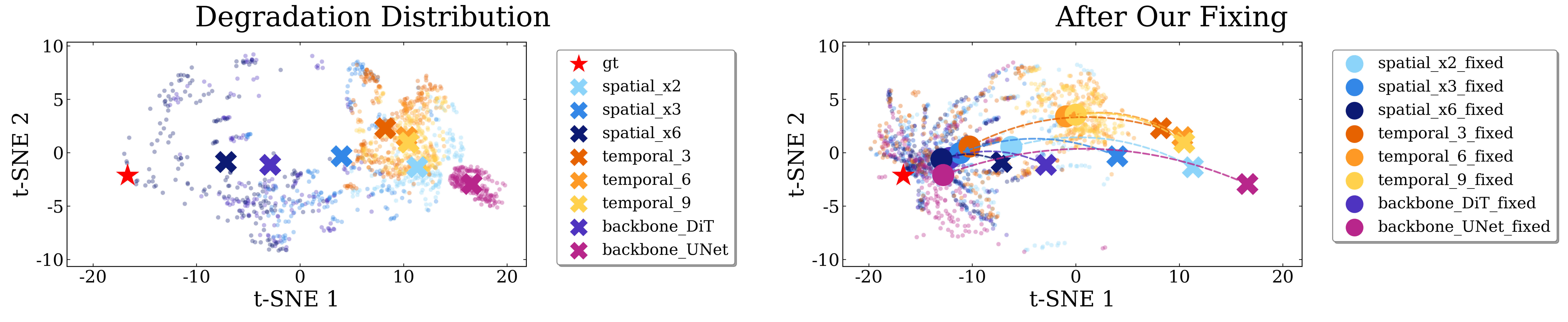}
  \begin{subfigure}{\imgWidth}
        \begin{tikzpicture}[spy using outlines={green,magnification=\ssmag,rectangle,width=\sizewidth, height=\sizeheight},inner sep=0]
            \node [align=center, img] {\includegraphics[width=\textwidth, trim={0 50 0 0}, clip]{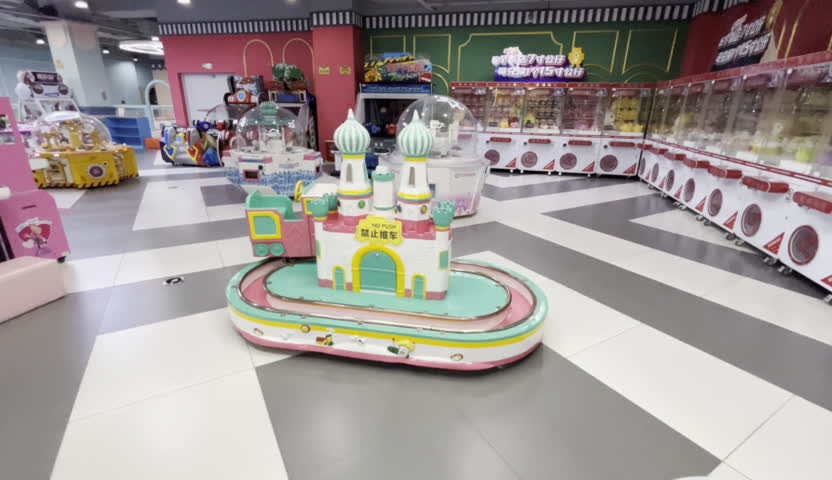}};
            \spy on \refour in node [left] at \rebigone;
    	\end{tikzpicture}
     \caption*{Ground truth}
    \end{subfigure}
    \begin{subfigure}{\imgWidth}
        \begin{tikzpicture}[spy using outlines={green,magnification=\ssmag,rectangle,width=\sizewidth, height=\sizeheight},inner sep=0]
            \node [align=center, img] {\includegraphics[width=\textwidth, trim={0 50 0 0}, clip]{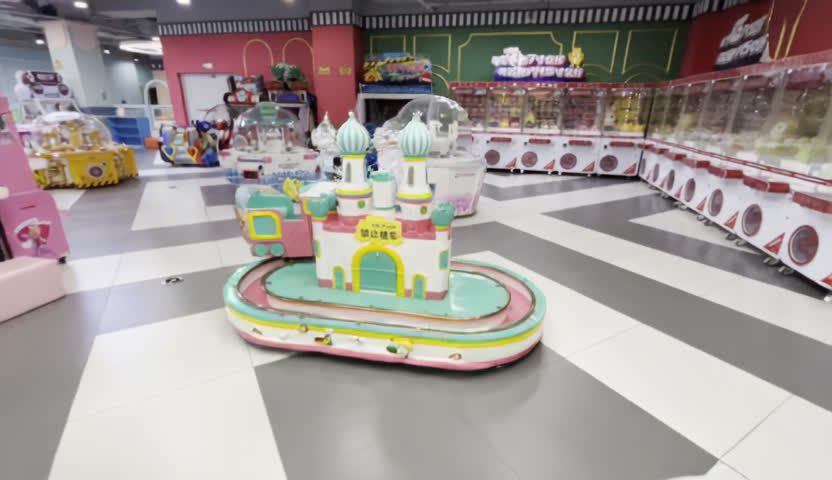}};
            \spy on \refour in node [left] at \rebigone;
    	\end{tikzpicture}
     \caption*{backbone\_DiT}
    \end{subfigure}
    \begin{subfigure}{\imgWidth}
        \begin{tikzpicture}[spy using outlines={green,magnification=\ssmag,rectangle,width=\sizewidth, height=\sizeheight},inner sep=0]
            \node [align=center, img] {\includegraphics[width=\textwidth, trim={0 50 0 0}, clip]{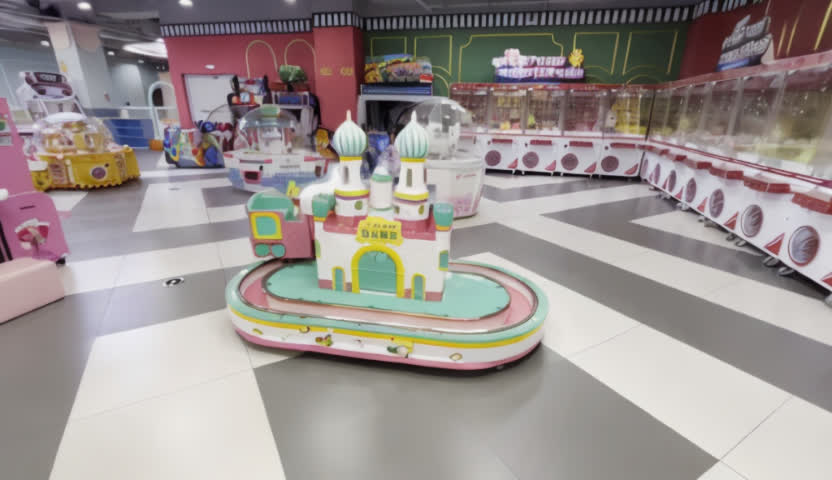}};
            \spy on \refour in node [left] at \rebigone;
    	\end{tikzpicture}
     \caption*{backbone\_UNet}
    \end{subfigure}
    \begin{subfigure}{\cropWidth}
        \begin{tikzpicture}[spy using outlines={green,magnification=\ssmag,rectangle,width=\sizewidth, height=\sizeheight},inner sep=0]
            \node [align=center, img] {\includegraphics[width=\textwidth, trim={348 228 418 212}, clip]{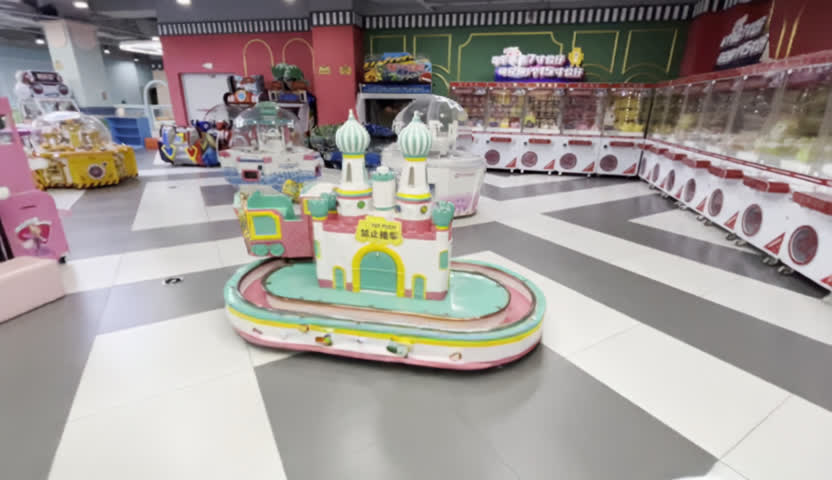}};
    	\end{tikzpicture}
     \caption*{spatial\_$\times6$}
    \end{subfigure}
    \begin{subfigure}{\cropWidth}
        \begin{tikzpicture}[spy using outlines={green,magnification=\ssmag,rectangle,width=\sizewidth, height=\sizeheight},inner sep=0]
            \node [align=center, img] {\includegraphics[width=\textwidth, trim={348 228 418 212}, clip]{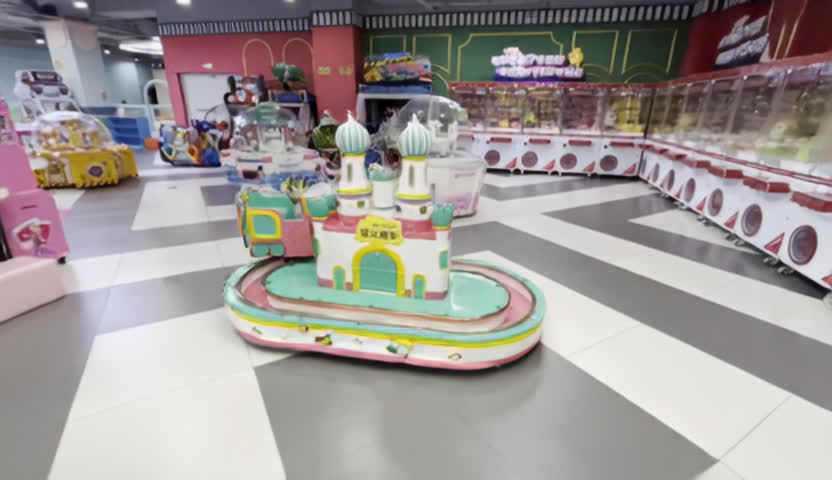}};
    	\end{tikzpicture}
     \caption*{spatial\_$\times3$}
    \end{subfigure}
    \begin{subfigure}{\cropWidth}
        \begin{tikzpicture}[spy using outlines={green,magnification=\ssmag,rectangle,width=\sizewidth, height=\sizeheight},inner sep=0]
            \node [align=center, img] {\includegraphics[width=\textwidth, trim={348 228 418 212}, clip]{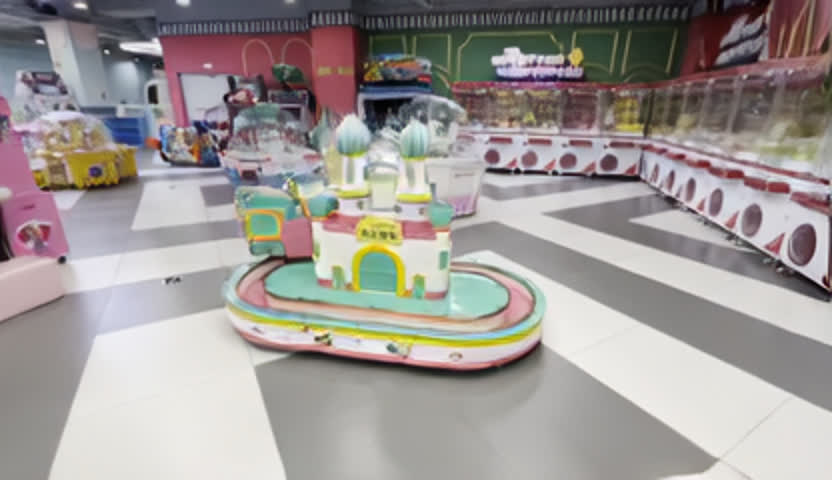}};
    	\end{tikzpicture}
     \caption*{spatial\_$\times2$}
    \end{subfigure}
    \begin{subfigure}{\cropWidth}
        \begin{tikzpicture}[spy using outlines={green,magnification=\ssmag,rectangle,width=\sizewidth, height=\sizeheight},inner sep=0]
            \node [align=center, img] {\includegraphics[width=\textwidth, trim={348 228 418 212}, clip]{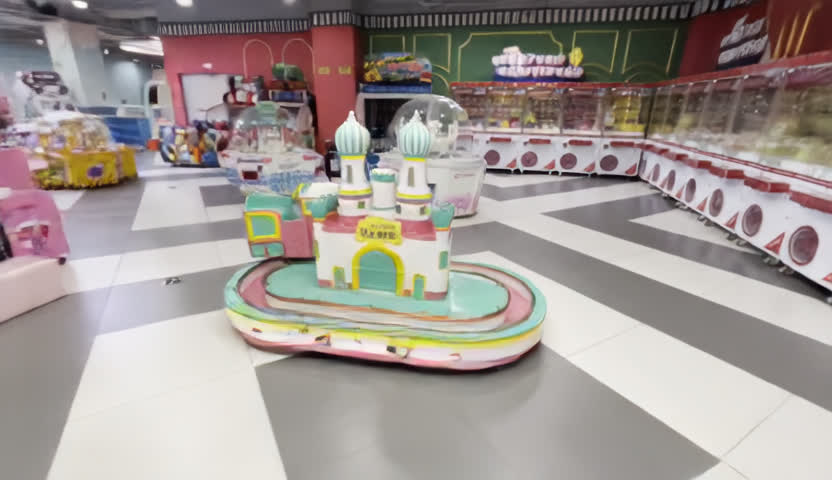}};
    	\end{tikzpicture}
     \caption*{temporal\_3}
    \end{subfigure}
    \begin{subfigure}{\cropWidth}
        \begin{tikzpicture}[spy using outlines={green,magnification=\ssmag,rectangle,width=\sizewidth, height=\sizeheight},inner sep=0]
            \node [align=center, img] {\includegraphics[width=\textwidth, trim={348 228 418 212}, clip]{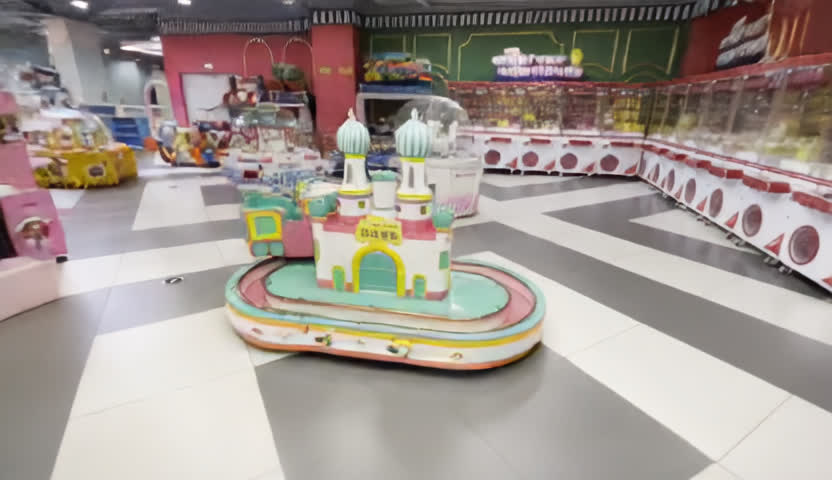}};
    	\end{tikzpicture}
     \caption*{temporal\_6}
    \end{subfigure}
    \begin{subfigure}{\cropWidth}
        \begin{tikzpicture}[spy using outlines={green,magnification=\ssmag,rectangle,width=\sizewidth, height=\sizeheight},inner sep=0]
            \node [align=center, img] {\includegraphics[width=\textwidth, trim={348 228 418 212}, clip]{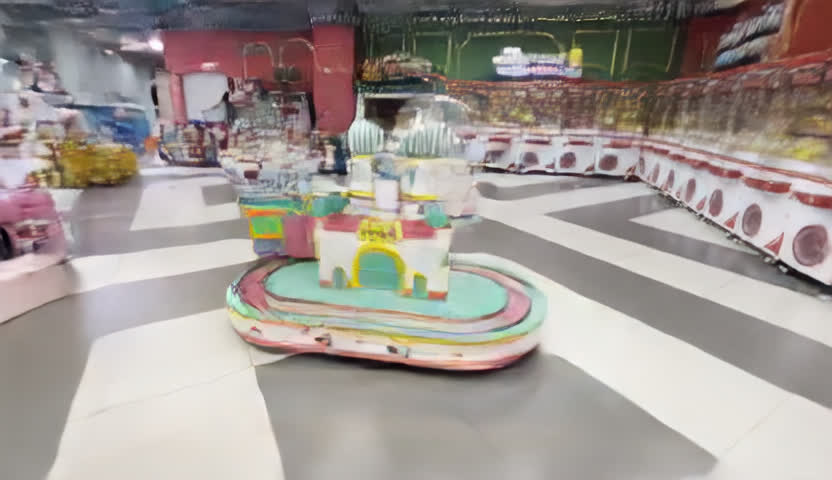}};
    	\end{tikzpicture}
     \caption*{temporal\_9}
    \end{subfigure}
  \caption{Degradation analysis with t-SNE feature visualization of spatial ($\times2 / \times3 / \times4 / \times6$), temporal (frame strides $\in\{1, 3, 6, 9\}$), and backbone-related (UNet~\cite{yu2024viewcrafter} and DiT~\cite{jiang2025vace}) degradations on a DL3DV~\cite{ling2024dl3dv} scene. The shared setting (spatial $\times4$, temporal stride 1, and DiT-based backbone) is plotted once as backbone\_DiT. Color families denote degradation types, and the red star marks the ground-truth. Crosses/circles denote cluster means before/after our fixing for each degradation type, with dashed arrows indicating the shift. Each cluster comprises 250 samples, with outliers beyond $1\sigma$ omitted for clarity. Visual examples of different diffusion degradations are provided.}
  \label{fig:degradation_tsne}
\end{figure*}
\FloatBarrier

\begin{equation}
    F_{deg} = \Phi(I_{deg})\in \mathbb{R}^{L\times D},\  F_{gt} = \Phi(I_{gt})\in \mathbb{R}^{L\times D},
\end{equation}
where $L$ and $D$ indicate the number and dimension of tokens. We only keep patch tokens in $F_{deg}$ and $F_{gt}$ to focus on local degradation details. Then, we aggregate patch tokens into compact per-image embeddings by concatenating the mean and standard deviation over all patch tokens, \ie,
$   
    \hat{F}_{i} = \left[ \mu(F_{i}), \sigma(F_{i}) \right] \in \mathbb{R}^{2D},\ i \in \{deg, gt\},
$
where $[\cdot, \cdot]$ denotes concatenation.
Finally, we compute the difference between $\hat{F}_{deg}$ and $\hat{F}_{gt}$ as the feature representation of diffusion degradations, \ie,
\begin{equation}
\hat{F}_{diff} = \lvert \hat{F}_{deg} - \hat{F}_{gt} \rvert \in \mathbb{R}^{2D}.
\label{eq:fdeg}
\end{equation}
\cref{eq:fdeg} mitigates the impact of the image content in the feature representation to facilitate the analysis of diffusion degradations.

\par 
Leveraging the feature visualization protocol, we investigate diffusion degradations from three key dimensions: spatial resolutions, temporal dynamics, and diffusion backbones:
\begin{itemize}
    \item \textbf{Spatial Degradation.} Spatial degradation is one of the most common degradation types for diffusion-based models. Since modern diffusion models typically project images into the latent space to alleviate computational overhead via a Variational Auto-Encoder (VAE)~\cite{kingma2013auto}, the latent spatial resolution directly affects the synthesis fidelity. Following~\cite{simeoni2025dinov3}, we keep the VAE compression ratio fixed and draw the feature distributions under varying input image resolutions (\ie, 240$\times$416 / 360$\times$624 / 480$\times$832 / 720$\times$1248, denoted by $\times2 / \times3 / \times4 / \times6$ in \cref{fig:degradation_tsne}). Due to the pixel-to-latent compression, latents with lower resolution inevitably attenuate high-frequency textures and sharp edges, leading to over-smoothed details (\cref{fig:degradation_tsne}). 
    While scaling up the input resolution preserves finer spatial details at the expense of increased computational cost, diffusion artifacts, \eg, hallucination, persist in the synthesized views due to the generative nature of diffusion models.

    \item \textbf{Temporal Degradation.} 
    Recent diffusion models design spatio-temporal VAEs and temporal attention mechanisms to model temporal dynamics and enforce inter-frame coherence~\cite{wan2025wan,blattmann2023svdiff}. However, these methods often struggle under rapid camera motions. Specifically, inter-frame correspondences weaken significantly under large viewpoint shifts, causing temporal priors to either over-smooth high-frequency textures or hallucinate inconsistent details (\cref{fig:degradation_tsne}). To investigate this, we simulate varying degrees of camera motion by modulating the temporal stride between sampled frames (\ie, frame stride $\in\{1, 3, 6, 9\}$). Larger intervals yield more significant camera spatial displacements, thereby exacerbating the severity of temporal degradation.
    
    \item \textbf{Backbone-related Degradation.} 
    Finally, we investigate backbone-specific degradations by comparing the two predominant generative architectures: convolutional UNets~\cite{ronneberger2015u} and transformer-based models (DiTs)~\cite{peebles2023scalable}. Despite both operating within similar latent diffusion paradigms, their distinct inductive biases result in different artifact distributions (\cref{fig:degradation_tsne}). Specifically, convolution-based UNets tend to over-smooth fine textures after repeated down-/up-sampling. Although DiT-based models exhibit superior performance over UNet-based ones, transformer backbones often introduce patch-wise inconsistencies or structure hallucinations under strong domain shifts.
\end{itemize}
\par
As shown in Fig.~\ref{fig:degradation_tsne}, degradations from different dimensions form separated distributions in the feature space, with varying scales inducing further distributional shifts. Despite the distributional divergence across degradation types and scales, our \myname\ consistently pushes each degradation cluster towards the ground-truth, demonstrating robust zero-shot generalization across diverse diffusion degradations.

\section{Method}
\label{sec:Method}

\begin{figure}[t]
  \centering
  \begin{tikzpicture}[inner sep=0]
    \node {\includegraphics[width=\linewidth]{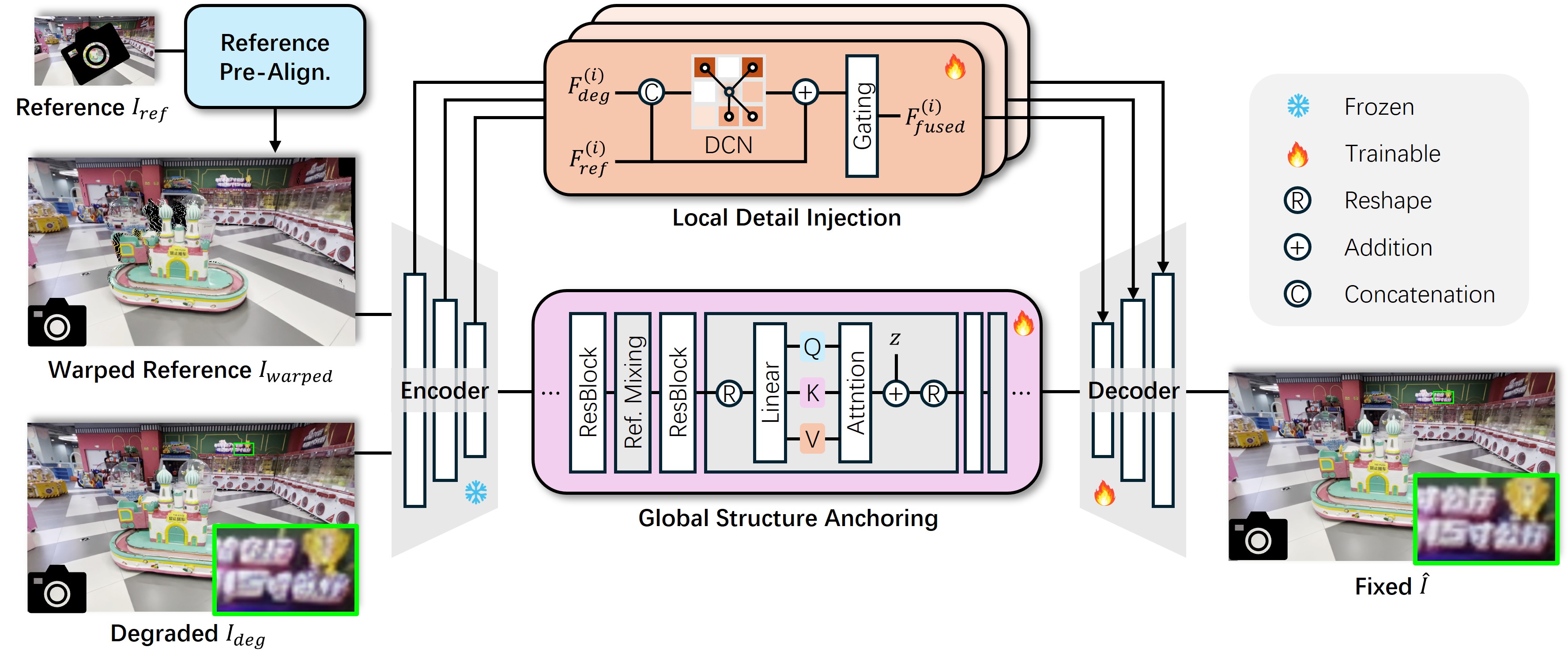}};
  \end{tikzpicture}
  \caption{Pipeline of \myname. Given a degraded novel view and a high-quality reference view, we first perform coarse alignment by the reference pre-alignment module. Leveraging the warped reference, we apply the global structure anchoring module to aggregate shared structures via reference-mixed attention. The local detail injection module adaptively fuses multi-scale features for fine-grained texture enhancement. }
  \label{fig:arch}
\end{figure}

Given a degraded novel view $I_{deg} \in \mathbb{R}^{3  \times H \times W}$ from view synthesis methods (\eg, NVS or SC) and a reference image $I_{ref} \in \mathbb{R}^{3 \times H \times W}$, our goal is to reconstruct a high-quality image $\hat{I} \in \mathbb{R}^{3 \times H \times W}$ that corrects distorted structures and restores texture details in the novel view. To handle diverse diffusion degradations with a universal fixer, we design a coarse-to-fine refinement framework as shown in \cref{fig:arch}. Specifically, we first perform reference pre-alignment (\cref{subsec:warp}) to warp reference images toward the target novel view, thereby reducing the spatial search space for subsequent feature extraction and correspondence matching. With the pre-aligned reference, we anchor shared structural information in deep features via reference-mixed self-attention (\cref{subsec:global}). To recover fine-grained texture details, we design local detail injection modules to adaptively fuse reference features for enhancement while filtering warping errors (\cref{subsec:local}). Finally, an efficient strategy is proposed for model training (\cref{subsec:train}).

\FloatBarrier %

\subsection{Reference Pre-Alignment}
\label{subsec:warp}
To alleviate the difficulty of long-range correspondence search, we first perform coarse alignment by warping $I_{ref}$ to the camera view of $I_{deg}$, generating the warped reference $I_{warped}$. This step brings relevant structural cues directly into the local spatial neighborhood of each pixel in $I_{deg}$, transforming an exhaustive global correspondence search into a tractable, localized refinement task.
Furthermore, by operating directly in the raw pixel space, this warping process preserves fine-grained texture details that are crucial for high-fidelity view synthesis. Specifically, the pre-alignment process can be formulated as
\begin{equation}
    I_{warped} = \mathcal{W}(I_{ref}, T),
\end{equation}
where $\mathcal{W}(\cdot)$ denotes the warping operator (\eg, softmax splatting~\cite{niklaus2020softmax}), and $T$ indicates the view transformation map.

\begin{wrapfigure}{r}{0.5\linewidth}
\centering
\begin{tikzpicture}[inner sep=0]
    \node {\includegraphics[width=\linewidth]{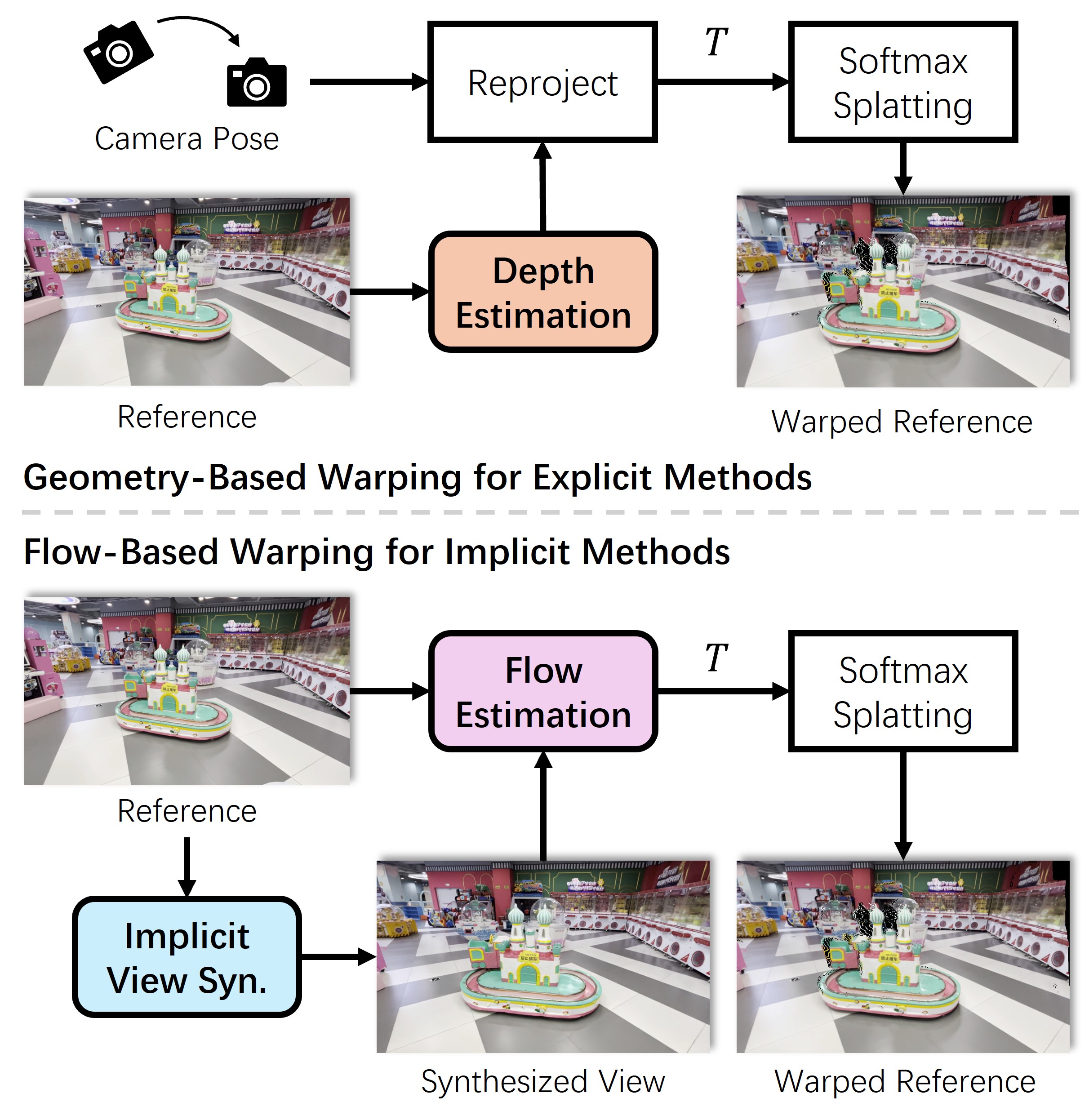}};
  \end{tikzpicture}
  \caption{Reference pre-alignment. For explicit view synthesis methods, \ie, DWI-based methods, we estimate depth information and perform geometry-based warping using camera poses. For implicit approaches, we perform flow-based warping with the optical flow estimated from the reference and the synthesized novel view.}
  \label{fig:warper}
\end{wrapfigure}

As shown in \cref{fig:warper}, the generation of $I_{warped}$ depends on the type of view synthesis approaches:
\begin{itemize}
    \item \textbf{Geometry-Based Warping for Explicit Methods:} For methods following the Depth-Warp-Inpaint (DWI) paradigm, the warped reference $I_{warped}$ is usually accessible through geometric transformations. Depending on the view synthesis task, this is typically achieved via perspective projection for NVS or disparity mapping for SC. Aligning with these established pipelines, we employ geometry-based warping to perform pre-alignment.
    \item \textbf{Flow-Based Warping for Implicit Methods:} Implicit view synthesis approaches usually render novel views in an end-to-end manner without relying on explicit geometry like depth maps~\cite{bai2025recammaster,shen2025stereopilot}, making geometry-based warping difficult. Nevertheless, recent implicit view synthesis methods have exhibited promising performance in achieving the overall appearance consistency across viewpoints, \eg, layout, illumination, and color distribution. Leveraging this characteristic, we perform flow-based warping with the optical flow estimated from the reference and the synthesized view (SEA-RAFT~\cite{wang2024sea} is used in our implementation) for reference pre-alignment.
\end{itemize}

\subsection{Global Structure Anchoring}
\label{subsec:global}
The goal of GSA is to leverage deep features from the reference view to rectify structural distortions, while simultaneously preserving the high-level semantics (such as the overall layout and appearance) of the degraded view. To this end, we implement \myname\ upon the pre-trained one-step diffusion model, SD-Turbo~\cite{sauer2024sdturbo}, to harness the robust feature extraction capabilities learned by its attention layers. To efficiently aggregate reference features, we follow DIFIX3D+~\cite{wu2025difix3d+} to implement the reference mixing mechanism in self-attention layers. Specifically, we first extract the latent space representations of $I_{deg}$ and $I_{warped}$, and concatenate them as input to GSA, denoted by $\mathbf{z}$:
\begin{equation}
    \mathbf{z} = [\mathcal{E}(I_{deg}), \mathcal{E}(I_{warped})] \in \mathbb{R}^{2C \times H \times W}
\end{equation}
where $\mathcal{E}(\cdot)$ denotes the frozen latent encoder. To capture the structural information in the reference view, we reshape the latents into a token sequence $\mathbf{z} \in \mathbb{R}^{(2HW) \times C}$. The anchoring operation is then defined as:
\begin{equation}
    \mathbf{z} := \mathbf{z} + \text{Softmax}\left(\frac{Q K^T}{\sqrt{d}}\right) V, \quad Q, K, V = \text{Linear}(\mathbf{z}).
\end{equation}
By computing attention across the concatenated $2 \times H \times W$ spatial-view tokens and applying a residual connection to the base latent representation, the features of $I_{deg}$ serve as anchors to aggregate shared structures from the reference view. Consequently, this process enforces structural alignment and regularizes the diffusion process to prevent content and color drift.

\subsection{Local Detail Injection}
\label{subsec:local}

Restoring high-fidelity textures requires fine-grained spatial cues. However, the spatial compression inherent in the pixel-to-latent conversion often fails to preserve high-frequency details in the bottleneck latent $\mathbf{z}$, making the recovery of high-fidelity textures highly ill-posed. To this end, we propose the LDI module to adaptively fuse fine-grained features from the latent encoder for local detail refinement.

The LDI operates across multiple feature scales. For each scale $i$, we extract the intermediate feature maps from the frozen encoder $\mathcal{E}$, \ie, 
\begin{equation}
\Bigl(F_{deg}^{(i)}\Bigr)_{i=1}^{n} \leftarrow \mathcal{E}(I_{deg}), \qquad
\Bigl(F_{ref}^{(i)}\Bigr)_{i=1}^{n} \leftarrow \mathcal{E}(I_{warped}),
\end{equation}
where $F^{(i)} \in \mathbb{R}^{C_i \times H_i \times W_i}$ denotes the feature at the $i$-th feature scale, and $n$ indicates the total number of feature scales. As these features are extracted before the deep bottleneck, they preserve rich low-level geometric cues and fine-grained texture information. However, since the coarse alignment in RPA (Sec.~\ref{subsec:warp}) often suffers from inaccuracies, either originating from depth estimation errors or from noisy optical flow, directly fusing these encoder features would introduce ghosting artifacts and blurriness. To address this, we perform detail injection through adaptive warping deformation and uncertainty-aware gated fusion.

\textbf{Adaptive Warping Deformation.} For each scale $i$, we employ Deformable Convolutional Networks (DCN) \cite{dai2017deformableconvolutionalnetworks} to rectify micro-misalignments. Since $I_{warped}$ is already in the spatial neighborhood of $I_{deg}$, the DCN only needs to predict small-range offsets $\Delta p^{(i)}$ and the associated mask $M^{(i)}$ to establish precise correspondence, \ie,
\begin{equation}
\bigl(\Delta p^{(i)},\, M^{(i)}\bigr) = \mathcal{P}_{\theta}^{(i)}\bigl([F_{deg}^{(i)}, F_{ref}^{(i)}]\bigr),
\end{equation}
\begin{equation}
    \tilde{F}_{ref}^{(i)} = F_{ref}^{(i)} + \text{DCN}(F_{ref}^{(i)}, \Delta p^{(i)}, M^{(i)}),
\end{equation}
where $\mathcal{P}_{\theta}^{(i)}$ denotes the offset and mask estimator of the i-th scale. By deforming the reference features, high-frequency details from the warped reference are accurately aligned to enhance the degraded view.

\textbf{Uncertainty-Aware Gated Fusion.} Despite the adaptive deformation of DCN, certain regions (\eg, occlusions or areas with extreme distortion) in $\tilde{F}_{ref}^{(i)}$ may remain unreliable due to warping errors. To prevent these artifacts from propagating through the restoration pipeline, we introduce a gated fusion mechanism to estimate the confidence of the reference cues:
\begin{equation}
    G^{(i)} = \sigma \left( \text{Conv} \left( [F_{deg}^{(i)}, \tilde{F}_{ref}^{(i)}] \right) \right),
\end{equation}
\begin{equation}
    F_{fused}^{(i)} = G^{(i)} \odot F_{deg}^{(i)} + (1 - G^{(i)}) \odot \tilde{F}_{ref}^{(i)},
\end{equation}
where $\text{Conv}(\cdot)$ is 2D convolution layer, $\sigma$ is the Sigmoid function, $\odot$ is the element-wise multiplication, and $G^{(i)} \in [0,1]^{C_i\times H_i\times W_i}$ is the estimated confidence map. This adaptive fusion allows the fixer to select high-fidelity features from $\tilde{F}_{ref}^{(i)}$ while filtering artifacts caused by warping errors. Finally, the fused feature $F_{fused}^{(i)}$ is injected into the corresponding decoder layer to guide the decoding of $\mathbf{z}$, \ie,
\begin{equation}\label{eq:decoder-output}
    \hat{I} = \mathcal{D}\bigl(\mathbf{z}, \mathbf{f}\bigr),
\end{equation}
where $\mathcal{D}(\cdot)$ denotes the latent decoder, and $\mathbf{f} = \bigl(F_{\mathrm{fused}}^{(i)}\bigr)_{i=1}^{n}$.

\subsection{Model Training}\label{subsec:train}
We propose a simple strategy to curate paired data $\{(I^{(i)}_{deg}, I^{(i)}_{warped}, I^{(i)}_{gt})\}_i$ for model training, where $I^{(i)}_{gt}$ indicates the ground-truth image. Given an image sequence $(I^{(i)}_{gt})_{i=1}^N$ from the training dataset with $N$ denoting the sequence length, we first estimate the scene geometry, \eg, depth and camera poses, via scene reconstruction methods (DA3~\cite{lin2025depth} is used in our implementation). Then, we select the middle frame as the reference image $I_{ref}:=I_{gt}^{({\lceil N/2 \rceil})}$ and generate the warped references $(I^{(i)}_{warped})_{i=1}^N$ by warping $I_{ref}$ to the other viewpoints (note that $I^{({\lceil N/2 \rceil})}_{warped}:=I_{ref}$ at the reference viewpoint). Following that, we synthesize the degraded images $(I^{(i)}_{deg})_{i=1}^N$ by applying diffusion models (VACE~\cite{jiang2025vace} is used in \myname) to re-generate the ground-truth images $(I^{(i)}_{gt})_{i=1}^N$, \ie,
\begin{equation}
    (I^{(i)}_{deg})_{i=1}^N = \operatorname{Diffusion}\left((I^{(i)}_{gt})_{i=1}^N\right).
\end{equation}
Consequently, we synthesize $(I^{(i)}_{deg})_{i=1}^N$ with diffusion degradations while ensuring that the underlying scene content remains semantically consistent with $(I^{(i)}_{gt})_{i=1}^N$. This design encourages the model to preserve the overall scene content, \eg, layout and appearance, of the novel view while fixing diffusion degradation artifacts.
Finally, we train \myname\ using $\ell_2$ loss $\mathcal{L}_2$ and perceptual loss $\mathcal{L}_{lpips}$~\cite{zhang2018unreasonable}:
\begin{equation}
     \mathcal{L} = \mathcal{L}_2(\hat{I},I_{gt}) +\mathcal{\lambda}_{lpips}\cdot\mathcal{L}_{lpips}(\hat{I},I_{gt}),
\end{equation}
where $\mathcal{\lambda}_{lpips}$ indicates the balancing weight.
\par 
Although diffusion degradation varies with different factors, \eg, spatial resolutions, temporal dynamics, and diffusion backbones (\cref{sec:degradation}), our training is performed under a fixed configuration (\ie, constant resolution, fixed motion speed, and a single diffusion backbone). By leveraging the coarse-to-fine alignment framework, we shift the learning objective from the degradation-specific modeling to local correspondence searching and reference aggregation. This design bypasses the challenge of explicitly modeling distribution shifts across diverse diffusion degradations, enabling robust zero-shot generalization without the need for extensive retraining or fine-tuning.

\begin{table*}[t]
\centering
\caption{Degradation fixing for novel view synthesis on the DL3DV dataset. Spatial, temporal, and backbone-related degradation denotes degradation from different input resolutions, camera motion speeds, and diffusion backbones, respectively (Sec.~\ref{sec:degradation}). \textcolor{red}{Best} results are marked.}
\label{tab:degradation_fixing_results}
\small
\setlength{\tabcolsep}{2.2pt}

\resizebox{\textwidth}{!}{%
\begin{tabular}{ll*{11}{c}}
\toprule
Degradation & Fixer & PSNR$\uparrow$ & SSIM$\uparrow$ & LPIPS$\downarrow$ & DISTS$\downarrow$ & FID$\downarrow$ & & PSNR$\uparrow$ & SSIM$\uparrow$ & LPIPS$\downarrow$ & DISTS$\downarrow$ & FID$\downarrow$ \\
\midrule[0.15em]
 &  & \multicolumn{5}{c}{$\times 2$ (240$\times$416)} & & \multicolumn{5}{c}{$\times 3$ (360$\times$624)} \\ \cmidrule(lr){3-7} \cmidrule(lr){9-13} 
\multirow{9}{*}{Spatial} & W/O fixer & 21.35 & 0.680 & 0.309 & 0.183 & 37.16 &  & 22.72 & 0.727 & 0.227 & 0.135 & 23.41 \\
 & DIFIX3D+~\cite{wu2025difix3d+} & 21.60 & 0.692 & 0.275 & 0.148 & 31.11 & & 22.86 & 0.732 & 0.195 & 0.103 & 20.81 \\
 & MaRINeR~\cite{bosiger2024mariner} & 21.67 & 0.706 & 0.238 & 0.140 & 33.73 & & 23.08 & 0.750 & 0.179 & 0.103 & 22.69 \\
\rowcolor{gray!15}\multicolumn{1}{c}{\cellcolor{white}{}} & \textbf{Ours} & \textcolor{red}{22.10} & \textcolor{red}{0.754} & \textcolor{red}{0.232} & \textcolor{red}{0.127} & \textcolor{red}{26.55} &  & \textcolor{red}{23.37} & \textcolor{red}{0.784} & \textcolor{red}{0.164} & \textcolor{red}{0.088} & \textcolor{red}{18.27} \\ \cline{2-13}
\addlinespace[0.25em]
\multicolumn{2}{l}{} & \multicolumn{5}{c}{\rule{0pt}{2.2ex}$\times 4$ (480$\times$832)} & & \multicolumn{5}{c}{\rule{0pt}{2.2ex}$\times 6$ (720$\times$1248)} \\ \cmidrule(lr){3-7}  \cmidrule(lr){9-13} 
 & W/O fixer & 23.43 & 0.750 & 0.175 & 0.105 & 19.96 & & 24.01 & 0.777 & 0.179 & 0.107 & 17.45 \\
 & DIFIX3D+~\cite{wu2025difix3d+} & 23.53 & 0.752 & 0.158 & 0.082 & 18.27 & & 23.48 & 0.762 & 0.174 & 0.088 & 17.87 \\
 & MaRINeR~\cite{bosiger2024mariner} & 23.75 & 0.769 & 0.157 & 0.089 & 19.94 &  & 23.61 & 0.772 & 0.157 & 0.087 & 18.84 \\
\rowcolor{gray!15}\multicolumn{1}{c}{\cellcolor{white}{}} & \textbf{Ours} & \textcolor{red}{24.03} & \textcolor{red}{0.793} & \textcolor{red}{0.138} & \textcolor{red}{0.072} & \textcolor{red}{16.66} & & \textcolor{red}{24.32} & \textcolor{red}{0.798} & \textcolor{red}{0.142} & \textcolor{red}{0.074} & \textcolor{red}{15.69} \\
\midrule[0.15em]
 &  & \multicolumn{5}{c}{Stride=1} & & \multicolumn{5}{c}{Stride=3} \\ \cmidrule(lr){3-7}  \cmidrule(lr){9-13} 
\multirow{9}{*}{Temporal} & W/O fixer & 23.43 & 0.750 & 0.175 & 0.105 & 19.96 & & 19.89 & 0.620 & 0.318 & 0.194 & 41.94 \\
 & DIFIX3D+~\cite{wu2025difix3d+} & 23.53 & 0.752 & 0.158 & 0.082 & 18.27 & & 20.00 & 0.626 & 0.289 & 0.156 & 33.55 \\
 & MaRINeR~\cite{bosiger2024mariner} & 23.75 & 0.769 & 0.157 & 0.089 & 19.94 & & 20.05 & 0.635 & 0.310 & 0.186 & 41.39 \\
\rowcolor{gray!15}\multicolumn{1}{c}{\cellcolor{white}{}} & \textbf{Ours} & \textcolor{red}{24.03} & \textcolor{red}{0.793} & \textcolor{red}{0.138} & \textcolor{red}{0.072} & \textcolor{red}{16.66} & & \textcolor{red}{20.23} & \textcolor{red}{0.656} & \textcolor{red}{0.276} & \textcolor{red}{0.150} & \textcolor{red}{33.01} \\
\cline{2-13}
\addlinespace[0.25em]
 &  & \multicolumn{5}{c}{\rule{0pt}{2.2ex}Stride=6} & &\multicolumn{5}{c}{\rule{0pt}{2.2ex}Stride=9} \\ \cmidrule(lr){3-7}  \cmidrule(lr){9-13} 
 & W/O fixer & 18.83 & 0.594 & 0.389 & 0.238 & 56.09 & & 18.16 & 0.585 & 0.435 & 0.264 & 66.09 \\
 & DIFIX3D+~\cite{wu2025difix3d+} & 18.98 & 0.604 & 0.355 & 0.200 & 43.69 & & 18.30 & 0.595 & 0.402 & 0.228 & 51.81 \\
 & MaRINeR~\cite{bosiger2024mariner} & 18.94 & 0.610 & 0.378 & 0.231 & 53.66 & & 18.22 & 0.599 & 0.425 & 0.258 & 62.95 \\
\rowcolor{gray!15}\multicolumn{1}{c}{\cellcolor{white}{}} & \textbf{Ours} & \textcolor{red}{19.23} & \textcolor{red}{0.644} & \textcolor{red}{0.334} & \textcolor{red}{0.190} & \textcolor{red}{42.49} & & \textcolor{red}{18.54} & \textcolor{red}{0.636} & \textcolor{red}{0.380} & \textcolor{red}{0.218} & \textcolor{red}{50.24} \\
\midrule[0.15em]
 &  & \multicolumn{5}{c}{UNet} & & \multicolumn{5}{c}{DiT} \\ \cmidrule(lr){3-7}  \cmidrule(lr){9-13} 
 \multirow{4}{*}{Backbone} & W/O fixer & 21.45 & 0.669 & 0.194 & 0.102 & 22.85 & & 23.43 & 0.750 & 0.175 & 0.105 & 19.96 \\
 & DIFIX3D+~\cite{wu2025difix3d+} & 21.38 & 0.666 & 0.182 & 0.095 & 23.90 & & 23.53 & 0.752 & 0.158 & 0.082 & 18.27 \\
 & MaRINeR~\cite{bosiger2024mariner} & 21.48 & 0.670 & 0.187 & 0.094 & 22.54 & & 23.75 & 0.769 & 0.157 & 0.089 & 19.94 \\
\rowcolor{gray!15}\multicolumn{1}{c}{\cellcolor{white}{}} & \textbf{Ours} & \textcolor{red}{22.19} & \textcolor{red}{0.739} & \textcolor{red}{0.170} & \textcolor{red}{0.086} & \textcolor{red}{20.84} & & \textcolor{red}{24.03} & \textcolor{red}{0.793} & \textcolor{red}{0.138} & \textcolor{red}{0.072} & \textcolor{red}{16.66} \\
\bottomrule
\end{tabular}%
}
\end{table*}

\section{Experiment and Analysis}

\subsection{Experimental Settings}

\textbf{Training.}
Our training set is curated from the first 1K scenes of the DL3DV dataset~\cite{ling2024dl3dv}, which we empirically find to be sufficient for achieving state-of-the-art performance. \myname\ is optimized using Adam~\cite{kingma2014adam} with a learning rate of $2\times10^{-5}$ and a batch size of 1 on $480\times832$ patches, with $\mathcal{\lambda}_{lpips} = 1.0$. The training process converges within 50K iterations, requiring approximately 14 hours on a single NVIDIA A100 GPU (40GB).

\textbf{Evaluation.} 
We employ the DL3DV test benchmark~\cite{ling2024dl3dv}, Mono2Stereo dataset~\cite{yu2025mono2stereo}, Spring dataset~\cite{mehl2023spring}, and the Spatial Video Dataset (SVD)~\cite{izadimehr2025svd} for evaluation. For explicit view synthesis methods, we report PSNR, SSIM~\cite{wang2004image}, LPIPS~\cite{zhang2018unreasonable}, DISTS~\cite{ding2022image}, and FID~\cite{heusel2017gans}. For implicit approaches, where the synthesized outputs may deviate spatially from the ground truth, we use non-reference metrics, including CLIP-IQA~\cite{wang2023exploring}, MUSIQ~\cite{ke2021musiq}, and MANIQA~\cite{yang2022maniqa}. FID is also computed between the fixed results and the source video frames to measure distribution differences. We mainly compare \myname\ with state-of-the-art novel view fixers DIFIX3D+~\cite{wu2025difix3d+} and MaRINeR~\cite{bosiger2024mariner}. To ensure fair comparisons, we retrain DIFIX3D+ and MaRINeR on our training dataset using their official training codes. For degradation analysis and ablation study, we conduct our experiments under the NVS setting.

\def\imgWidth{0.185\textwidth} %

\tikzset{img/.style={rectangle, minimum width=\imgWidth, draw=black, inner sep=0, outer sep=0}}

\providecommand{\ZoomPicBySampling}[6]{}
\renewcommand{\ZoomPicBySampling}[6][north east]{%
  \begin{subfigure}{\imgWidth}
    \centering
    \begin{tikzpicture}[
      spy using outlines={rectangle, green, magnification=#2, size=#3},
      inner sep=0
    ]
      \node (img) [img] {\includegraphics[width=\textwidth]{#5}};
      \spy on #4 in node [anchor=#1] at (img.#1);
    \end{tikzpicture}
    \if\relax\detokenize{#6}\relax\else\caption*{#6}\fi
  \end{subfigure}%
}

\begin{figure*}[!t]
\centering
\setlength{\tabcolsep}{0.6pt}

\newcommand{\SetSceneZoom}[4]{%
  \global\def\ZoomPos{#1}%
  \global\def\ZoomMag{#2}%
  \global\def\ZoomSize{#3}%
  \global\def\GTPos{#4}%
}

\newcommand{\SceneAZoom}{%
  \SetSceneZoom{(-0.8,0.12)}{4}{1.0cm}{(-0.9,0.12)}%
}

\newcommand{\SceneBZoom}{%
  \SetSceneZoom{(-0.4,0.4)}{6}{1cm}{(-0.4,0.4)}%
}

\newcommand{\SceneCZoom}{%
  \SetSceneZoom{(0.4,0.0)}{3}{1cm}{(0.7,0.0)}%
}

\newcommand{\SceneDZoom}{%
  \SetSceneZoom{(0.05,0.48)}{6}{1.0cm}{(0.05,0.48)}%
}

\newcommand{\FiveCols}[6][north east]{%
  \ZoomPicBySampling[#1]{\ZoomMag}{\ZoomSize}{\GTPos}{#2}{} &
  \ZoomPicBySampling[#1]{\ZoomMag}{\ZoomSize}{\ZoomPos}{#3}{} &
  \ZoomPicBySampling[#1]{\ZoomMag}{\ZoomSize}{\ZoomPos}{#4}{} &
  \ZoomPicBySampling[#1]{\ZoomMag}{\ZoomSize}{\ZoomPos}{#5}{} &
  \ZoomPicBySampling[#1]{\ZoomMag}{\ZoomSize}{\ZoomPos}{#6}{} \\
}

\begin{tabular}{@{}c c c c c@{}}
  \multicolumn{5}{c}{\vspace{-0.2em}{Applied on VACE (explicit NVS)}} \\
  \SceneAZoom\FiveCols[north east]{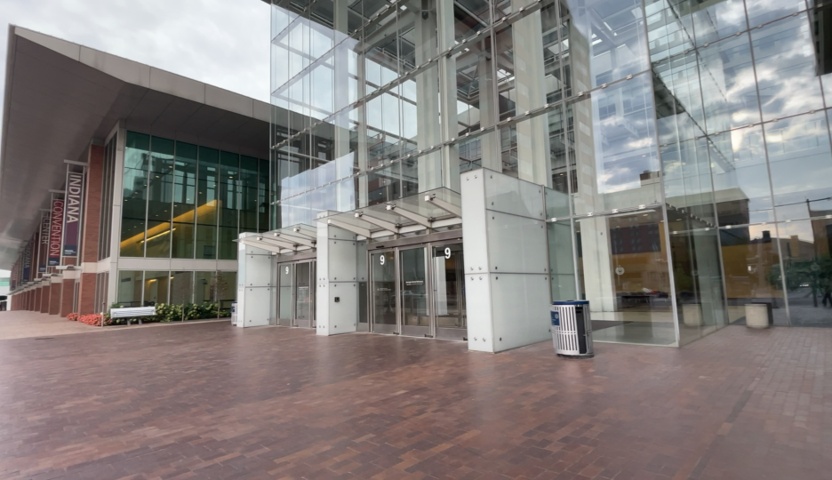}{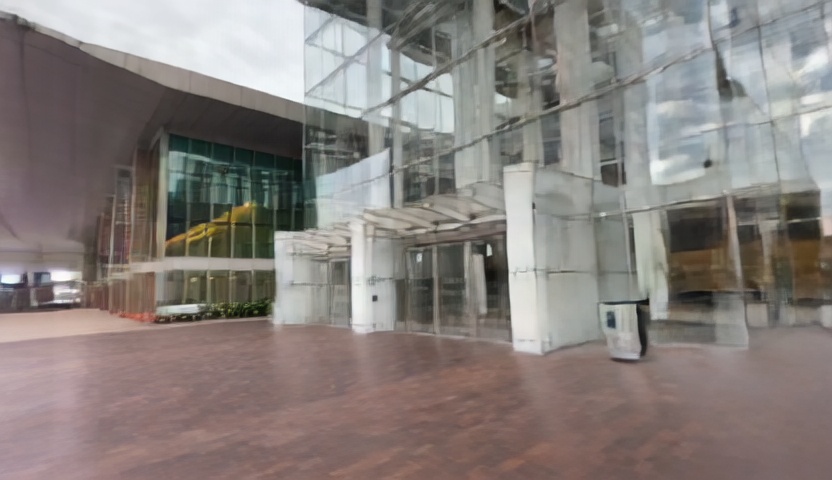}{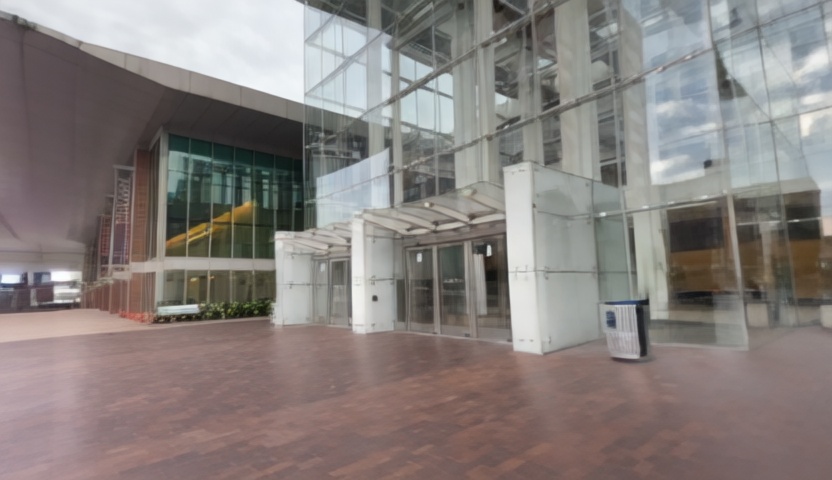}{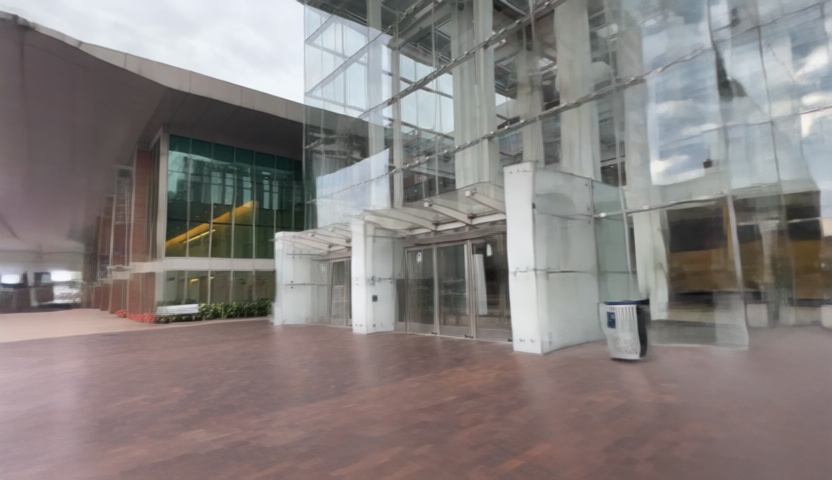}{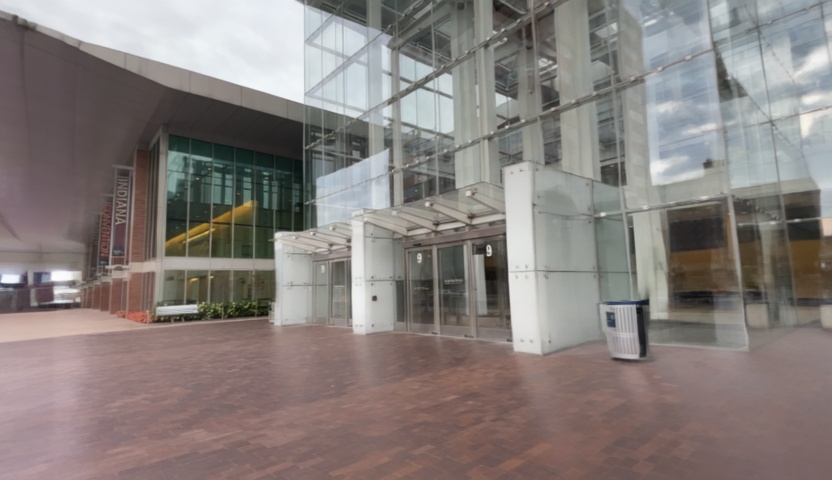}

  \multicolumn{5}{c}{\vspace{-0.2em}{Applied on StereoCrafter (explicit SC)}} \\
  \SceneBZoom\FiveCols[north east]{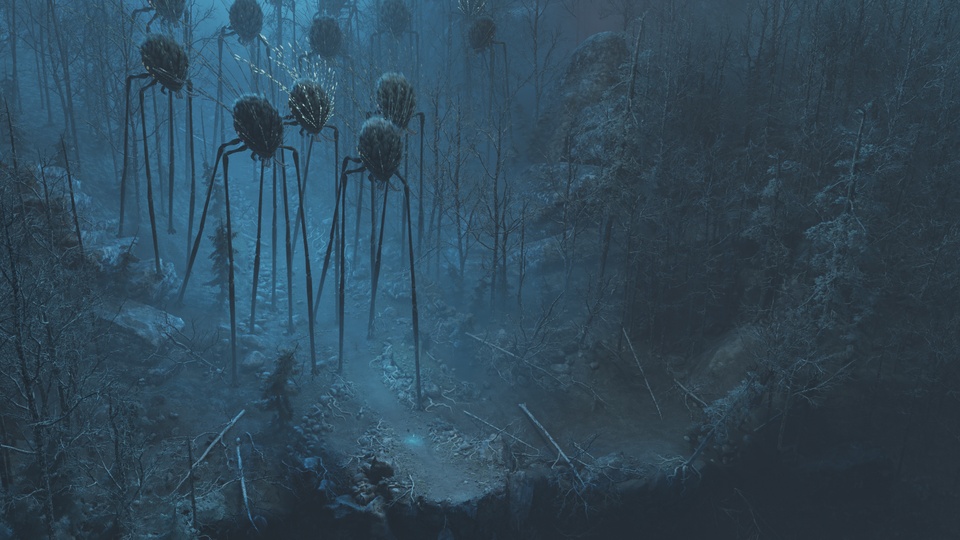}{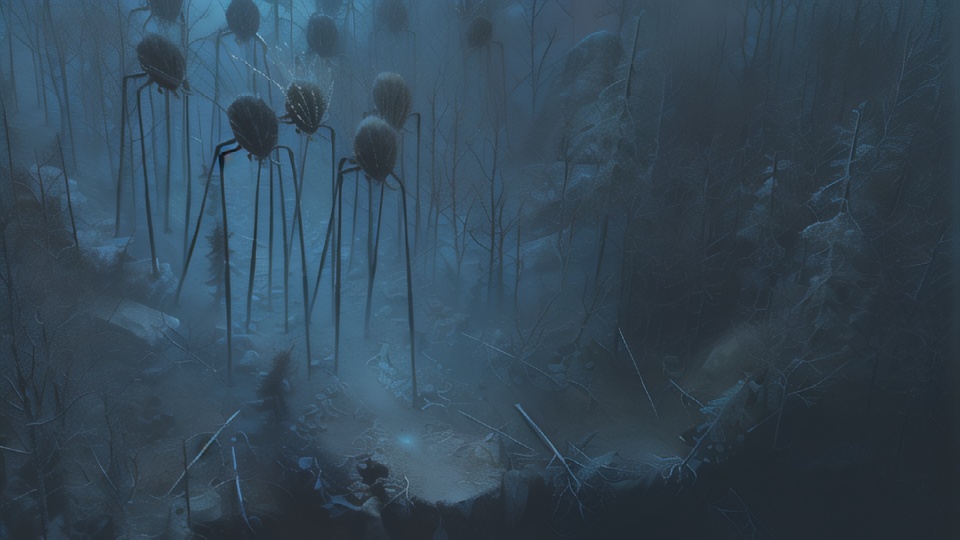}{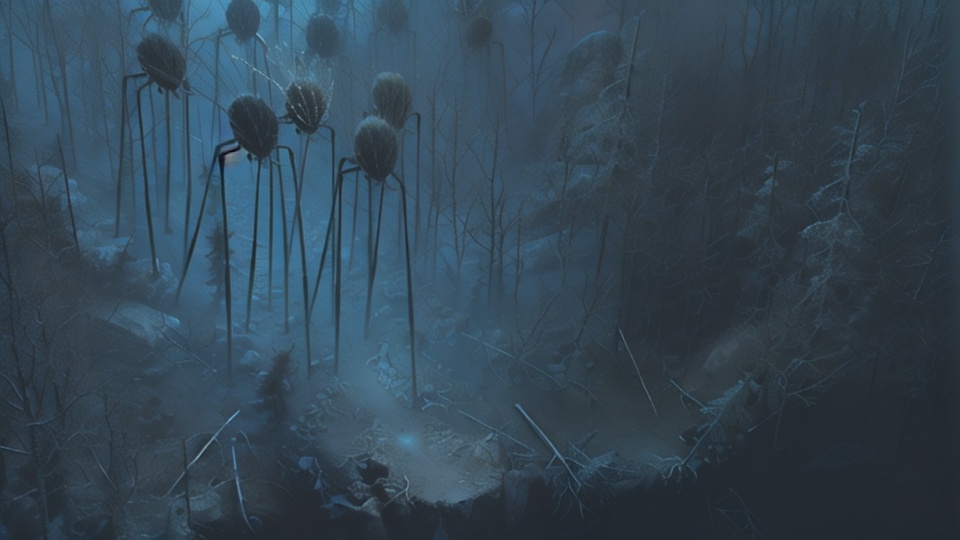}{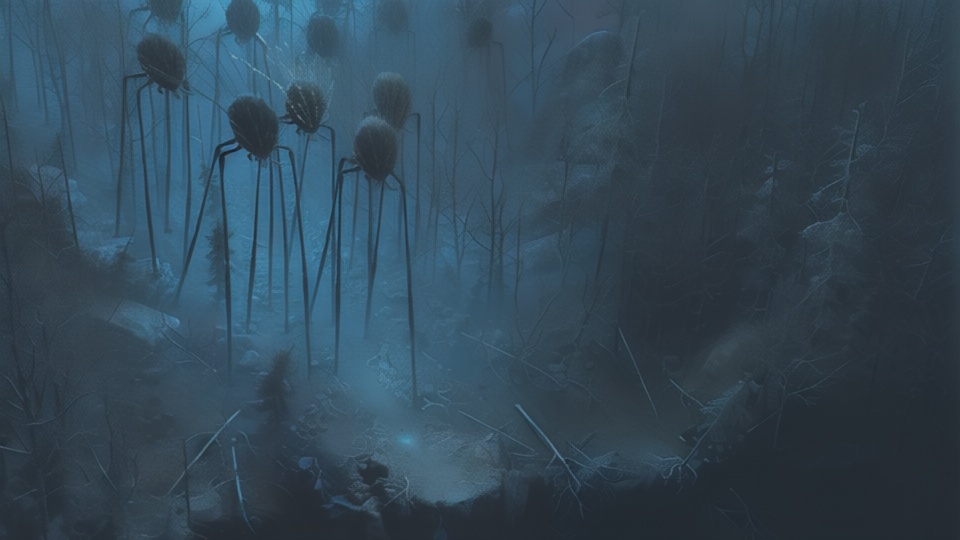}{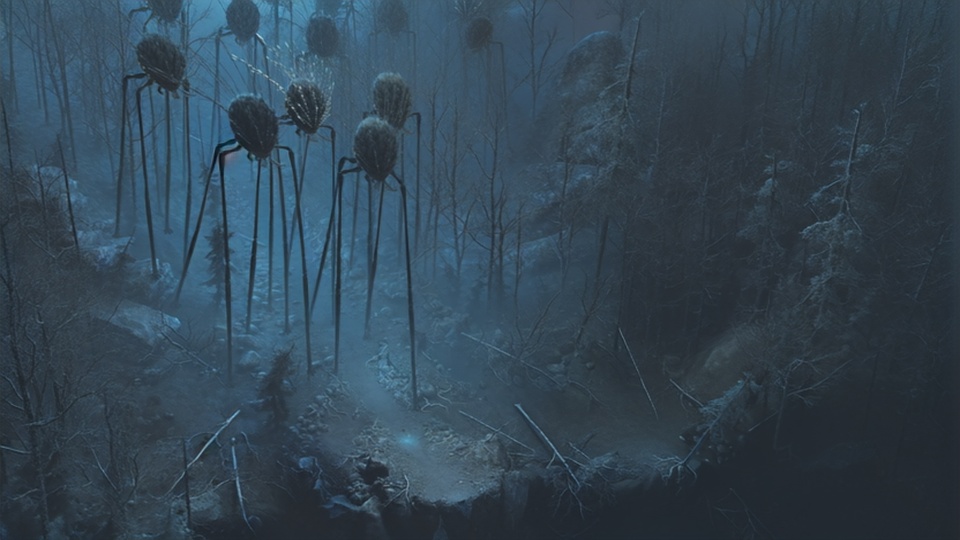}

  \multicolumn{5}{c}{\vspace{-0.2em}{Applied on ReCamMaster (implicit NVS)}} \\
  \SceneCZoom\FiveCols[north west]{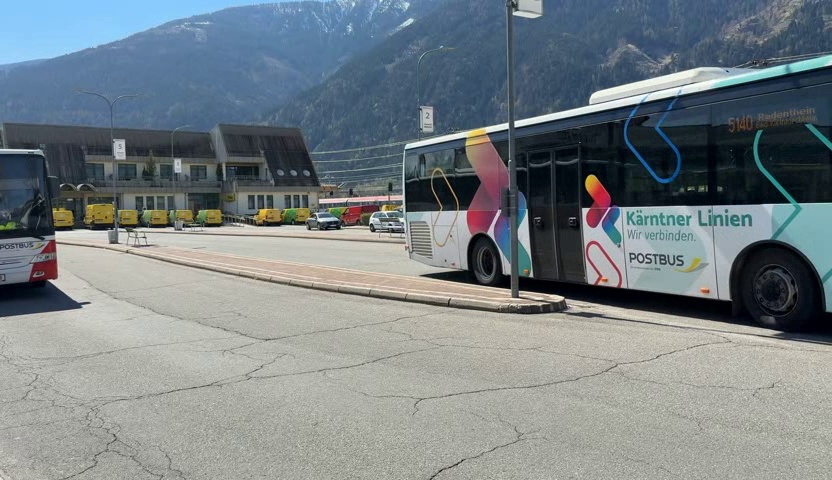}{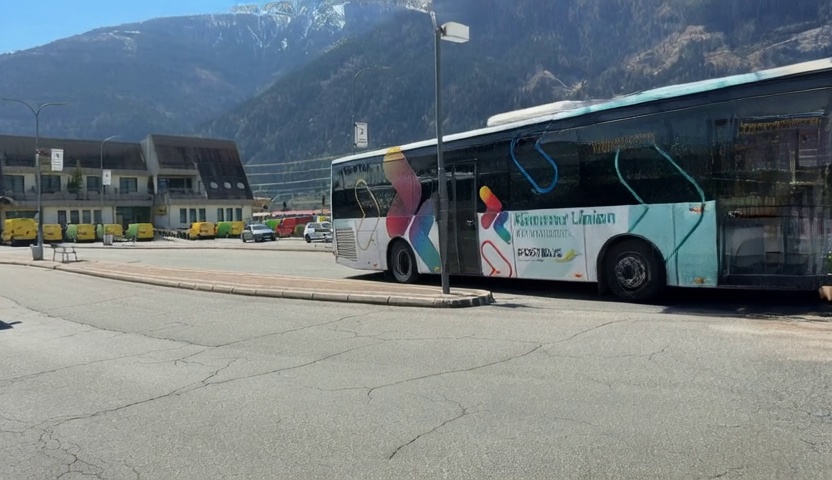}{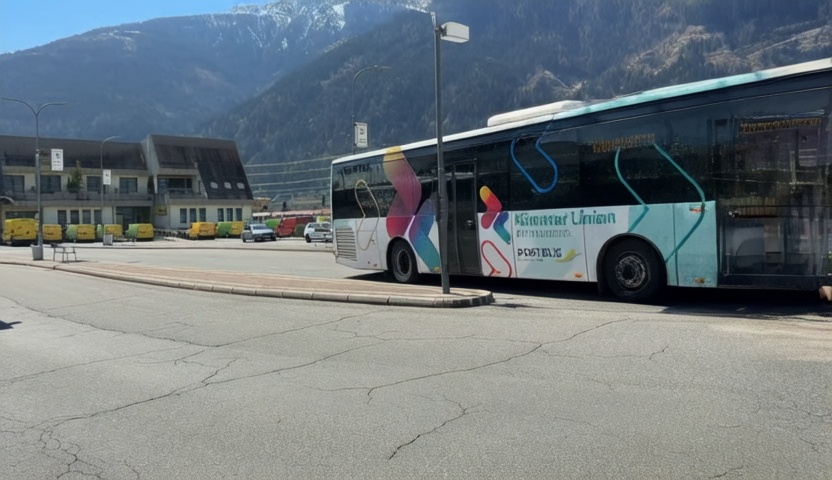}{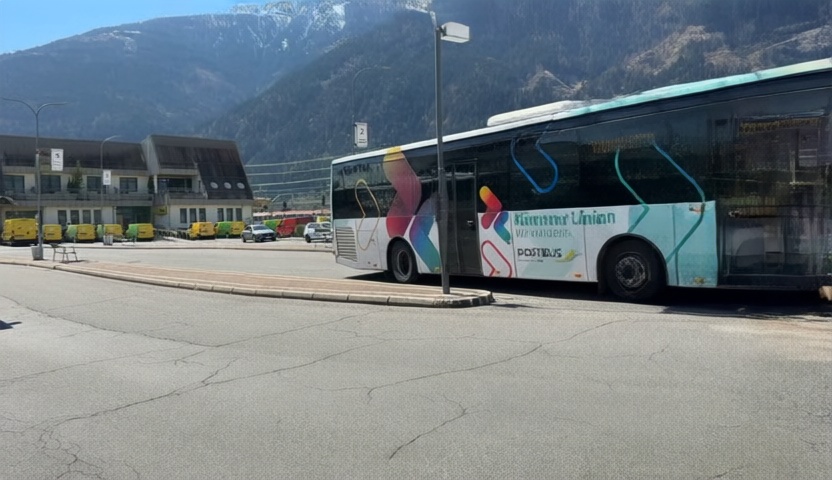}{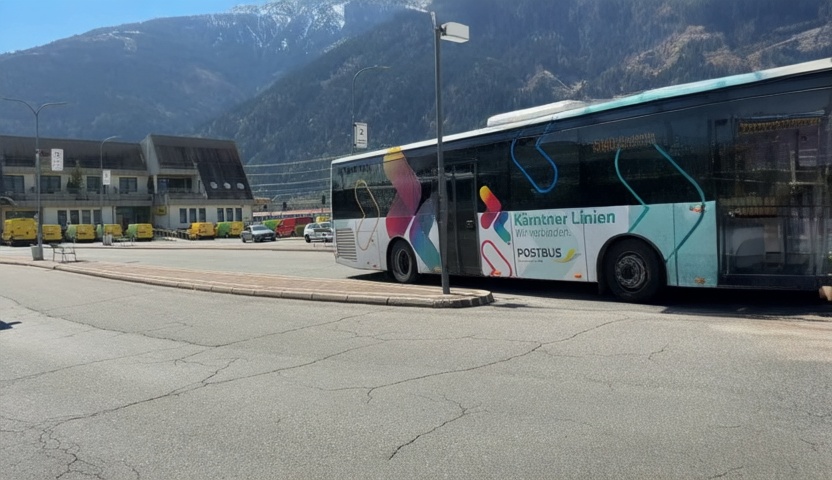}

  \multicolumn{5}{c}{\vspace{-0.2em}{Applied on StereoPilot (implicit SC)}} \\
  \SceneDZoom\FiveCols[north west]{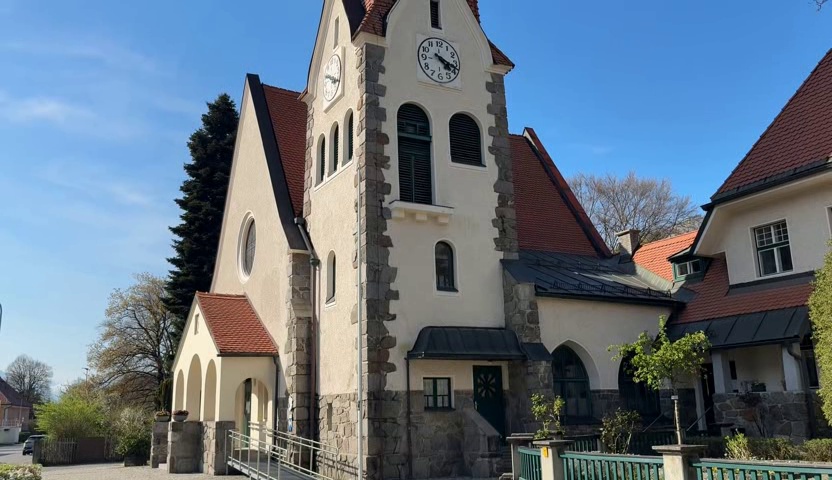}{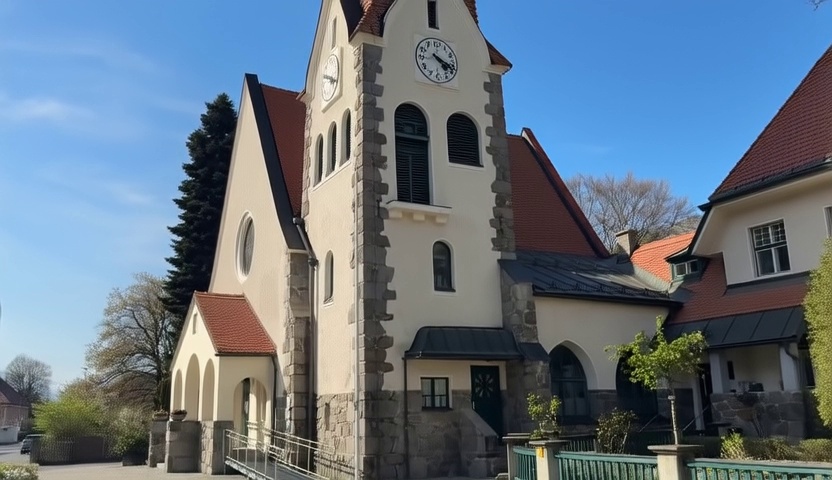}{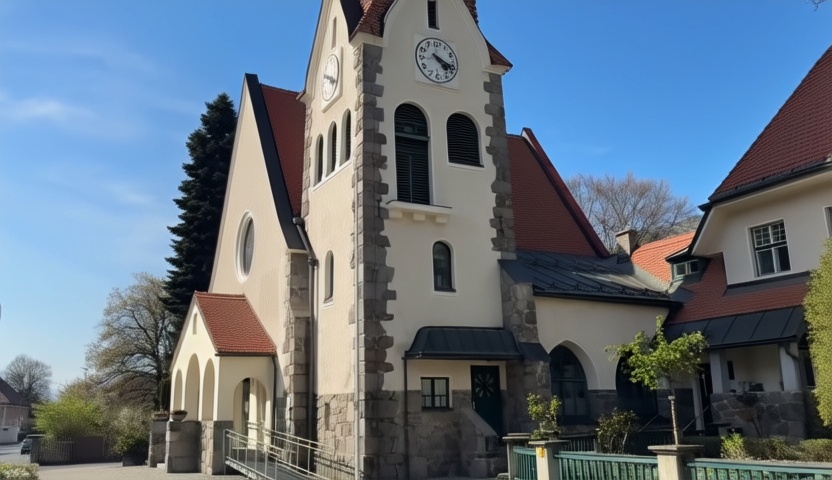}{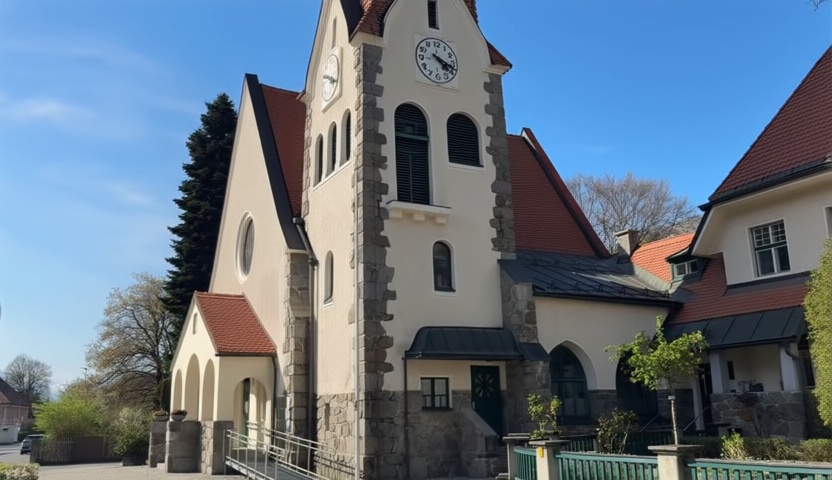}{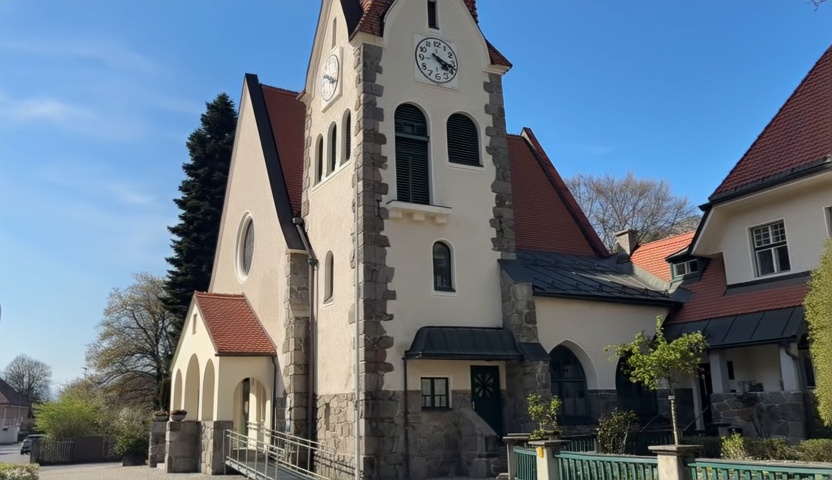}

  {{Source view}} & {{W/O fixer}} & {{DIFIX3D+}} & {{MaRINeR}} & {\textbf{Ours}} \\
\end{tabular}

\caption{Visual results of applying novel view fixers (including DIFIX3D+~\cite{wu2025difix3d+}, MaRINeR~\cite{bosiger2024mariner}, and ours) to improve diffusion-based view synthesis methods.}
\label{fig:vis_comparison}
\end{figure*}

\begin{table*}[t]
\centering
\caption{Stereo conversion on Mono2Stereo~\cite{yu2025mono2stereo} and Spring~\cite{mehl2023spring} datasets. We perform stereo conversion using the state-of-the-art methdos Mono2Stereo~\cite{yu2025mono2stereo} and StereoCrafter~\cite{zhao2024stereocrafter}. \textcolor{red}{Best} results are marked.}
\label{tab:sc_fixing_results}
\small
\setlength{\tabcolsep}{2.2pt}
\resizebox{\textwidth}{!}{%
\begin{tabular}{llcccccccccc}
\toprule
\multirow{2}{*}{Dataset} & \multirow{2}{*}{Fixer} & \multicolumn{5}{c}{Mono2Stereo~\cite{yu2025mono2stereo} output} & \multicolumn{5}{c}{StereoCrafter~\cite{zhao2024stereocrafter} output} \\
\cmidrule(lr){3-7} \cmidrule(lr){8-12}
 &  & PSNR$\uparrow$ & SSIM$\uparrow$ & LPIPS$\downarrow$ & DISTS$\downarrow$ & FID$\downarrow$ & PSNR$\uparrow$ & SSIM$\uparrow$ & LPIPS$\downarrow$ & DISTS$\downarrow$ & FID$\downarrow$ \\
\midrule[0.15em]
\multirow{4}{*}{Mono2Stereo~\cite{yu2025mono2stereo}} 
& W/O fixer  & 29.52 & 0.847 & 0.101 & 0.071 & 10.66 & 30.33 & 0.893 & 0.108 & 0.086 & 14.68 \\
& DIFIX3D+ \cite{wu2025difix3d+}  & 29.20 & 0.852 & 0.098 & 0.071 & 9.90 & 30.31 & 0.896 & 0.106 & 0.087 & 13.82 \\
& MaRINeR~\cite{bosiger2024mariner} & 29.91 & 0.861 & 0.115 & 0.081 & 12.12 & 30.61 & 0.903 & 0.107 & 0.086 & 14.66 \\
\rowcolor{gray!15} & \textbf{Ours} & \textcolor{red}{30.01} & \textcolor{red}{0.882} & \textcolor{red}{0.088} & \textcolor{red}{0.065} & \textcolor{red}{9.29} & \textcolor{red}{30.84} & \textcolor{red}{0.916} & \textcolor{red}{0.087} & \textcolor{red}{0.073} & \textcolor{red}{10.50} \\
\midrule[0.15em]
\multirow{4}{*}{Spring~\cite{mehl2023spring}}
& W/O fixer  & 27.43 & 0.645 & 0.167 & 0.092 & 34.53 & 26.82 & 0.702 & 0.257 & 0.153 & 91.87 \\
& DIFIX3D+ \cite{wu2025difix3d+}  & 27.46 & 0.654 & 0.174 & 0.102 & 32.64 & 26.64 & 0.706 & 0.240 & 0.145 & 72.77 \\
& MaRINeR~\cite{bosiger2024mariner} & 27.66 & 0.661 & 0.216 & 0.134 & 41.28 & 26.87 & 0.703 & 0.303 & 0.179 & 89.25 \\
\rowcolor{gray!15} & \textbf{Ours} & \textcolor{red}{29.35} & \textcolor{red}{0.827} & \textcolor{red}{0.127} & \textcolor{red}{0.086} & \textcolor{red}{29.43} & \textcolor{red}{28.04} & \textcolor{red}{0.843} & \textcolor{red}{0.167} & \textcolor{red}{0.113} & \textcolor{red}{47.22} \\
\bottomrule
\end{tabular}%
}
\end{table*}

\begin{table*}[t]
\centering
\caption{Implicit view synthesis on the Spring~\cite{mehl2023spring} and SVD~\cite{izadimehr2025svd} datasets. We employ the state-of-the-art implicit methods for view synthesis (ReCamMaster~\cite{bai2025recammaster} for NVS and StereoPilot~\cite{shen2025stereopilot} for SC). \textcolor{red}{Best} results are marked.}
\label{tab:implicit_fixing_results}
\small
\setlength{\tabcolsep}{2.2pt}
\resizebox{\textwidth}{!}{%
\begin{tabular}{llcccccccc}
\toprule
\multirow{2}{*}{Dataset} & \multirow{2}{*}{Method} & \multicolumn{4}{c}{ReCamMaster~\cite{bai2025recammaster} output} & \multicolumn{4}{c}{StereoPilot~\cite{shen2025stereopilot} output} \\
\cmidrule(lr){3-6} \cmidrule(lr){7-10}
 & & CLIP-IQA$\uparrow$ & MUSIQ$\uparrow$ & MANIQA$\uparrow$ & FID$\downarrow$ & CLIP-IQA$\uparrow$ & MUSIQ$\uparrow$ & MANIQA$\uparrow$ & FID$\downarrow$ \\
\midrule[0.15em]

\multirow{4}{*}{Spring~\cite{mehl2023spring}} & W/O fixer & 0.347 & 54.38 & 0.272 & 72.21 & 0.344 & 49.79 & \textcolor{red}{0.331} & 90.44 \\
 & DIFIX3D+~\cite{wu2025difix3d+} & 0.453 & 54.26 & 0.260 & 76.88 & 0.389 & 48.77 & 0.311 & 94.15 \\
 & MaRINeR~\cite{bosiger2024mariner} & 0.429 & 53.13 & 0.276 & 73.38 & 0.378 & 48.41 & 0.296 & 83.93 \\
\rowcolor{gray!15}  & \textbf{Ours} & \textcolor{red}{0.482} & \textcolor{red}{55.75} & \textcolor{red}{0.282} & \textcolor{red}{70.12} & \textcolor{red}{0.452} & \textcolor{red}{50.62} & 0.319 & \textcolor{red}{77.67} \\
\midrule[0.15em]
\multirow{4}{*}{SVD~\cite{izadimehr2025svd}} & W/O fixer & 0.417 & 71.72 & 0.396 & 23.69 & 0.405 & \textcolor{red}{71.67} & \textcolor{red}{0.442} & 25.18 \\
 & DIFIX3D+~\cite{wu2025difix3d+} & 0.494 & 70.99 & 0.372 & 24.57 & 0.484 & 70.08 & 0.397 & 24.82 \\
 & MaRINeR~\cite{bosiger2024mariner} & 0.524 & 71.19 & 0.392 & 23.53 & 0.525 & 70.89 & 0.411 & 19.30 \\
\rowcolor{gray!15}  & \textbf{Ours} & \textcolor{red}{0.531} & \textcolor{red}{72.00} & \textcolor{red}{0.410} & \textcolor{red}{22.36} & \textcolor{red}{0.538} & \textcolor{red}{71.67} & 0.436 & \textcolor{red}{18.42} \\
\bottomrule
\end{tabular}%
}
\end{table*}

\subsection{Degradation Fixing}
\label{subsec:degradation-fixing}
Following the same experimental setting in Sec.~\ref{sec:degradation} and \cref{fig:degradation_tsne}, we evaluate the degradation fixing performance of \myname\ across three prevalent degradation types in diffusion-based view synthesis, including spatial, temporal, and backbone-related degradations. Although our method is trained under a fixed configuration (\ie, spatial resolution $480\times832$, temporal stride equal to 1, and DiT-based backbone), \myname\ achieves state-of-the-art performance and shows strong zero-shot generalization across different degradation types and difficulty levels, as shown in \cref{tab:degradation_fixing_results}.

\subsection{View Synthesis Fixing}
We evaluate the performance of \myname\ on the NVS and SC tasks. As shown in \cref{fig:vis_comparison}, our method outperforms previous novel view fixers in recovering fine-grained structure and texture details. In the SC task, we employ the recent Mono2Stereo~\cite{yu2025mono2stereo} and StereoCrafter~\cite{zhao2024stereocrafter} to generate right views from left-view inputs, and \myname\ achieves significant improvements in the perceptual quality of novel views, as shown in \cref{fig:vis_comparison} and \cref{tab:sc_fixing_results}. We also apply \myname\ to enhance implicit view synthesis approaches (\ie, ReCamMaster~\cite{bai2025recammaster} for NVS and StereoPilot~\cite{shen2025stereopilot} for SC) in a plug-and-play manner, and our proposed method consistently shows superior results over existing novel view fixers (\cref{tab:implicit_fixing_results}).

\subsection{Ablation Study}
\begin{table*}[t]
\centering
\caption{Ablation study on (a) component designs and (b) referencing mechanisms.}
\captionsetup[subtable]{skip=1pt}
\begin{subtable}{\textwidth}
\centering
\vspace{-1em}
\subcaption{Component design ablation.}
\vspace{-1em}
\label{tab:ablation_components}
\setlength{\tabcolsep}{3.5pt}
{\scriptsize
\begin{tabular}{c ccc ccccc}
\toprule
 & \multicolumn{3}{c}{Components} & \multicolumn{5}{c}{Metrics} \\
\cmidrule(lr){2-4} \cmidrule(lr){5-9}
ID & GSA & RPA & LDI & PSNR$\uparrow$ & SSIM$\uparrow$ & LPIPS$\downarrow$ & DISTS$\downarrow$ & FID$\downarrow$ \\
\midrule[0.15em]
\#1 & $\checkmark$ &   &  & 26.19 & 0.812 & 0.113 & 0.066 & 24.23 \\
\#2 & $\checkmark$ & $\checkmark$  &  & 27.09 & 0.866 & 0.092 & 0.055 & 20.62 \\
\#3 &  & $\checkmark$ & $\checkmark$ & 25.86 & 0.859 & 0.114 & 0.068 & 24.95 \\

\rowcolor{gray!15} \#4 & $\checkmark$ & $\checkmark$ & $\checkmark$ & \textcolor{red}{27.37} & \textcolor{red}{0.875} & \textcolor{red}{0.084} & \textcolor{red}{0.051} & \textcolor{red}{20.05} \\
\bottomrule
\end{tabular}%
}
\end{subtable}
\begin{subtable}{\textwidth}
\centering
\vspace{1em}
\subcaption{Referencing mechanism ablation.}
\vspace{-1em}
\label{tab:ablation_ref_mech}
\setlength{\tabcolsep}{3pt}
{\scriptsize
\begin{tabular}{cllccccc}
\toprule
ID & Type & Method & PSNR$\uparrow$ & SSIM$\uparrow$ & LPIPS$\downarrow$ & DISTS$\downarrow$ & FID$\downarrow$ \\
\midrule[0.15em]
\#1 & Convolution & Skip connection~\cite{wu2025difix3d+} & 26.19 & 0.812 & 0.113 & 0.066 & 24.23 \\
\#2 & Attention & Cross attention~\cite{dosovitskiy2020image} & 25.87 & 0.798 & 0.118 & 0.070 & 26.89 \\
\#3 & Matching & Feature matching~\cite{bosiger2024mariner} & 26.14 & 0.811 & 0.106 & 0.060 & 24.54 \\
\rowcolor{gray!15} \#4 & \textbf{Warping} & \textbf{Ours} & \textcolor{red}{27.37} & \textcolor{red}{0.875} & \textcolor{red}{0.084} & \textcolor{red}{0.051} & \textcolor{red}{20.05} \\
\bottomrule
\end{tabular}%
}
\end{subtable}
\end{table*}

\textbf{Component design.} 
\cref{tab:ablation_components} demonstrates the contributions of each component in our \myname\ pipeline. Compared with the variant that directly utilizes features from the reference view (\#1), RPA alleviates the difficulty of correspondence search via pre-alignment, yielding significant performance improvement (\#2 \vs \#1). GSA aggregates the shared structures between reference and target views, providing a crucial foundation for degradation removal (\#4 \vs \#3). Finally, incorporating the LDI module enables the adaptive extraction of high-fidelity reference details while suppressing warping artifacts (\#4 \vs \#2).

\par
\textbf{Referencing mechanism.}
Many referencing mechanisms have been designed to leverage reference features for enhancement. We compare our warping-based referencing method against three representative paradigms: skip connection (convolution-based)~\cite{wu2025difix3d+}, cross attention (attention-based)~\cite{dosovitskiy2020image}, and feature matching (matching-based)~\cite{bosiger2024mariner}. For fair comparisons, we use the same backbone (SD-Turbo~\cite{sauer2024sdturbo}) and replace our RPA and LDI modules with different referencing mechanisms. All variants are trained under the same setting. The proposed warping-based referencing facilitates the recovery of fine-grained structures and textures (see supp. for the visual comparisons), and \cref{tab:ablation_ref_mech} verifies our superior performance over previous referencing methods.

\FloatBarrier %

\section{Conclusion}
In this paper, we systematically analyze three prevalent degradation types in diffusion-based view synthesis and propose a universal reference-guided framework to address them in a coarse-to-fine manner. We first perform reference pre-alignment to facilitate correspondence search, then aggregate deep features for global structure anchoring, and finally inject fine-grained texture details to achieve high-quality view synthesis. Extensive experiments demonstrate the state-of-the-art performance of \myname\ in NVS and SC tasks and verify our zero-shot generalization across different diffusion degradations.

\clearpage
\section*{Supplementary Material}

\section*{Overview}
This supplementary material provides additional details that complement the main paper, including more ablation experiments, degradation analysis details, limitation analysis, discussion and visual results.

\section{Ablation study details}

\subsection{More ablation on local detail injection}
\begin{table*}[h]
\centering
\caption{Ablation study on component designs in the proposed LDI module.}
\label{tab:ablation_components}
\setlength{\tabcolsep}{3.5pt}
{\scriptsize
\begin{tabular}{c cc ccccc}
\toprule
 & \multicolumn{2}{c}{Components} & \multicolumn{5}{c}{Metrics} \\
\cmidrule(lr){2-3} \cmidrule(lr){4-8}
ID & AWD & UAGF & PSNR$\uparrow$ & SSIM$\uparrow$ & LPIPS$\downarrow$ & DISTS$\downarrow$ & FID$\downarrow$ \\
\midrule[0.15em]
\#1 &  &  & 27.09 & 0.866 & 0.092 & 0.055 & 20.62 \\
\#2 & $\checkmark$ &  & 27.18 & 0.870 & 0.089 & 0.053 & 20.83 \\
\rowcolor{gray!15} \#3 & $\checkmark$ & $\checkmark$ & \textcolor{red}{27.37} & \textcolor{red}{0.875} & \textcolor{red}{0.084} & \textcolor{red}{0.051} & \textcolor{red}{20.05} \\
\bottomrule
\end{tabular}
}
\end{table*}

\noindent Our Local Detail Injection (LDI) module consists of two key designs: (i) \,\emph{Adaptive Warping Deformation} (AWD) for better local alignment, and (ii) \,\emph{Uncertainty-Aware Gated Fusion} (UAGF) for selectively injecting reliable reference details.
While the main paper ablates LDI as a whole module, here we further disentangle LDI and study the contribution of each design byThis supplementary material provides progressively enabling AWD and UAGF, as summarized in Table~\ref{tab:ablation_components}.
Specifically, \#1 removes both component, \#2 enables AWD only, \#3 enables both components. Both AWD and UAGF consistently contribute to LDI, and combining them yields the best performance, validating the effectiveness of each design.

\subsection{Visual results on ablation study}

\def\imgWidth{0.185\textwidth} %

\tikzset{img/.style={rectangle, minimum width=\imgWidth, draw=black, inner sep=0, outer sep=0}}

\providecommand{\ZoomPicBySampling}[6]{}
\renewcommand{\ZoomPicBySampling}[6][north east]{%
  \begin{subfigure}{\imgWidth}
    \centering
    \begin{tikzpicture}[
      spy using outlines={rectangle, green, magnification=#2, size=#3},
      inner sep=0
    ]
      \node (img) [img] {\includegraphics[width=\textwidth]{#5}};
      \spy on #4 in node [anchor=#1] at (img.#1);
    \end{tikzpicture}
    \if\relax\detokenize{#6}\relax\else\caption*{#6}\fi
  \end{subfigure}%
}

\newcommand{\SetSceneZoom}[4]{%
  \global\def\ZoomPos{#1}%
  \global\def\ZoomMag{#2}%
  \global\def\ZoomSize{#3}%
  \global\def\GTPos{#4}%
}

\newcommand{\FiveCols}[6][north east]{%
  \ZoomPicBySampling[#1]{\ZoomMag}{\ZoomSize}{\GTPos}{#2}{} &
  \ZoomPicBySampling[#1]{\ZoomMag}{\ZoomSize}{\ZoomPos}{#3}{} &
  \ZoomPicBySampling[#1]{\ZoomMag}{\ZoomSize}{\ZoomPos}{#4}{} &
  \ZoomPicBySampling[#1]{\ZoomMag}{\ZoomSize}{\ZoomPos}{#5}{} &
  \ZoomPicBySampling[#1]{\ZoomMag}{\ZoomSize}{\ZoomPos}{#6}{} \\
}

\begin{figure*}[h]
\centering
\setlength{\tabcolsep}{0.6pt}
\begin{tabular}{@{}c c c c c@{}}
  \multicolumn{5}{c}{\vspace{-0.2em}{Ablation study on component design (Tab. 4a in the main paper)}} \\
  \SetSceneZoom{(-0.1,-0.0)}{4}{1.0cm}{(-0.1,-0.0)}\FiveCols[north east]{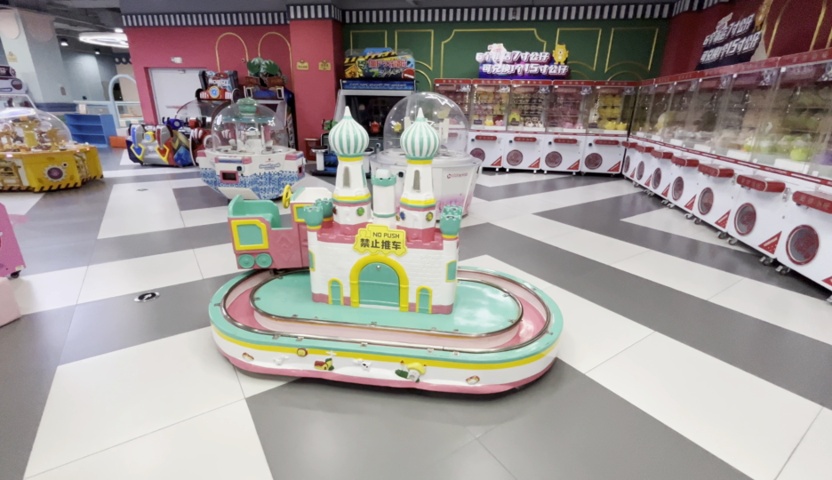}{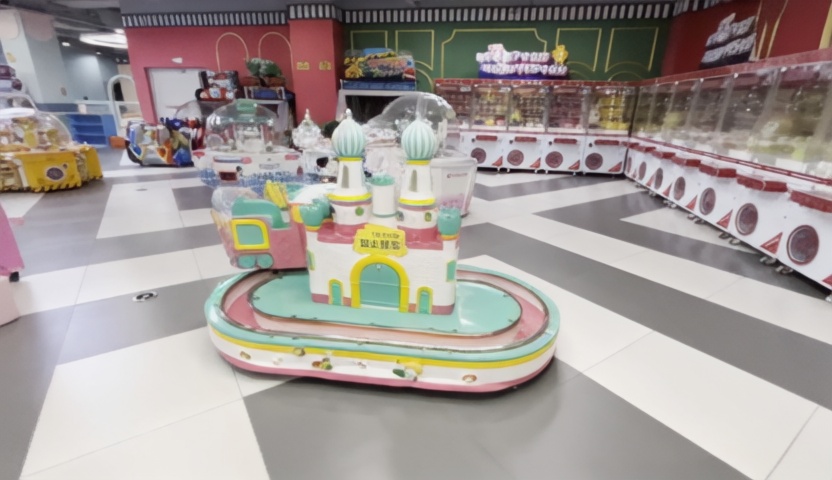}{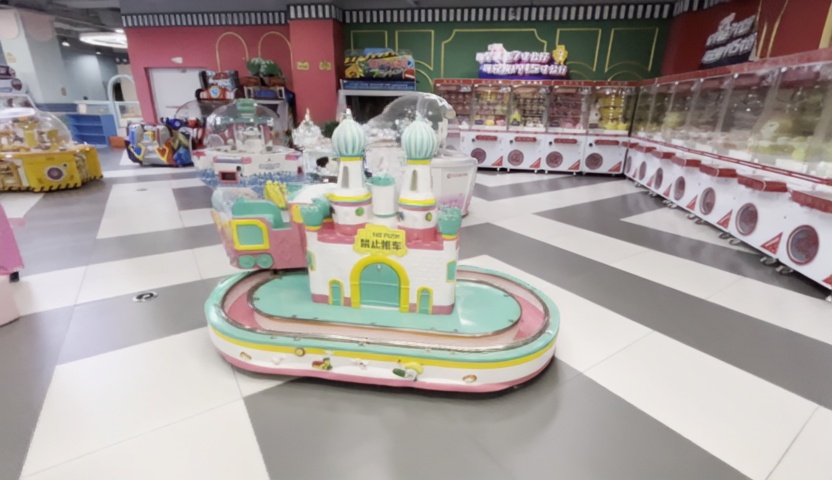}{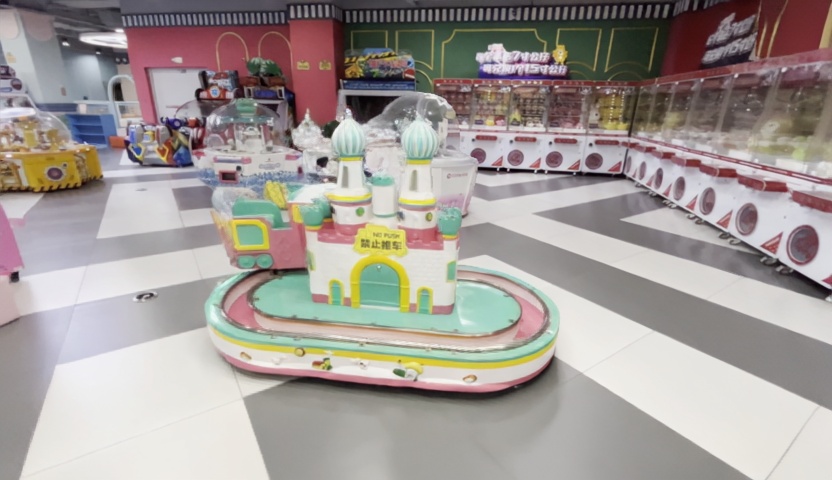}{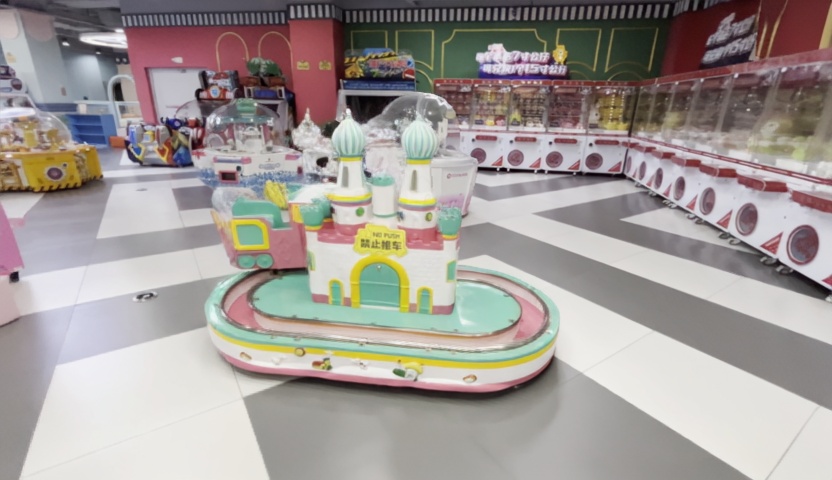}
  {GT} & {\#1} & {\#2} & {\#3} & {\textbf{\#4}} \\
  
  \multicolumn{5}{c}{\vspace{-0.2em}{Ablation study on referencing mechanism (Tab. 4b in the main paper)}} \\
  \SetSceneZoom{(-0.65,0.4)}{3}{1.0cm}{(-0.65,0.4)}\FiveCols[north east]{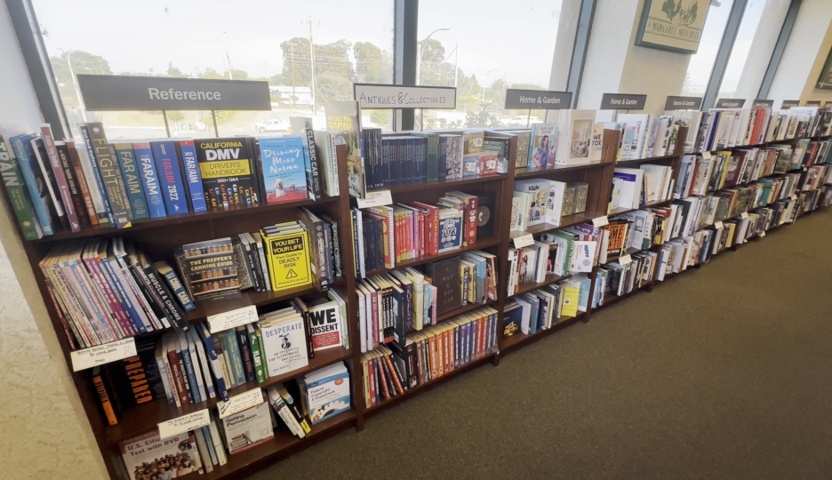}{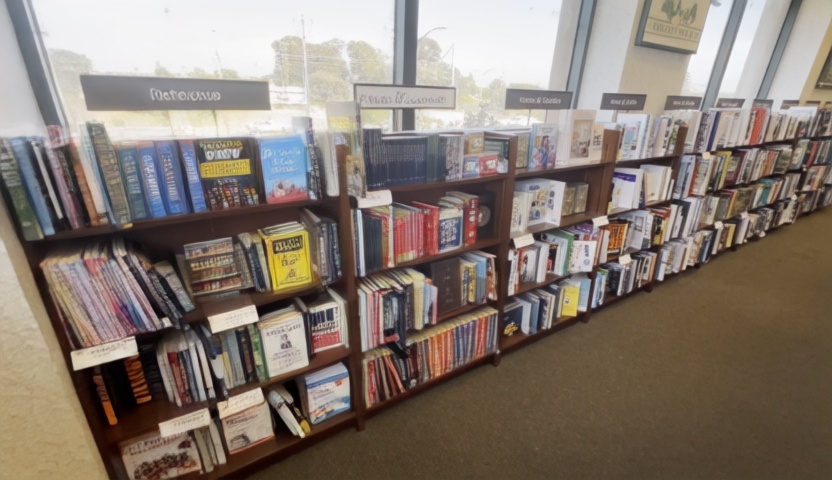}{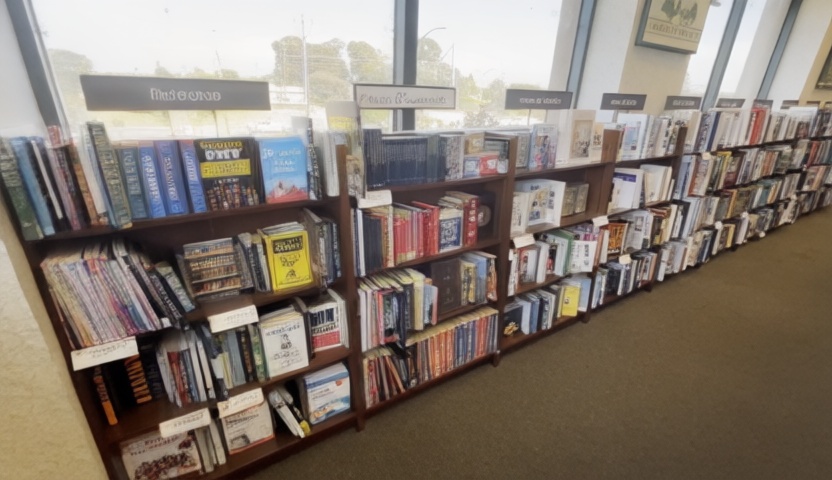}{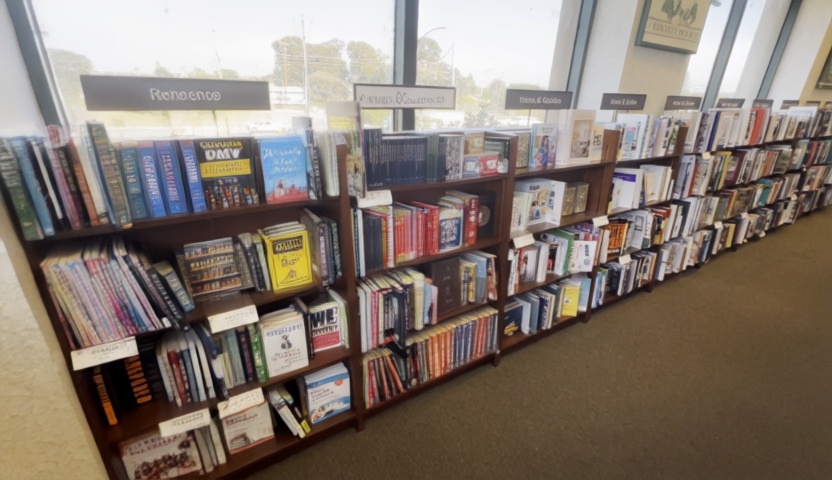}{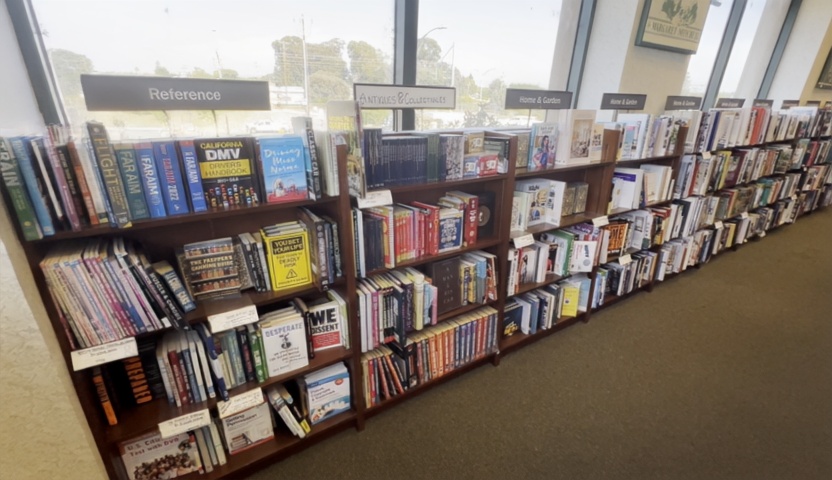}

  {GT} & {\#1} & {\#2} & {\#3} & {\textbf{\#4}} \\
\end{tabular}

\caption{Visual results of ablation study on component design and referencing mechanism, respectively. The ID  corresponds to the ID in Tab.~4a and~4b of the main paper.}
\label{fig:ablation_comparison}
\end{figure*}

To complement the quantitative results reported in Tab.~4 of the main paper, we provide additional visual comparisons for the ablation study (\cref{fig:ablation_comparison}). For component design ablation, our RPA makes the correspondence search easier (\#2 \vs \#1 in Tab. 4a), LDI leads to more fine-grained details recovering (\#4 \vs \#2 in Tab. 4a), and GSA contributes to better global structure anchoring (\#4 \vs \#3 in Tab. 4a). For the referencing mechanism ablation, our warping-based referencing restores the finest details among all variants (\#4 in Tab. 4b).

\section{Degradation analysis details}

\subsection{DINOv3 map visualization}

\newcommand{\DinoDir}{image/Exp/final_vis_jpg/dino}

\newcommand{\DinoInputImg}{\DinoDir/input.jpg}

\newcommand{\DinoInputVShift}{0.06\textwidth}

\newcommand{\DinoFeatXFour}{\DinoDir/feat_map_x4.jpg}
\newcommand{\DinoFeatXEight}{\DinoDir/feat_map_x8.jpg}
\newcommand{\DinoFeatXSixteen}{\DinoDir/feat_map_x16.jpg}
\newcommand{\DinoAttnXFour}{\DinoDir/attn_map_x4.jpg}
\newcommand{\DinoAttnXEight}{\DinoDir/attn_map_x8.jpg}
\newcommand{\DinoAttnXSixteen}{\DinoDir/attn_map_x16.jpg}

\begin{figure*}[h]
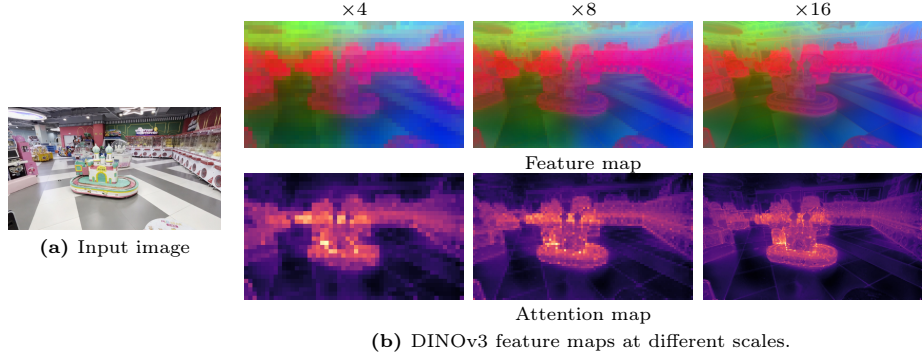

  \centering

  \begin{subfigure}[c]{0.23\textwidth}
    \centering
    \vspace*{\DinoInputVShift}
    \includegraphics[width=\linewidth]{\DinoInputImg}
    \caption{Input image}
    \label{fig:dino:input}
  \end{subfigure}\hspace{0.02\textwidth}%
  \begin{subfigure}[c]{0.74\textwidth}
    \centering
    \setlength{\tabcolsep}{2pt}
    \renewcommand{\arraystretch}{1.0}

    {\scriptsize
    \begin{tabular}{c c c}
      $\times 4$ & $\times 8$ & $\times 16$ \\
      \includegraphics[width=0.32\linewidth]{\DinoFeatXFour} &
      \includegraphics[width=0.32\linewidth]{\DinoFeatXEight} &
      \includegraphics[width=0.32\linewidth]{\DinoFeatXSixteen} \\
      \multicolumn{3}{c}{\vspace{-0.3ex}\scriptsize Feature map}\\
      \includegraphics[width=0.32\linewidth]{\DinoAttnXFour} &
      \includegraphics[width=0.32\linewidth]{\DinoAttnXEight} &
      \includegraphics[width=0.32\linewidth]{\DinoAttnXSixteen} \\
      \multicolumn{3}{c}{\vspace{-0.3ex}\scriptsize Attention map}\\
    \end{tabular}}

    \caption{DINOv3 feature maps at different scales.}
    \label{fig:dino:grid}
  \end{subfigure}

  \caption{DINOv3 feature map visualization.}
  \label{fig:dino}
\end{figure*}

\noindent To further justify using DINOv3 patch tokens to extract degradation features for the t-SNE analysis, we visualize the token representations of a single degraded image. Specifically, we feed the same image at three input resolutions (\ie, 480$\times$832 / 960$\times$1664 / 1920$\times$3328, denoted by $\times 4$/$\times 8$/$\times 16$ in \cref{fig:dino}). We then (1) reshape the patch tokens back to a spatial grid and apply PCA to project the features to 3 dimensions for RGB rendering, producing the \emph{feature maps}; and (2) use the CLS token to compute attention scores over patch tokens and visualize them as a heatmap, producing the \emph{attention maps}. Finally, for fair comparison across input resolutions, we resize all visualizations to a standard resolution of 480$\times$832. From the visualizations, we observe that the feature maps from patch tokens preserve rich fine-grained details, indicating that they are well suited for deriving degradation features in our subsequent analysis. Meanwhile, the attention maps between the CLS token and patch tokens highlight semantically related regions, further supporting the semantic consistency and representational validity of the patch tokens.

\subsection{Degradation fixing results}

\begin{figure*}[!t]
  \centering
  \vspace{1mm}

  \begin{subfigure}[t]{\linewidth}
    \centering
    \includegraphics[width=\linewidth]{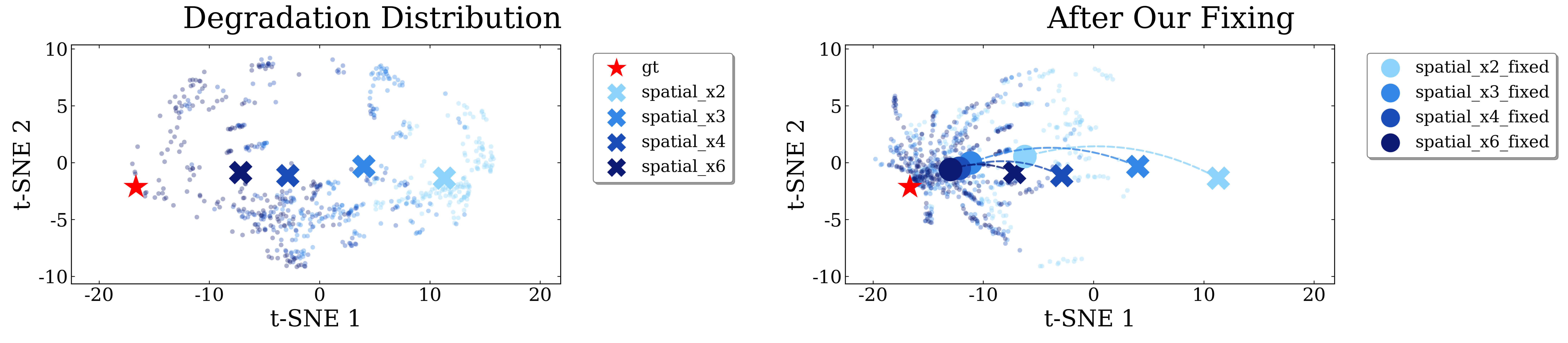}
    \caption{Spatial combined}
  \end{subfigure}

  \vspace{2mm}

  \begin{subfigure}[t]{\linewidth}
    \centering
    \includegraphics[width=\linewidth]{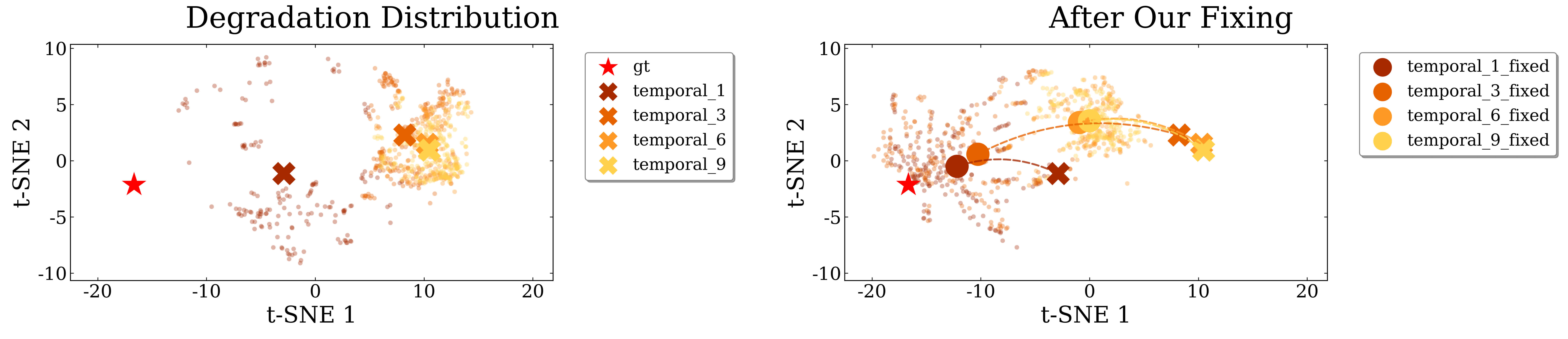}
    \caption{Temporal combined}
  \end{subfigure}

  \vspace{2mm}

  \begin{subfigure}[t]{\linewidth}
    \centering
    \includegraphics[width=\linewidth]{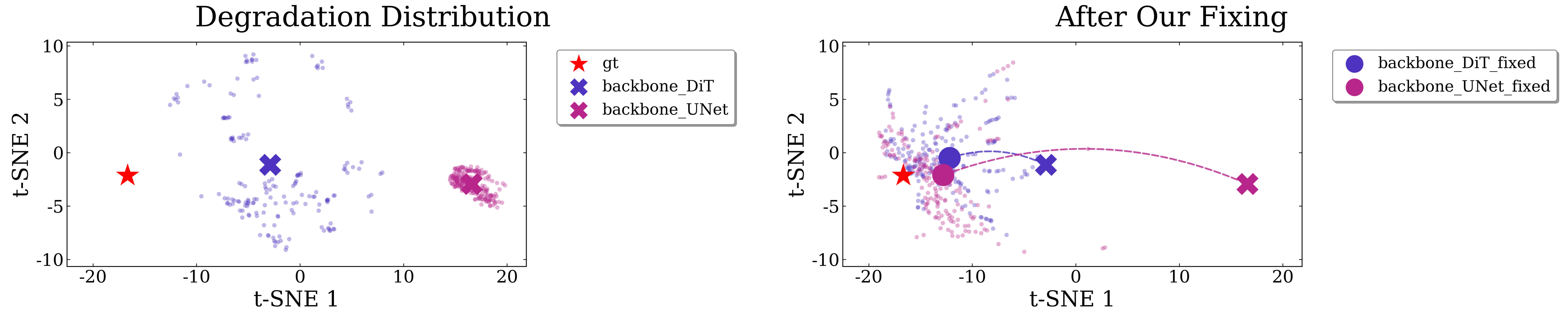}
    \caption{Backbone combined}
  \end{subfigure}

  \vspace{2mm}

\caption{t-SNE visualization under different degradation dimensions. Lighter colors indicate more severe degradation.}
  \label{fig:tsne_all_supp}
\end{figure*}

\def\imgWidth{0.185\textwidth} %

\tikzset{img/.style={rectangle, minimum width=\imgWidth, draw=black, inner sep=0, outer sep=0}}

\newcommand{\DegradationSceneZoom}{%
  \SetSceneZoom{(0.35,0.45)}{4.0}{1.0cm}{(0.35,0.45)}%
}
\begin{figure*}[!t]
\centering
\setlength{\tabcolsep}{0.6pt}
\begin{tabular}{@{}c c c c c c@{}}
  \multicolumn{6}{c}{\vspace{-0.2em}{Spatial degradation}} \\
  \texttt{x2} & \DegradationSceneZoom\FiveCols[south west]{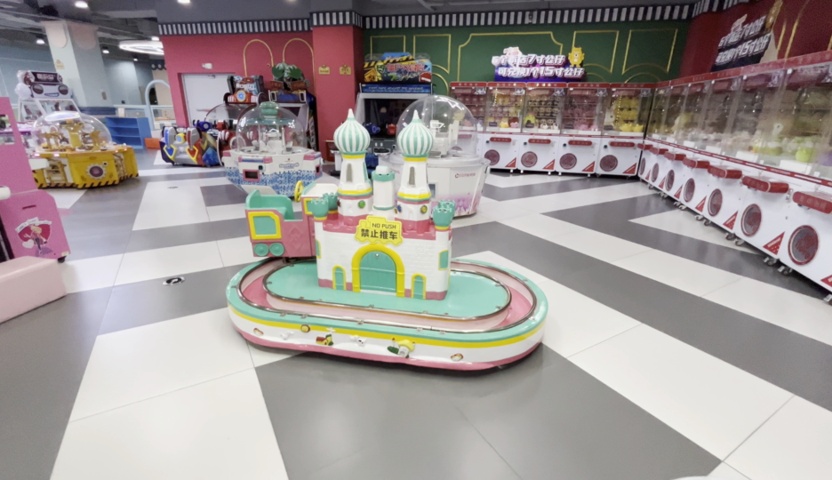}{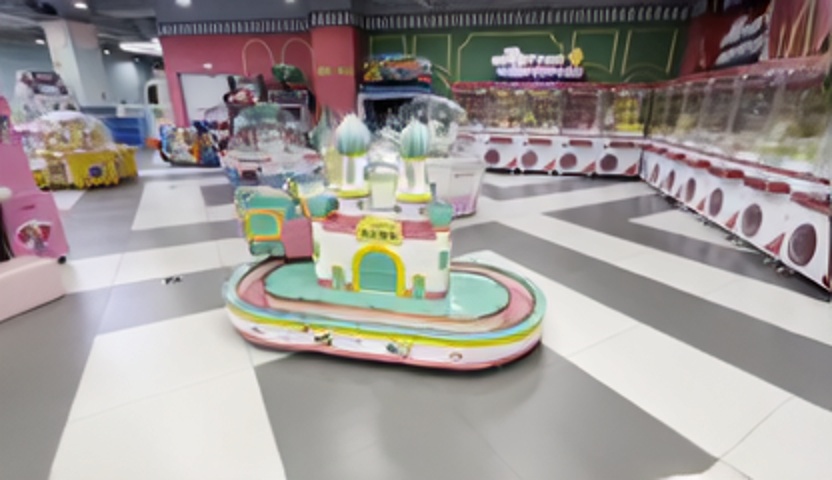}{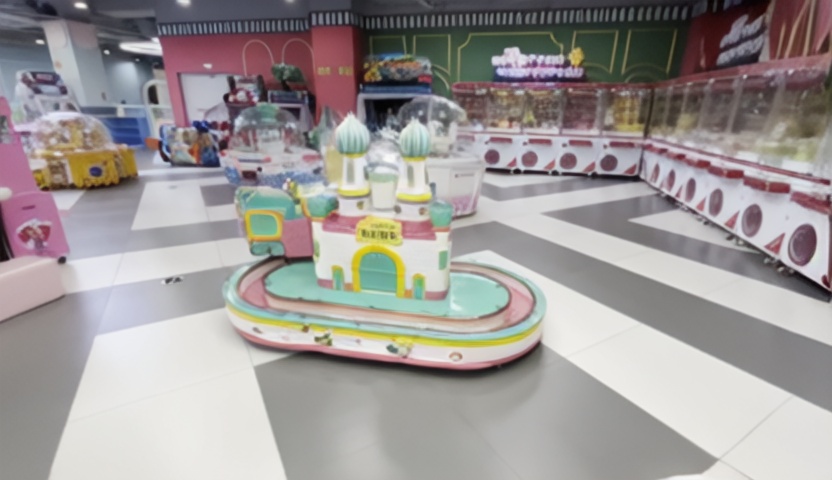}{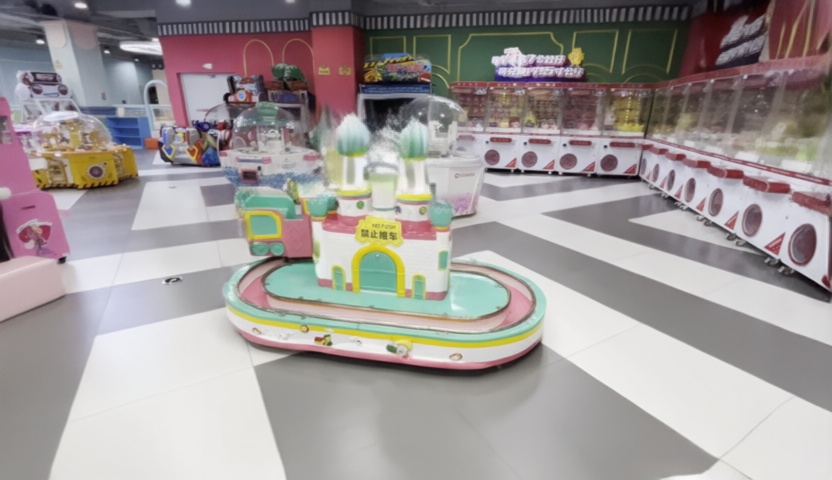}{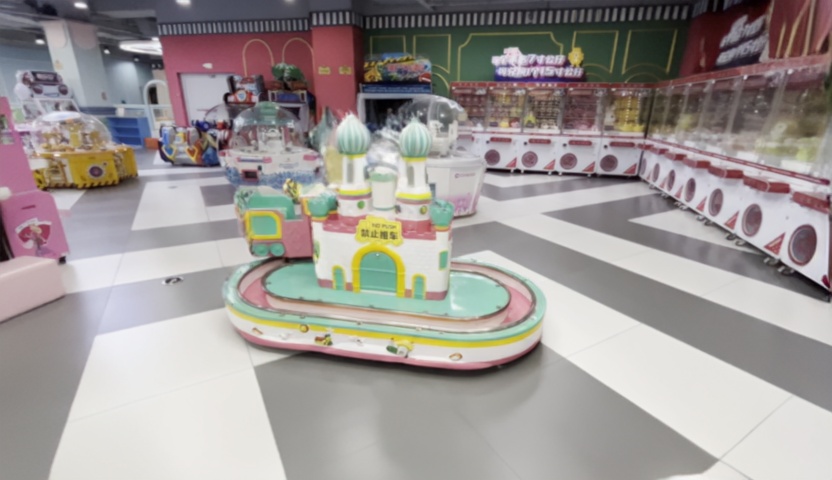}

  \texttt{x3} & \DegradationSceneZoom\FiveCols[south west]{image/Exp/final_vis_jpg/degradation/gt/0a1b7c20a92c43c6b8954b1ac909fb2f0fa8b2997b80604bc8bbec80a1cb2da3_common_common.jpg}{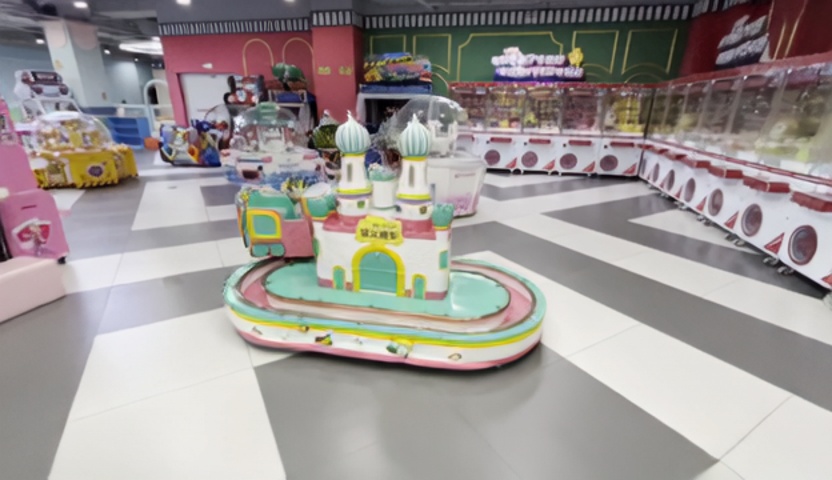}{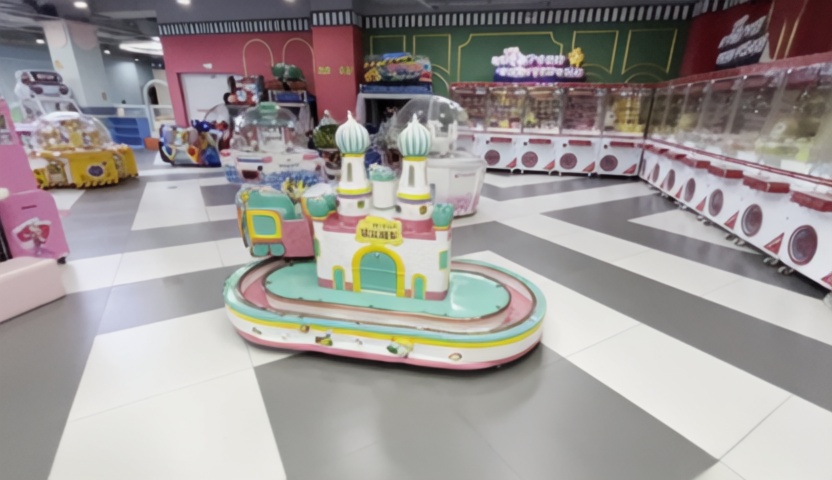}{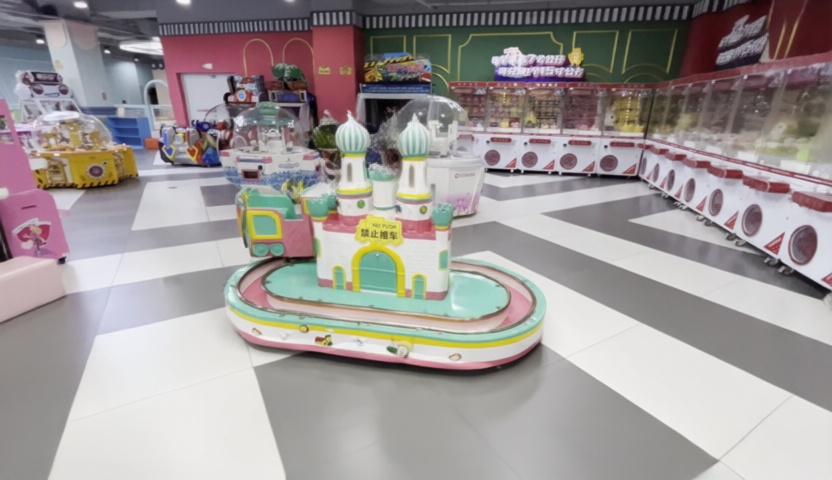}{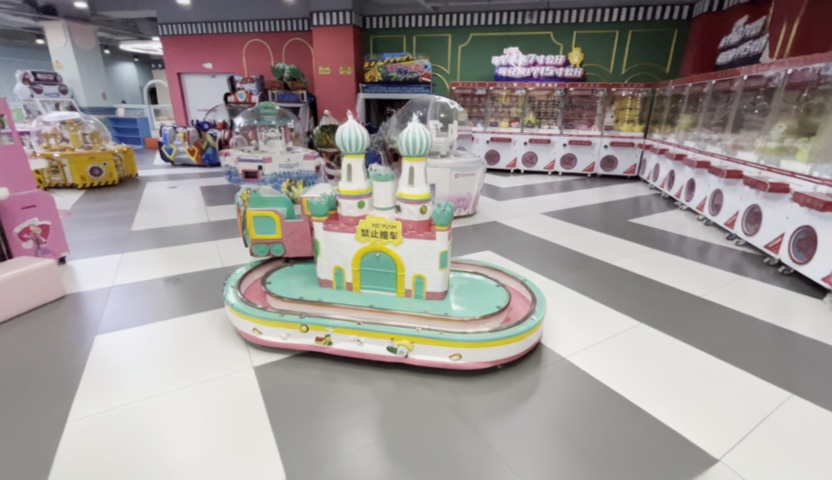}

  \texttt{x4} & \DegradationSceneZoom\FiveCols[south west]{image/Exp/final_vis_jpg/degradation/gt/0a1b7c20a92c43c6b8954b1ac909fb2f0fa8b2997b80604bc8bbec80a1cb2da3_common_common.jpg}{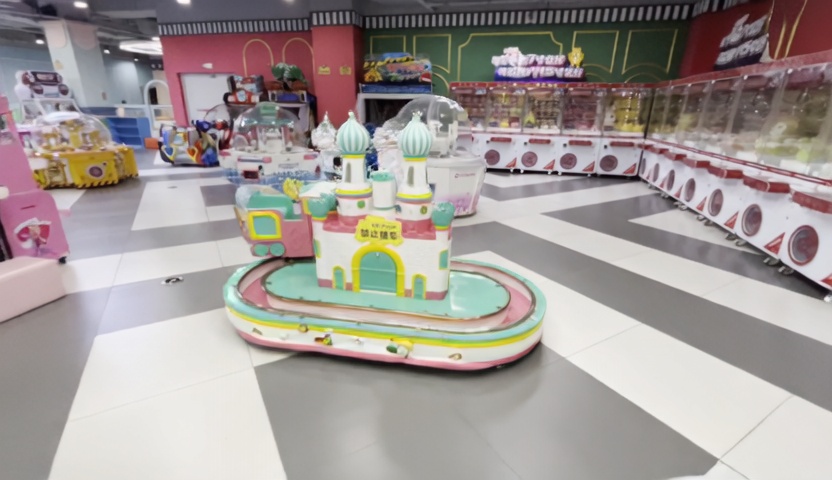}{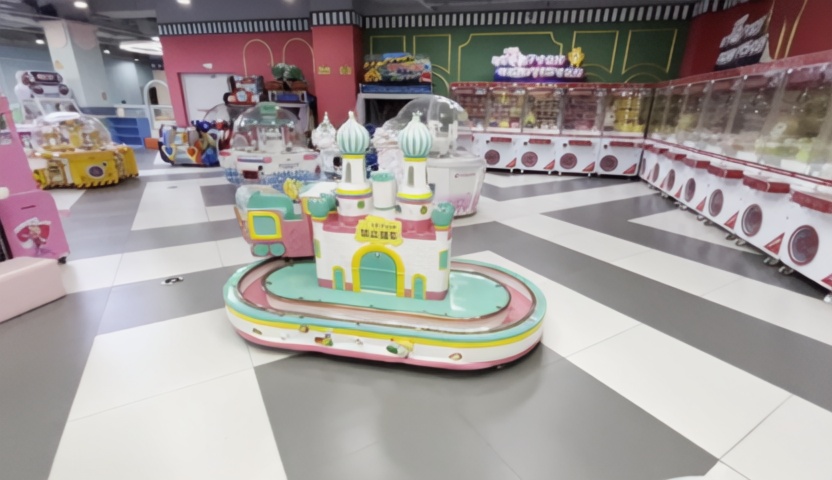}{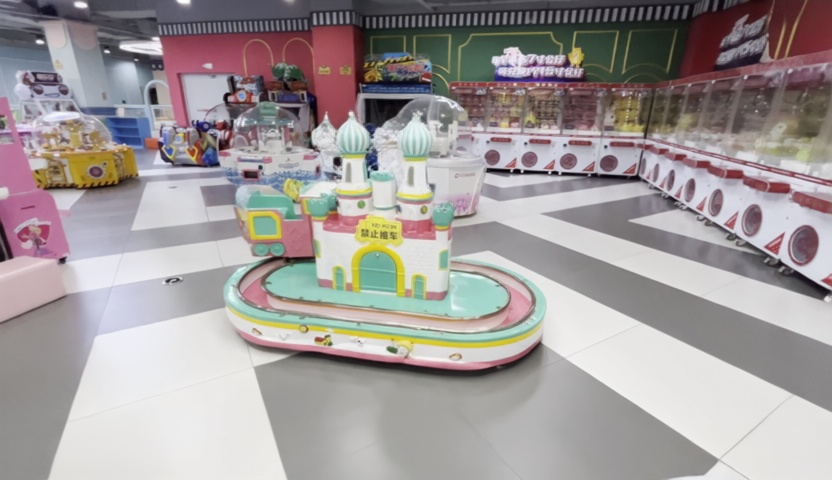}{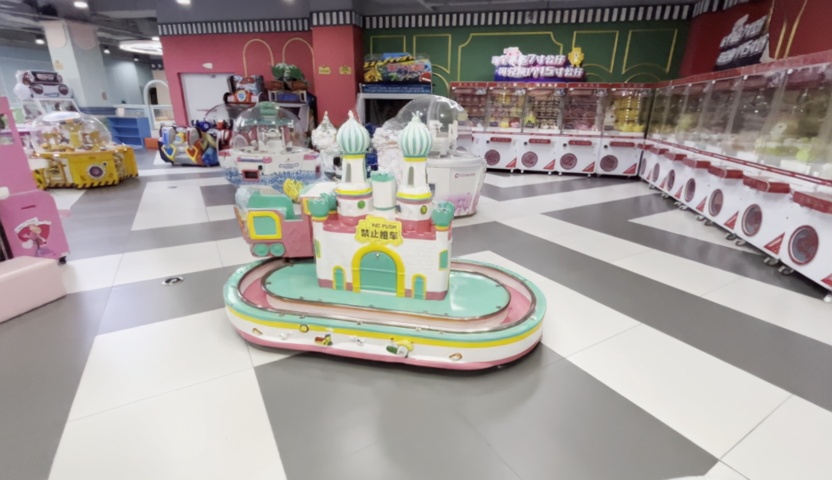}

  \texttt{x6} & \DegradationSceneZoom\FiveCols[south west]{image/Exp/final_vis_jpg/degradation/gt/0a1b7c20a92c43c6b8954b1ac909fb2f0fa8b2997b80604bc8bbec80a1cb2da3_common_common.jpg}{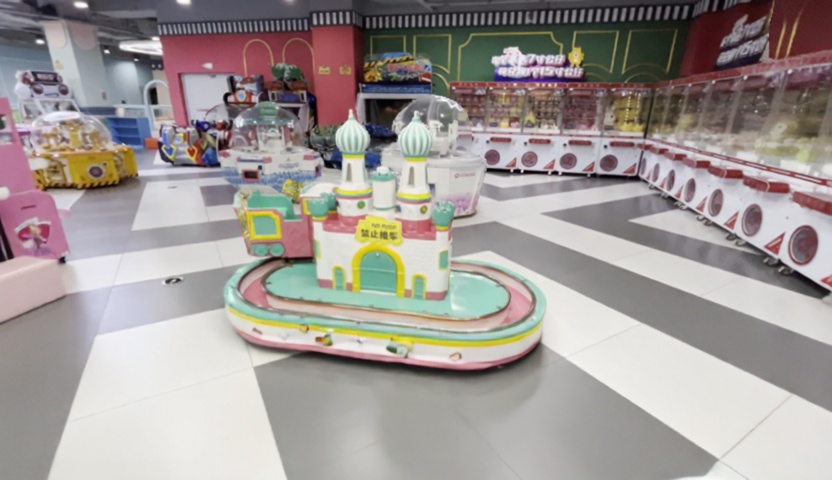}{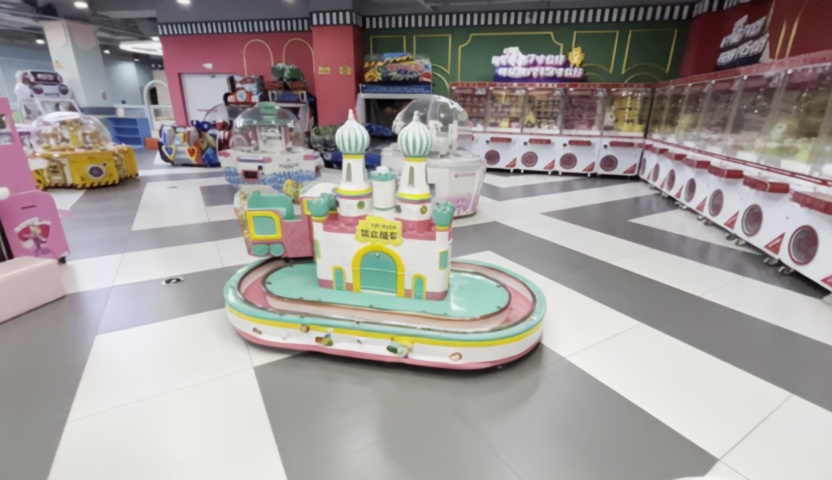}{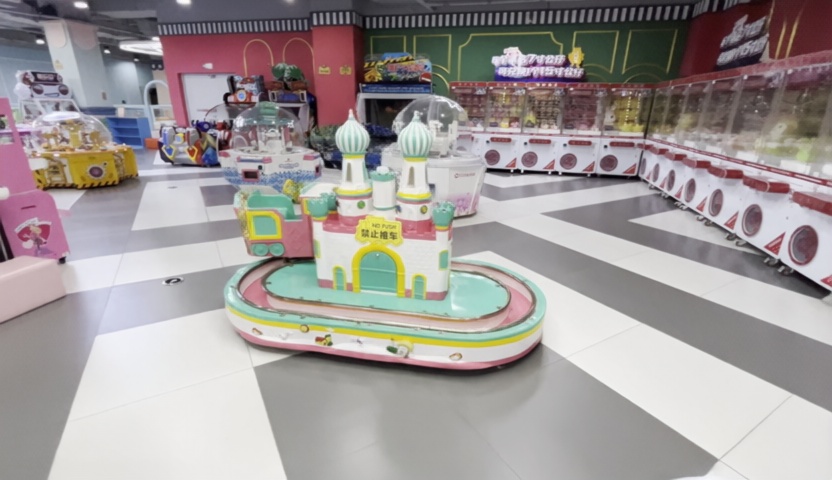}{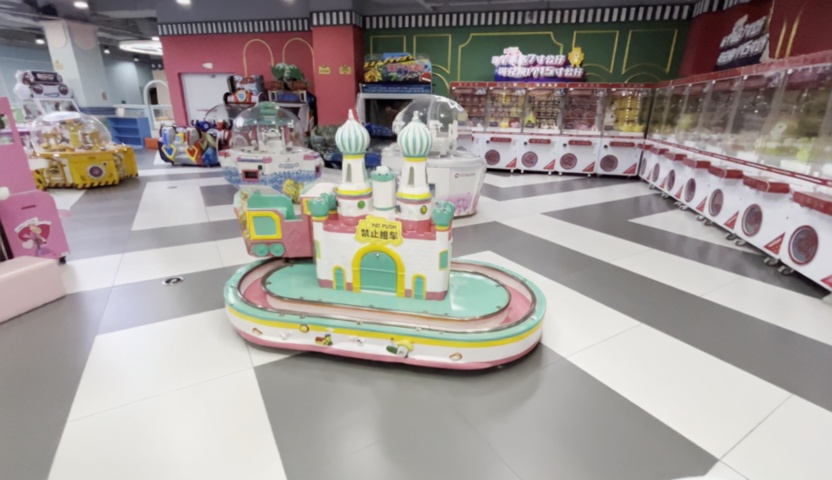}
  {Scale} & {{GT}} & {{W/O fixer}} & {{DIFIX3D+}} & {{MaRINeR}} & {\textbf{Ours}} \\
  
  \multicolumn{6}{c}{\vspace{-0.2em}{Temporal degradation}} \\
  \texttt{9} & \DegradationSceneZoom\FiveCols[south west]{image/Exp/final_vis_jpg/degradation/gt/0a1b7c20a92c43c6b8954b1ac909fb2f0fa8b2997b80604bc8bbec80a1cb2da3_common_common.jpg}{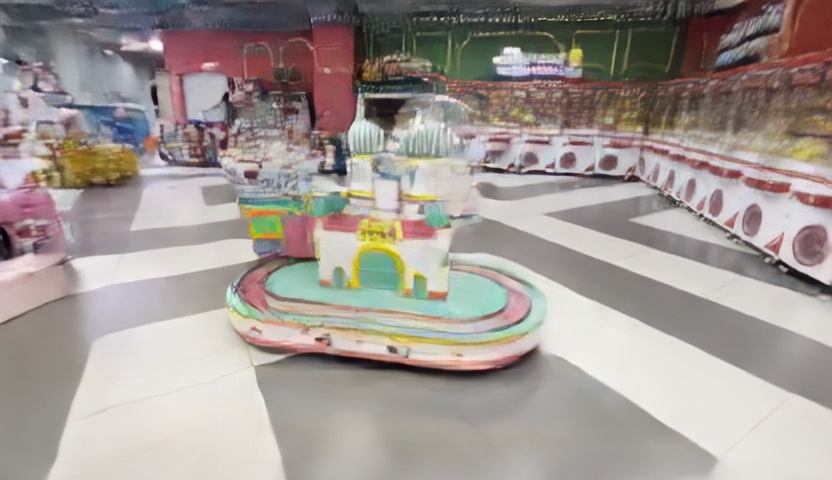}{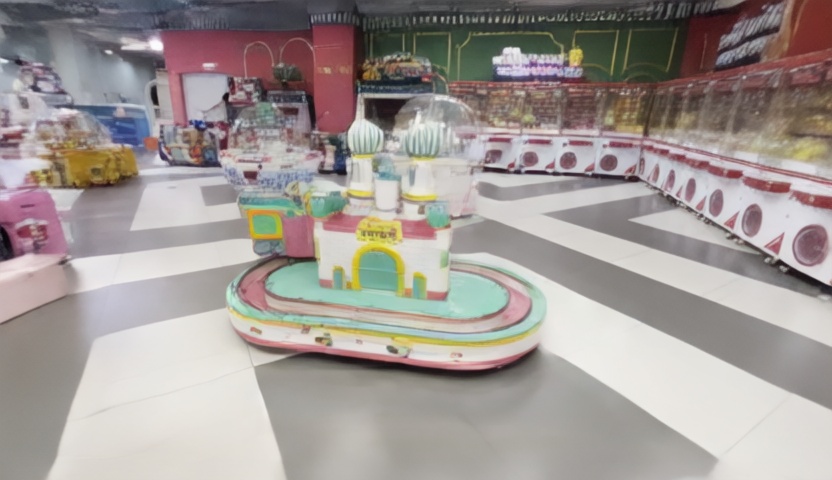}{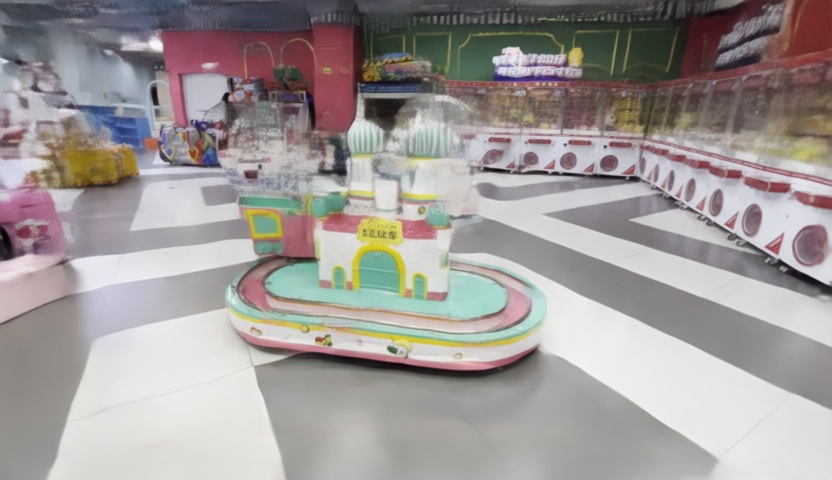}{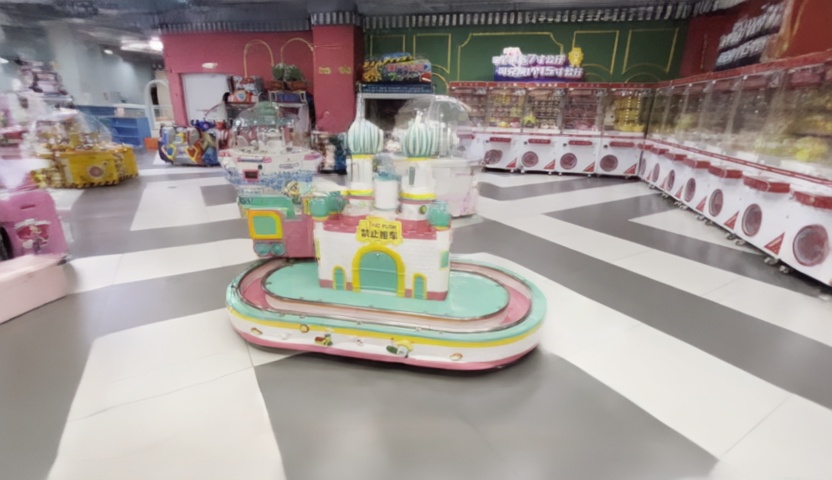}

  \texttt{6} & \DegradationSceneZoom\FiveCols[south west]{image/Exp/final_vis_jpg/degradation/gt/0a1b7c20a92c43c6b8954b1ac909fb2f0fa8b2997b80604bc8bbec80a1cb2da3_common_common.jpg}{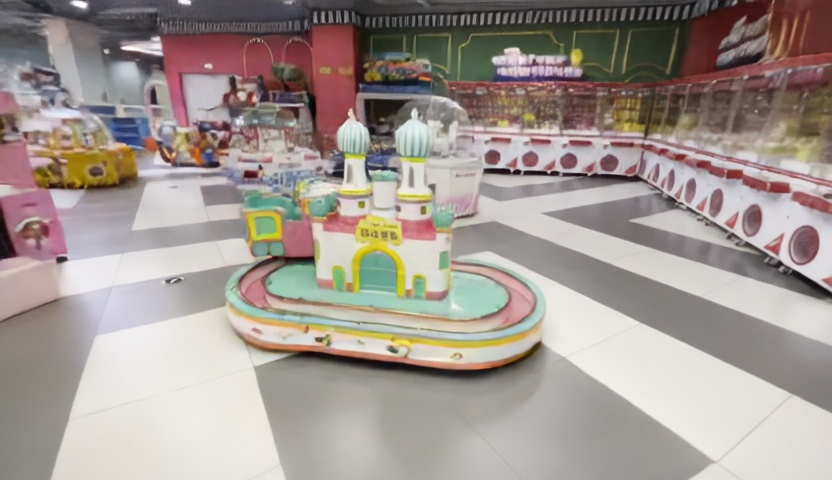}{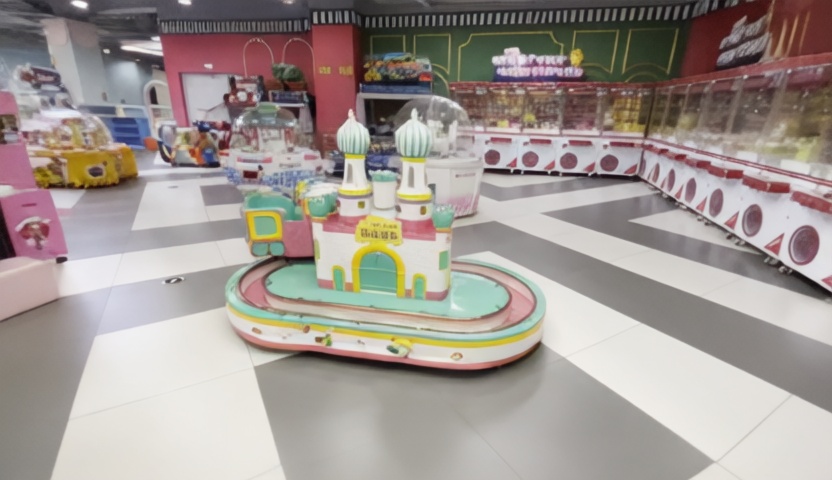}{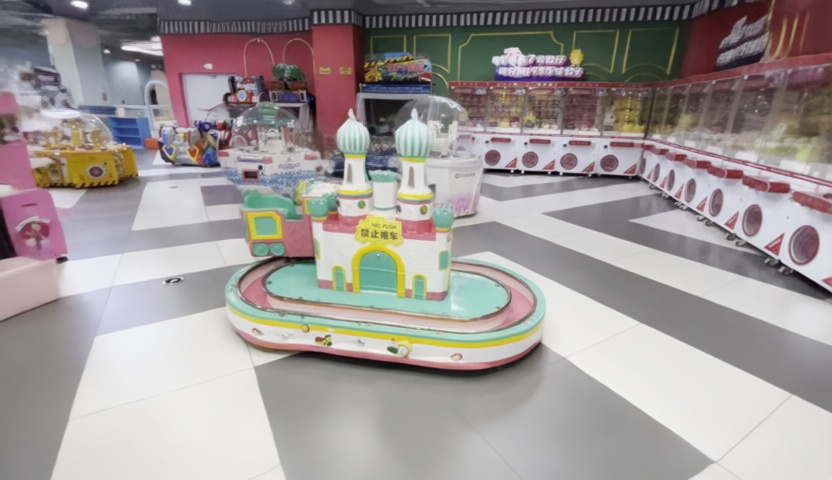}{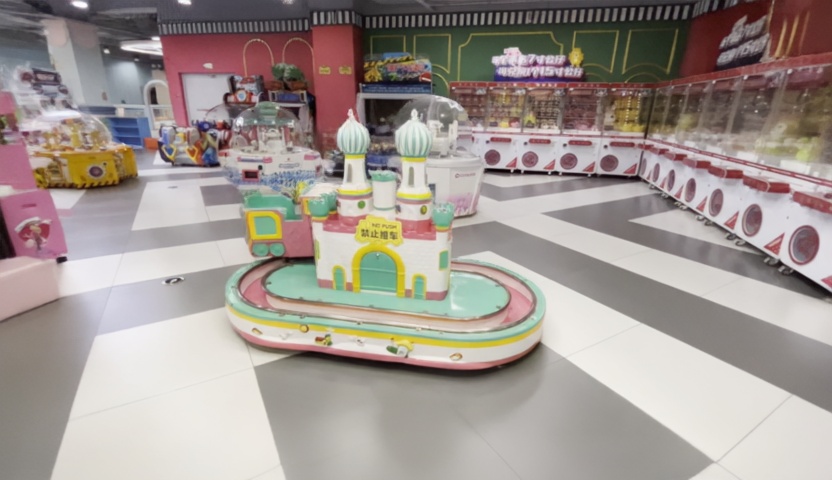}

  \texttt{3} & \DegradationSceneZoom\FiveCols[south west]{image/Exp/final_vis_jpg/degradation/gt/0a1b7c20a92c43c6b8954b1ac909fb2f0fa8b2997b80604bc8bbec80a1cb2da3_common_common.jpg}{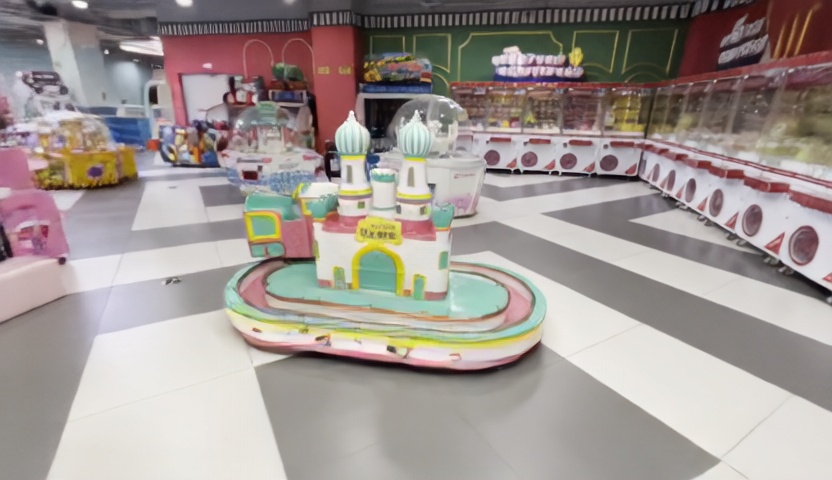}{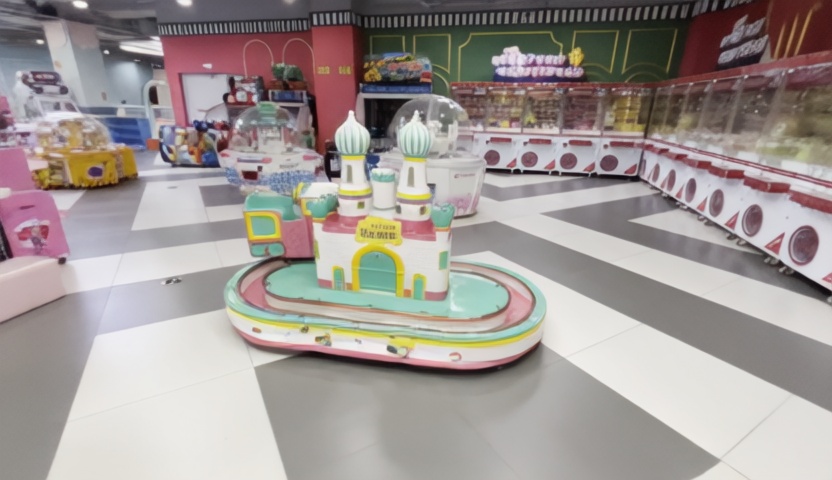}{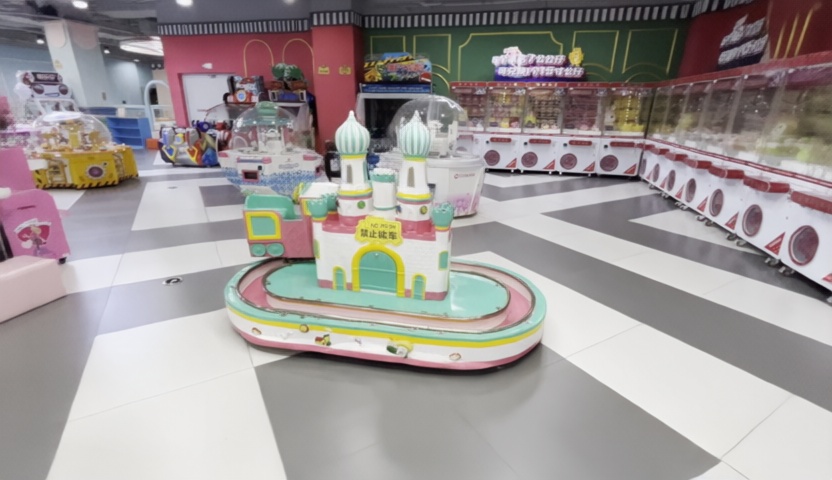}{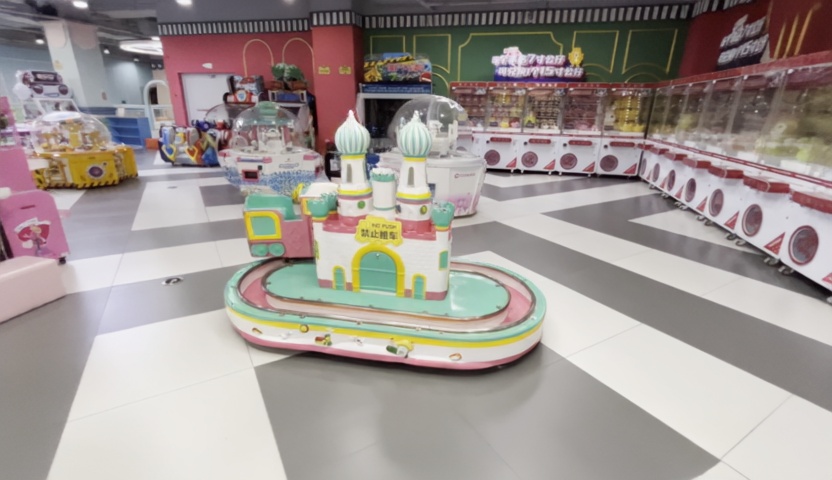}

  \texttt{1} & \DegradationSceneZoom\FiveCols[south west]{image/Exp/final_vis_jpg/degradation/gt/0a1b7c20a92c43c6b8954b1ac909fb2f0fa8b2997b80604bc8bbec80a1cb2da3_common_common.jpg}{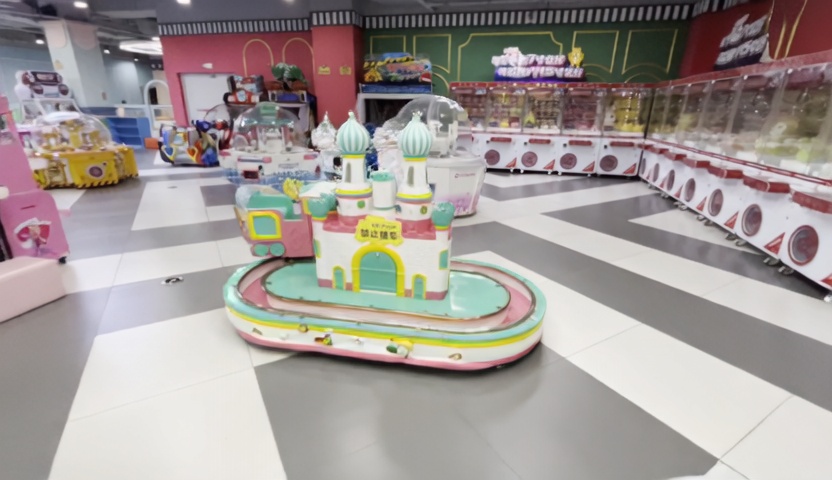}{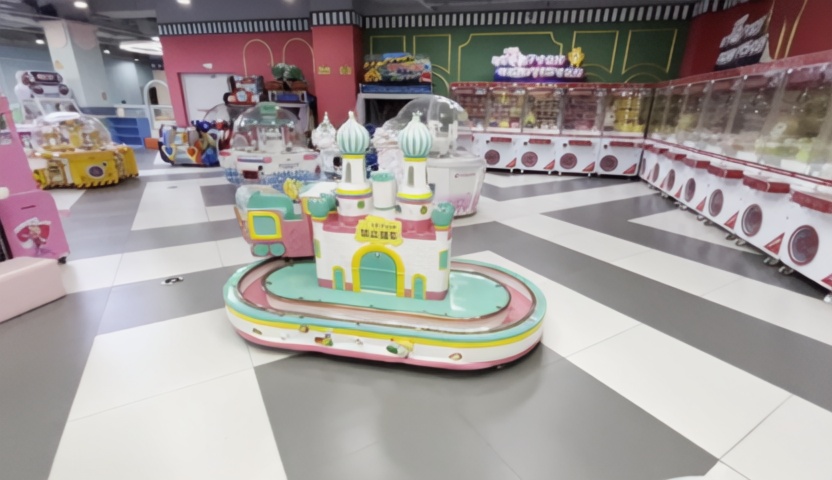}{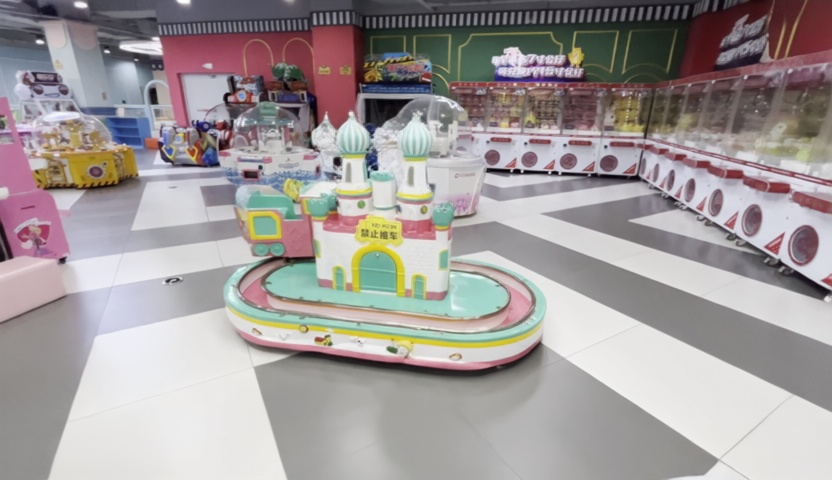}{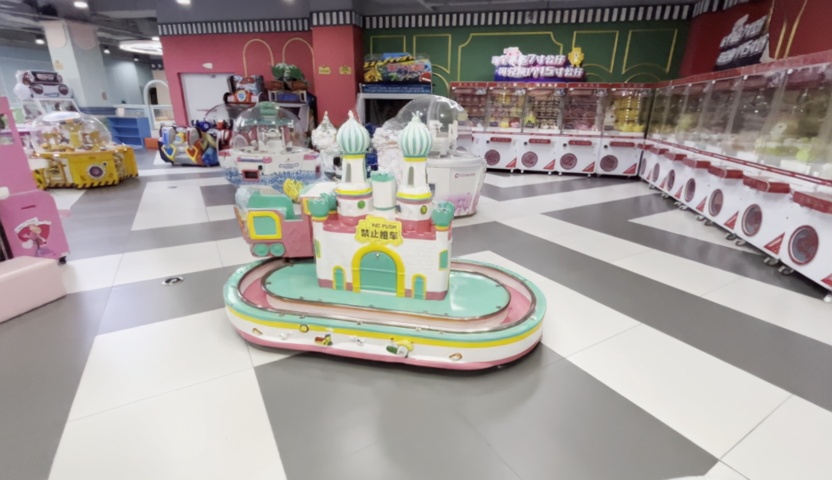}
  {Scale} & {{GT}} & {{W/O fixer}} & {{DIFIX3D+}} & {{MaRINeR}} & {\textbf{Ours}} \\
  
  \multicolumn{6}{c}{\vspace{-0.2em}{Backbone-related degradation}} \\
  \texttt{UNet} & \DegradationSceneZoom\FiveCols[south west]{image/Exp/final_vis_jpg/degradation/gt/0a1b7c20a92c43c6b8954b1ac909fb2f0fa8b2997b80604bc8bbec80a1cb2da3_common_common.jpg}{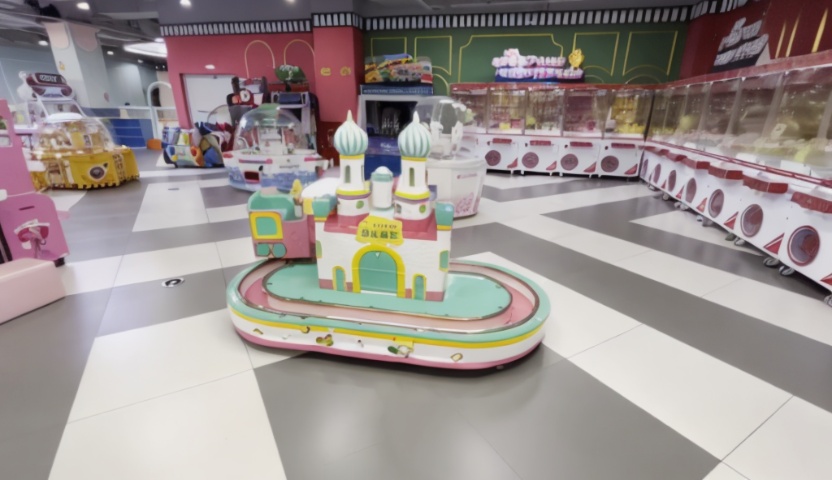}{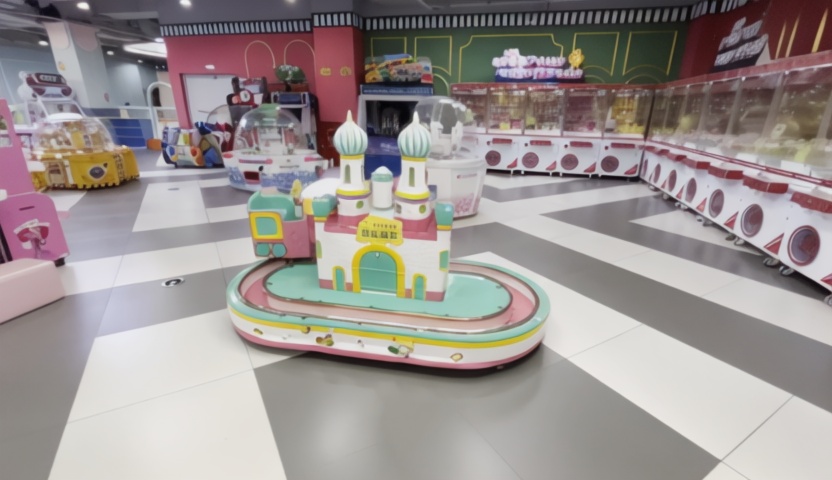}{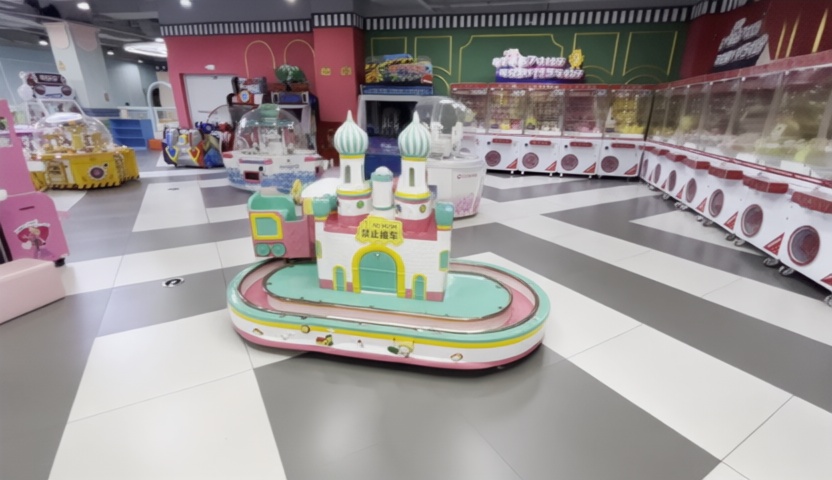}{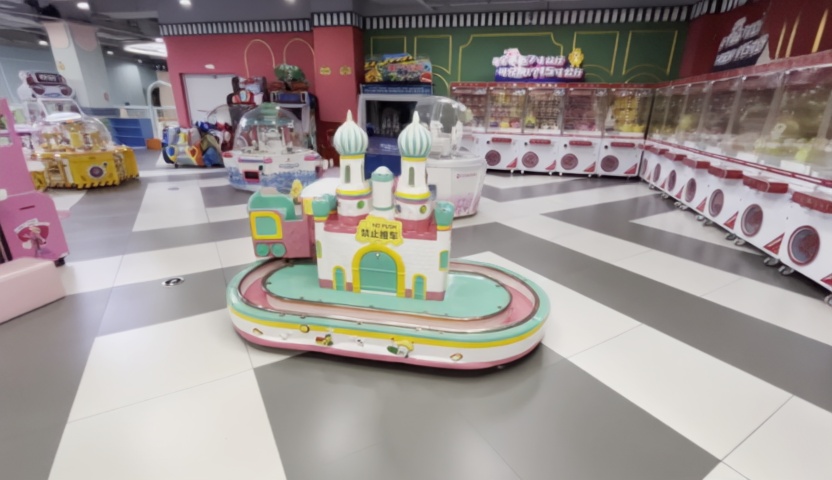}

  \texttt{DiT} & \DegradationSceneZoom\FiveCols[south west]{image/Exp/final_vis_jpg/degradation/gt/0a1b7c20a92c43c6b8954b1ac909fb2f0fa8b2997b80604bc8bbec80a1cb2da3_common_common.jpg}{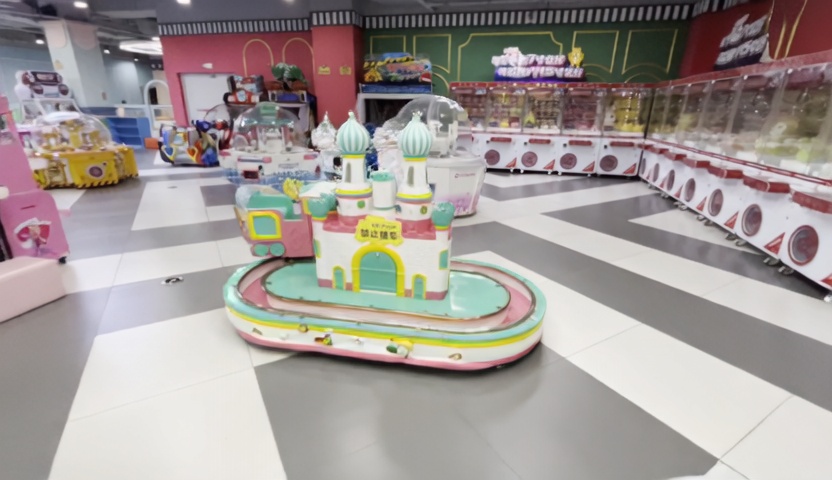}{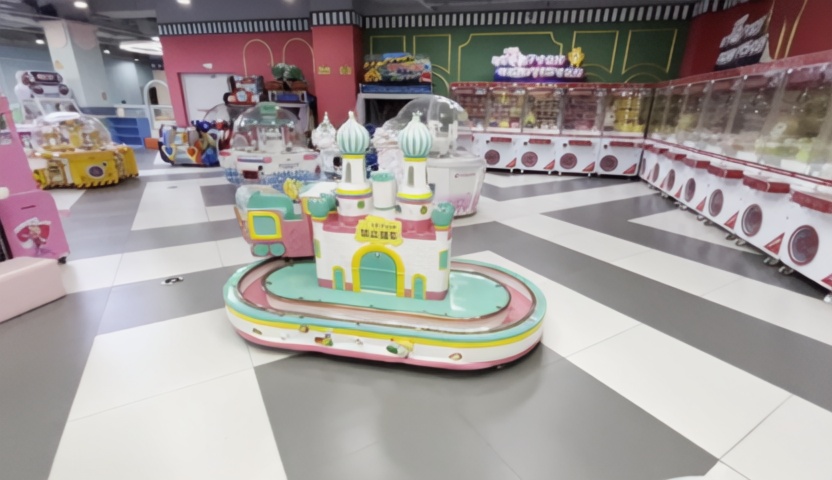}{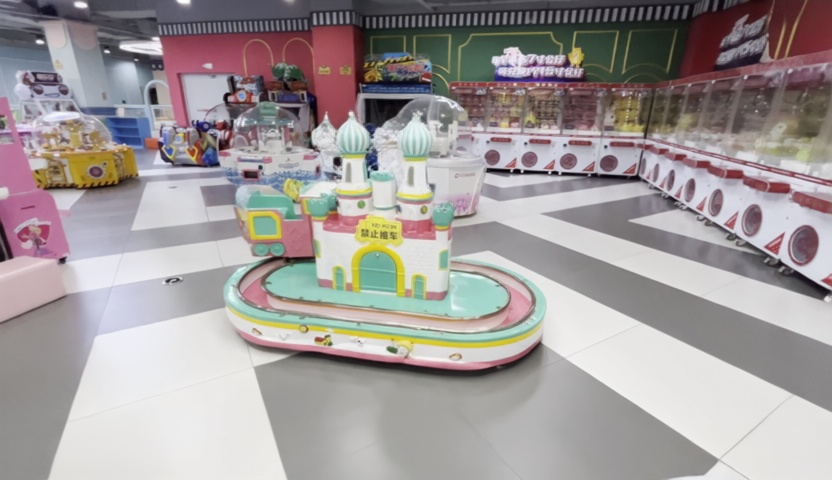}{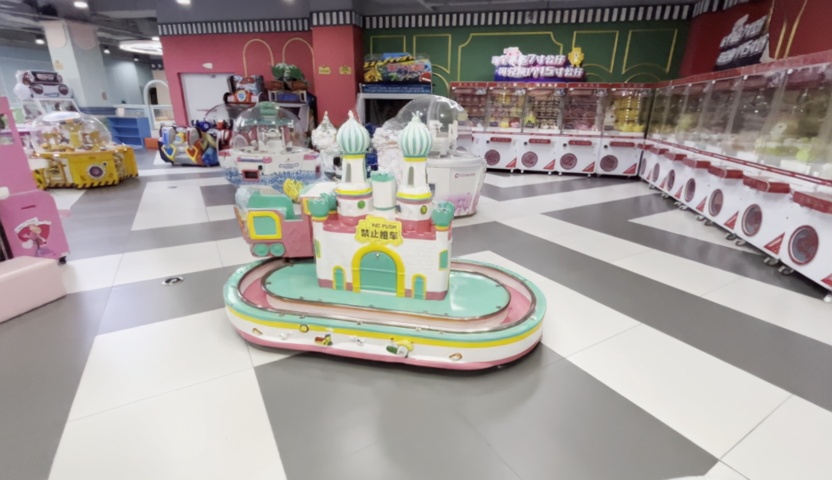}

  {Scale} & {{GT}} & {{W/O fixer}} & {{DIFIX3D+}} & {{MaRINeR}} & {\textbf{Ours}} \\
\end{tabular}

\caption{Additional degradation and fixing visual results. The visual quality consistently improves as the degradation severity decreases (across rows) and as more effective fixers are applied (across columns). These results are a subset of the samples used to produce the t-SNE visualization.}
\label{fig:degradation}
\end{figure*}

We further split the all-in-one t-SNE figure (Fig.~2 in the main paper) into three subfigures corresponding to different degradation dimensions (\ie, spatial, temporal, and backbone-related) in \cref{fig:tsne_all_supp}. We also provide the corresponding degradation and fixed visual results in \cref{fig:degradation}, where the visual quality consistently improves as the degradation severity decreases (across rows) and as more effective fixing methods are applied (across columns). These results are a subset of the samples used to produce the t-SNE visualization.

\section{Limitation analysis}

\def\imgWidthFour{0.235\textwidth} %
\def\imgWidthFive{0.185\textwidth} %
\def\imgWidth{\imgWidthFive} %

\providecommand{\SetImgWidth}[1]{}%
\renewcommand{\SetImgWidth}[1]{%
  \def\imgWidth{#1}%
  \tikzset{img/.style={rectangle, minimum width=\imgWidth, draw=black, inner sep=0, outer sep=0}}%
}

\SetImgWidth{\imgWidthFive}

\providecommand{\FourCols}[5]{}%
\renewcommand{\FourCols}[5][north east]{%
  \ZoomPicBySampling[#1]{\ZoomMag}{\ZoomSize}{\GTPos}{#2}{} &
  \ZoomPicBySampling[#1]{\ZoomMag}{\ZoomSize}{\ZoomPos}{#3}{} &
  \ZoomPicBySampling[#1]{\ZoomMag}{\ZoomSize}{\ZoomPos}{#4}{} &
  \ZoomPicBySampling[#1]{\ZoomMag}{\ZoomSize}{\ZoomPos}{#5}{} \\
}

\providecommand{\ThreeColsSrc}[4]{}%
\renewcommand{\ThreeColsSrc}[4][north east]{%
  \ZoomPicBySampling[#1]{\ZoomMag}{\ZoomSize}{\GTPos}{#2}{} &
  \ZoomPicBySampling[#1]{\ZoomMag}{\ZoomSize}{\RefPos}{#3}{} &
  \ZoomPicBySampling[#1]{\ZoomMag}{\ZoomSize}{\ZoomPos}{#4}{} \\
}

\providecommand{\FiveColsSrc}[6]{}%
\renewcommand{\FiveColsSrc}[6][north east]{%
  \ZoomPicBySampling[#1]{\ZoomMag}{\ZoomSize}{\GTPos}{#2}{} &
  \ZoomPicBySampling[#1]{\ZoomMag}{\ZoomSize}{\RefPos}{#3}{} &
  \ZoomPicBySampling[#1]{\ZoomMag}{\ZoomSize}{\ZoomPos}{#4}{} &
  \ZoomPicBySampling[#1]{\ZoomMag}{\ZoomSize}{\ZoomPos}{#5}{} &
  \ZoomPicBySampling[#1]{\ZoomMag}{\ZoomSize}{\ZoomPos}{#6}{} \\
}

\providecommand{\SetSceneZoomSrc}[5]{}%
\renewcommand{\SetSceneZoomSrc}[5]{%
  \SetSceneZoom{#1}{#3}{#4}{#5}%
  \global\def\RefPos{#2}%
}

\providecommand{\ZoomPicBySamplingTwo}[8]{}%
\renewcommand{\ZoomPicBySamplingTwo}[8][north east]{%
  \begin{subfigure}{\imgWidth}
    \centering
    \begin{tikzpicture}[
      spy using outlines={rectangle},
      inner sep=0
    ]
      \node (img) [img] {\includegraphics[width=\textwidth]{#8}};
      \spy[draw=green,magnification=#2,size=#3] on #4 in node [anchor=#1] at (img.#1);
      \spy[draw=orange,magnification=#5,size=#6] on #7 in node [anchor=south east] at (img.south east);
    \end{tikzpicture}
  \end{subfigure}%
}

\providecommand{\SetSceneZoomTwoSepSrc}[8]{}%
\renewcommand{\SetSceneZoomTwoSepSrc}[8]{%
  \global\def\ZoomPosA{#1}%
  \global\def\RefPosA{#2}%
  \global\def\ZoomMagA{#3}%
  \global\def\ZoomSizeA{#4}%
  \global\def\GTPosA{#1}%
  \global\def\ZoomPosB{#5}%
  \global\def\RefPosB{#6}%
  \global\def\ZoomMagB{#7}%
  \global\def\ZoomSizeB{#8}%
  \global\def\GTPosB{#5}%
}
\providecommand{\ThreeColsSrcTwo}[4]{}%
\renewcommand{\ThreeColsSrcTwo}[4][north east]{%
  \ZoomPicBySamplingTwo[#1]{\ZoomMagA}{\ZoomSizeA}{\GTPosA}{\ZoomMagB}{\ZoomSizeB}{\GTPosB}{#2} &
  \ZoomPicBySamplingTwo[#1]{\ZoomMagA}{\ZoomSizeA}{\RefPosA}{\ZoomMagB}{\ZoomSizeB}{\RefPosB}{#3} &
  \ZoomPicBySamplingTwo[#1]{\ZoomMagA}{\ZoomSizeA}{\ZoomPosA}{\ZoomMagB}{\ZoomSizeB}{\ZoomPosB}{#4} \\
}

\begin{figure}[h]
\centering
\SetImgWidth{0.32\textwidth}
\setlength{\tabcolsep}{0.6pt}
\begin{tabular}{@{}c c c@{}}
  \multicolumn{3}{c}{\vspace{-0.2em}} \\
  \SetSceneZoomTwoSepSrc{(-1.45,0.6)}{(-1.2,0.6)}{1}{1.0cm}{(1.4,0.6)}{(1.4,0.6)}{1}{1.0cm}%
  \ThreeColsSrcTwo[south west]{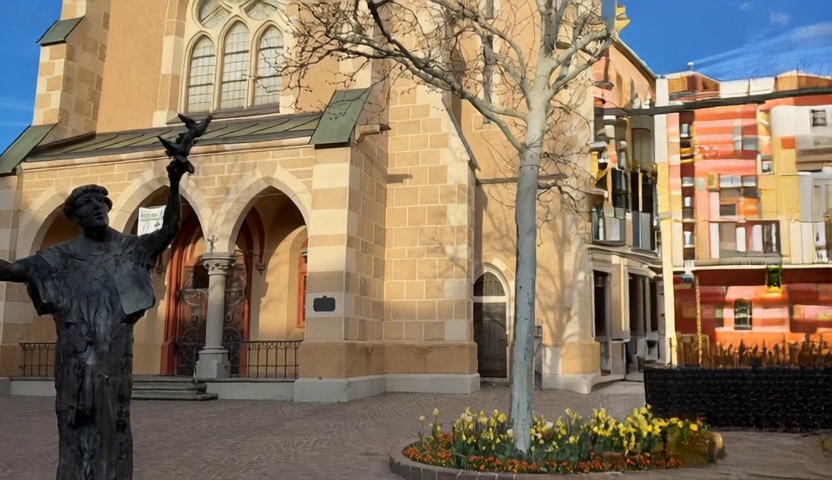}{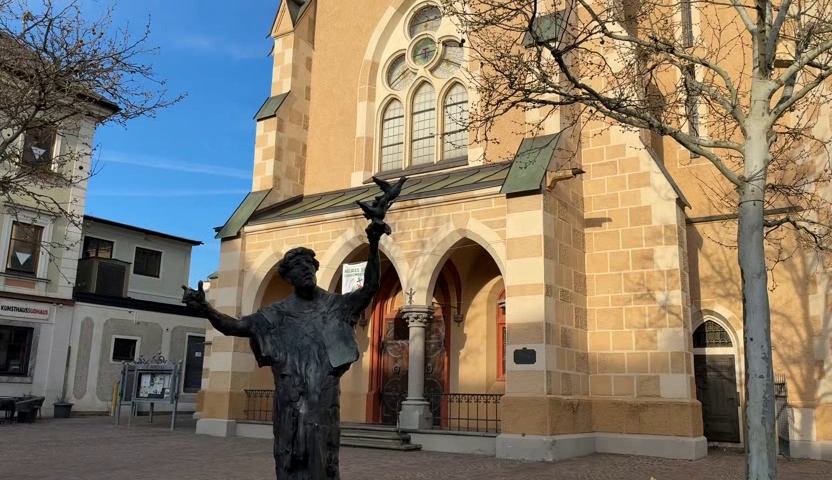}{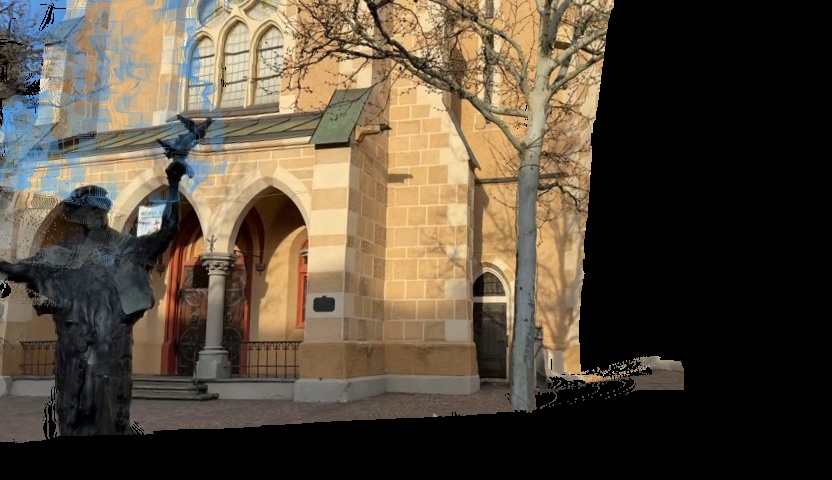}
  {Target} & {Source} & {Warping} \\
\end{tabular}
\caption{Flow-based warping failure case.}
\label{fig:flow_failure_case}
\end{figure}

\noindent For implicit end-to-end approaches, we estimate the optical flow via SEA-RAFT~\cite{wang2024sea} between the diffusion-synthesized target image and the source image; however, when the viewpoint change between target and source view is large, SEA-RAFT can struggle to estimate accurate correspondences, which further leads to erroneous warping (\cref{fig:flow_failure_case}).

\section{Discussion}

\def\imgWidthFour{0.235\textwidth} %
\def\imgWidthFive{0.185\textwidth} %
\def\imgWidth{\imgWidthFive} %

\providecommand{\SetImgWidth}[1]{}%
\renewcommand{\SetImgWidth}[1]{%
  \def\imgWidth{#1}%
  \tikzset{img/.style={rectangle, minimum width=\imgWidth, draw=black, inner sep=0, outer sep=0}}%
}

\SetImgWidth{\imgWidthFive}

\providecommand{\FourCols}[5]{}%
\renewcommand{\FourCols}[5][north east]{%
  \ZoomPicBySampling[#1]{\ZoomMag}{\ZoomSize}{\GTPos}{#2}{} &
  \ZoomPicBySampling[#1]{\ZoomMag}{\ZoomSize}{\ZoomPos}{#3}{} &
  \ZoomPicBySampling[#1]{\ZoomMag}{\ZoomSize}{\ZoomPos}{#4}{} &
  \ZoomPicBySampling[#1]{\ZoomMag}{\ZoomSize}{\ZoomPos}{#5}{} \\
}

\providecommand{\ThreeColsSrc}[4]{}%
\renewcommand{\ThreeColsSrc}[4][north east]{%
  \ZoomPicBySampling[#1]{\ZoomMag}{\ZoomSize}{\GTPos}{#2}{} &
  \ZoomPicBySampling[#1]{\ZoomMag}{\ZoomSize}{\RefPos}{#3}{} &
  \ZoomPicBySampling[#1]{\ZoomMag}{\ZoomSize}{\ZoomPos}{#4}{} \\
}

\providecommand{\FiveColsSrc}[6]{}%
\renewcommand{\FiveColsSrc}[6][north east]{%
  \ZoomPicBySampling[#1]{\ZoomMag}{\ZoomSize}{\GTPos}{#2}{} &
  \ZoomPicBySampling[#1]{\ZoomMag}{\ZoomSize}{\RefPos}{#3}{} &
  \ZoomPicBySampling[#1]{\ZoomMag}{\ZoomSize}{\ZoomPos}{#4}{} &
  \ZoomPicBySampling[#1]{\ZoomMag}{\ZoomSize}{\ZoomPos}{#5}{} &
  \ZoomPicBySampling[#1]{\ZoomMag}{\ZoomSize}{\ZoomPos}{#6}{} \\
}

\providecommand{\SetSceneZoomSrc}[5]{}%
\renewcommand{\SetSceneZoomSrc}[5]{%
  \SetSceneZoom{#1}{#3}{#4}{#5}%
  \global\def\RefPos{#2}%
}

\providecommand{\ZoomPicBySamplingTwo}[8]{}%
\renewcommand{\ZoomPicBySamplingTwo}[8][north east]{%
  \begin{subfigure}{\imgWidth}
    \centering
    \begin{tikzpicture}[
      spy using outlines={rectangle},
      inner sep=0
    ]
      \node (img) [img] {\includegraphics[width=\textwidth]{#8}};
      \spy[draw=green,magnification=#2,size=#3] on #4 in node [anchor=#1] at (img.#1);
      \spy[draw=orange,magnification=#5,size=#6] on #7 in node [anchor=south east] at (img.south east);
    \end{tikzpicture}
  \end{subfigure}%
}

\providecommand{\SetSceneZoomTwoSepSrc}[8]{}%
\renewcommand{\SetSceneZoomTwoSepSrc}[8]{%
  \global\def\ZoomPosA{#1}%
  \global\def\RefPosA{#2}%
  \global\def\ZoomMagA{#3}%
  \global\def\ZoomSizeA{#4}%
  \global\def\GTPosA{#1}%
  \global\def\ZoomPosB{#5}%
  \global\def\RefPosB{#6}%
  \global\def\ZoomMagB{#7}%
  \global\def\ZoomSizeB{#8}%
  \global\def\GTPosB{#5}%
}
\providecommand{\ThreeColsSrcTwo}[4]{}%
\renewcommand{\ThreeColsSrcTwo}[4][north east]{%
  \ZoomPicBySamplingTwo[#1]{\ZoomMagA}{\ZoomSizeA}{\GTPosA}{\ZoomMagB}{\ZoomSizeB}{\GTPosB}{#2} &
  \ZoomPicBySamplingTwo[#1]{\ZoomMagA}{\ZoomSizeA}{\RefPosA}{\ZoomMagB}{\ZoomSizeB}{\RefPosB}{#3} &
  \ZoomPicBySamplingTwo[#1]{\ZoomMagA}{\ZoomSizeA}{\ZoomPosA}{\ZoomMagB}{\ZoomSizeB}{\ZoomPosB}{#4} \\
}

\begin{figure*}[!h]
\centering
\setlength{\tabcolsep}{0.6pt}

\begin{subfigure}{\textwidth}
\centering
\SetImgWidth{\imgWidthFive}
\begin{tabular}{@{}c c c c c@{}}
  \multicolumn{5}{c}{\vspace{-0.2em}} \\
  \SetSceneZoomSrc{(0.2,-0.2)}{(0.15,-0.15)}{4}{1.0cm}{(0.2,-0.2)}\FiveColsSrc[north west]{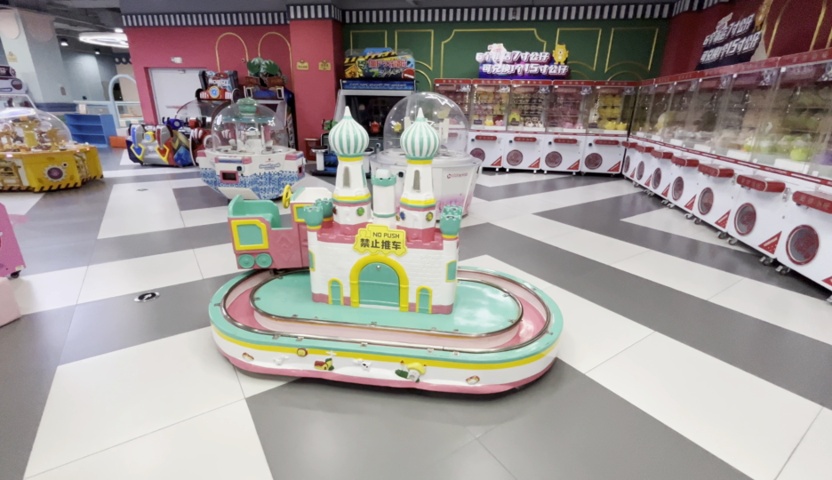}{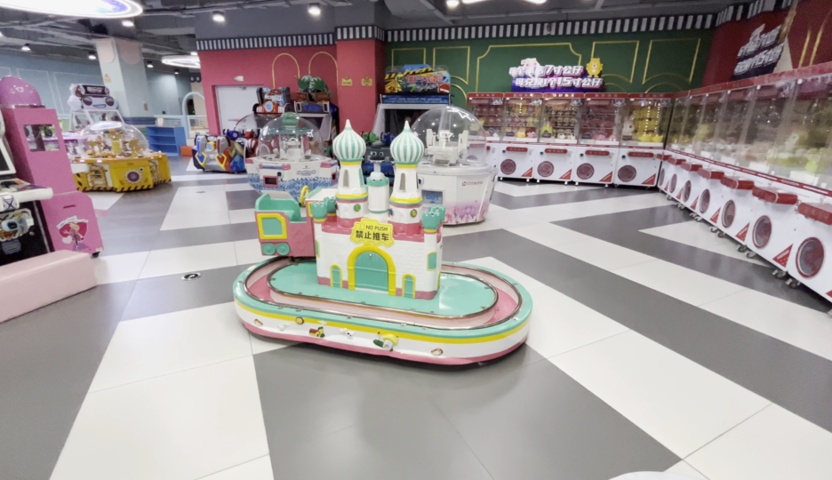}{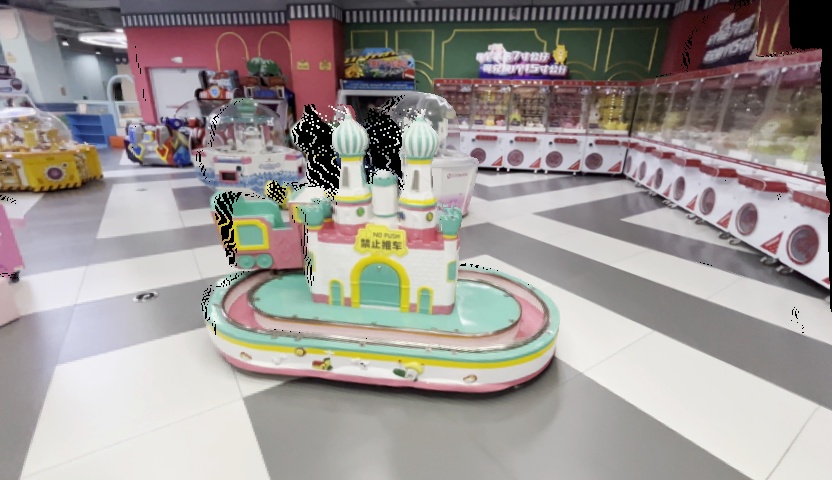}{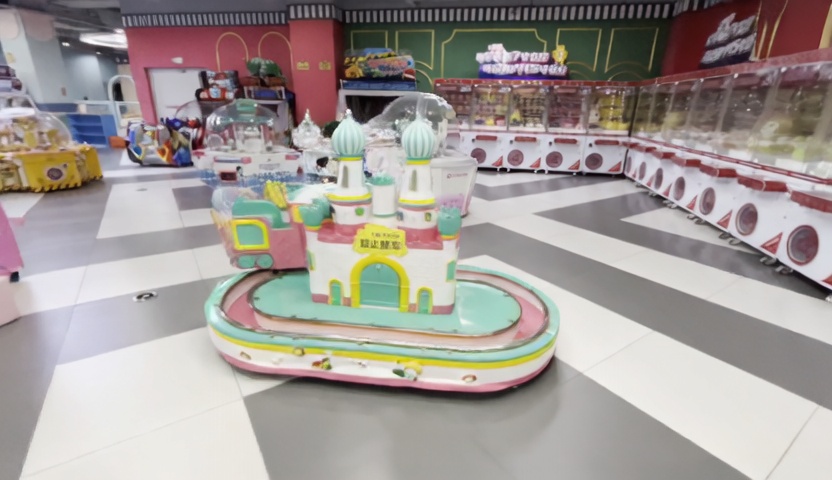}{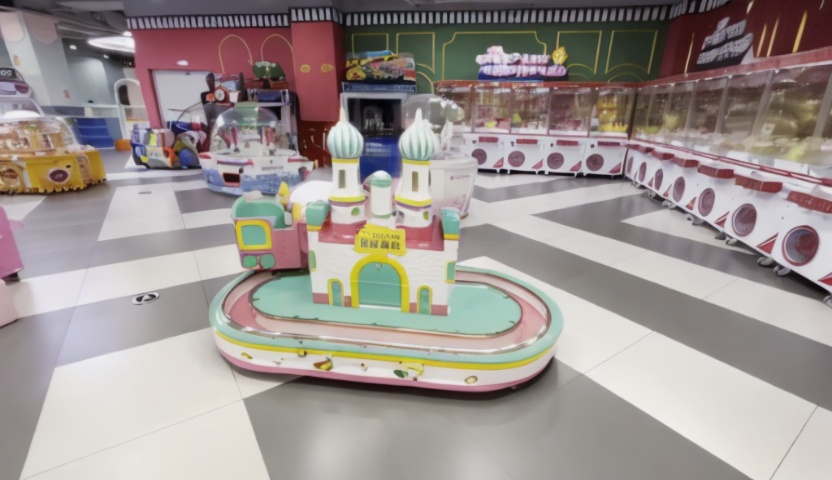}
  
  \multicolumn{5}{c}{\vspace{-1.5em}} \\
  \SetSceneZoomSrc{(-0.8,0.2)}{(-0.5,0.1)}{4}{1.0cm}{(-0.8,0.2)}\FiveColsSrc[north east]{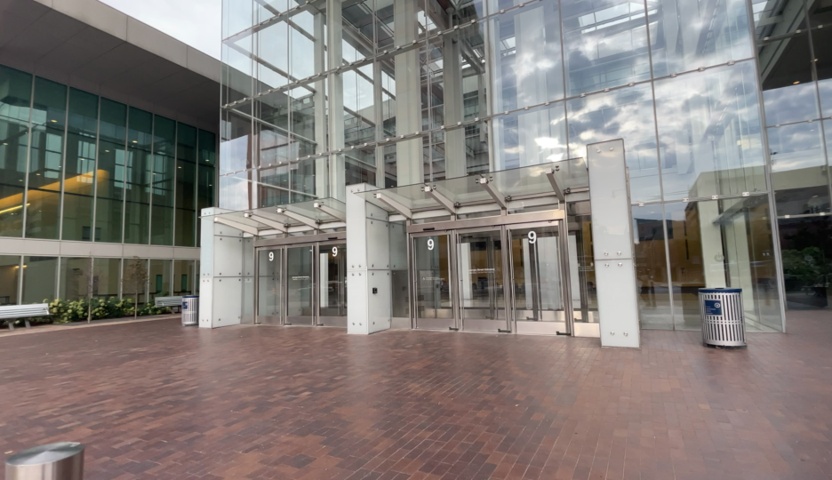}{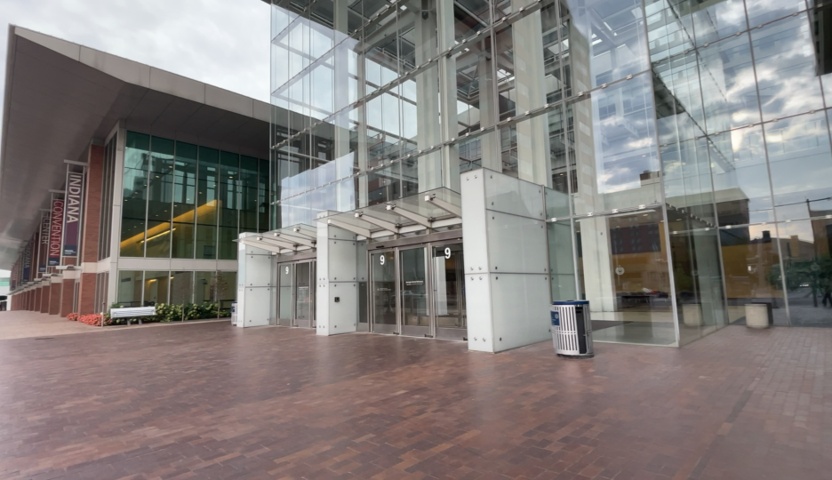}{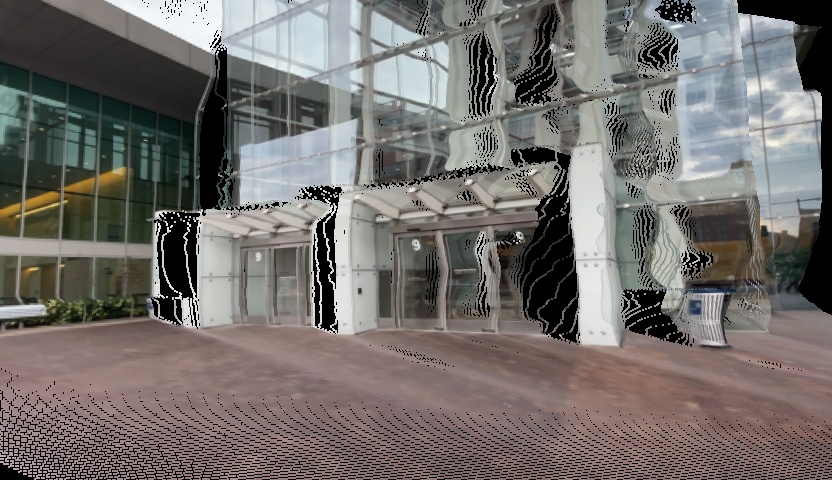}{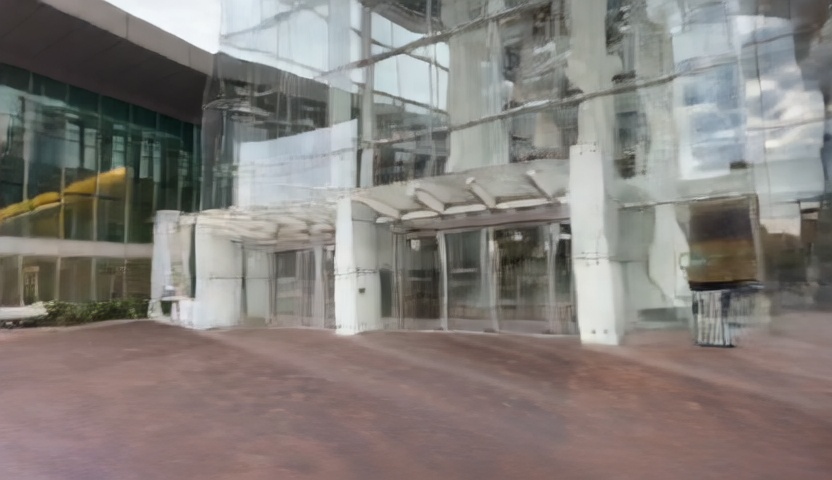}{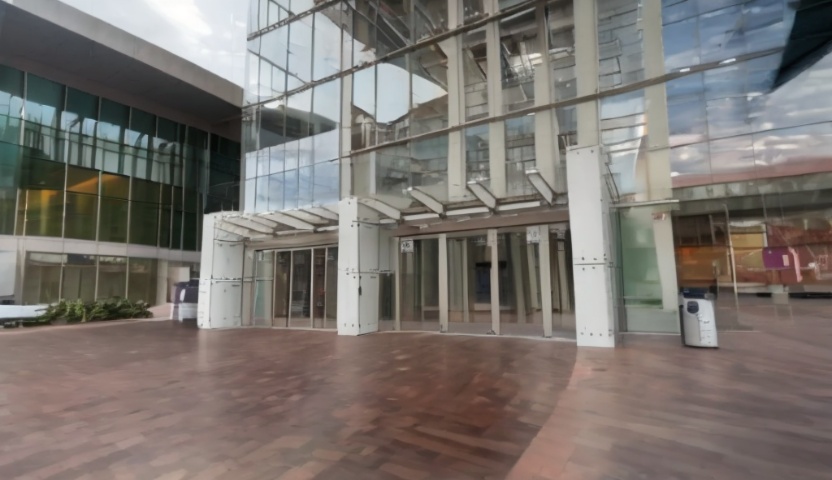}

  {Target} & {Source} & {Warping} & {VACE} & {Viewcrafter} \\
\end{tabular}
\caption{Explicit DWI issues.}
\label{fig:warping_failure_case}
\end{subfigure}

\vspace{0.5em}

\begin{subfigure}{\textwidth}
\centering
\SetImgWidth{0.32\textwidth}
\begin{tabular}{@{}c c c@{}}
  \multicolumn{3}{c}{\vspace{-0.2em}} \\
  \SetSceneZoomSrc{(0.5,-0.45)}{(0.5,-0.45)}{3}{1.0cm}{(0.5,-0.45)}%
  \ThreeColsSrc[north west]{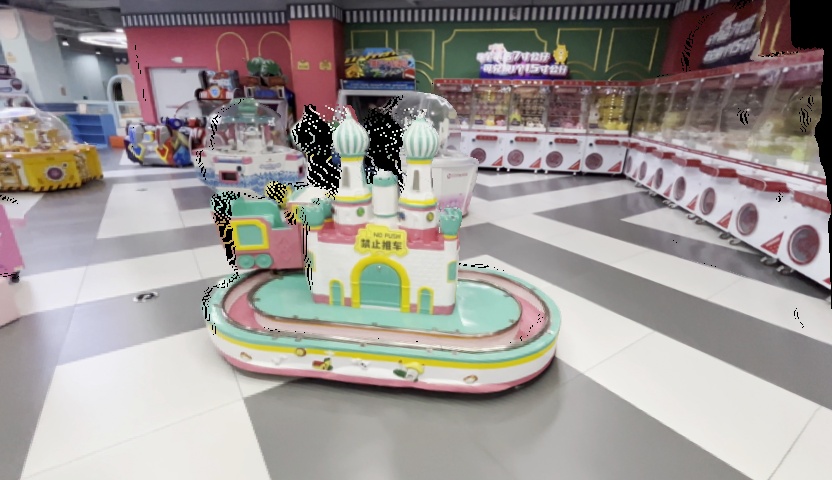}{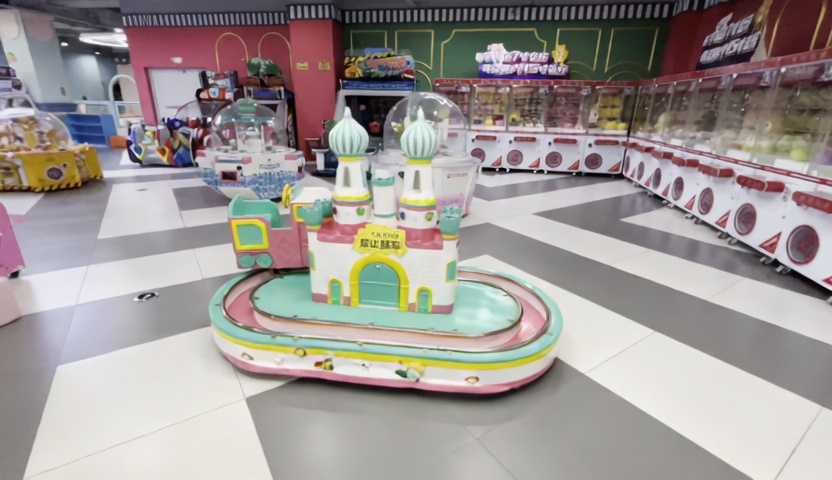}{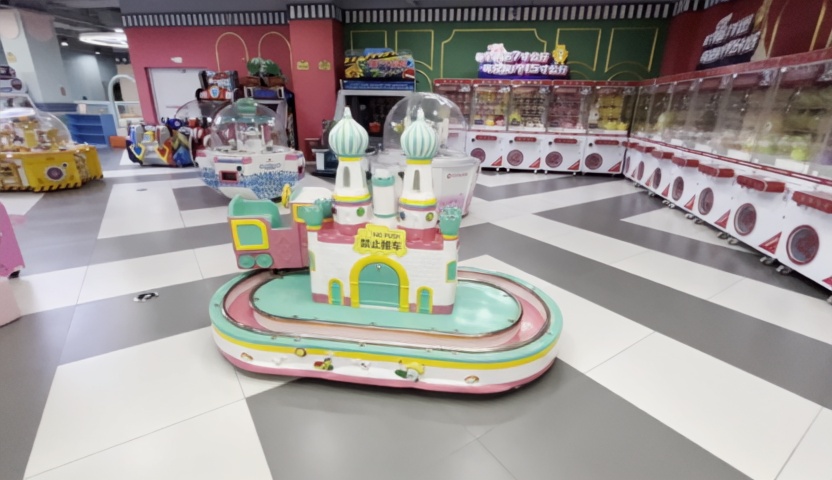}

  {Warping} & {W/O fixer} & {\textbf{Ours}} \\
\end{tabular}
\caption{View-dependent effect preservation case.}
\label{fig:vde_case}
\end{subfigure}

\caption{Warping and view-dependent effects discussion.}
\label{fig:failure}
\end{figure*}

For explicit DWI methods, we observe two main issues: (1) view-dependent effects can be baked into the estimated geometry and thus propagated by warping; (2) a single surface point is typically constrained to have a single depth, which cannot represent multi-depth structures (e.g., semi-transparent or layered regions), leading to warping errors. These errors are then propagated to the inpainting stage, often leading DWI methods to produce corrupted or unreal outputs (\cref{fig:warping_failure_case}). However, for the case that upstream models already generate possible view-dependent effects (VDE), although our \myname\ relies on RPA, the proposed UAGF enables us to preserve such VDE when the warped regions are assessed to be \emph{unreliable}, instead of blindly overwriting them using warping during fixing (\cref{fig:vde_case}).

\FloatBarrier %

\section{More visual results}

\subsection{Benchmark fixing results}

\def\imgWidth{0.185\textwidth} %

\tikzset{img/.style={rectangle, minimum width=\imgWidth, draw=black, inner sep=0, outer sep=0}}

\providecommand{\SetSceneZoom}[4]{}%
\renewcommand{\SetSceneZoom}[4]{%
  \global\def\ZoomPos{#1}%
  \global\def\ZoomMag{#2}%
  \global\def\ZoomSize{#3}%
  \global\def\GTPos{#4}%
}

\providecommand{\FiveCols}[6][north east]{}%
\renewcommand{\FiveCols}[6][north east]{%
  \ZoomPicBySampling[#1]{\ZoomMag}{\ZoomSize}{\GTPos}{#2}{} &
  \ZoomPicBySampling[#1]{\ZoomMag}{\ZoomSize}{\ZoomPos}{#3}{} &
  \ZoomPicBySampling[#1]{\ZoomMag}{\ZoomSize}{\ZoomPos}{#4}{} &
  \ZoomPicBySampling[#1]{\ZoomMag}{\ZoomSize}{\ZoomPos}{#5}{} &
  \ZoomPicBySampling[#1]{\ZoomMag}{\ZoomSize}{\ZoomPos}{#6}{} \\
}

\begin{figure*}[h]
\centering
\setlength{\tabcolsep}{0.6pt}
\begin{tabular}{@{}c c c c c@{}}
  \multicolumn{5}{c}{\vspace{-0.2em}{Applied on VACE (explicit NVS)}} \\
  \SetSceneZoom{(-0.4,-0.5)}{4}{1.0cm}{(-0.4,-0.5)}\FiveCols[north east]{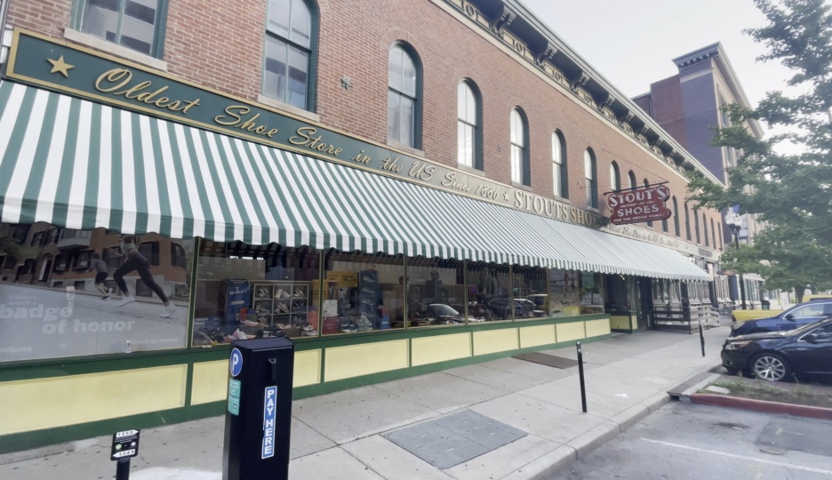}{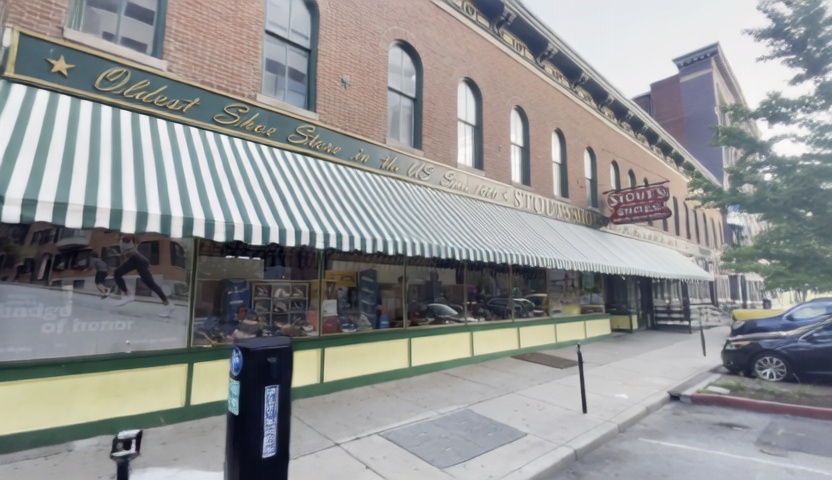}{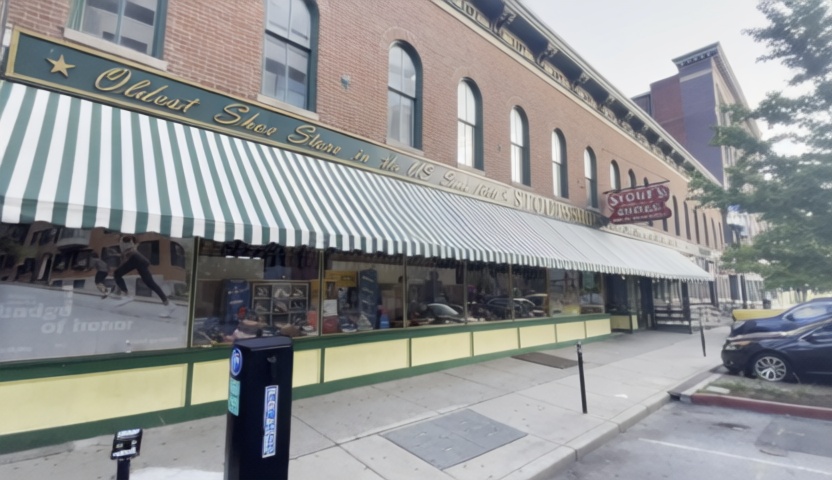}{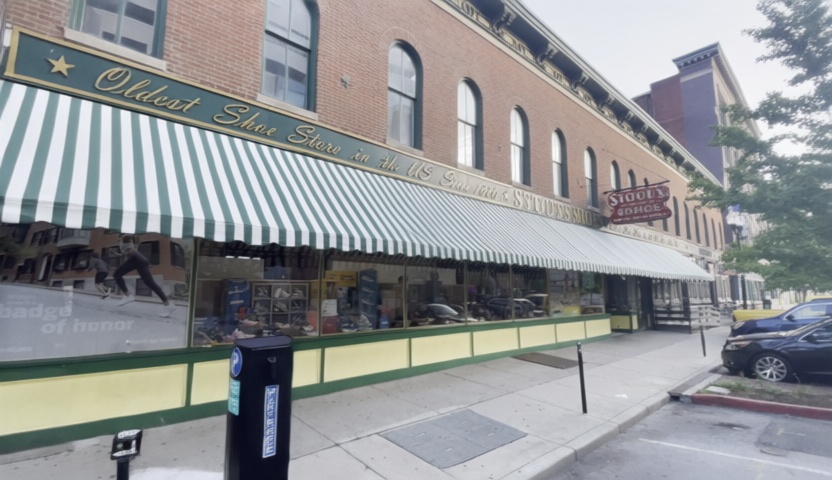}{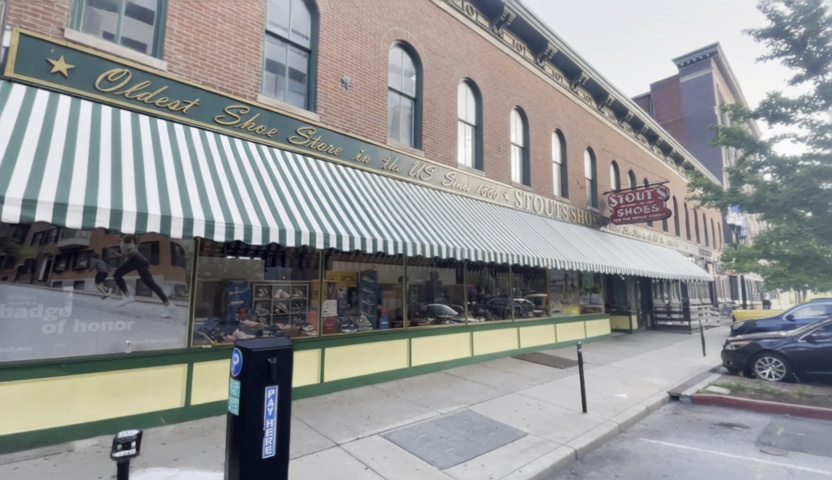}

  \multicolumn{5}{c}{\vspace{-1.5em}} \\
  \SetSceneZoom{(-0.14,0.25)}{3}{1.0cm}{(-0.14,0.25)}\FiveCols[north east]{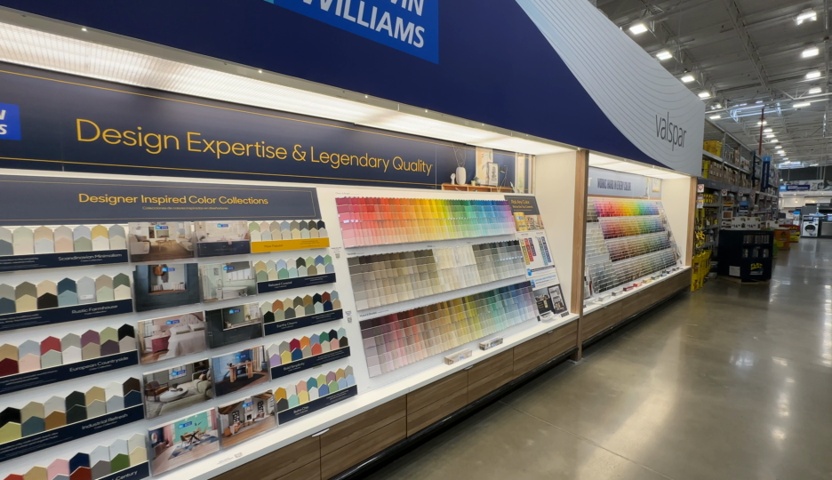}{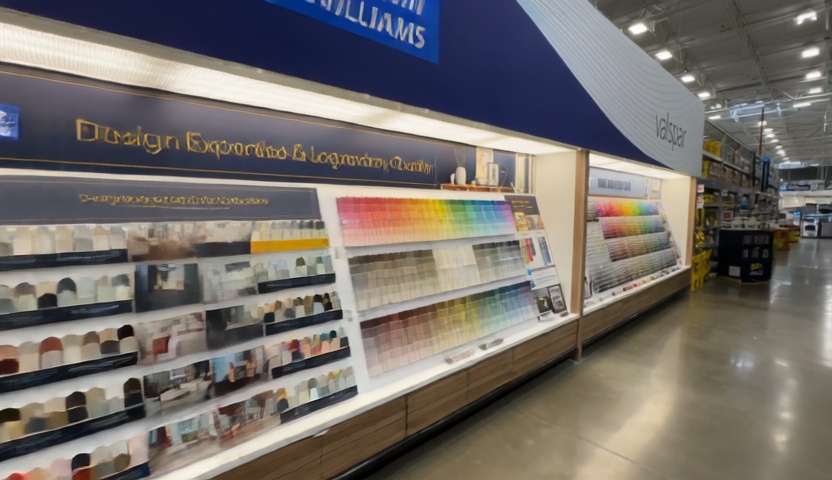}{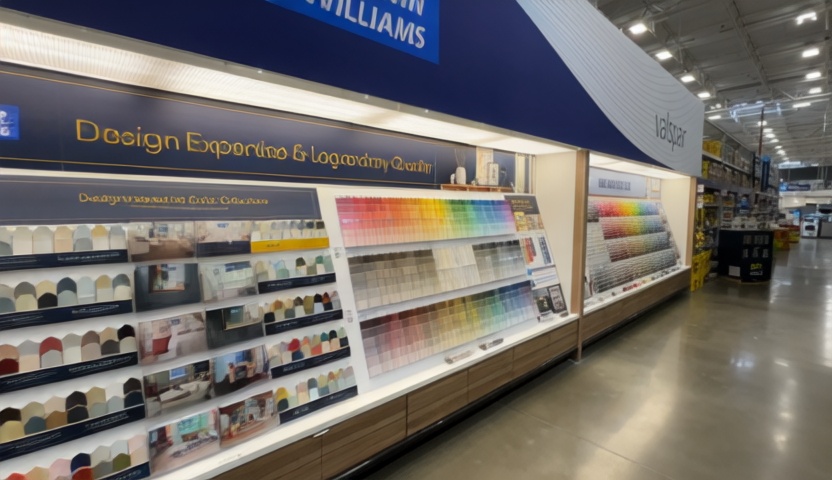}{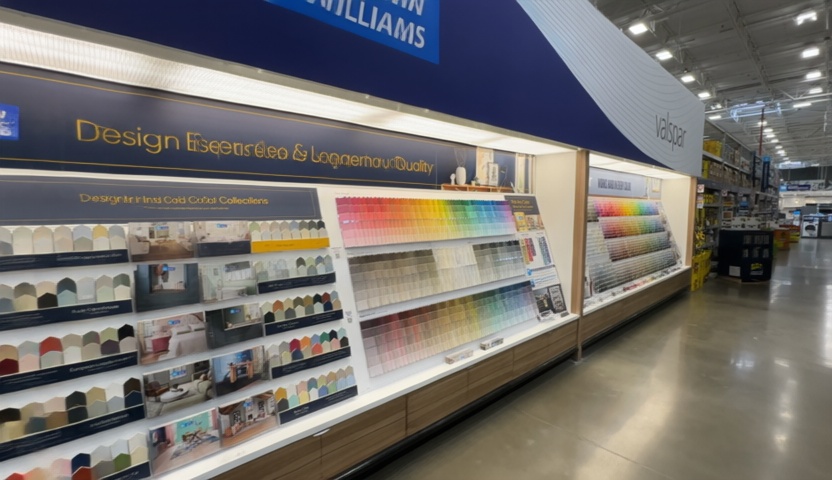}{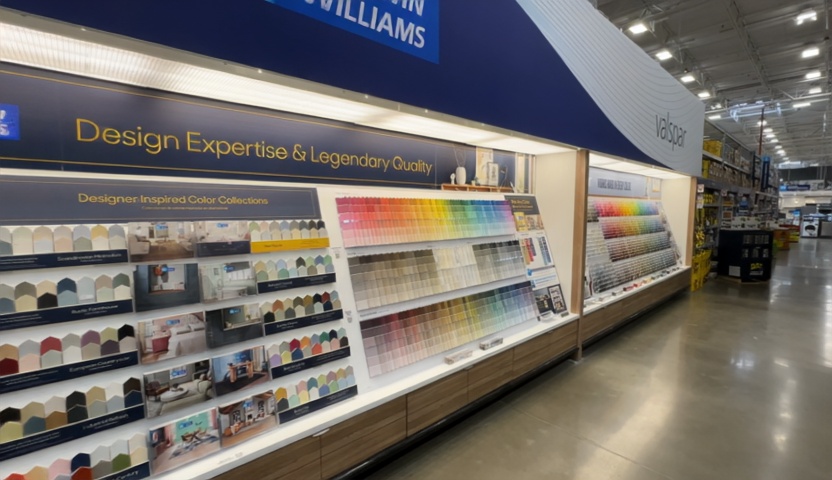}
  
  \multicolumn{5}{c}{\vspace{-1.5em}} \\
  \SetSceneZoom{(-0.6,0.25)}{5}{1.0cm}{(-0.6,0.25)}\FiveCols[north east]{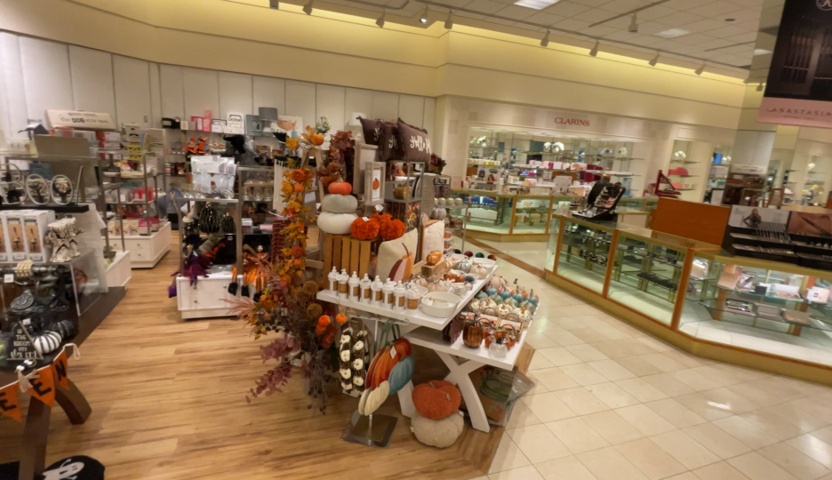}{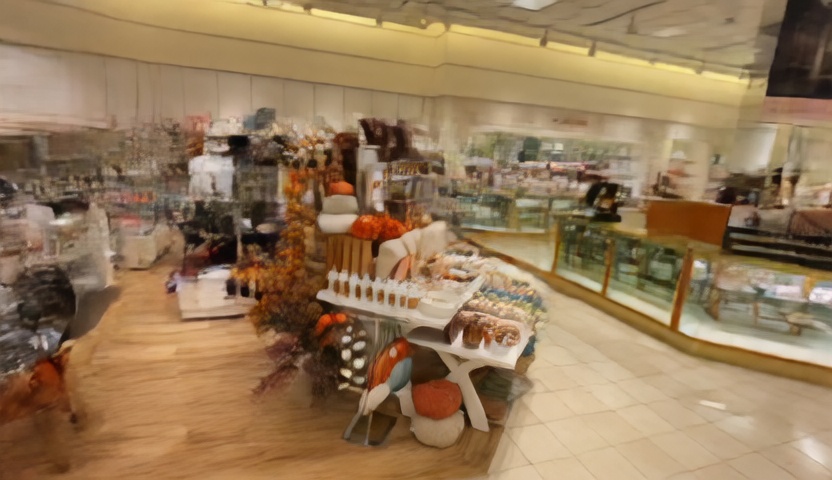}{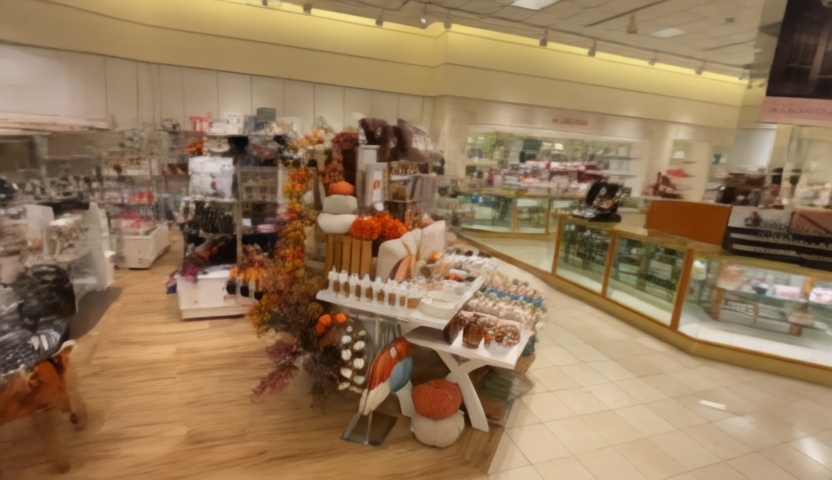}{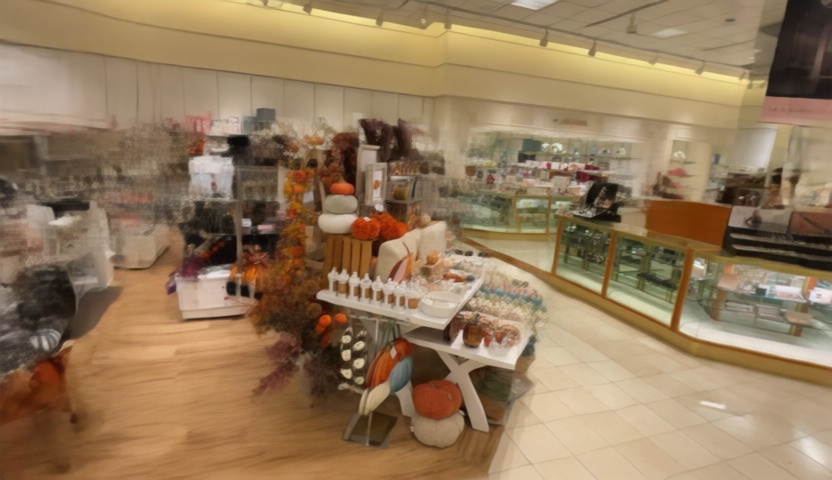}{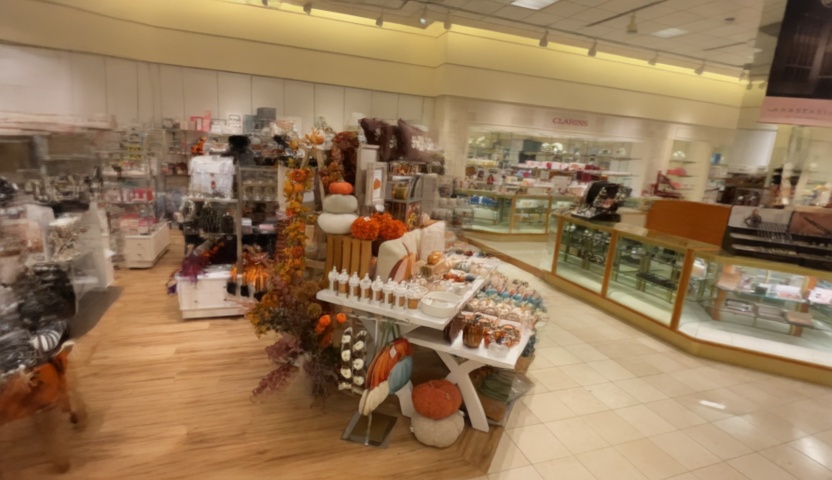}
  
  \multicolumn{5}{c}{\vspace{-1.5em}} \\
  \SetSceneZoom{(-0.15,0.3)}{4}{1.0cm}{(-0.15,0.3)}\FiveCols[north east]{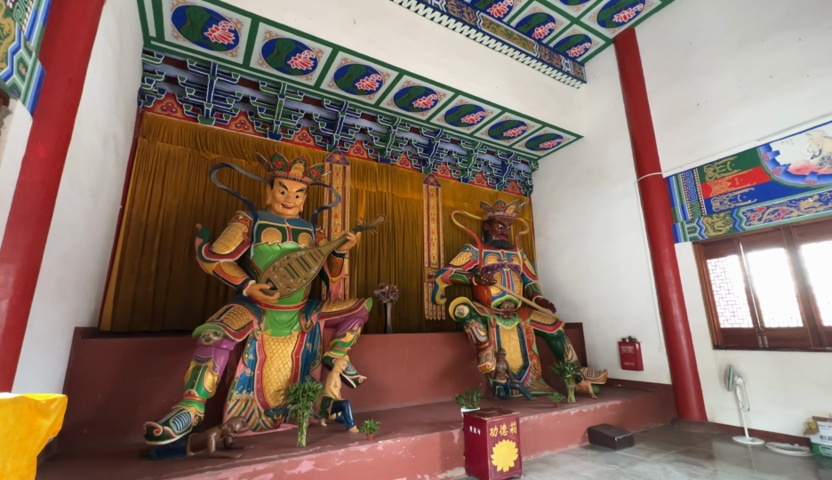}{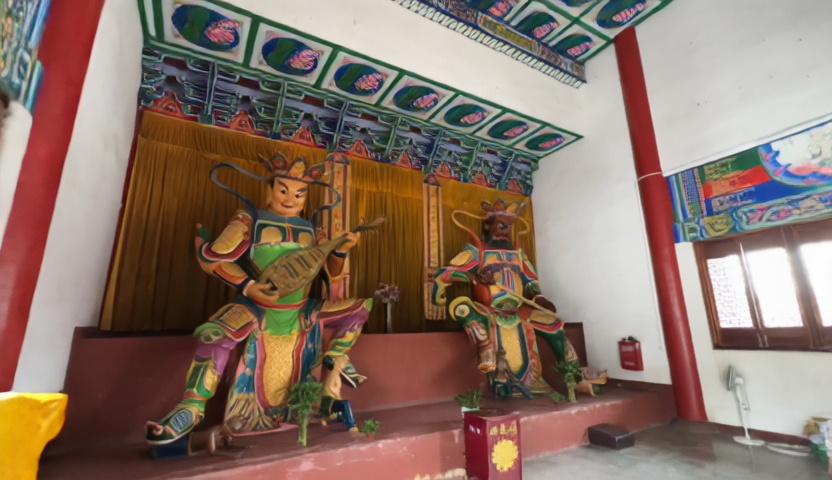}{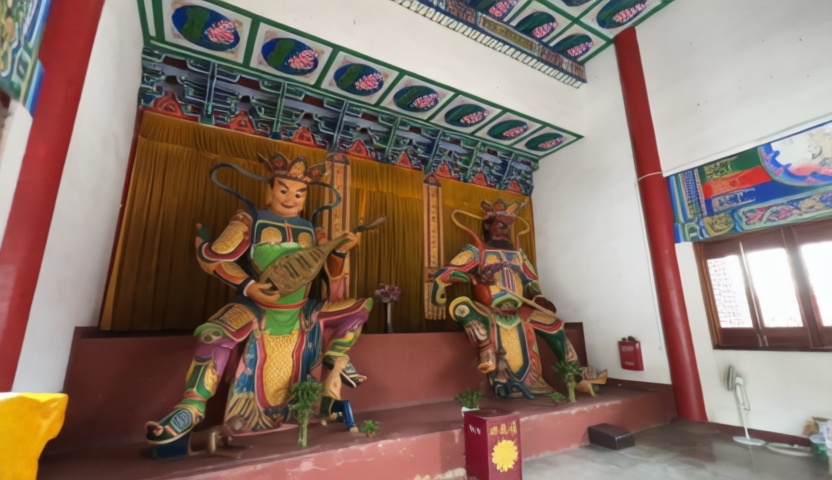}{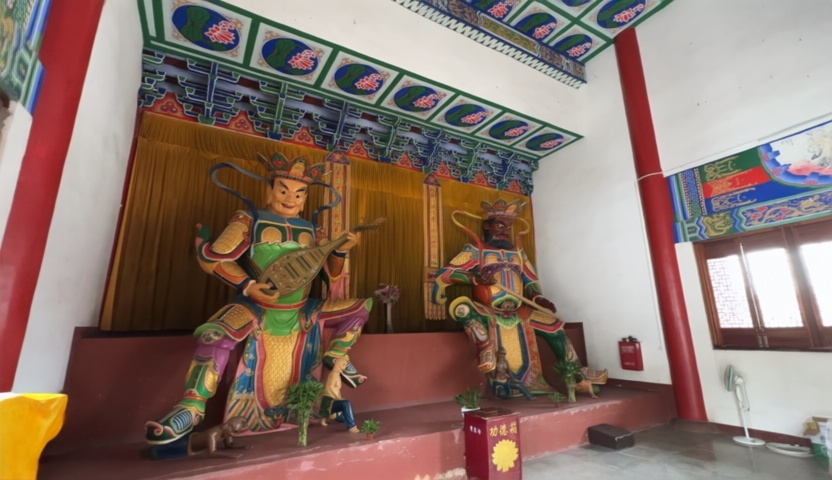}{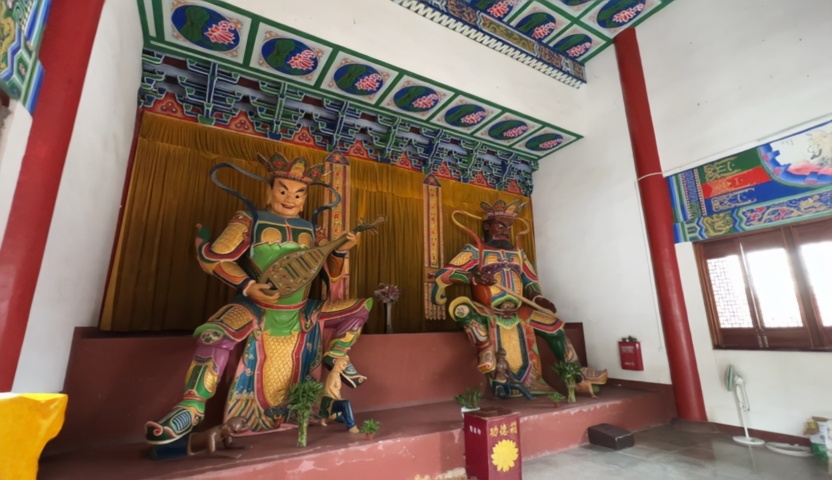}
  {GT} & {W/O fixer} & {DIFIX3D+} & {{MaRINeR}} & {\textbf{Ours}} \\
    
  \multicolumn{5}{c}{\vspace{-0.2em}{Applied on Viewcrafter (explicit NVS)}} \\
  \SetSceneZoom{(-0.7,0.25)}{4}{1.0cm}{(-0.7,0.25)}\FiveCols[north east]{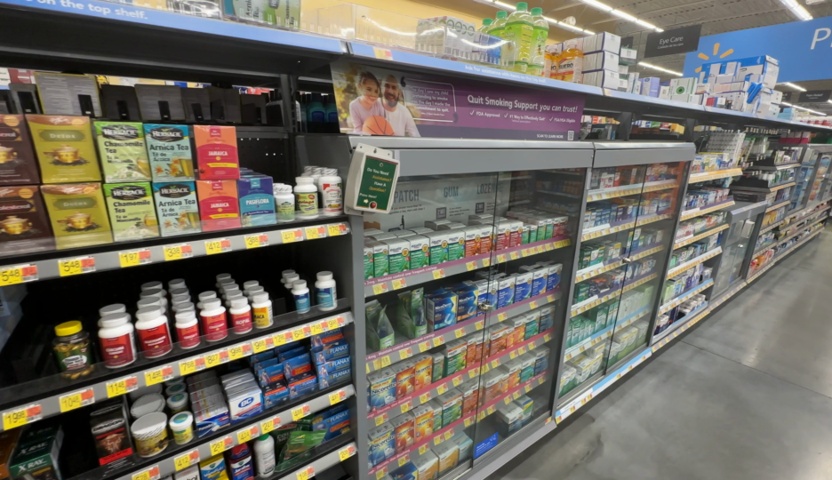}{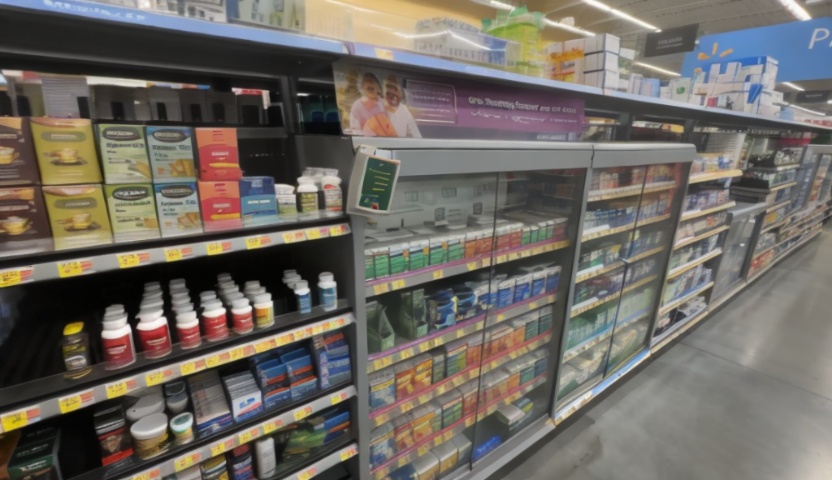}{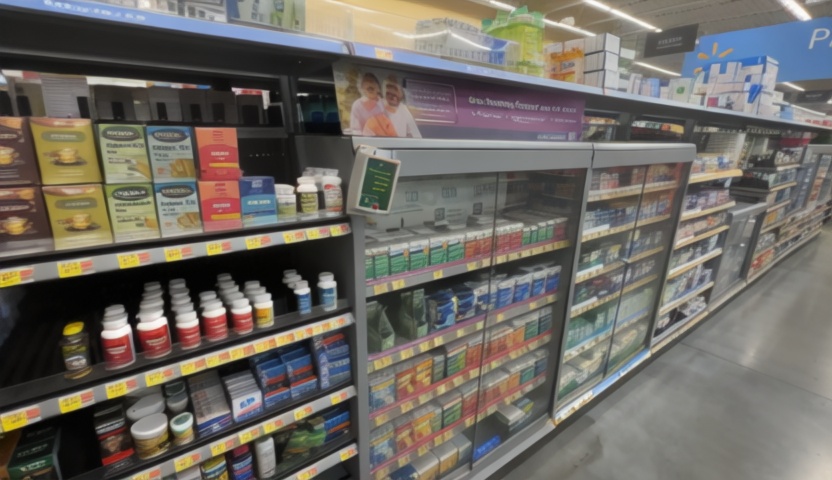}{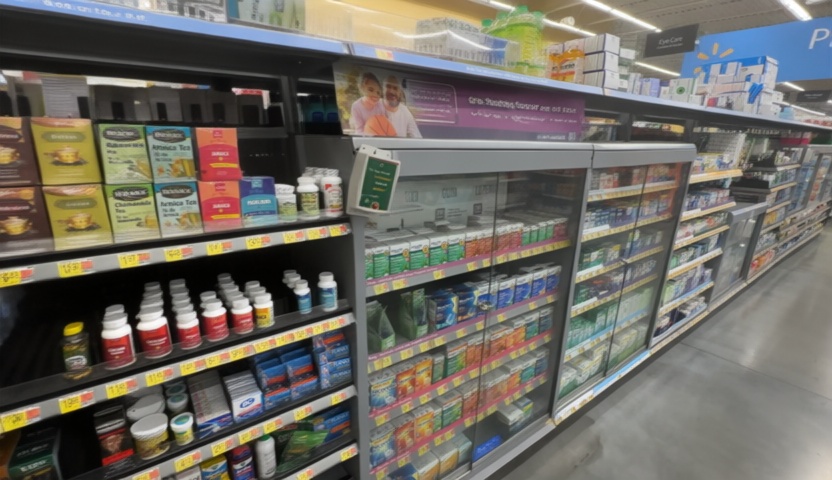}{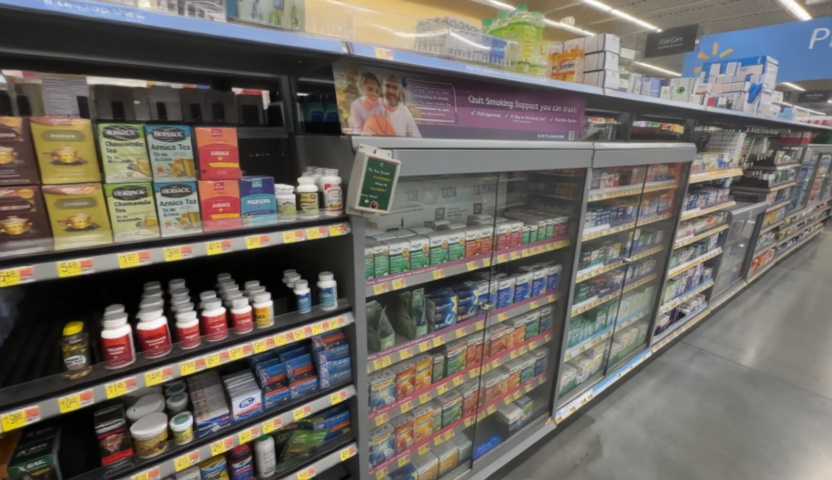}
  
  \multicolumn{5}{c}{\vspace{-1.5em}} \\
  \SetSceneZoom{(0.45,0.53)}{4}{1.0cm}{(0.45,0.53)}\FiveCols[north west]{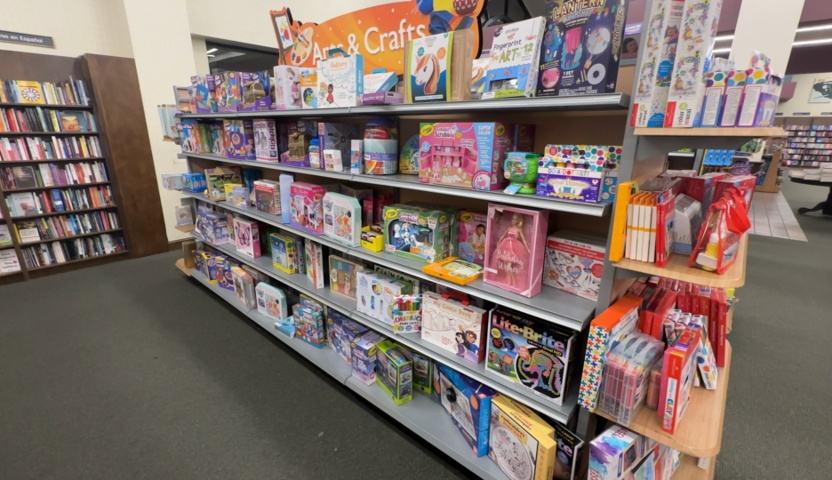}{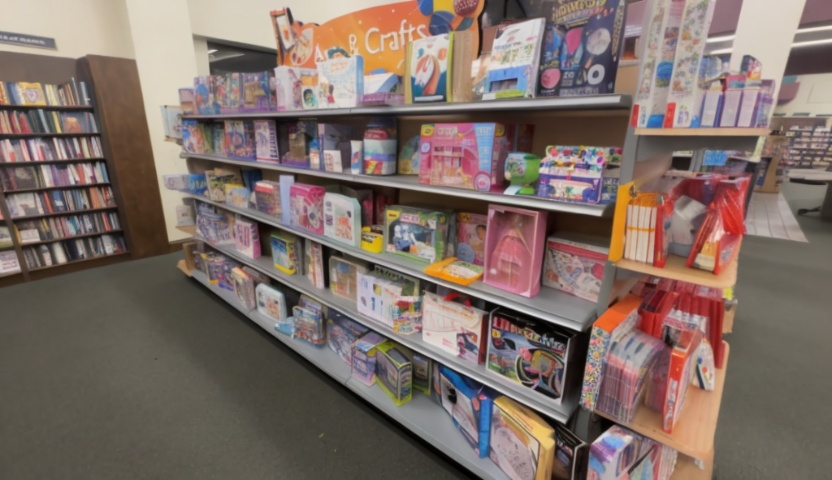}{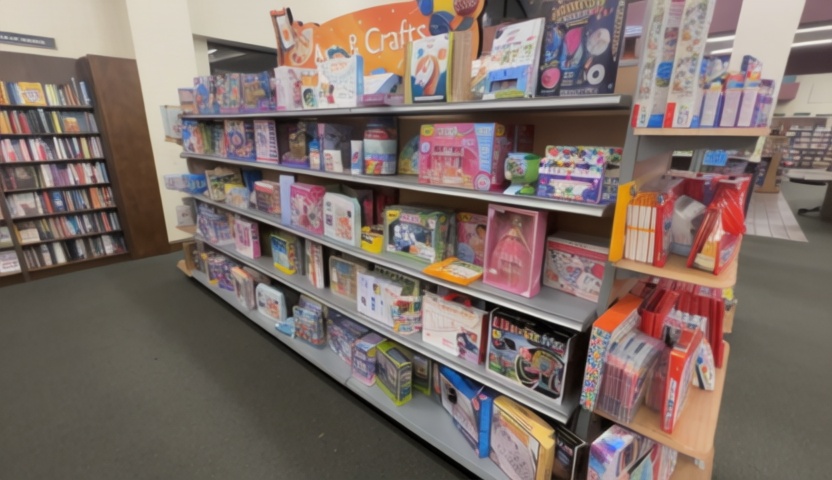}{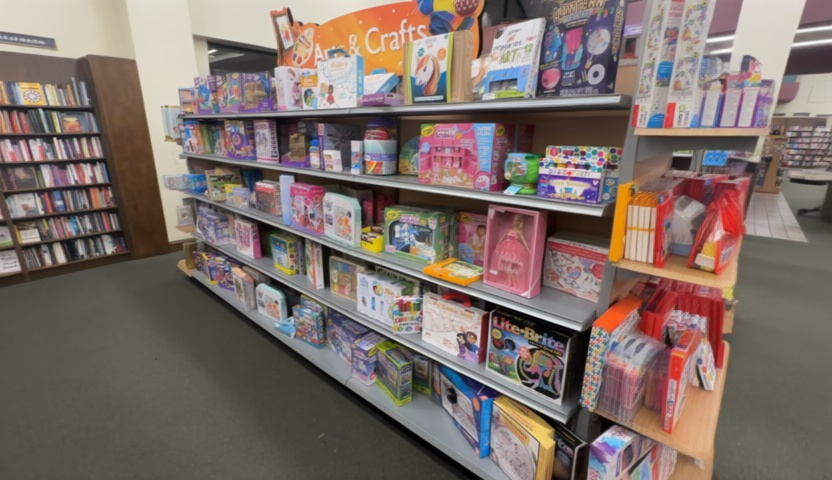}{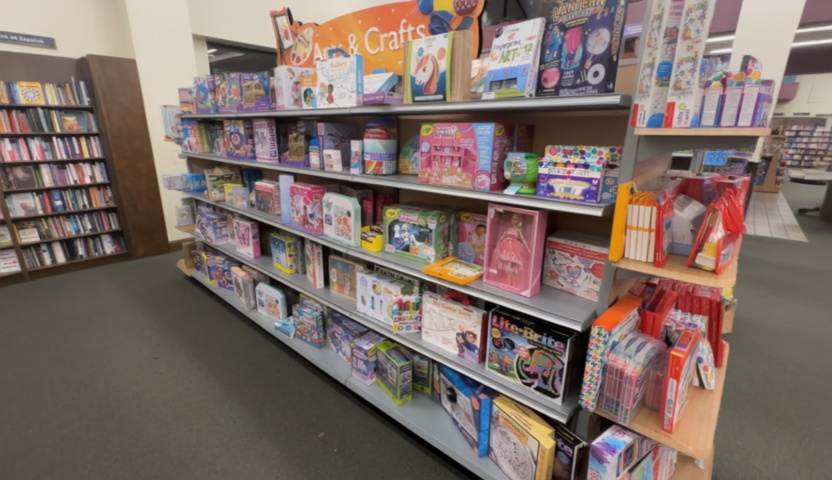}

  \multicolumn{5}{c}{\vspace{-1.5em}} \\
  \SetSceneZoom{(-0.0,-0.4)}{2}{1.0cm}{(-0.0, -0.4)}\FiveCols[north east]{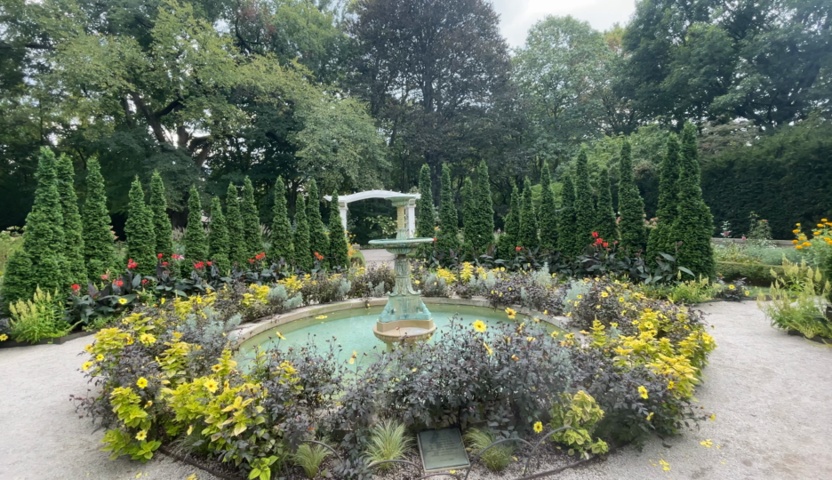}{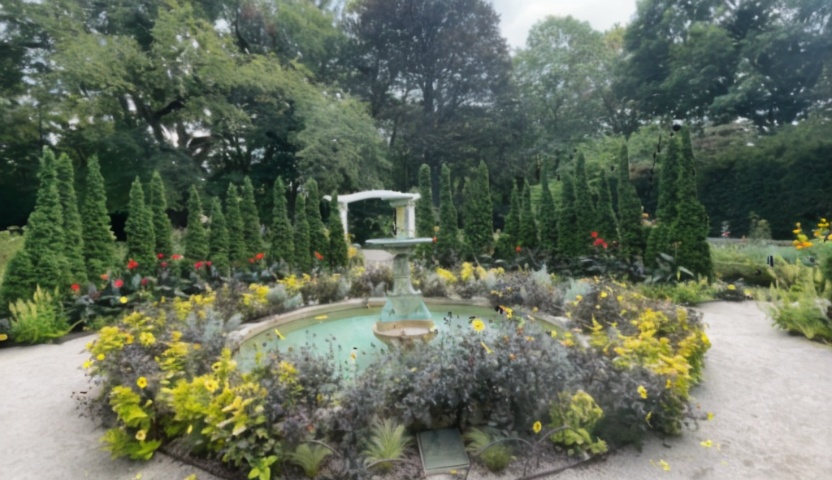}{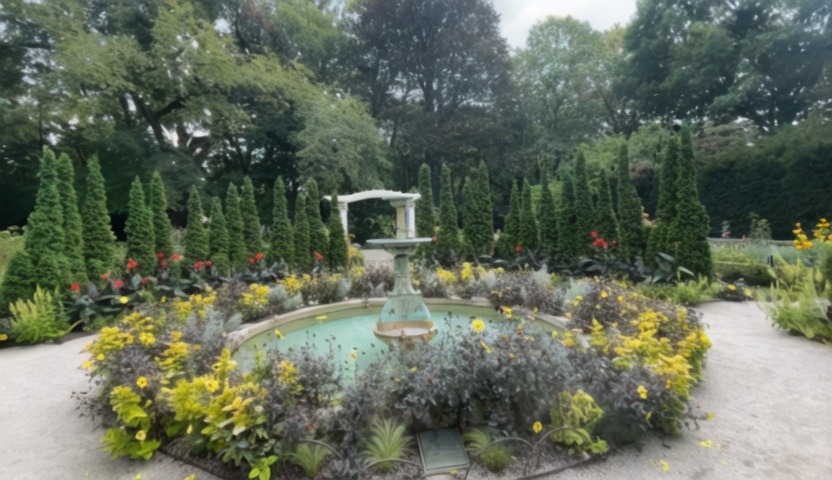}{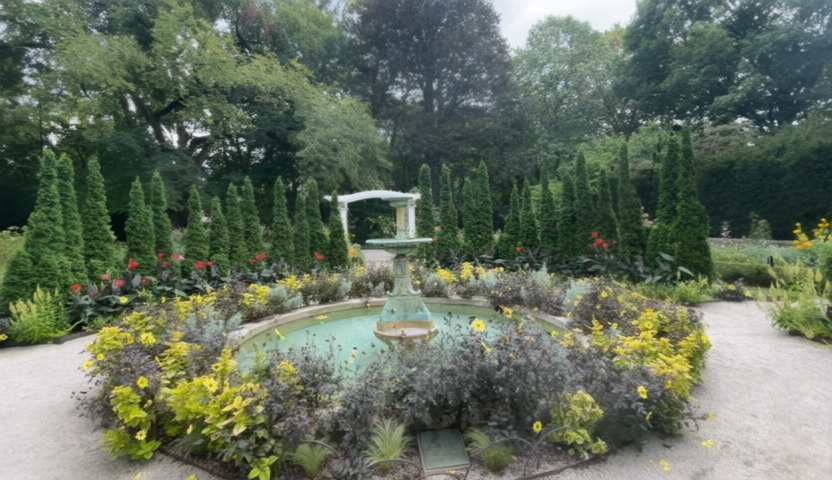}{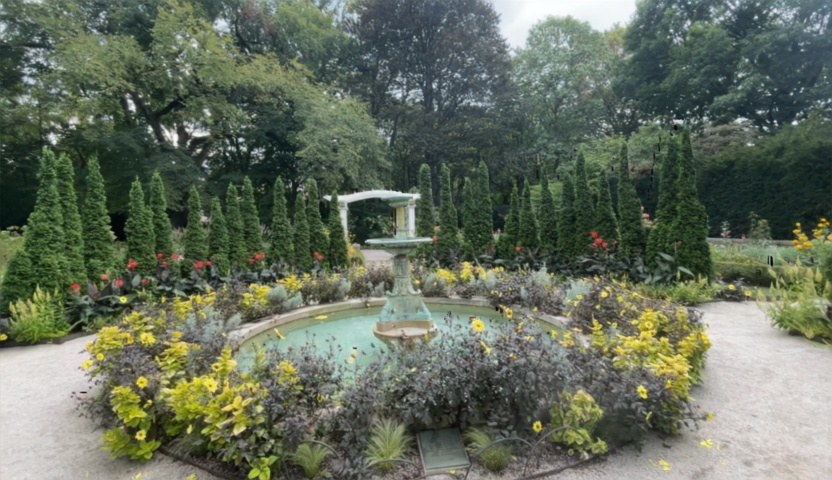}

\multicolumn{5}{c}{\vspace{-1.5em}} \\
  \SetSceneZoom{(-0.1,-0.1)}{3}{1.0cm}{(-0.1, -0.1)}\FiveCols[north east]{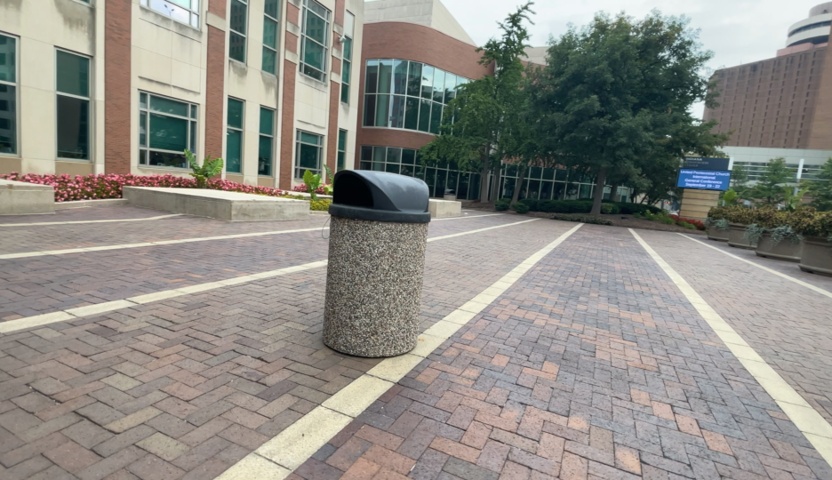}{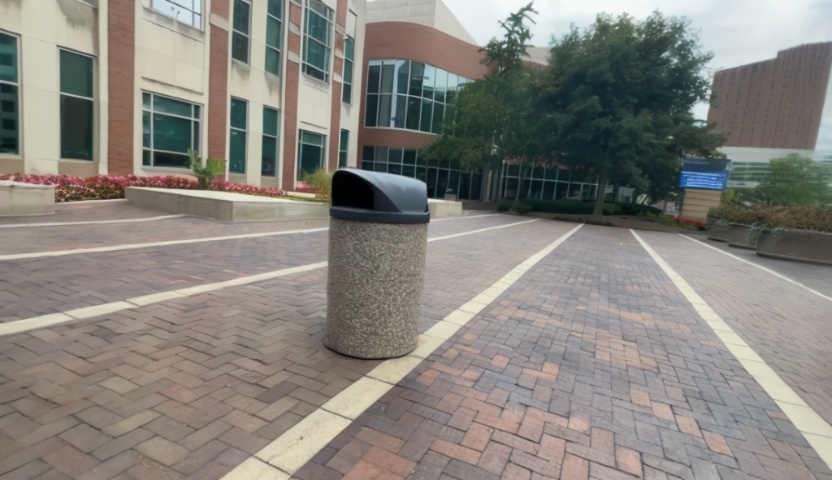}{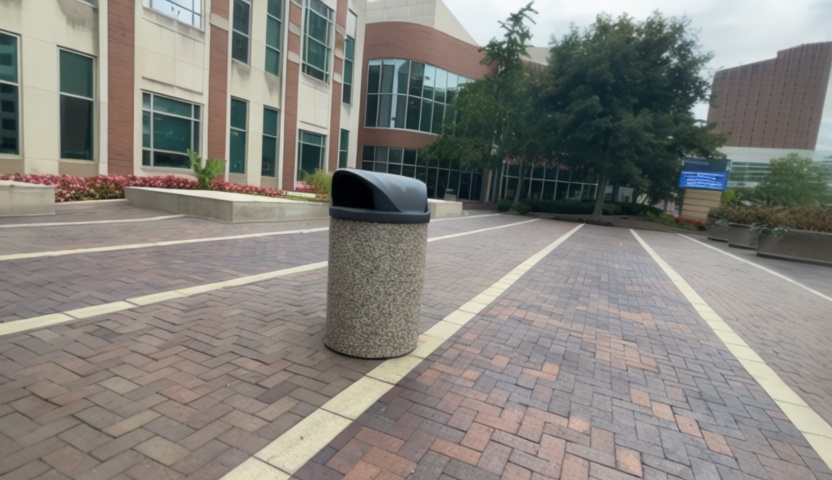}{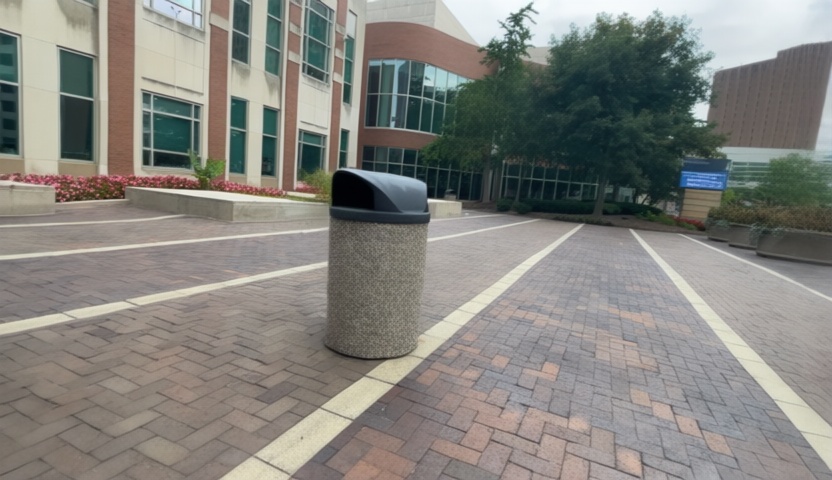}{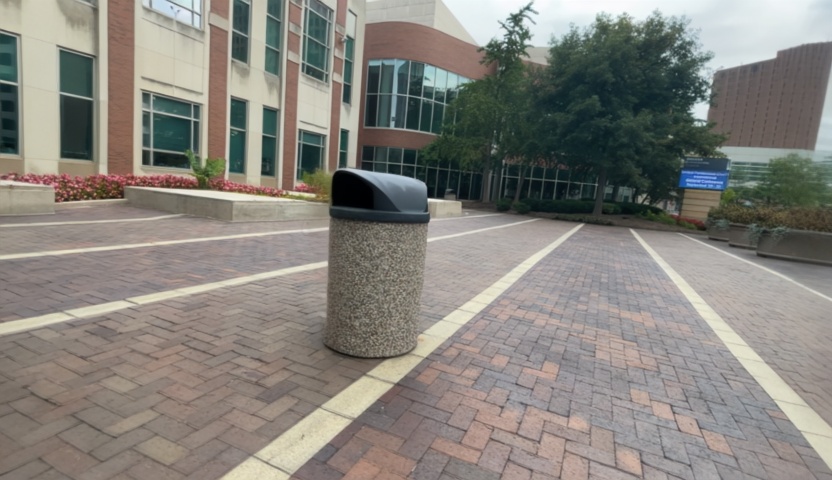}
  {GT} & {W/O fixer} & {DIFIX3D+} & {{MaRINeR}} & {\textbf{Ours}} \\
    
  \multicolumn{5}{c}{\vspace{-0.2em}{Applied on Mono2Stereo (explicit SC)}} \\
  \SetSceneZoom{(-0.75,0.3)}{3}{1cm}{(-0.75,0.3)}\FiveCols[north east]{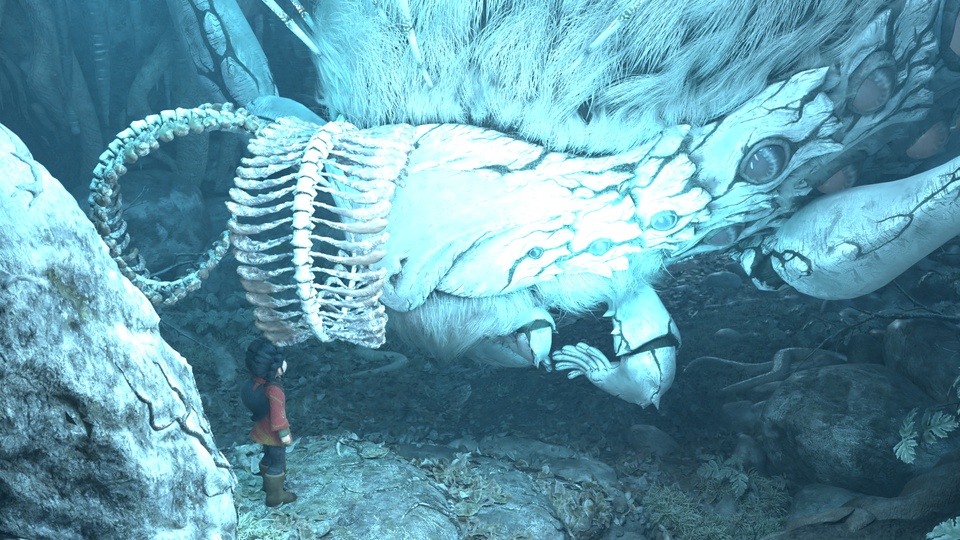}{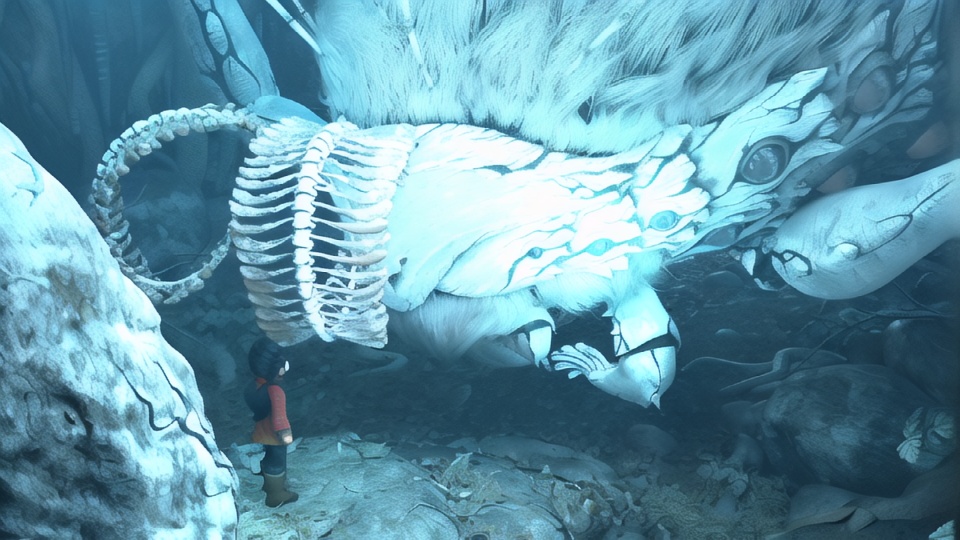}{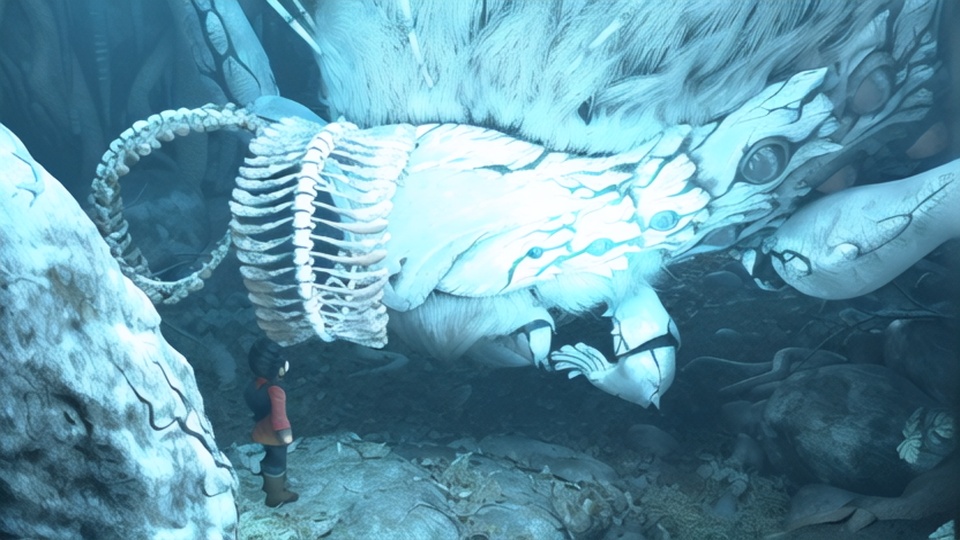}{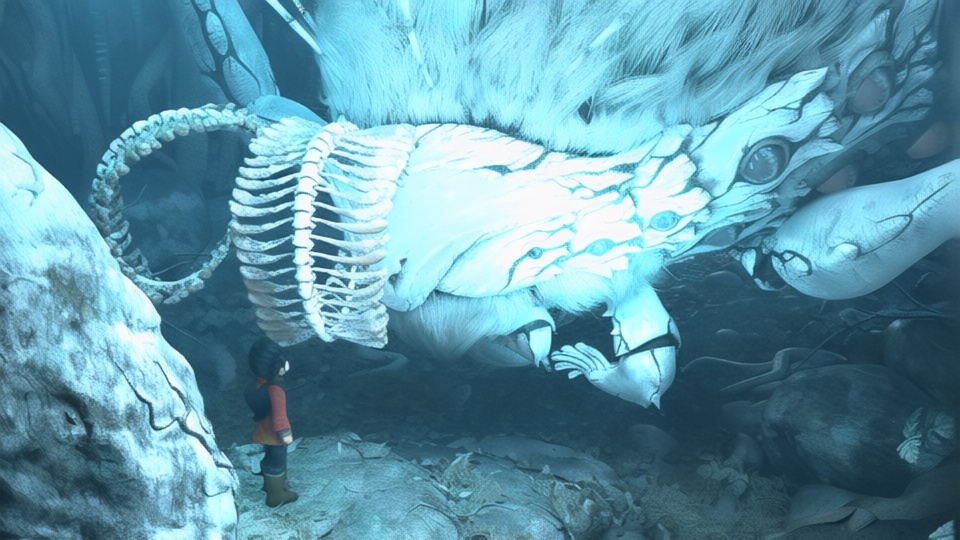}{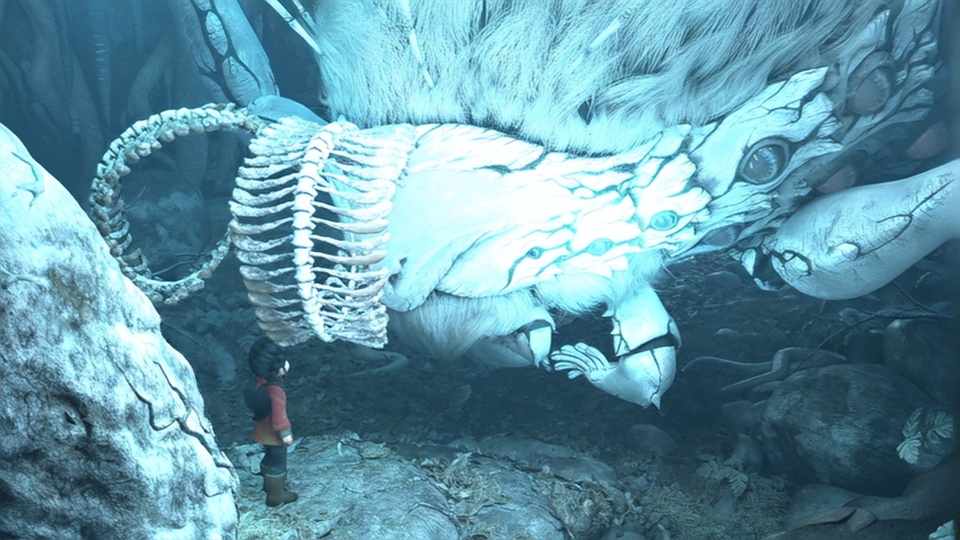}
  
  \multicolumn{5}{c}{\vspace{-1.5em}} \\
  \SetSceneZoom{(-0.4,0.2)}{2}{1cm}{(-0.4,0.2)}\FiveCols[north east]{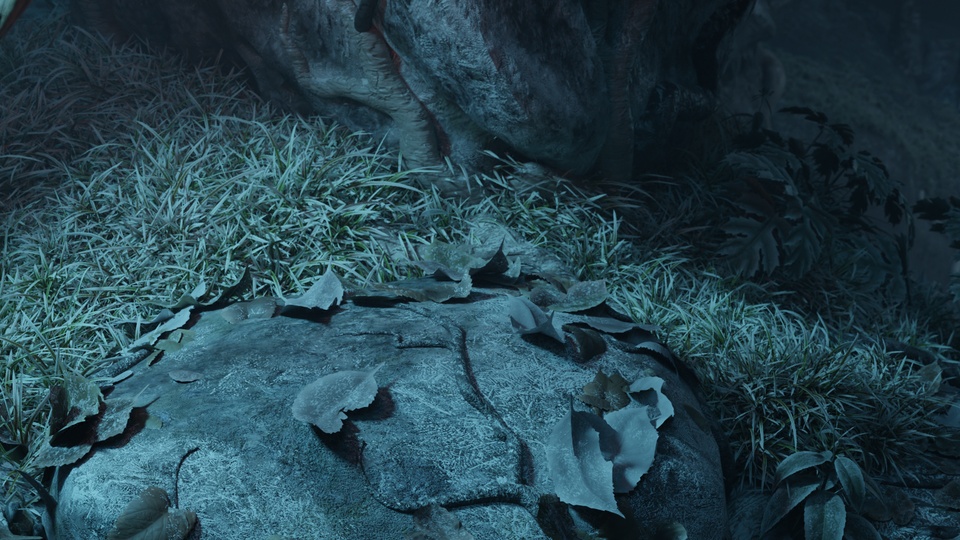}{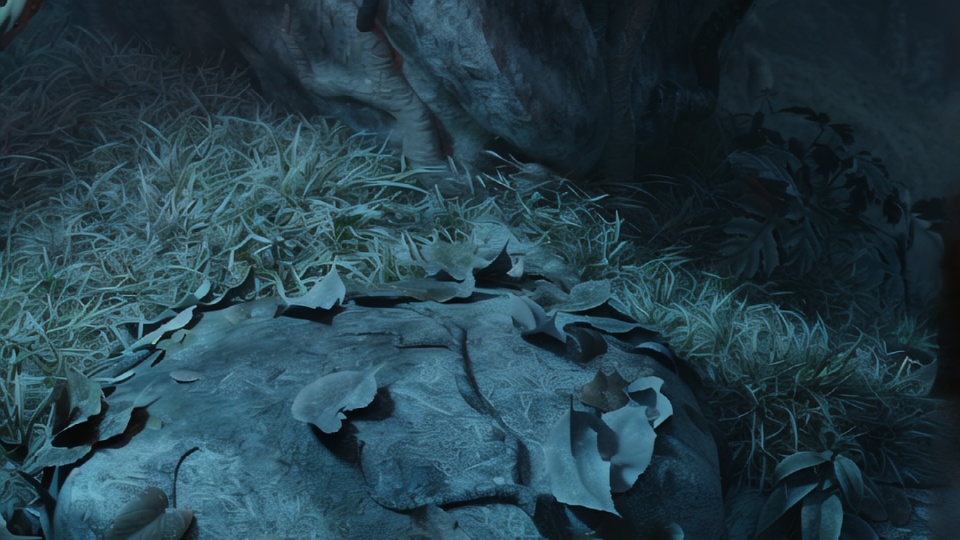}{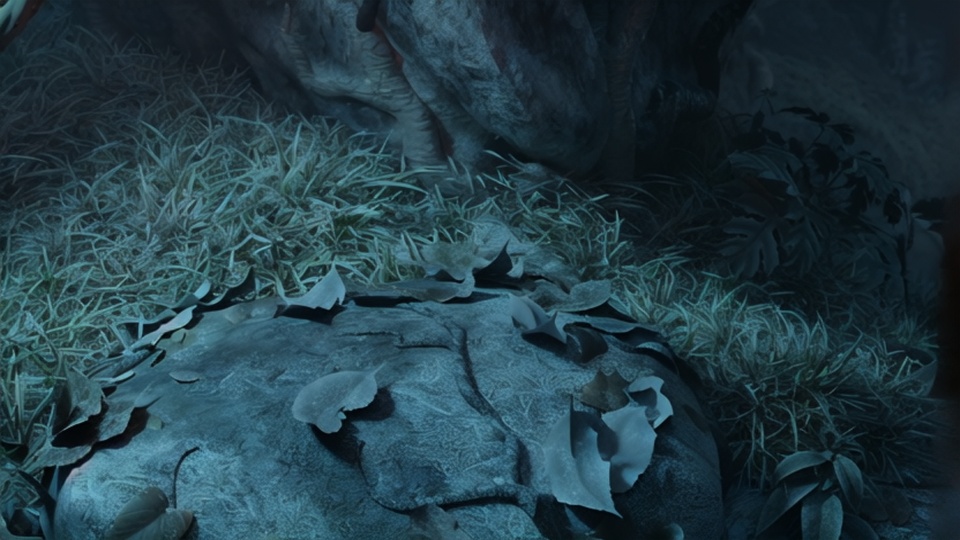}{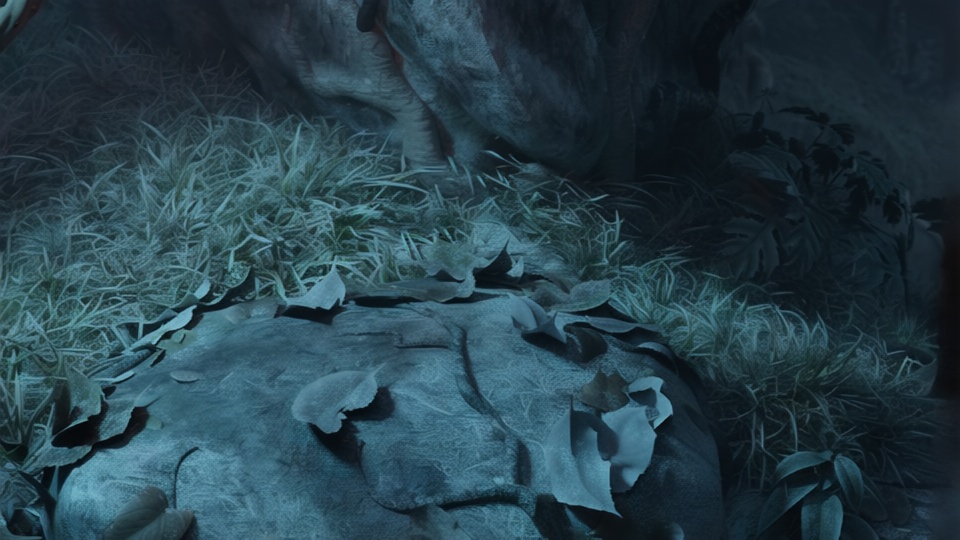}{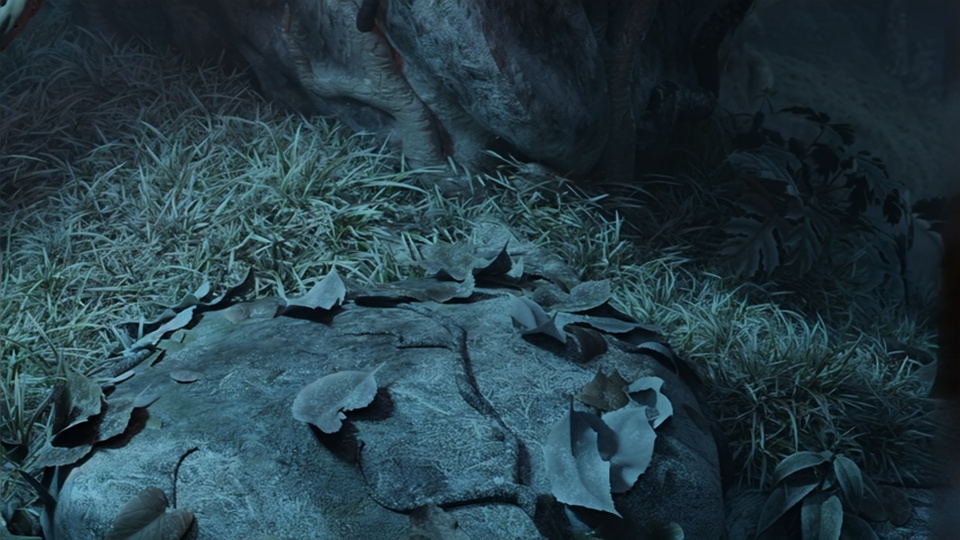}

  \multicolumn{5}{c}{\vspace{-1.5em}} \\
  \SetSceneZoom{(0.0,0.3)}{2}{1cm}{(0.0,0.3)}\FiveCols[north west]{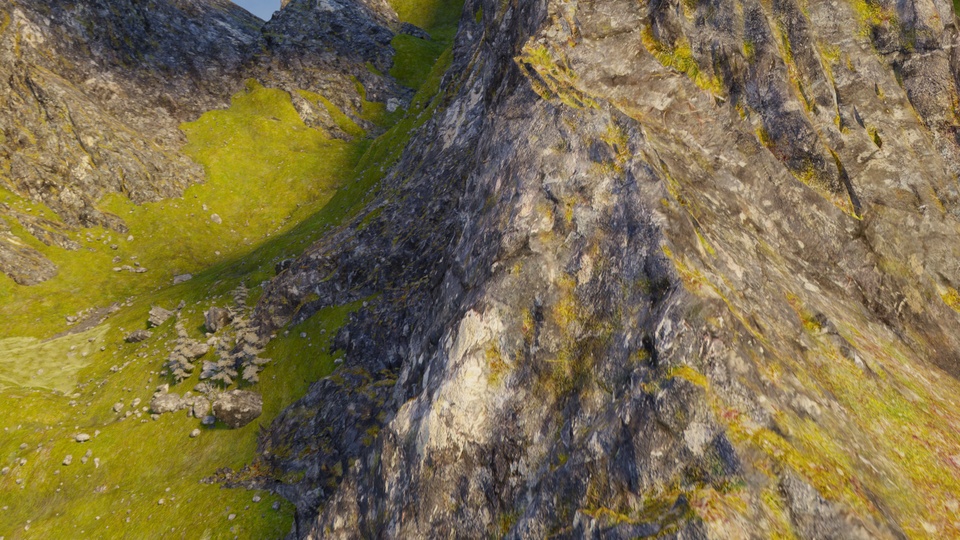}{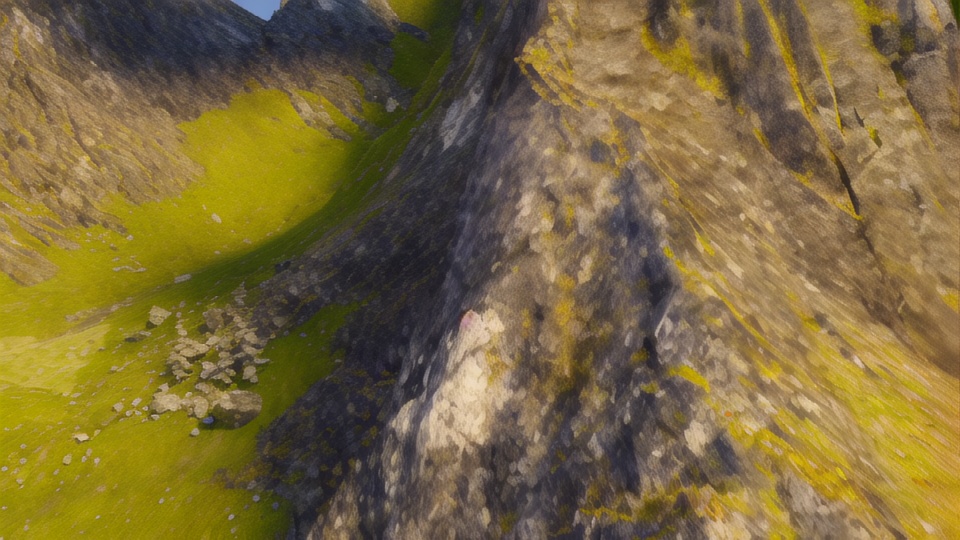}{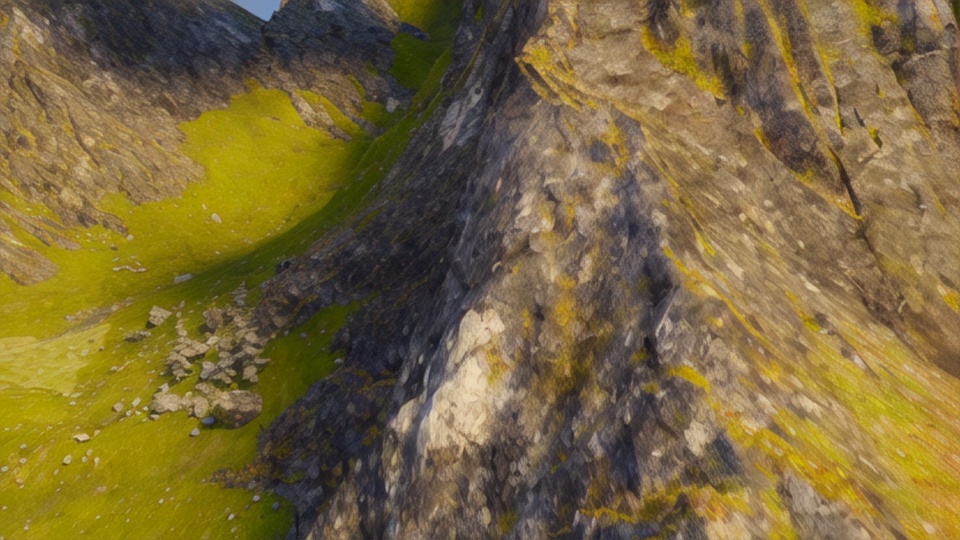}{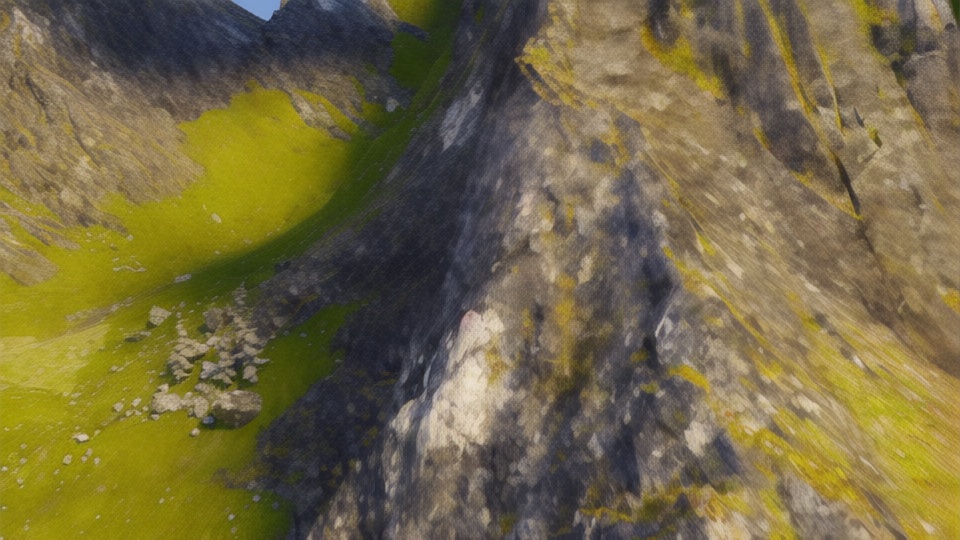}{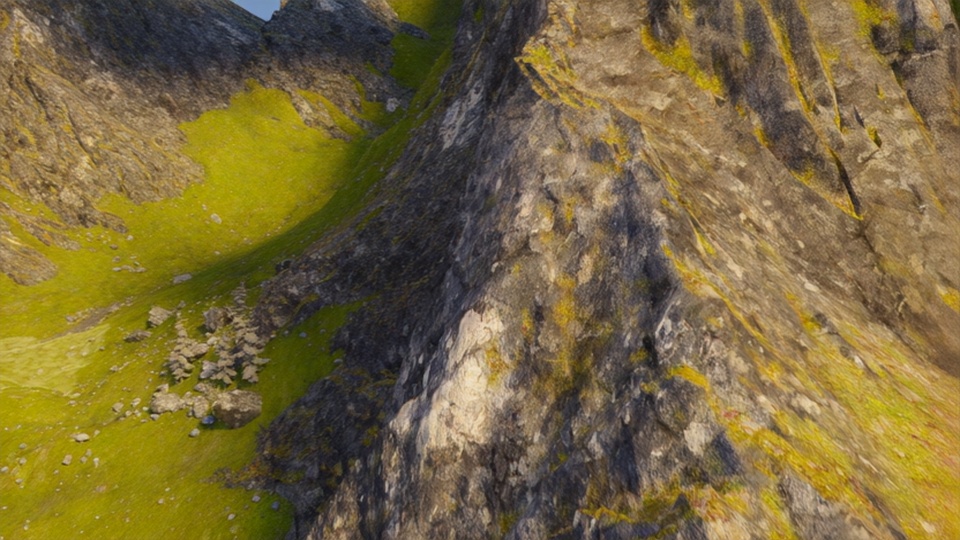}
  {GT} & {W/O fixer} & {DIFIX3D+} & {{MaRINeR}} & {\textbf{Ours}} \\
\end{tabular}

\caption{Visual results of applying plug-and-play fixers (including DIFIX3D+~\cite{wu2025difix3d+}, MaRINeR~\cite{bosiger2024mariner}, and ours) to improve diffusion-based view synthesis methods.}
\label{fig:vis_comparison}
\end{figure*}

\begin{figure*}[h]
\ContinuedFloat
\centering
\setlength{\tabcolsep}{0.6pt}
\begin{tabular}{@{}c c c c c@{}}
  \multicolumn{5}{c}{\vspace{-0.2em}{Applied on StereoCrafter (explicit SC)}} \\
  \SetSceneZoom{(-0.6,-0.3)}{2}{1cm}{(-0.6,-0.3)}\FiveCols[north east]{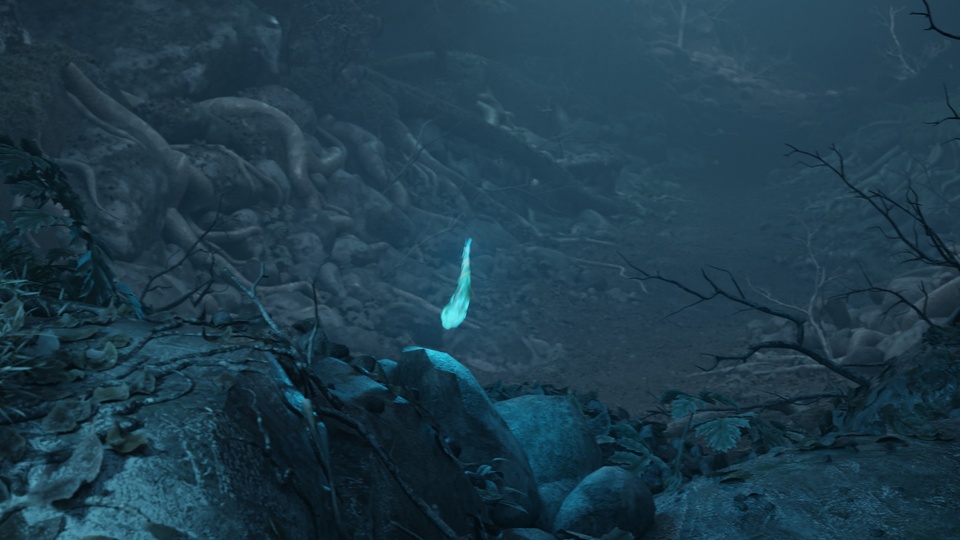}{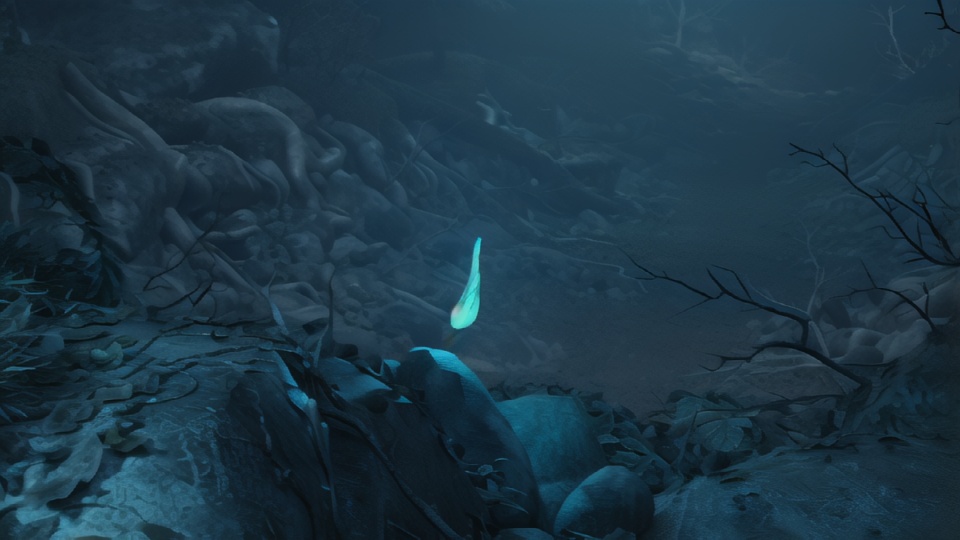}{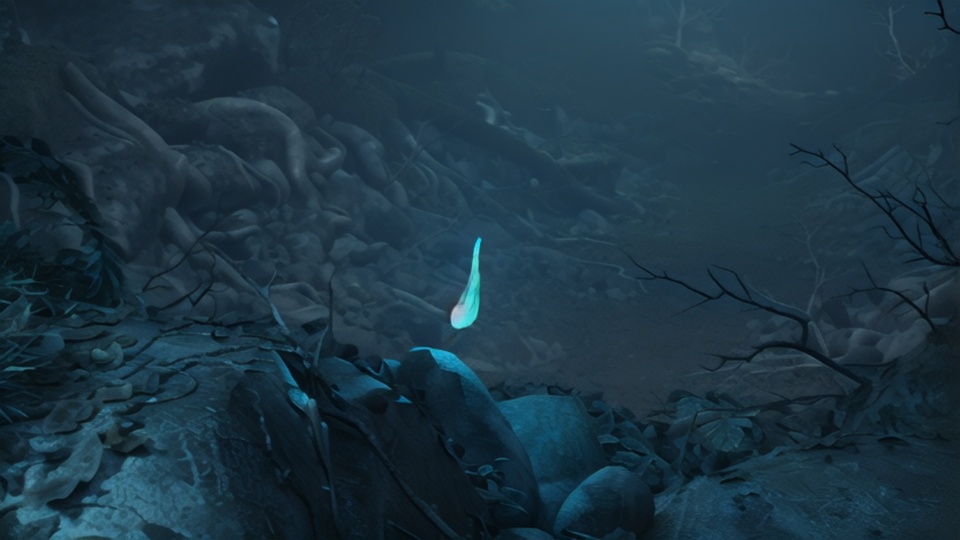}{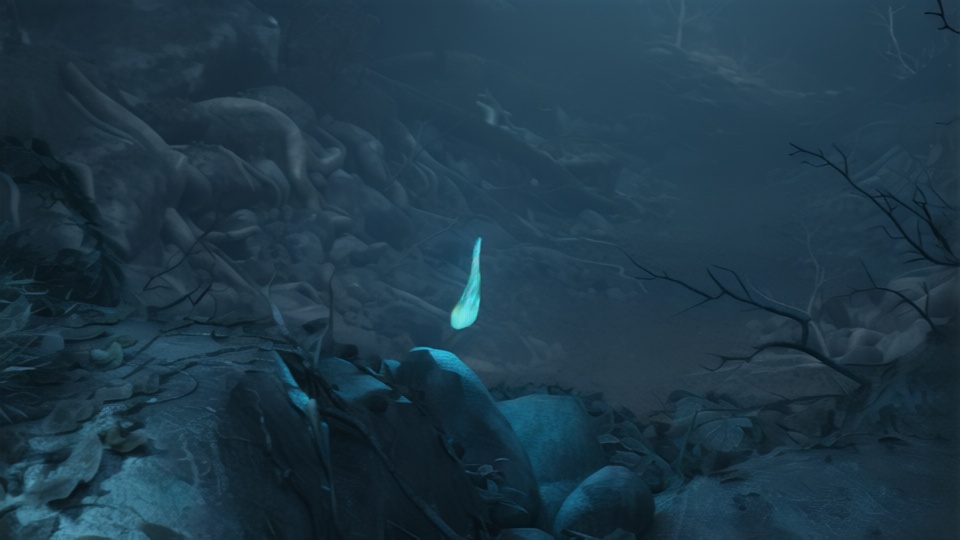}{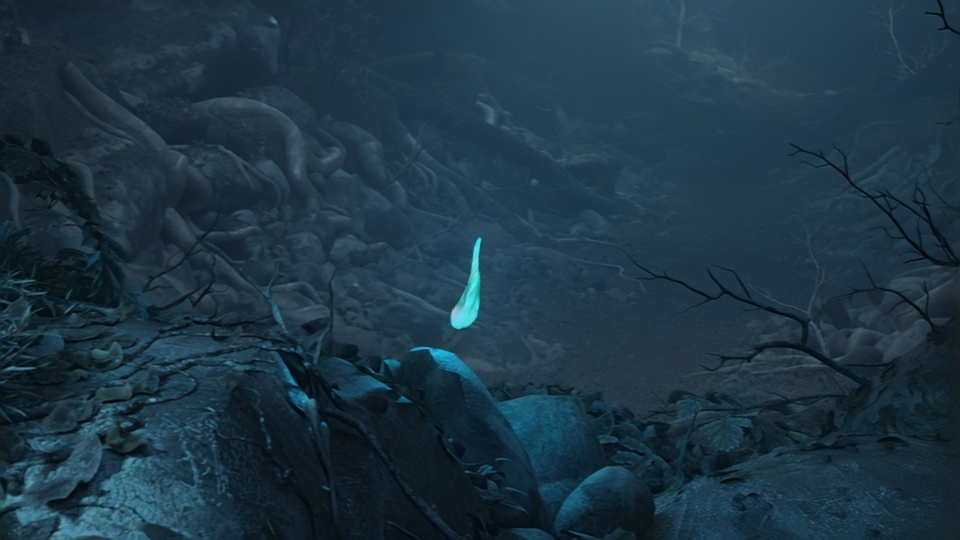}
  
  \multicolumn{5}{c}{\vspace{-1.5em}} \\
  \SetSceneZoom{(-0.4, -0.4)}{2}{1cm}{(-0.4,-0.4)}\FiveCols[north east]{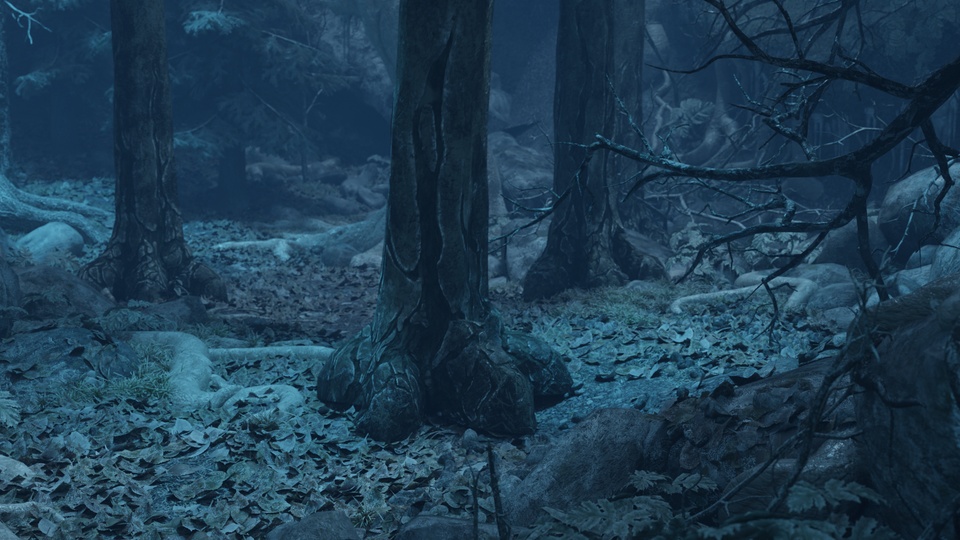}{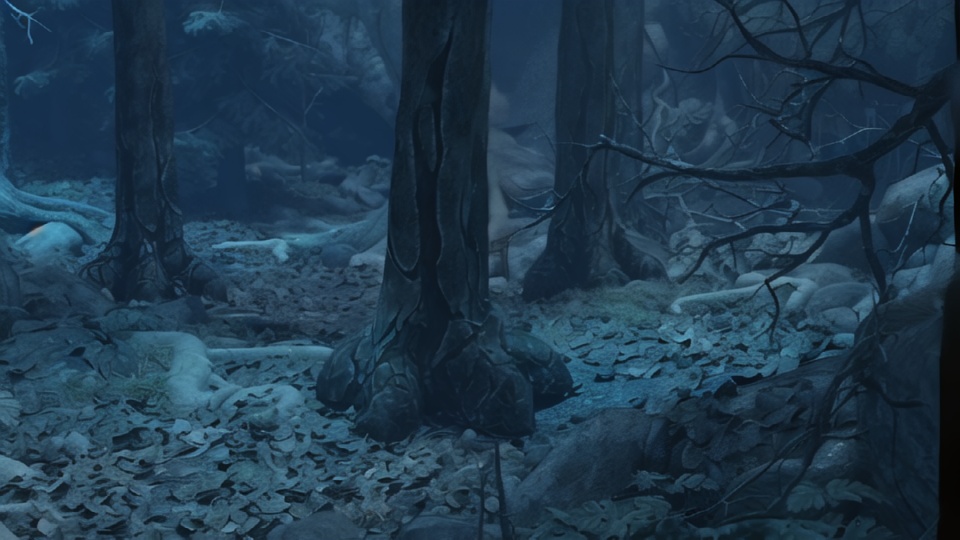}{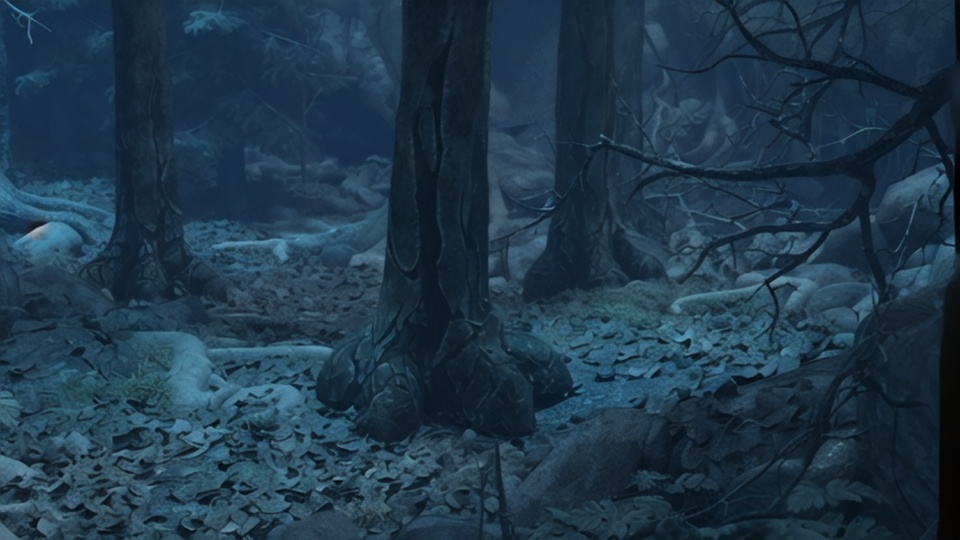}{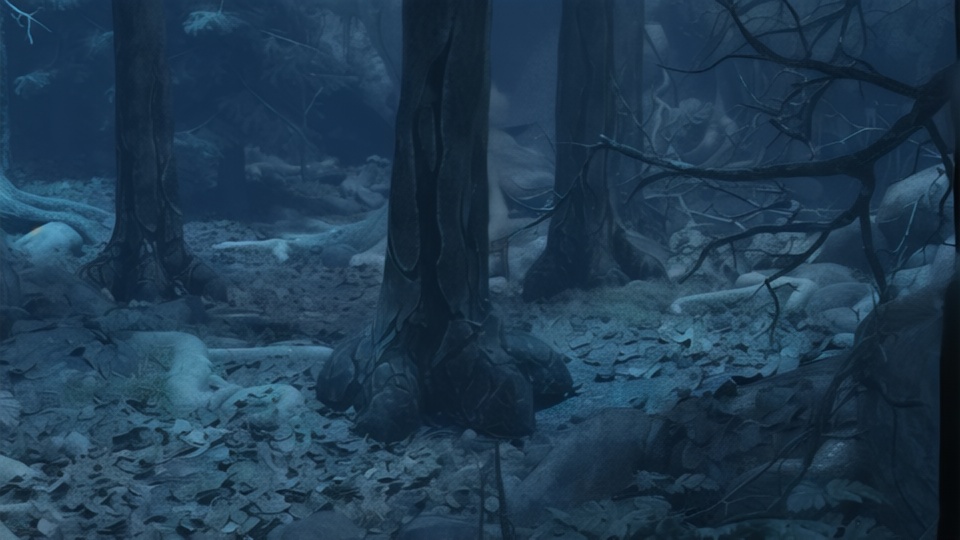}{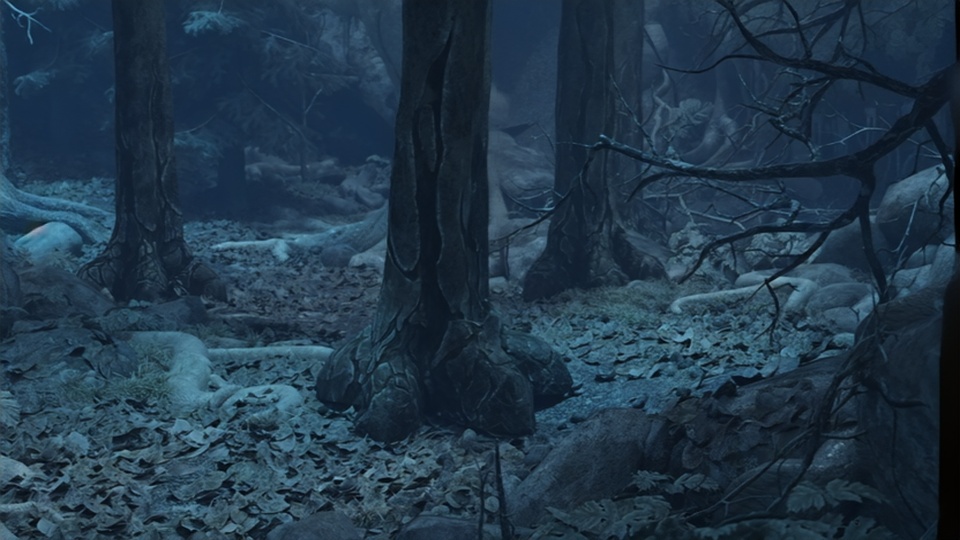}

  \multicolumn{5}{c}{\vspace{-1.5em}} \\
  \SetSceneZoom{(-0.7, -0.2)}{2}{1cm}{(-0.7,-0.2)}\FiveCols[north east]{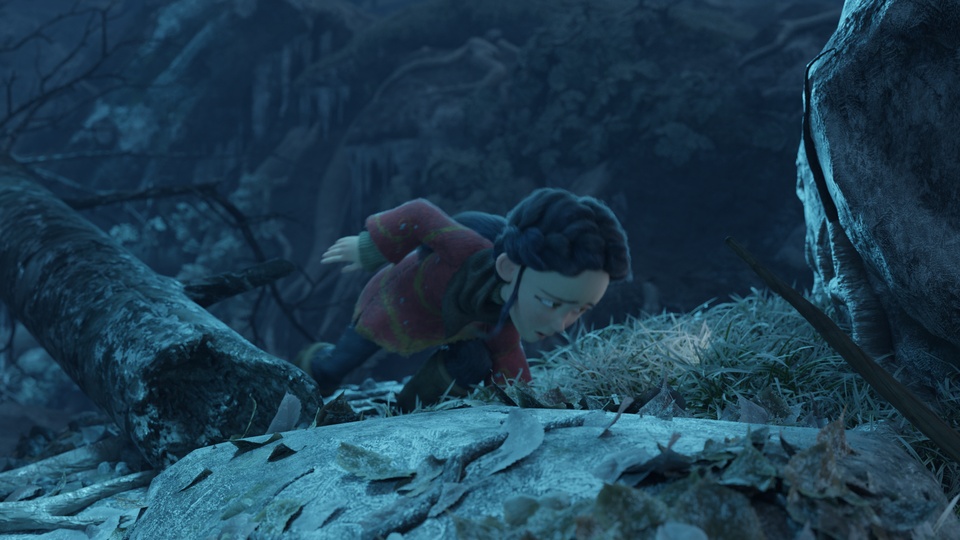}{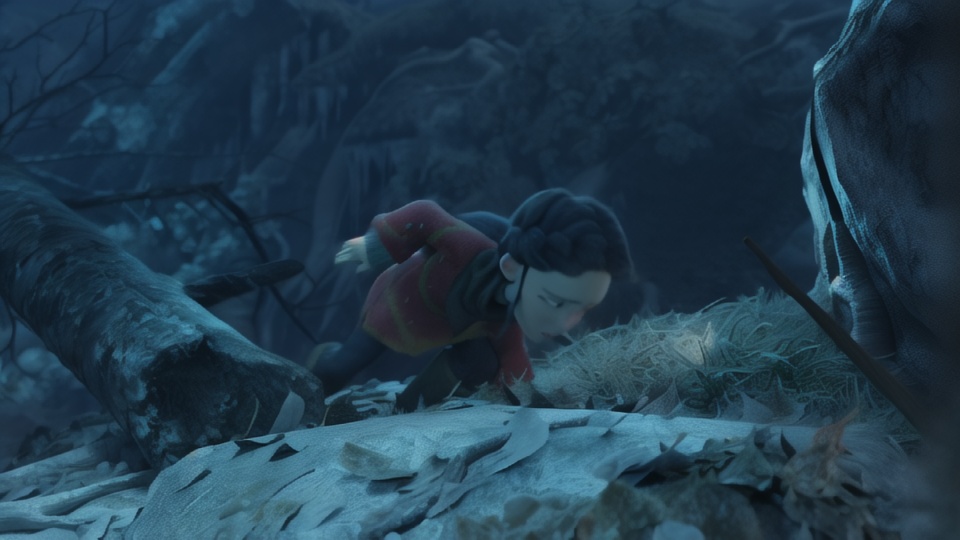}{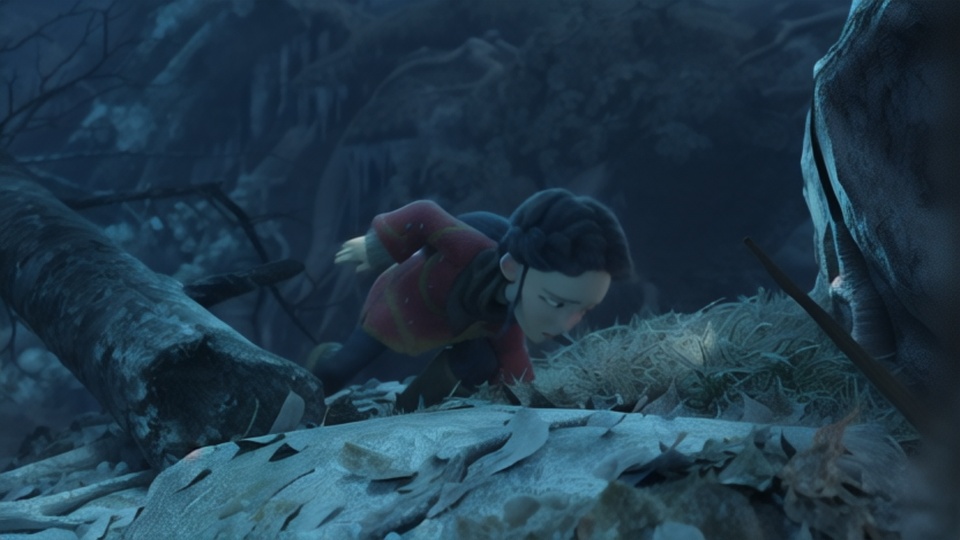}{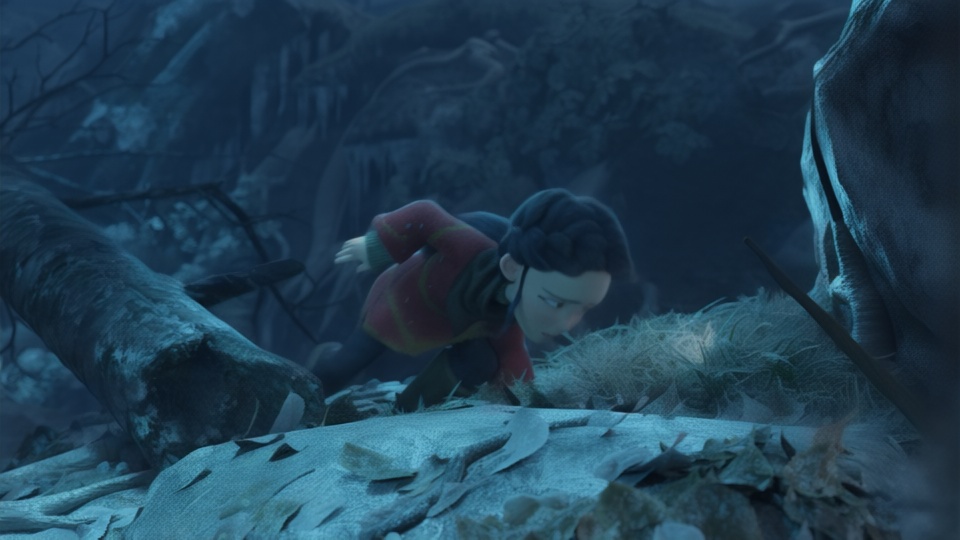}{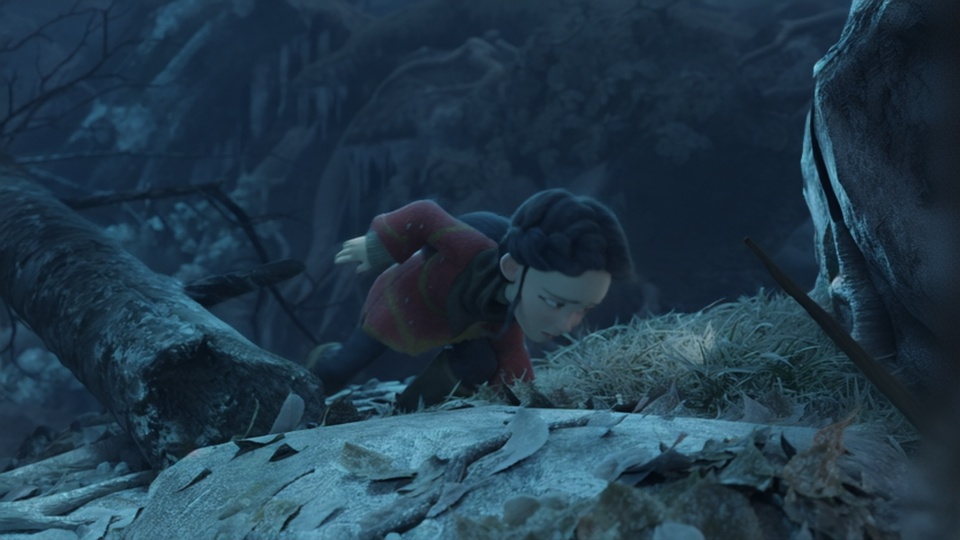}
  {GT} & {W/O fixer} & {DIFIX3D+} & {{MaRINeR}} & {\textbf{Ours}} \\
   
  \multicolumn{5}{c}{\vspace{-0.2em}{Applied on ReCamMaster (implicit NVS)}} \\
  \SetSceneZoom{(0.55,-0.5)}{3}{1cm}{(0.7,-0.5)}\FiveCols[north west]{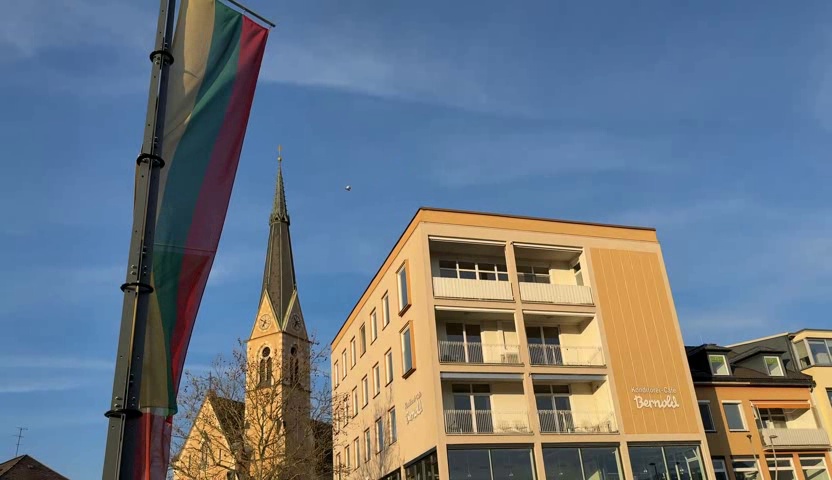}{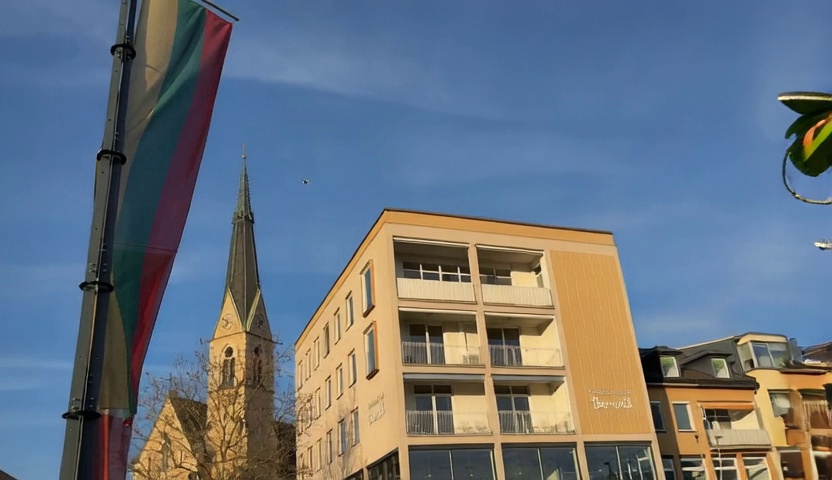}{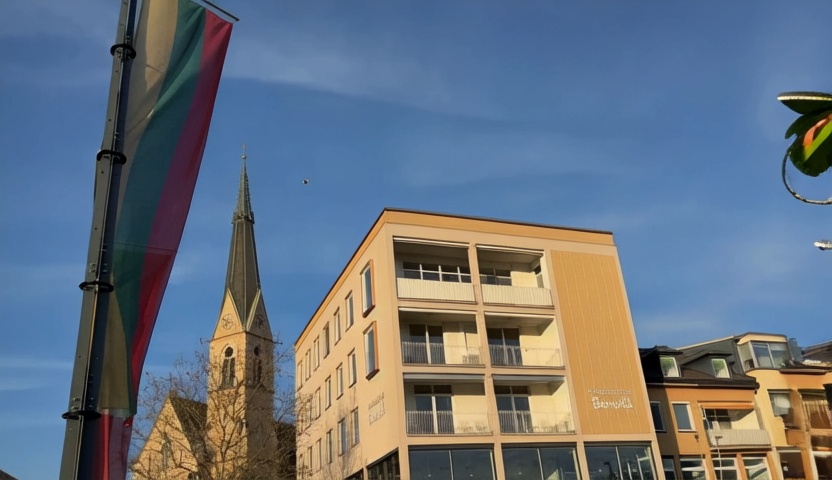}{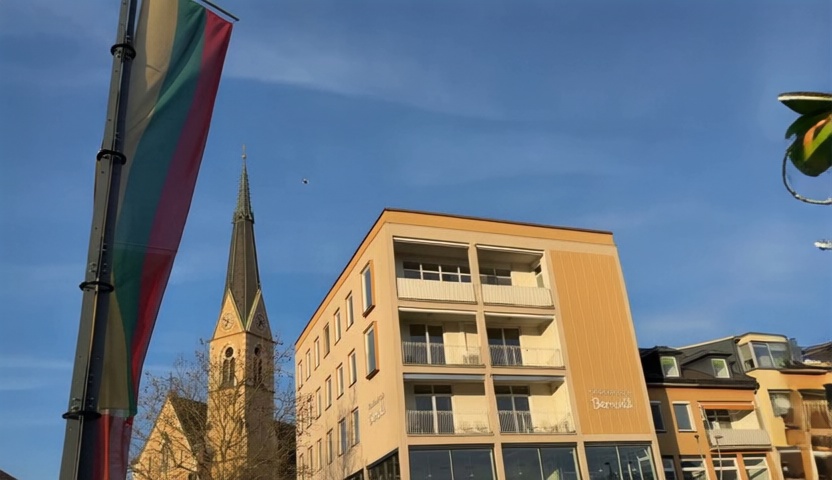}{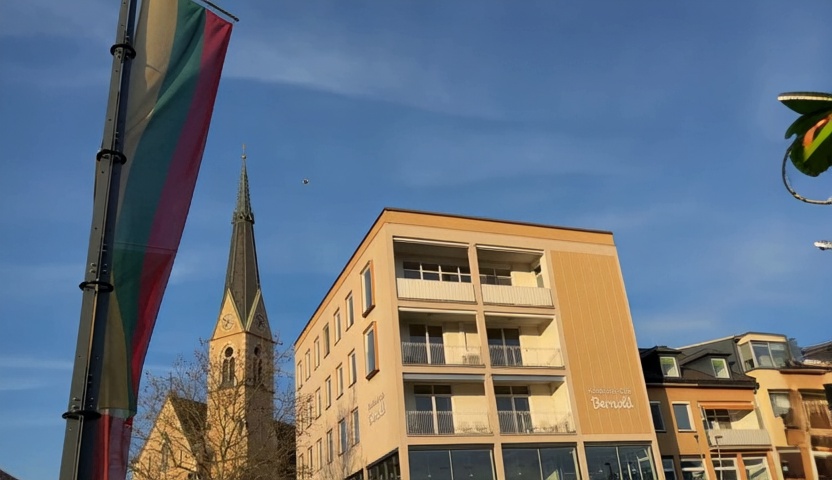}
  
  \multicolumn{5}{c}{\vspace{-1.5em}} \\
  \SetSceneZoom{(0.15,0.13)}{3}{1cm}{(0.65,0.1)}\FiveCols[north west]{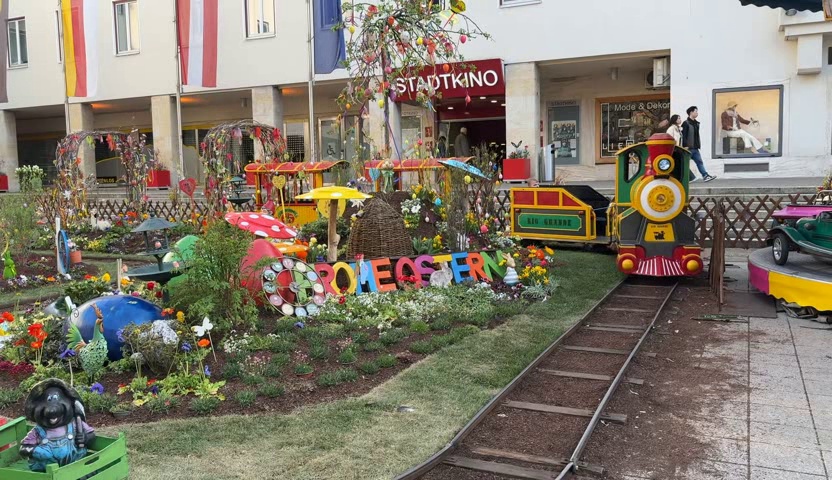}{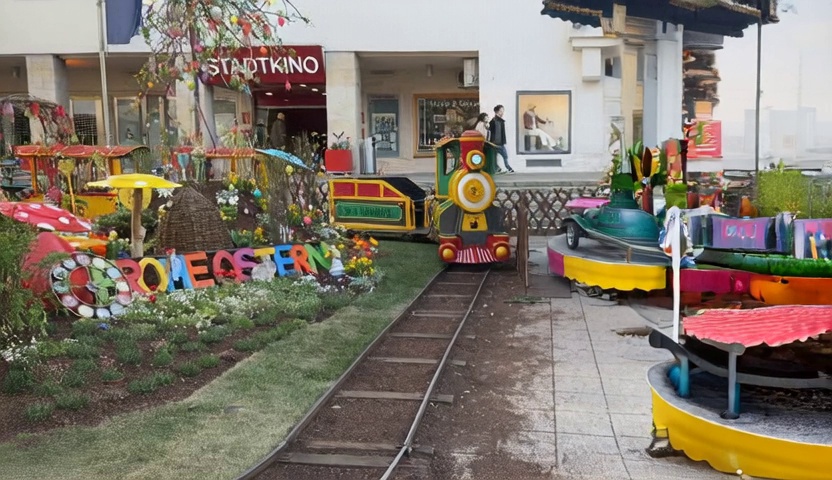}{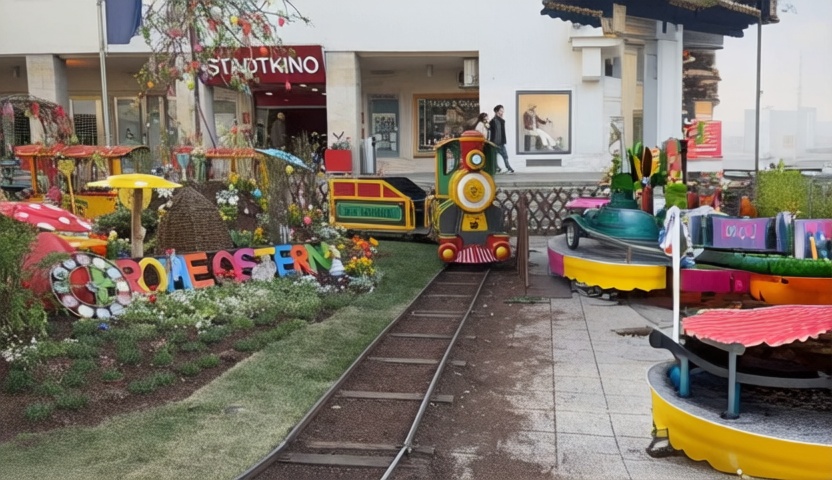}{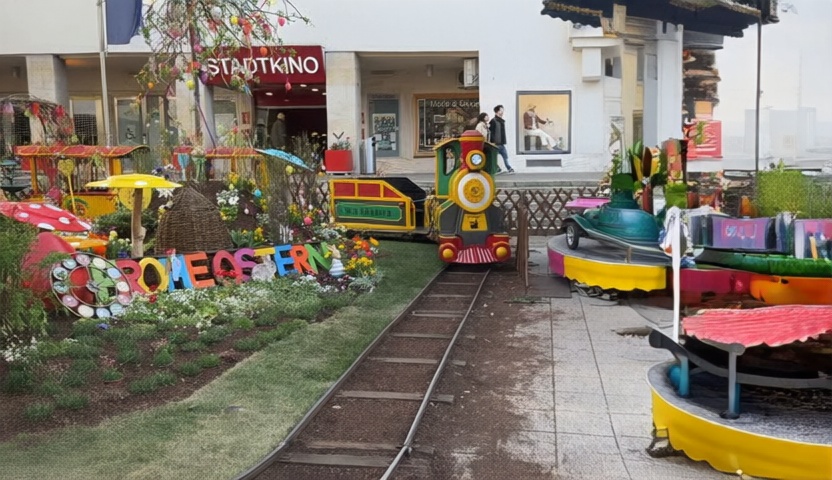}{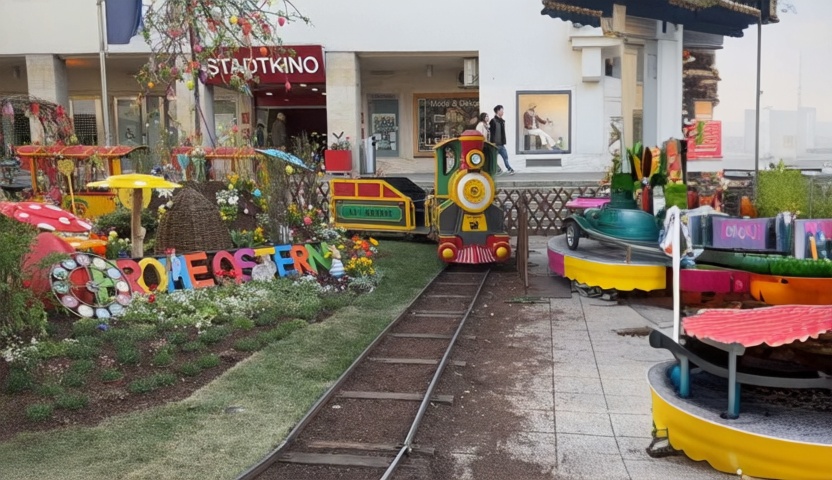}
  
  \multicolumn{5}{c}{\vspace{-1.5em}} \\
  \SetSceneZoom{(-0.35,-0.1)}{2.5}{0.8cm}{(0.05,-0.1)}\FiveCols[north east]{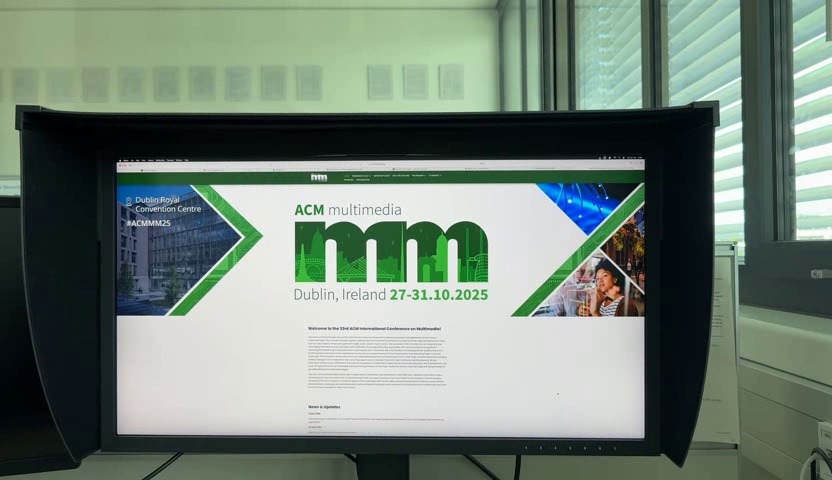}{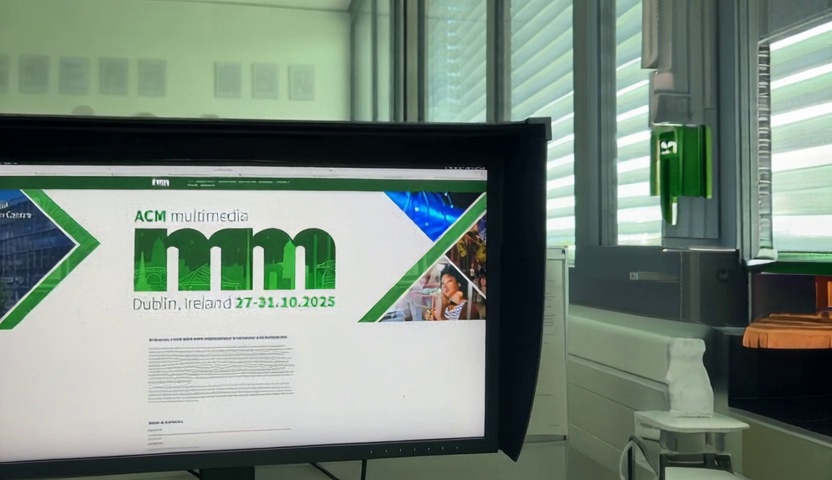}{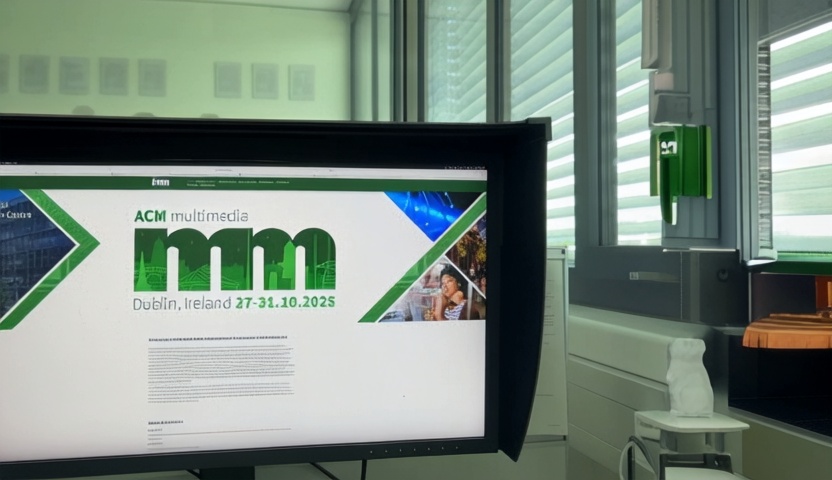}{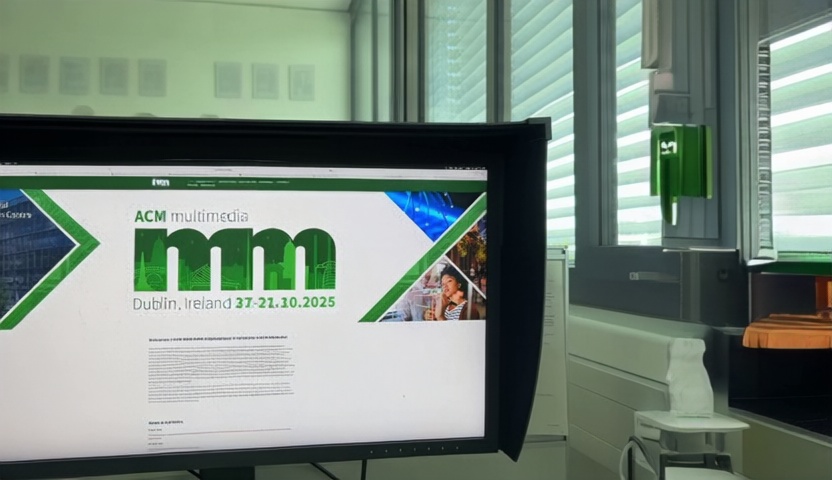}{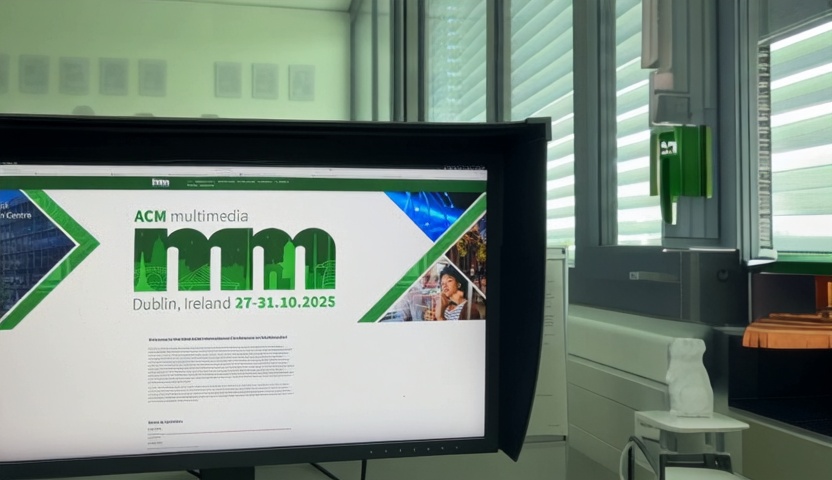}
  
 \multicolumn{5}{c}{\vspace{-1.5em}} \\
 \SetSceneZoom{(-0.08,0.5)}{5}{0.8cm}{(0.06,0.48)}\FiveCols[north west]{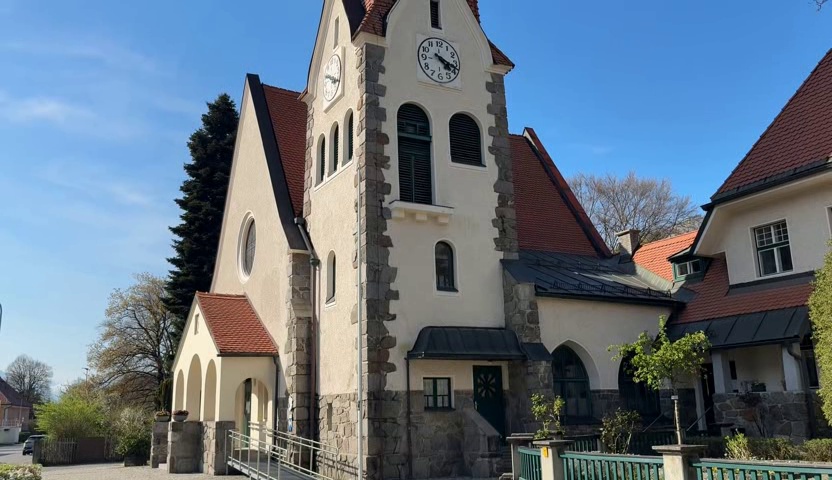}{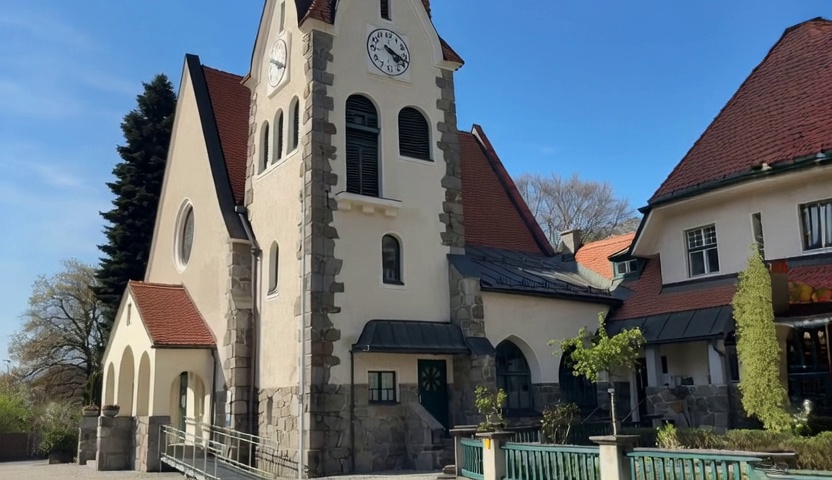}{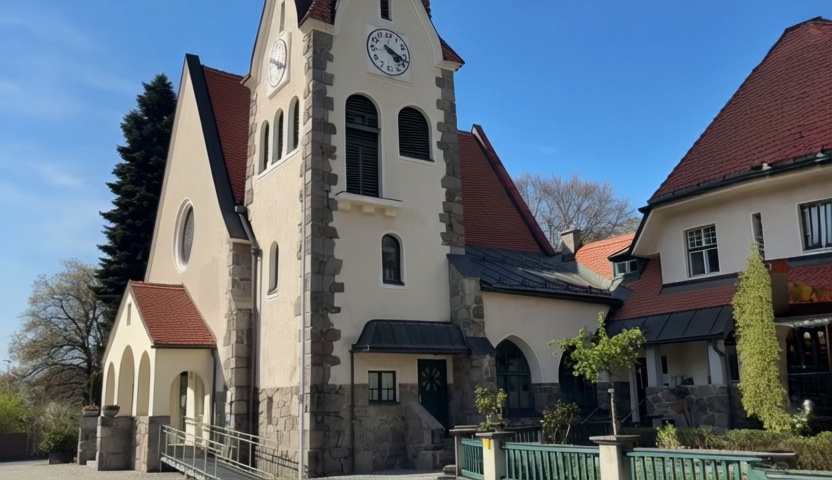}{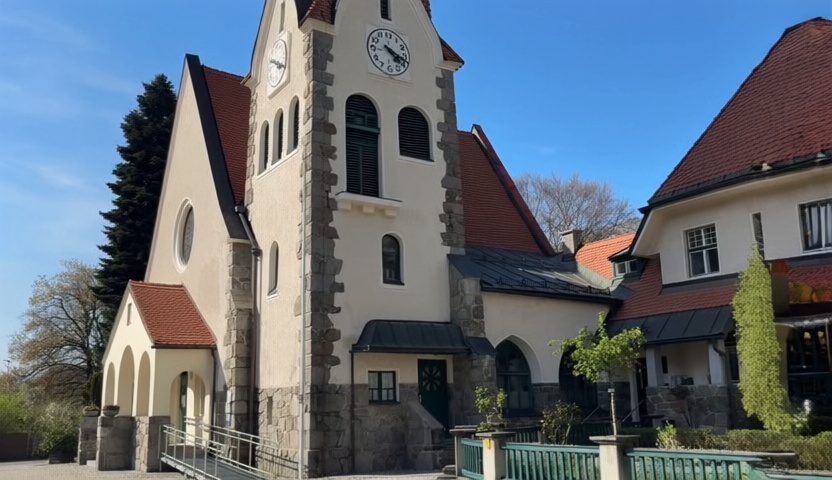}{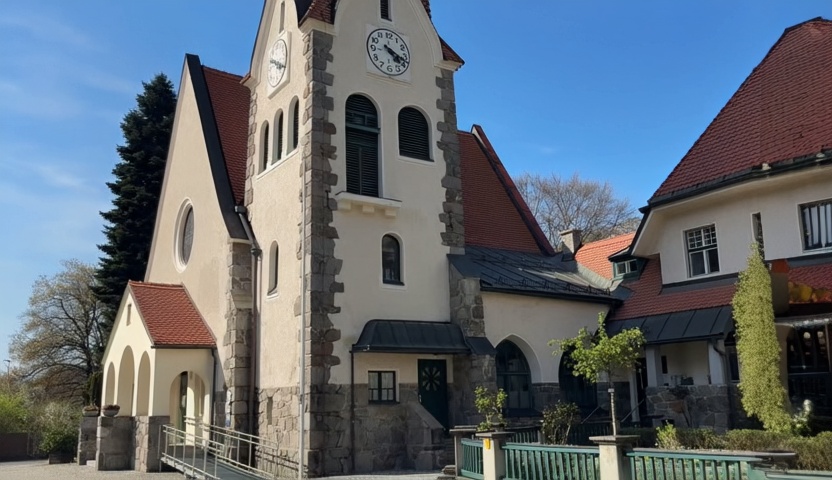}
 {Source view} & {W/O fixer} & {DIFIX3D+} & {{MaRINeR}} & {\textbf{Ours}} \\
  
  \multicolumn{5}{c}{\vspace{-0.2em}{Applied on StereoPilot (implicit SC)}} \\
  \SetSceneZoom{(-1.01,0.05)}{4}{1.0cm}{(-0.98, 0.05)}\FiveCols[north east]{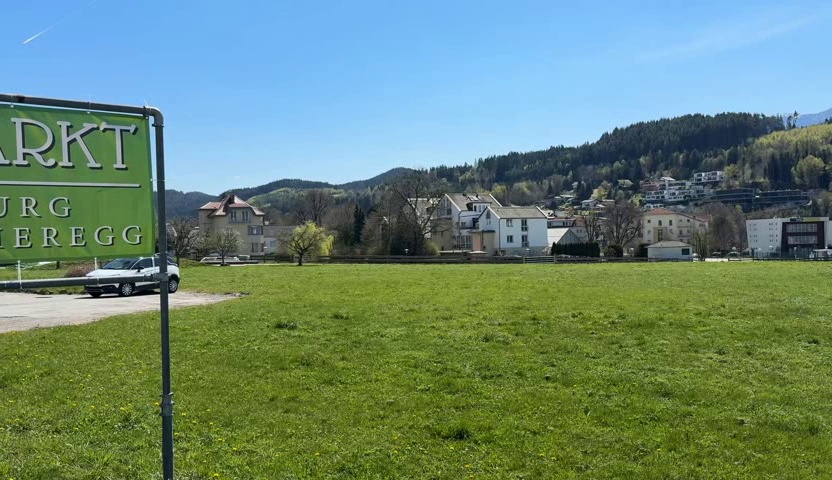}{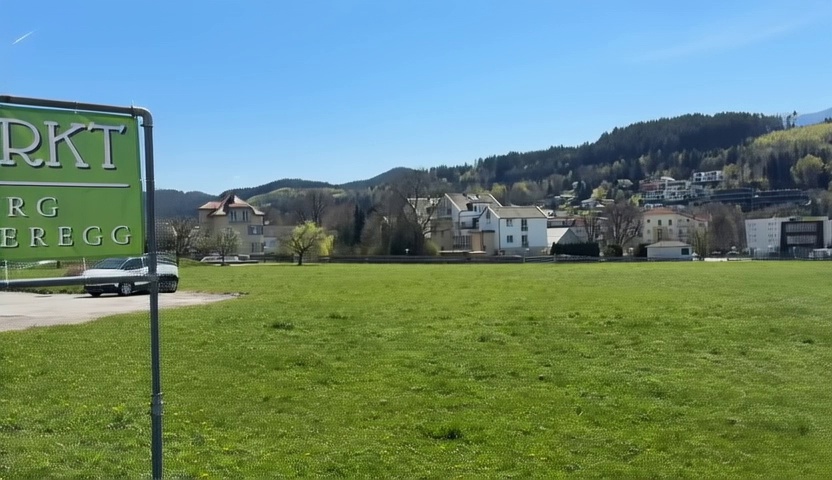}{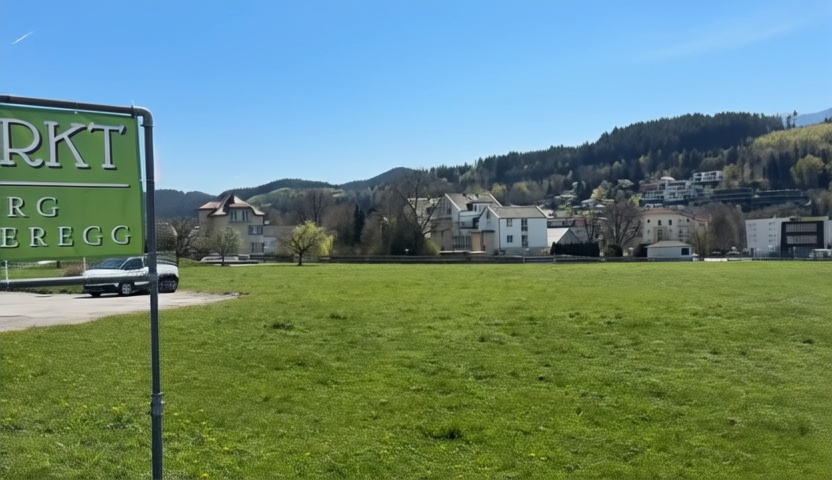}{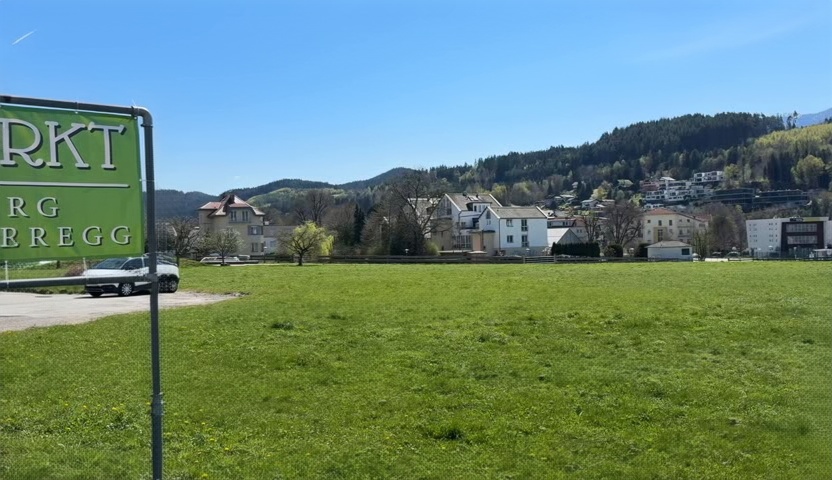}{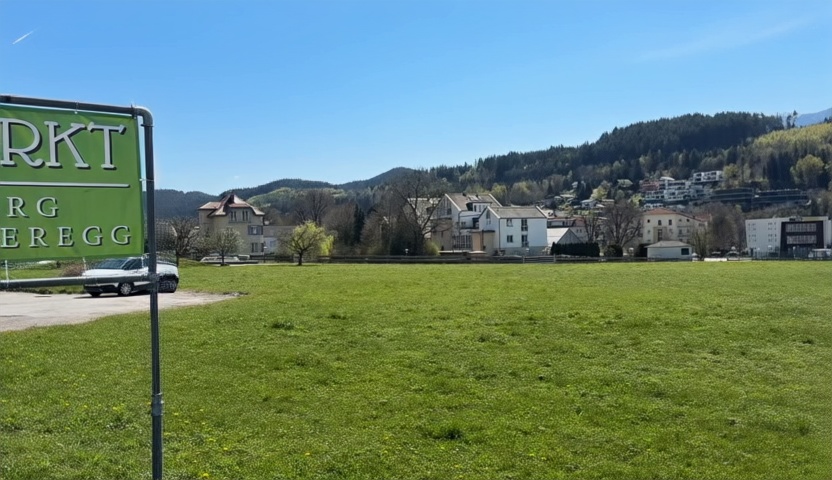}

  \multicolumn{5}{c}{\vspace{-1.5em}} \\
  \SetSceneZoom{(0.7,0.05)}{2}{1.0cm}{(0.7, 0.05)}\FiveCols[north west]{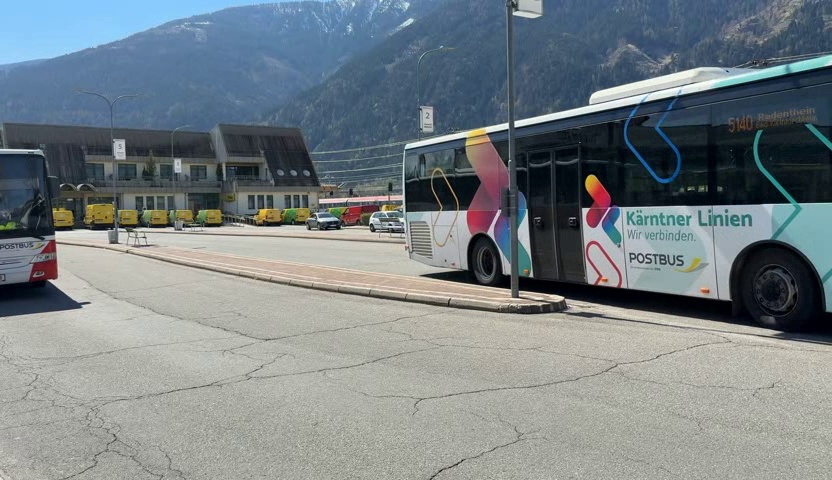}{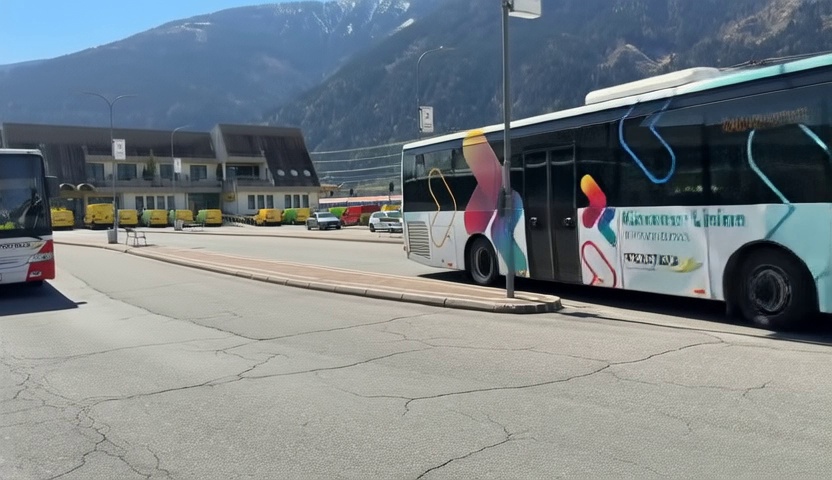}{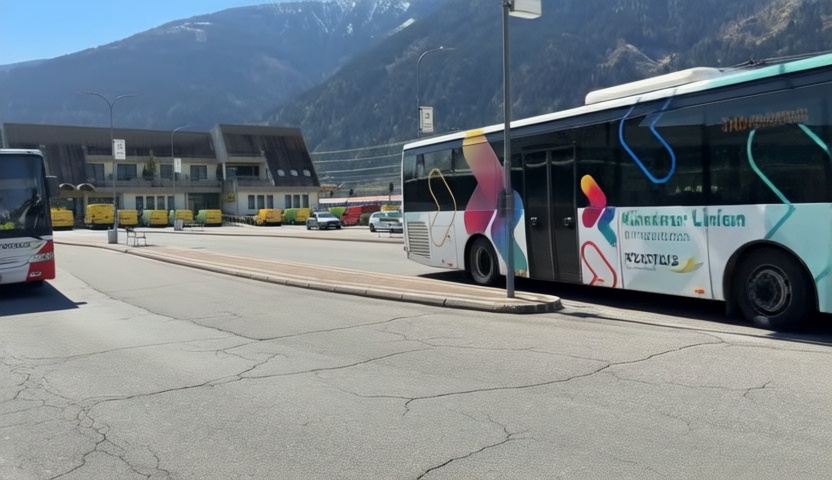}{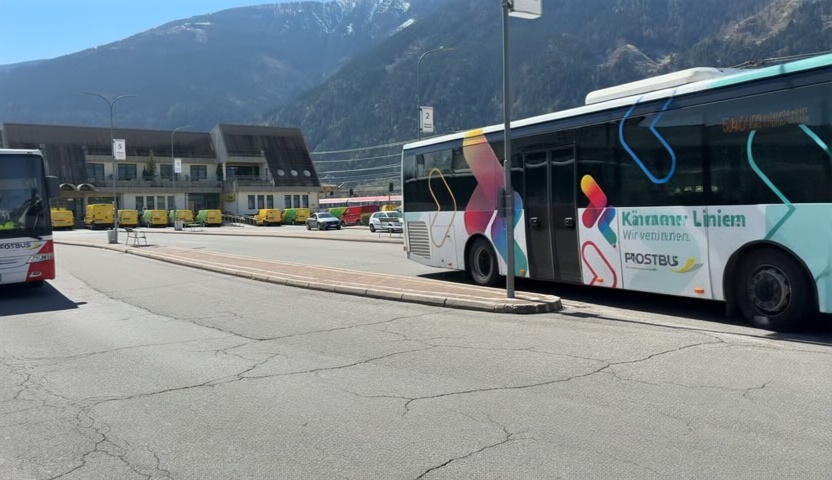}{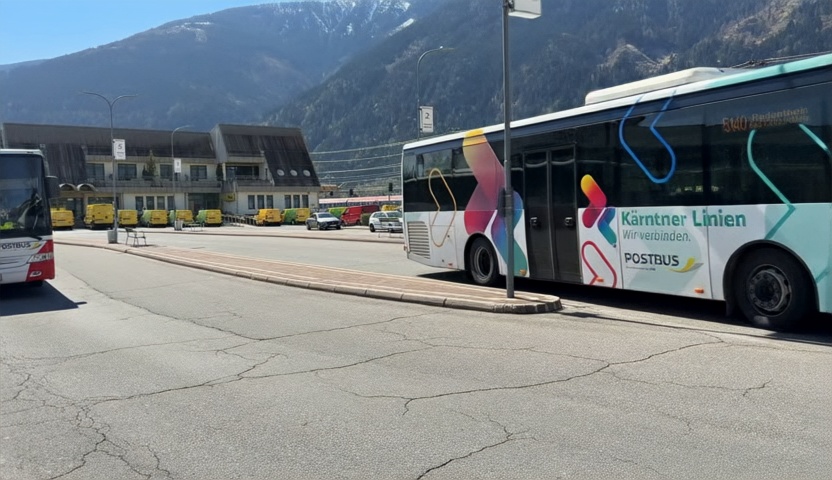}
  
  \multicolumn{5}{c}{\vspace{-1.5em}} \\
  \SetSceneZoom{(0.2,0.05)}{2}{1.0cm}{(0.2, 0.05)}\FiveCols[north west]{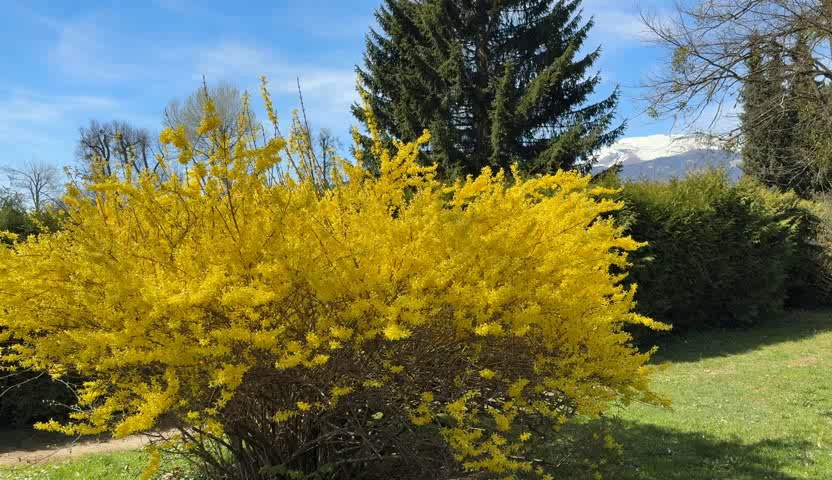}{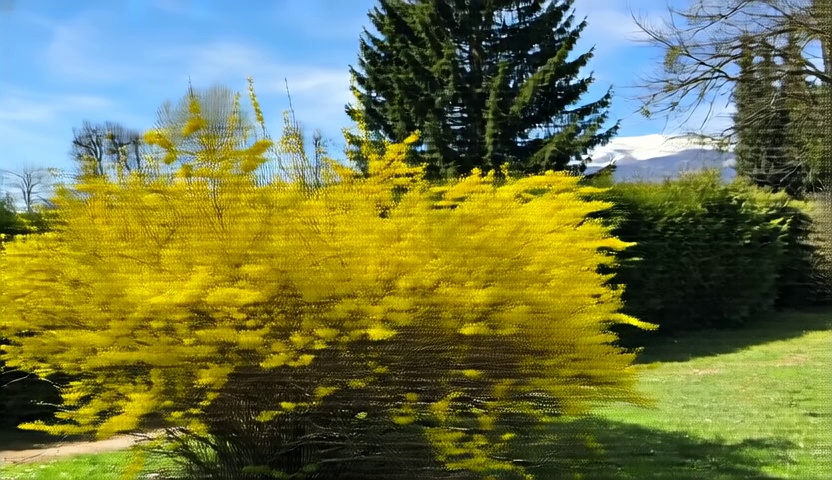}{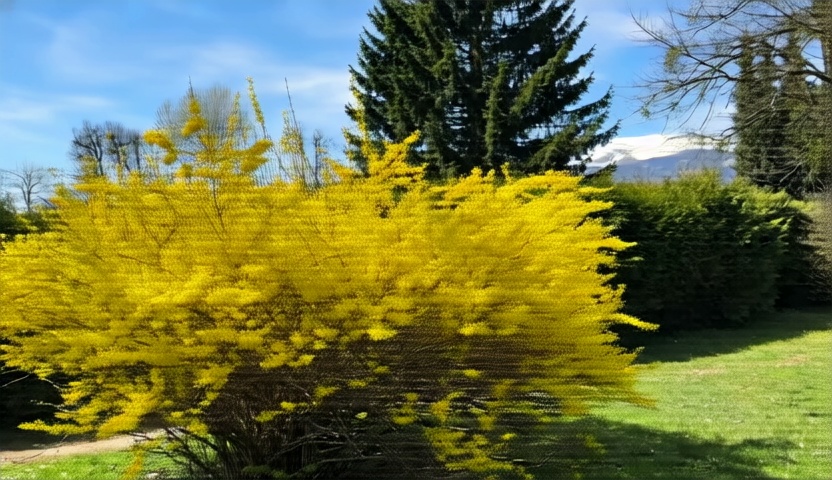}{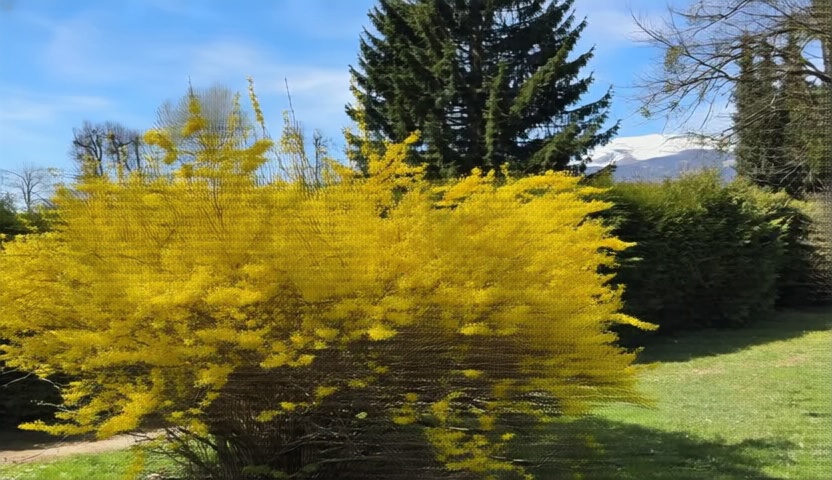}{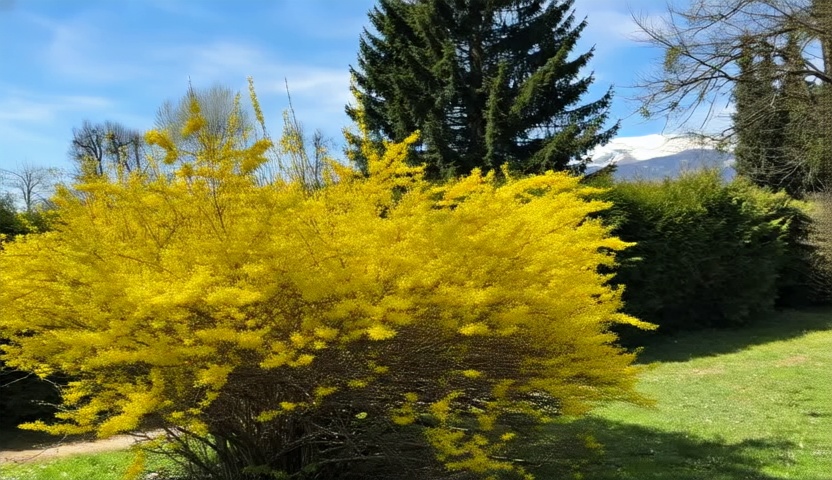}
  
  \multicolumn{5}{c}{\vspace{-1.5em}} \\
  \SetSceneZoom{(-0.4,-0.4)}{2}{1.0cm}{(-0.4, -0.4)}\FiveCols[north east]{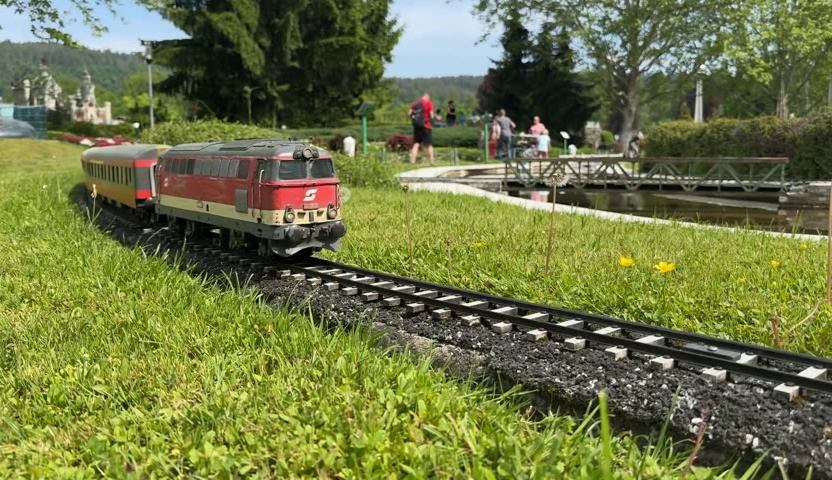}{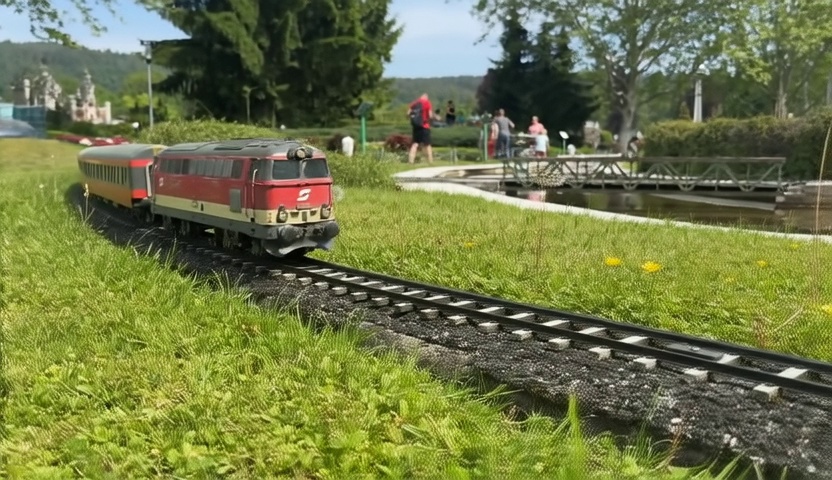}{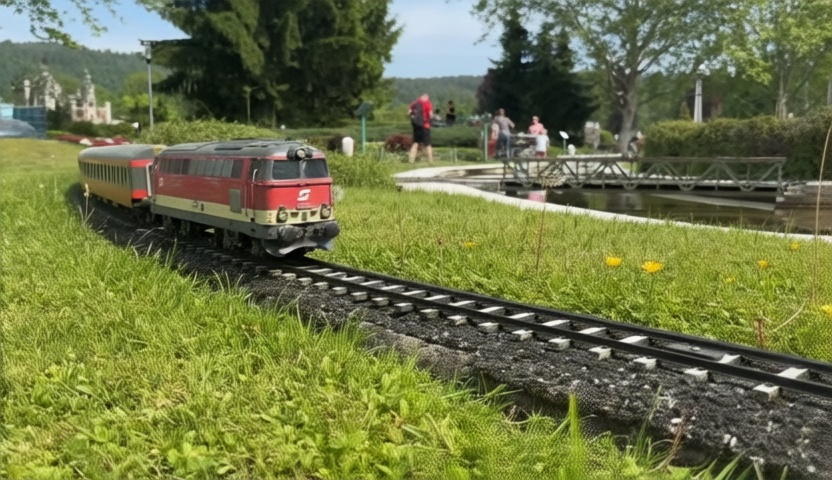}{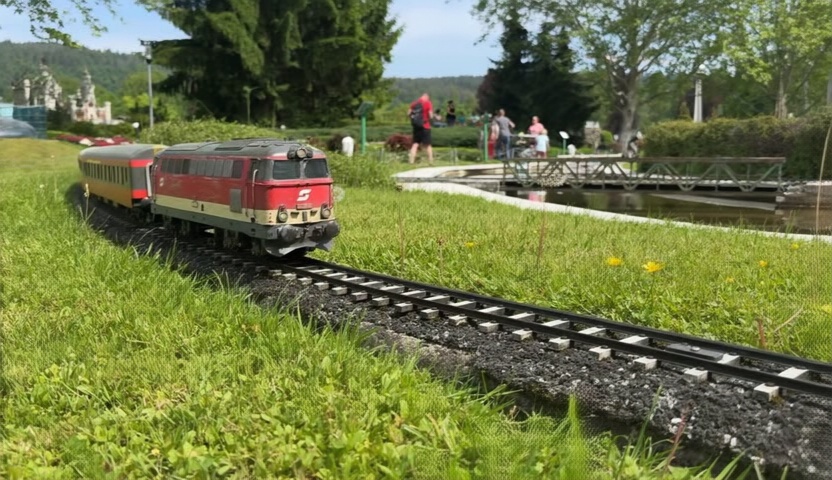}{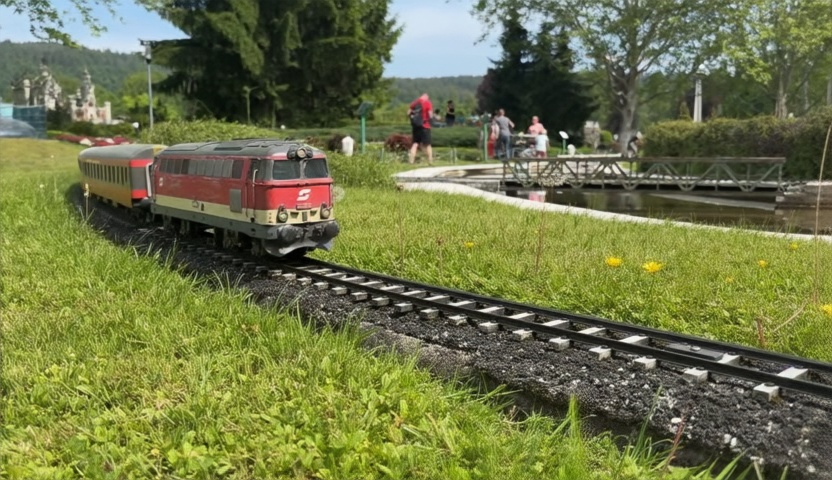}
  
  {Source view} & {W/O fixer} & {DIFIX3D+} & {{MaRINeR}} & {\textbf{Ours}} \\
\end{tabular}

\captionsetup{list=false}
\caption{Visual results of applying plug-and-play fixers (continued).}
\end{figure*}

We evaluate our method on six upstream baseline models in total for both NVS and SC tasks (\ie, VACE~\cite{jiang2025vace}, ViewCrafter~\cite{yu2024viewcrafter}, Mono2Stereo~\cite{yu2025mono2stereo}, StereoCrafter~\cite{zhao2024stereocrafter}, RecamMaster~\cite{bai2025recammaster}, and StereoPilot~\cite{shen2025stereopilot}) on DL3DV~\cite{ling2024dl3dv}, Spring~\cite{mehl2023spring}, Mono2stereo~\cite{yu2025mono2stereo} and Spatial Video Dataset~\cite{izadimehr2025svd}. As shown in Fig.~\ref{fig:vis_comparison}, our approach consistently fixed the view synthesis degradation in the outputs of the upstream models and outperforms state-of-the-art novel view fixers (\ie, DIFIX3D+~\cite{wu2025difix3d+} and MaRINeR~\cite{bosiger2024mariner}).

\subsection{In-the-wild fixing results}

\noindent We perform more in-the-wild tests. Specifically, we use VACE~\cite{jiang2025vace}, serving as the representative example of explicit DWI pipeline. By simply applying our plug-and-play fixer, we can significantly remove the degradation and improve the visual quality (\cref{fig:in-the-wild}).

\def\imgWidthFour{0.235\textwidth} %
\def\imgWidthFive{0.185\textwidth} %
\def\imgWidth{\imgWidthFive} %

\providecommand{\SetImgWidth}[1]{}%
\renewcommand{\SetImgWidth}[1]{%
  \def\imgWidth{#1}%
  \tikzset{img/.style={rectangle, minimum width=\imgWidth, draw=black, inner sep=0, outer sep=0}}%
}

\SetImgWidth{\imgWidthFive}

\providecommand{\FourCols}[5]{}%
\renewcommand{\FourCols}[5][north east]{%
  \ZoomPicBySampling[#1]{\ZoomMag}{\ZoomSize}{\GTPos}{#2}{} &
  \ZoomPicBySampling[#1]{\ZoomMag}{\ZoomSize}{\ZoomPos}{#3}{} &
  \ZoomPicBySampling[#1]{\ZoomMag}{\ZoomSize}{\ZoomPos}{#4}{} &
  \ZoomPicBySampling[#1]{\ZoomMag}{\ZoomSize}{\ZoomPos}{#5}{} \\
}

\providecommand{\ThreeColsSrc}[4]{}%
\renewcommand{\ThreeColsSrc}[4][north east]{%
  \ZoomPicBySampling[#1]{\ZoomMag}{\ZoomSize}{\GTPos}{#2}{} &
  \ZoomPicBySampling[#1]{\ZoomMag}{\ZoomSize}{\RefPos}{#3}{} &
  \ZoomPicBySampling[#1]{\ZoomMag}{\ZoomSize}{\ZoomPos}{#4}{} \\
}

\providecommand{\FiveColsSrc}[6]{}%
\renewcommand{\FiveColsSrc}[6][north east]{%
  \ZoomPicBySampling[#1]{\ZoomMag}{\ZoomSize}{\GTPos}{#2}{} &
  \ZoomPicBySampling[#1]{\ZoomMag}{\ZoomSize}{\RefPos}{#3}{} &
  \ZoomPicBySampling[#1]{\ZoomMag}{\ZoomSize}{\ZoomPos}{#4}{} &
  \ZoomPicBySampling[#1]{\ZoomMag}{\ZoomSize}{\ZoomPos}{#5}{} &
  \ZoomPicBySampling[#1]{\ZoomMag}{\ZoomSize}{\ZoomPos}{#6}{} \\
}

\providecommand{\SetSceneZoomSrc}[5]{}%
\renewcommand{\SetSceneZoomSrc}[5]{%
  \SetSceneZoom{#1}{#3}{#4}{#5}%
  \global\def\RefPos{#2}%
}

\begin{figure*}[h]
\centering
\setlength{\tabcolsep}{0pt}

\centering
\SetImgWidth{0.25\textwidth}
\begin{tabular}{@{}c c c c@{}}
  \multicolumn{4}{c}{\vspace{-0.2em}} \\
  \SetSceneZoom{(0.3,0.4)}{1.5}{1.0cm}{(0.3,0.4)}%
  \FourCols[north west]{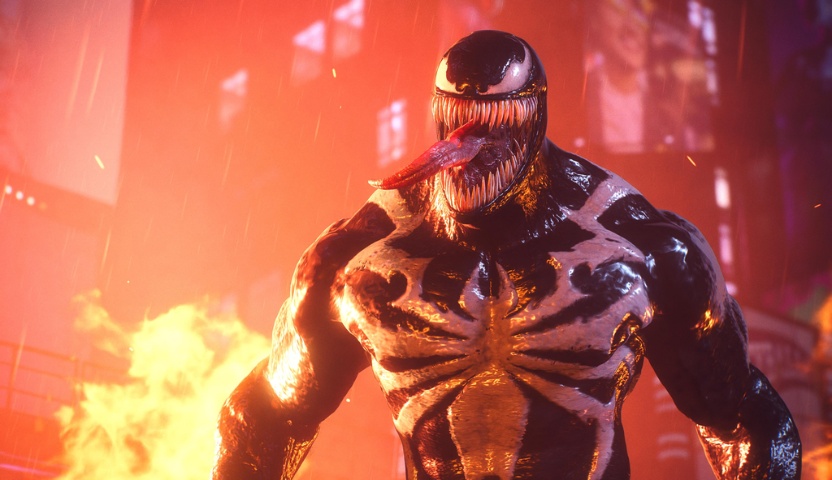}{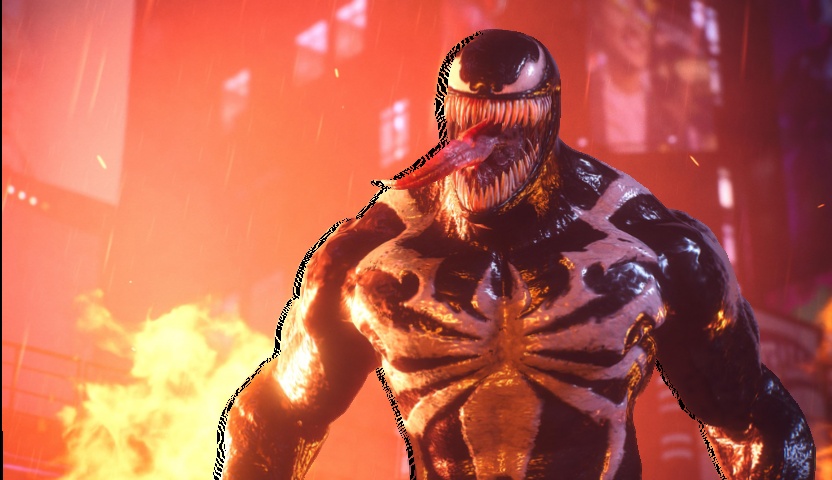}{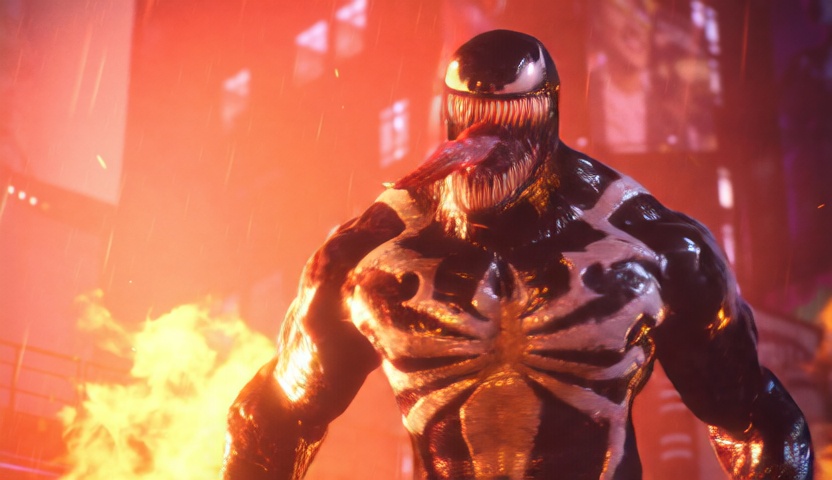}{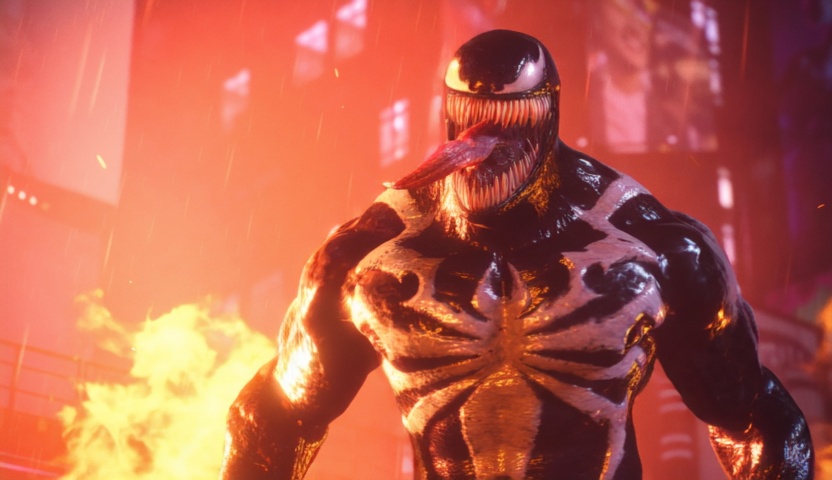}

  \multicolumn{4}{c}{\vspace{-1.55em}} \\
  \SetSceneZoom{(0.8,-0.6)}{2}{1.0cm}{(0.6,-0.6)}%
  \FourCols[north west]{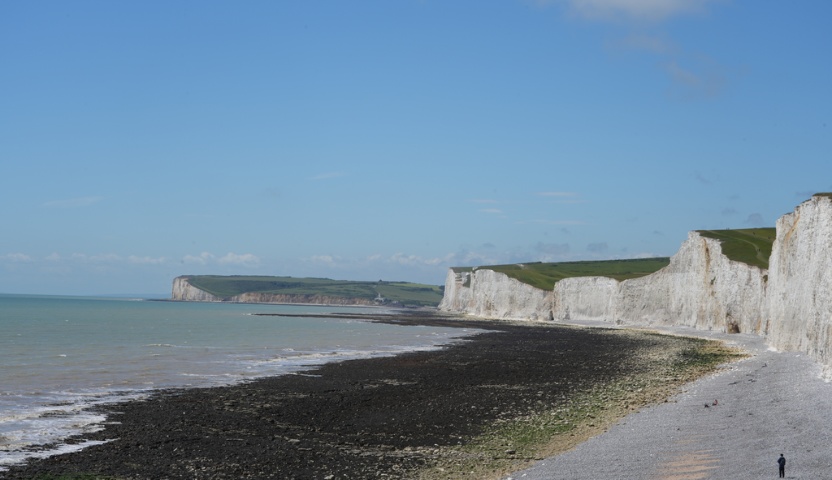}{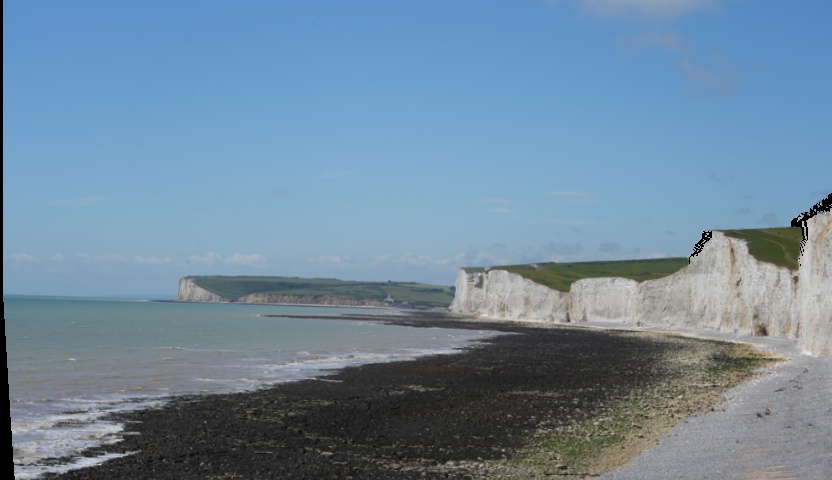}{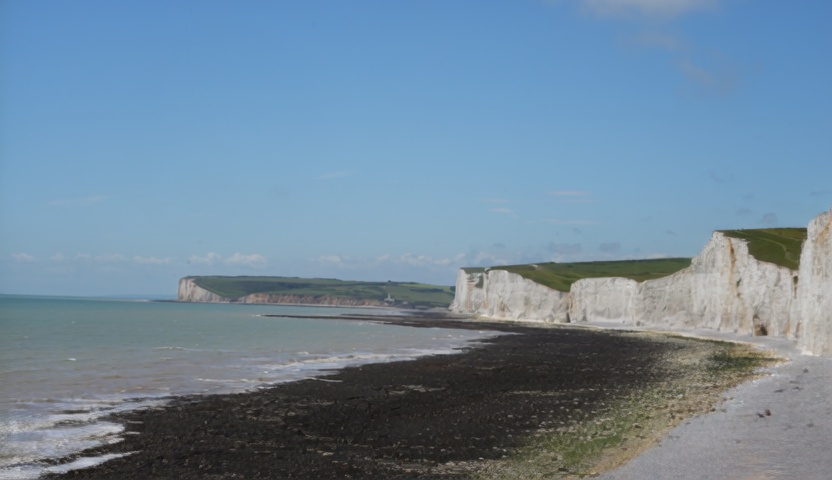}{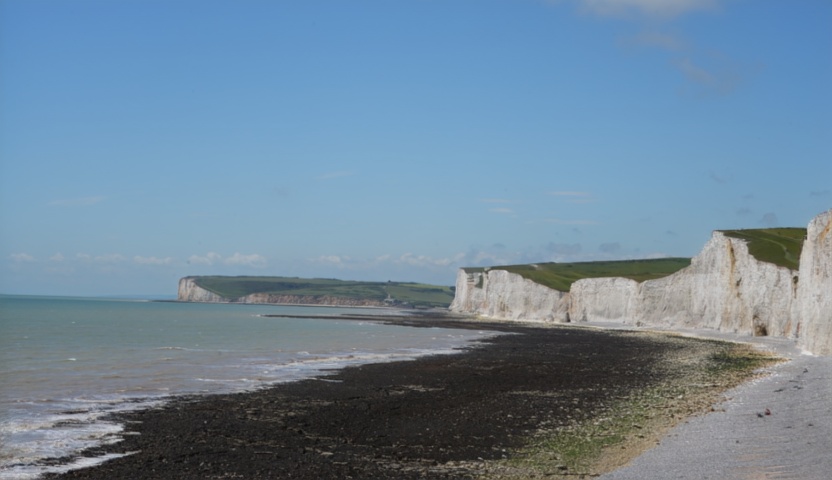}

  \multicolumn{4}{c}{\vspace{-1.55em}} \\
  \SetSceneZoom{(-0.2,-0.5)}{3}{1.0cm}{(-0.35,-0.4)}%
  \FourCols[north east]{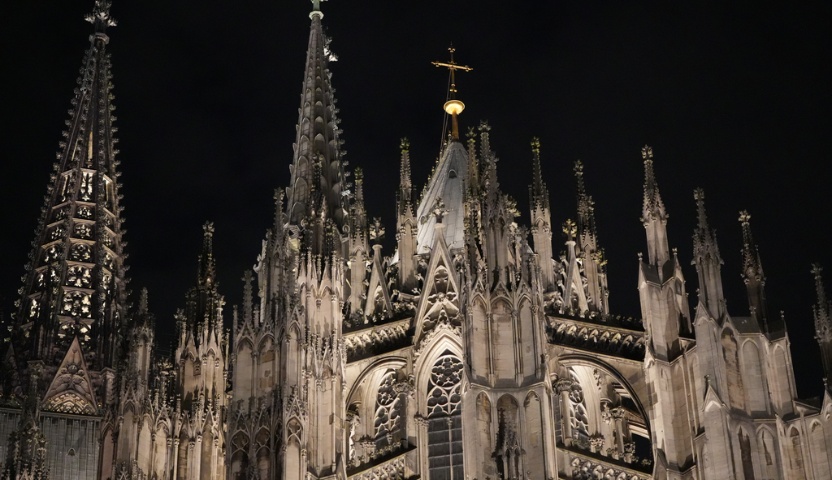}{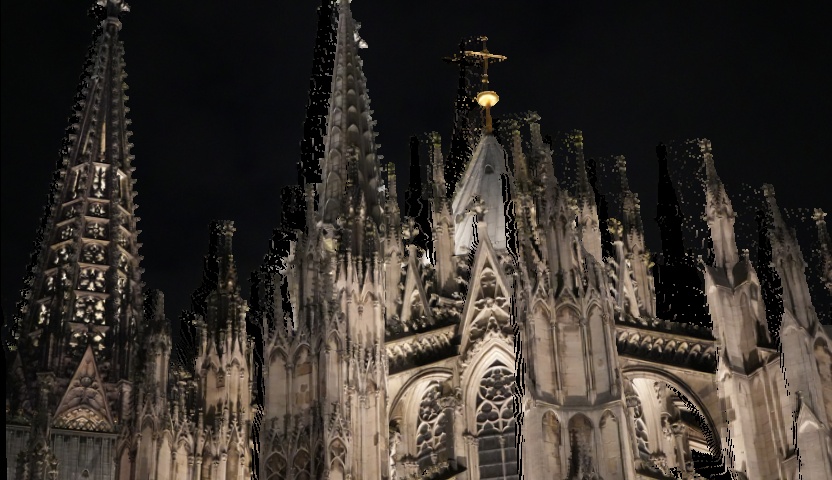}{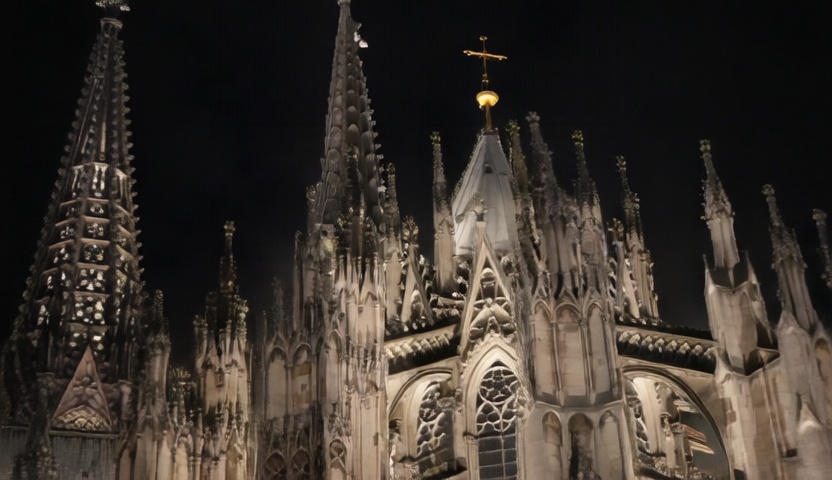}{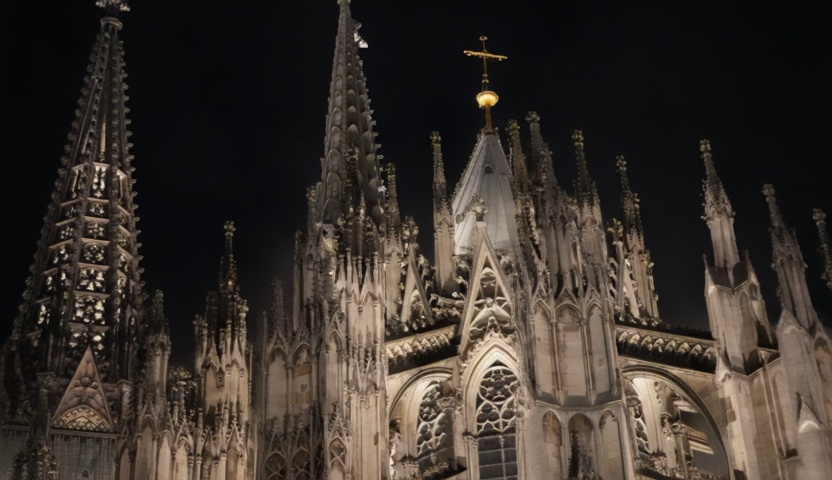}

  \multicolumn{4}{c}{\vspace{-1.55em}} \\
  \SetSceneZoom{(-0.5,0.3)}{2}{1.0cm}{(-0.6,0.3)}%
  \FourCols[north east]{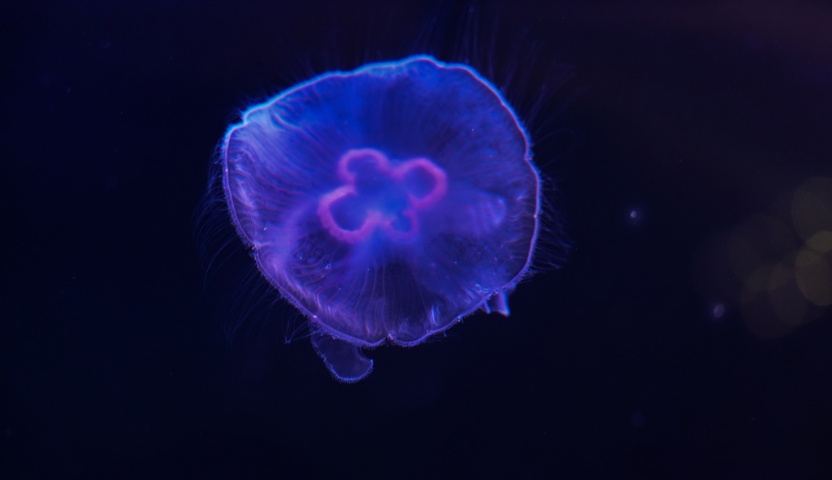}{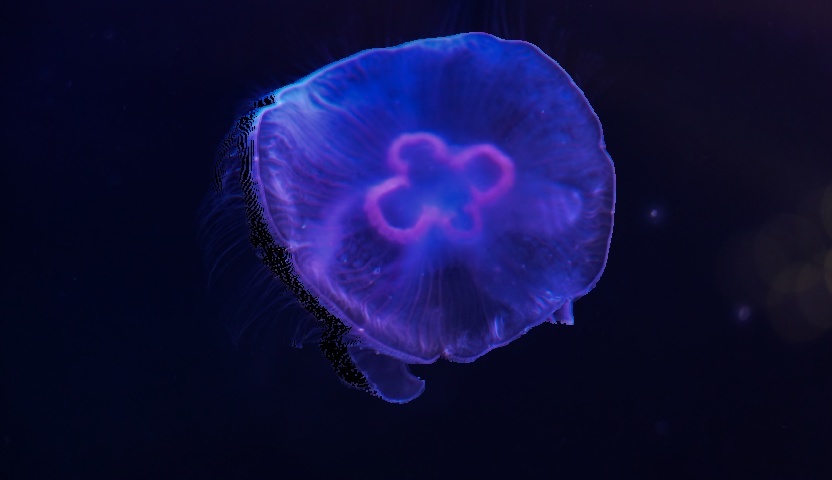}{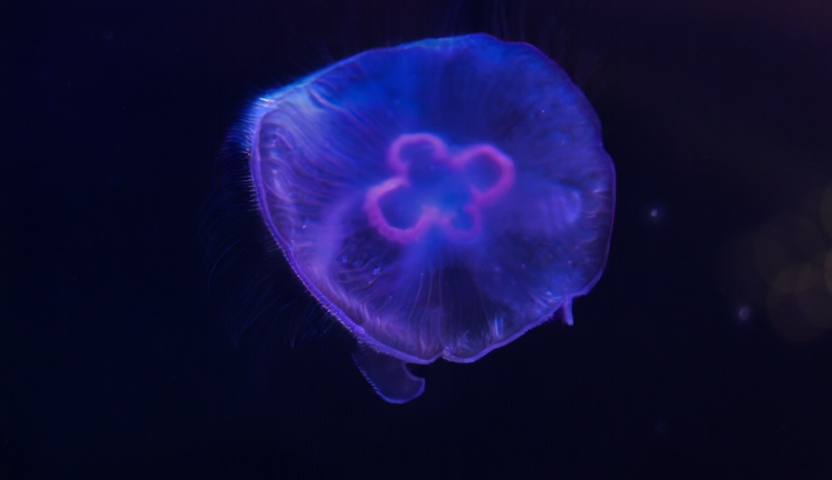}{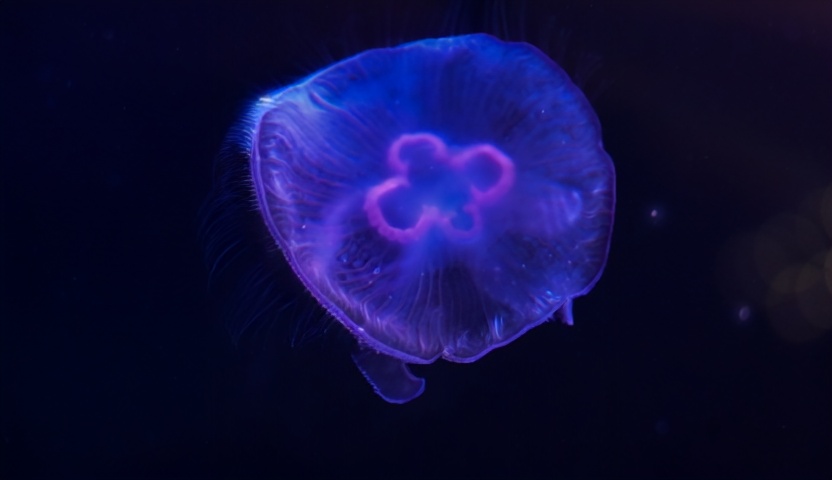}

  \multicolumn{4}{c}{\vspace{-1.55em}} \\
  \SetSceneZoom{(-0.8,0.2)}{6}{1.0cm}{(-0.83,0.15)}%
  \FourCols[north east]{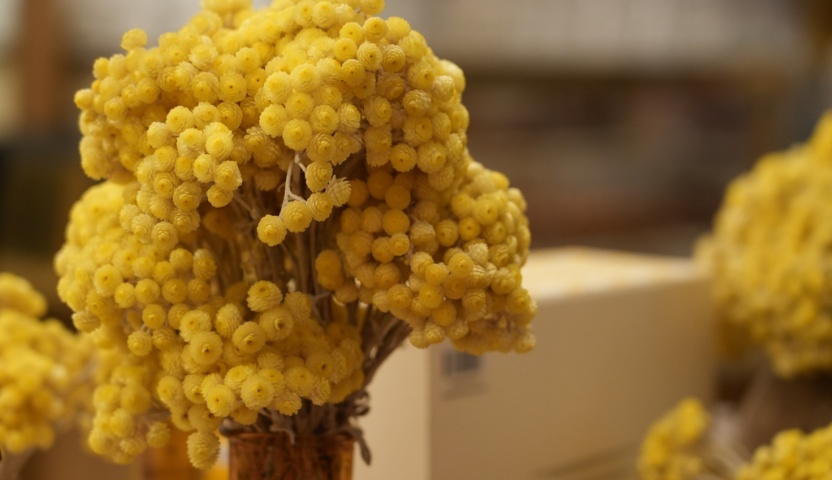}{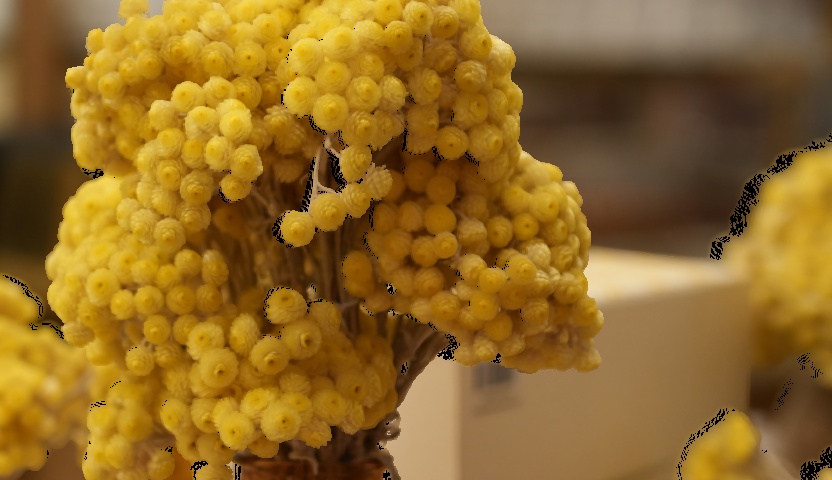}{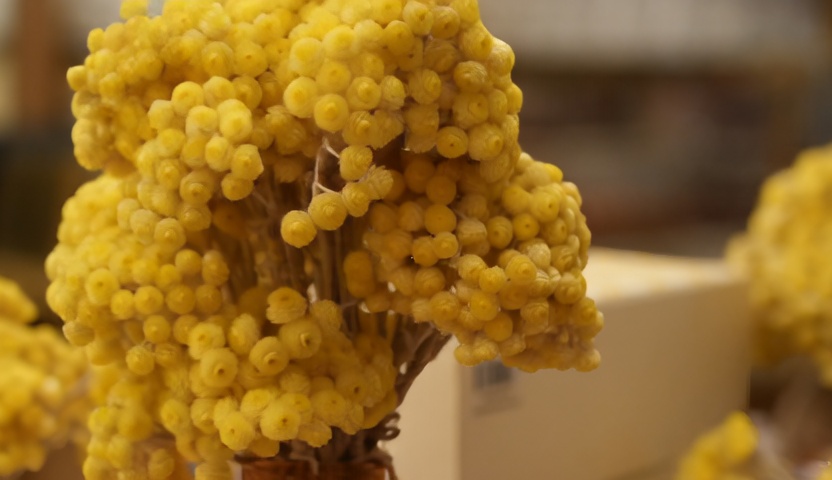}{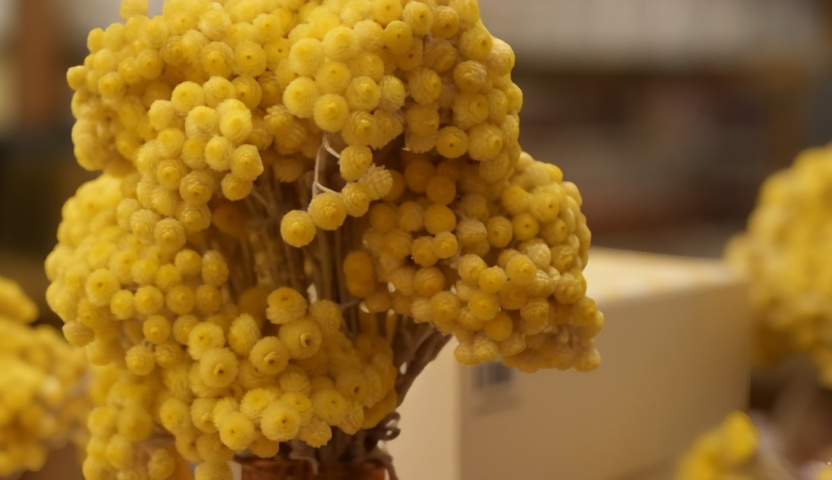}

  \multicolumn{4}{c}{\vspace{-1.55em}} \\
  \SetSceneZoom{(-0.6, -0.5)}{3}{1.0cm}{(-0.6,-0.4)}%
  \FourCols[north east]{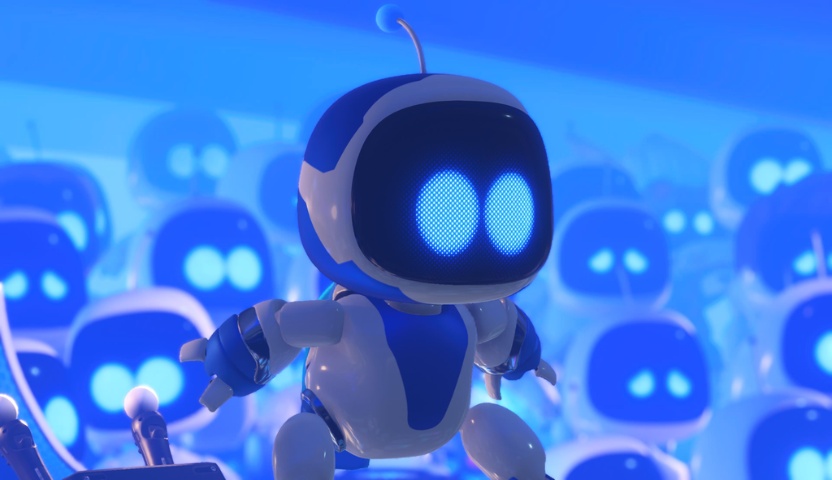}{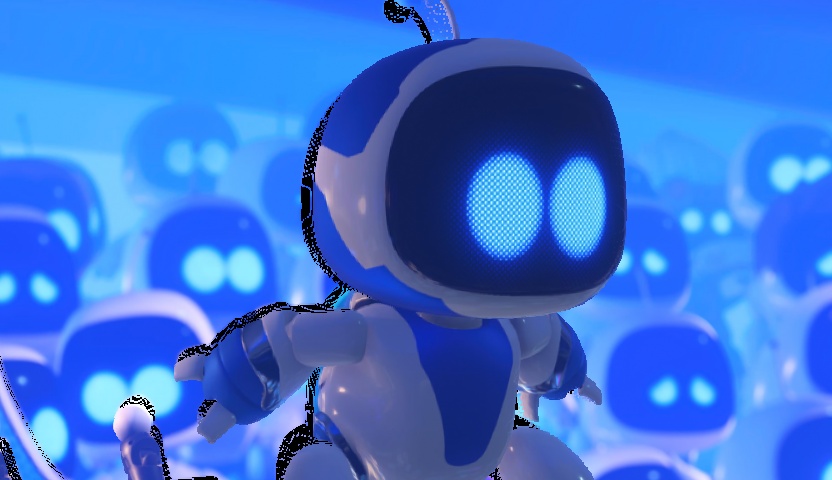}{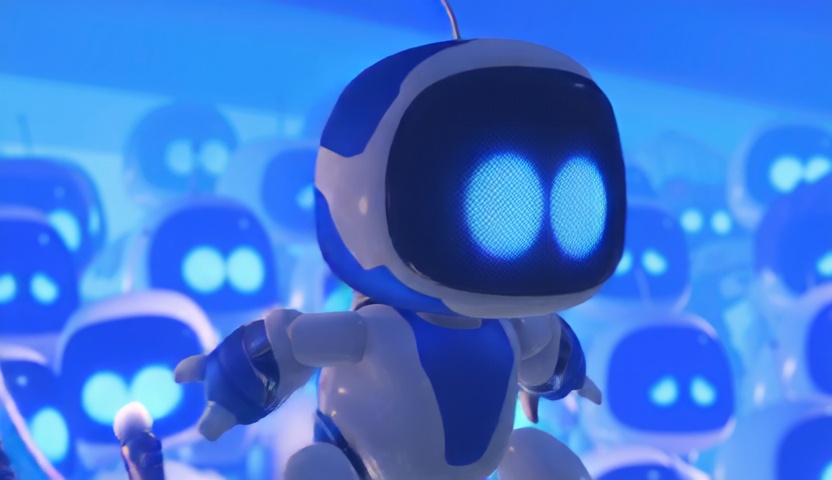}{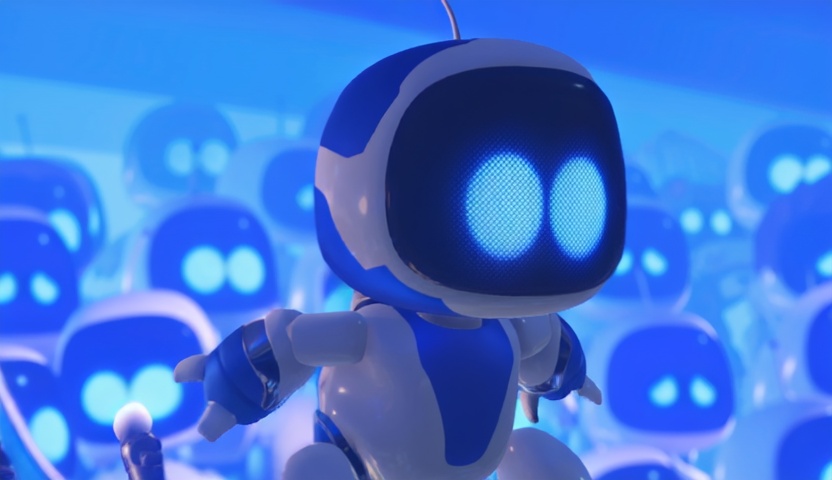}

  \multicolumn{4}{c}{\vspace{-1.55em}} \\
  \SetSceneZoom{(0.5,-0.2)}{2}{1.0cm}{(0.0,-0.2)}%
  \FourCols[south west]{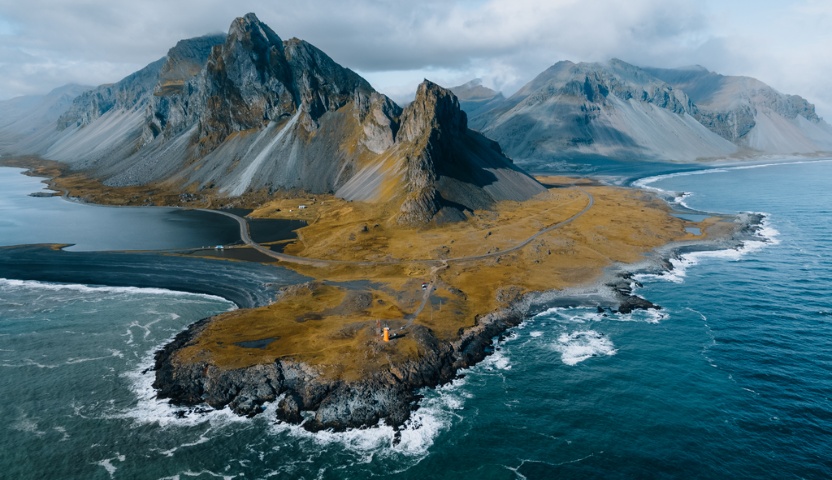}{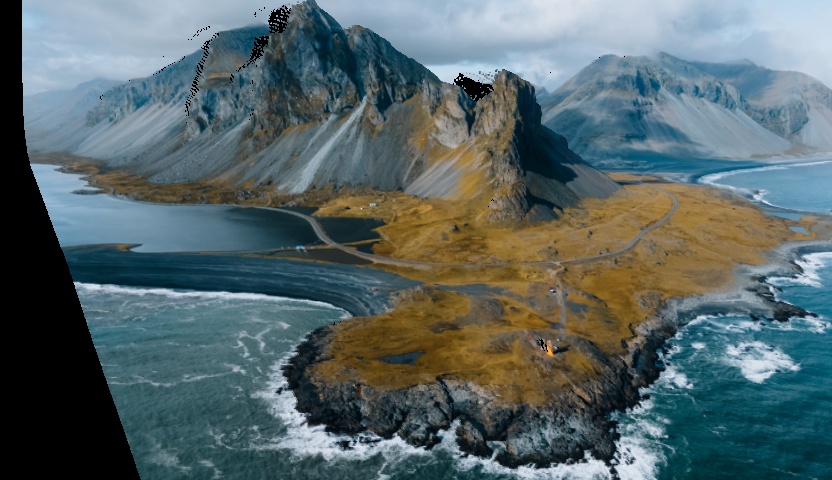}{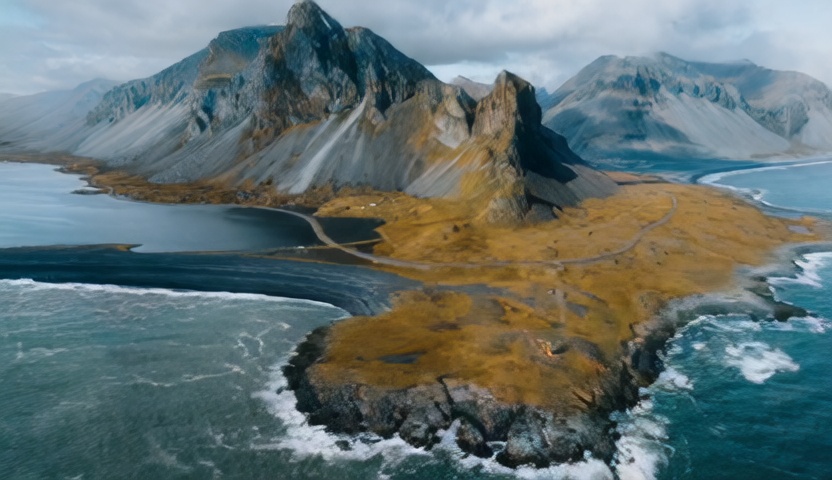}{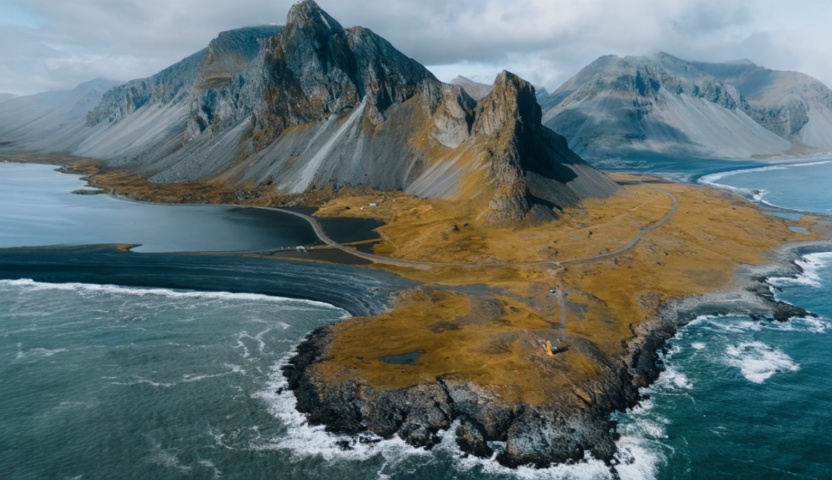}

    \multicolumn{4}{c}{\vspace{-1.55em}} \\
  \SetSceneZoom{(0.05,0.5)}{2}{1.0cm}{(-0.25,0.45)}%
  \FourCols[south west]{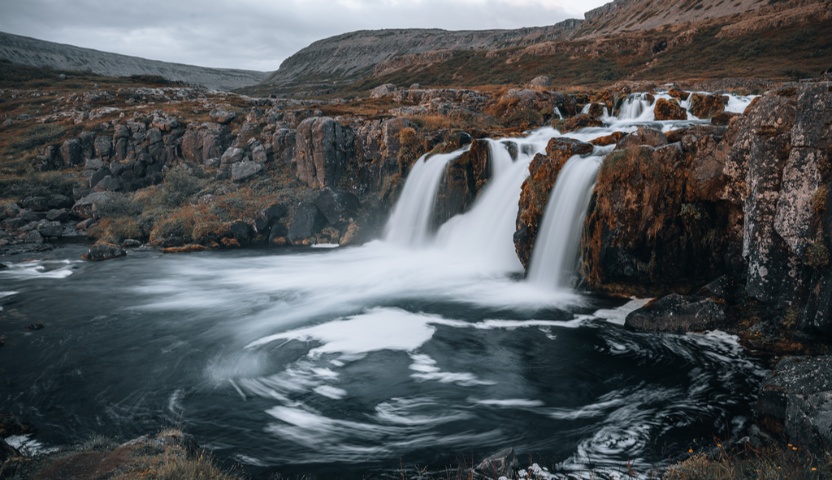}{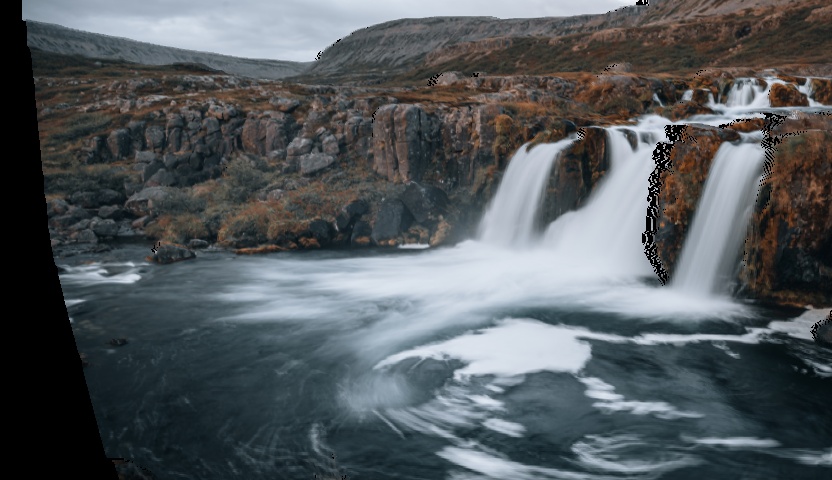}{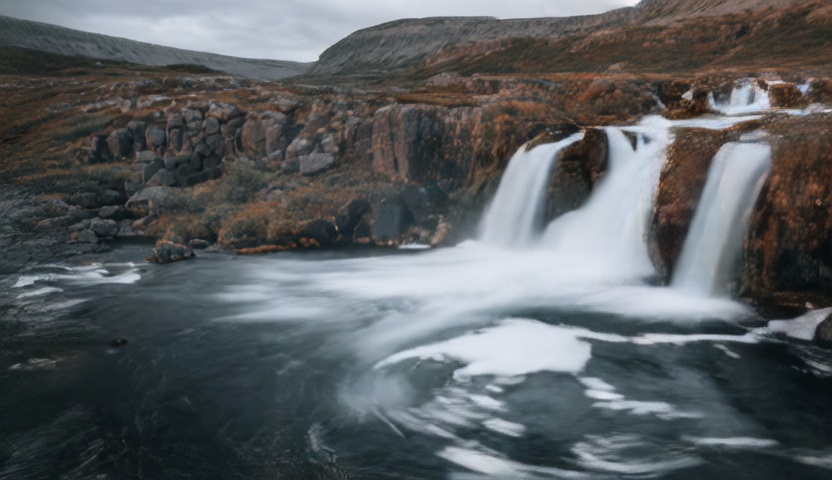}{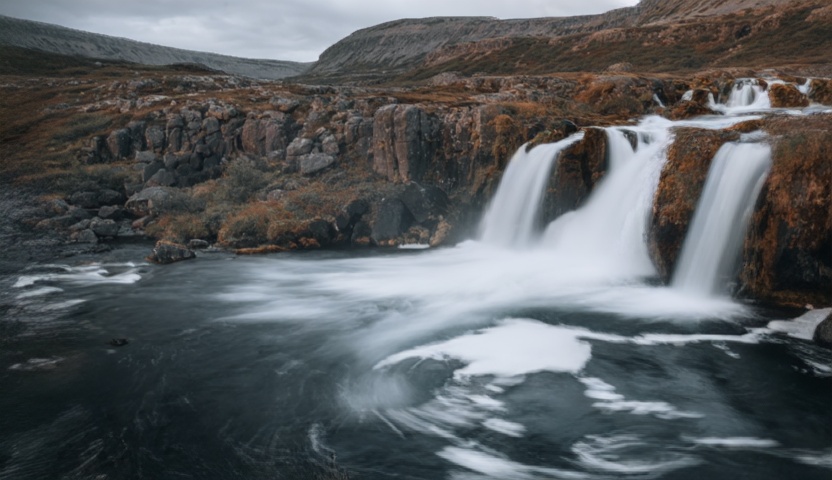}

  \multicolumn{4}{c}{\vspace{-1.55em}} \\
  \SetSceneZoom{(-0.4,-0.6)}{3}{1.0cm}{(-0.2,-0.5)}%
  \FourCols[south east]{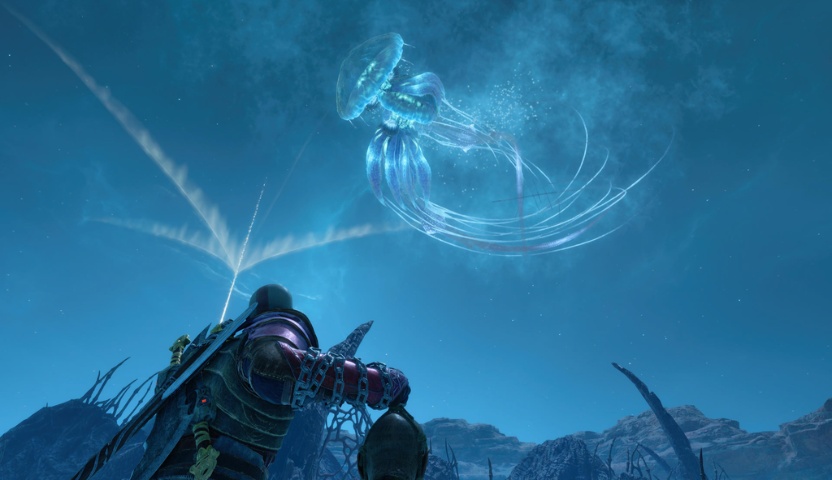}{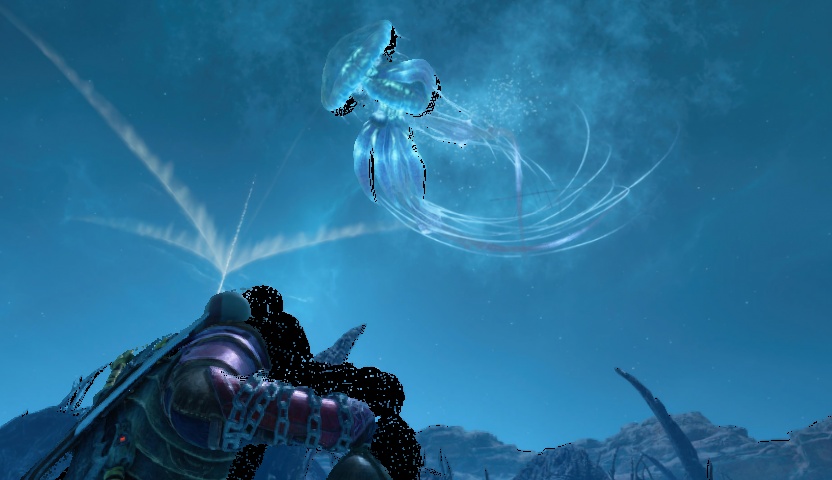}{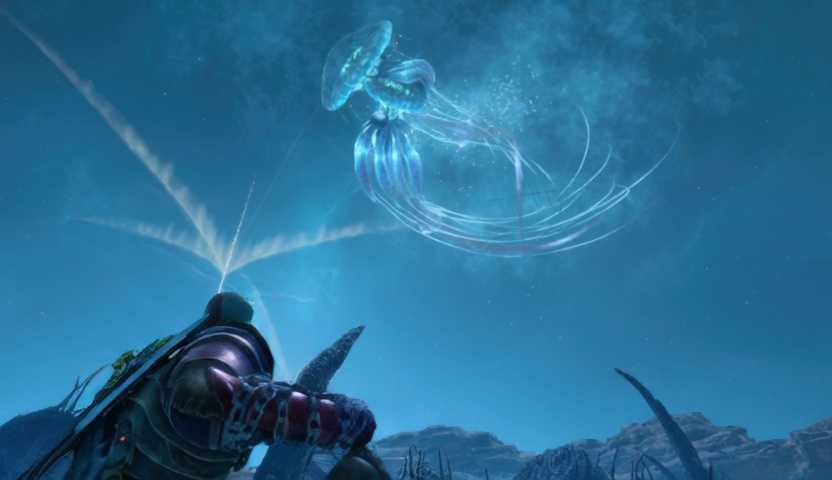}{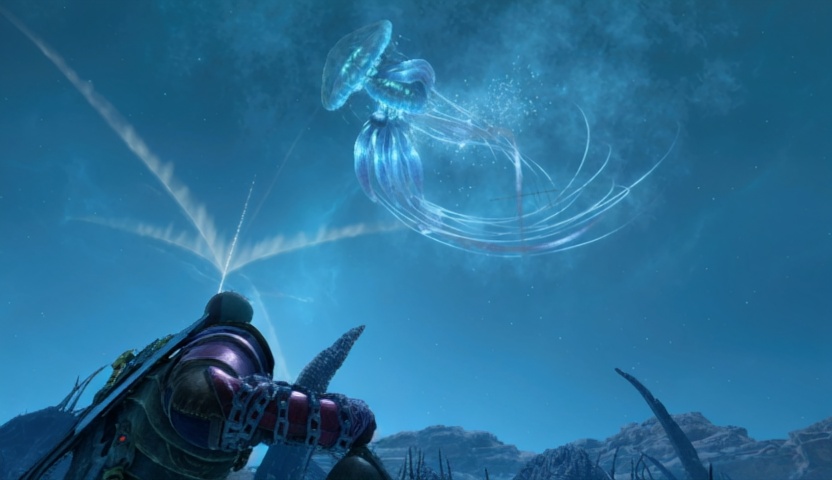}

  \multicolumn{4}{c}{\vspace{-1.55em}} \\
  \SetSceneZoom{(0.8,-0.72)}{3}{1.0cm}{(0.83,-0.72)}%
  \FourCols[north west]{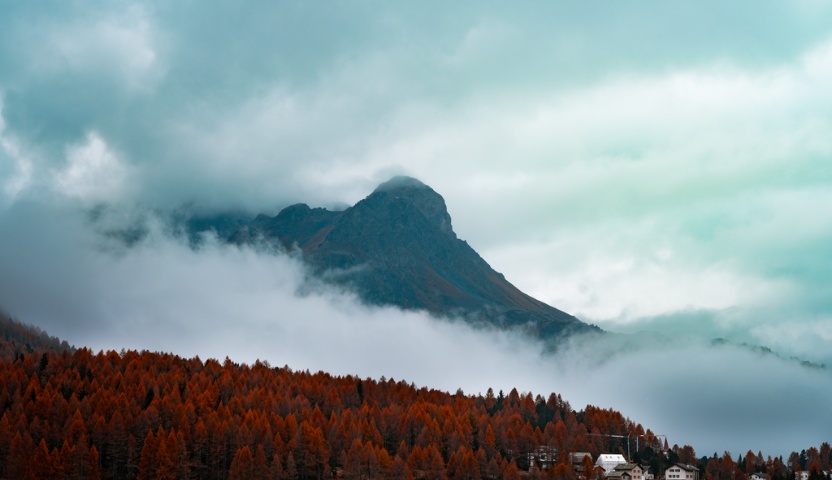}{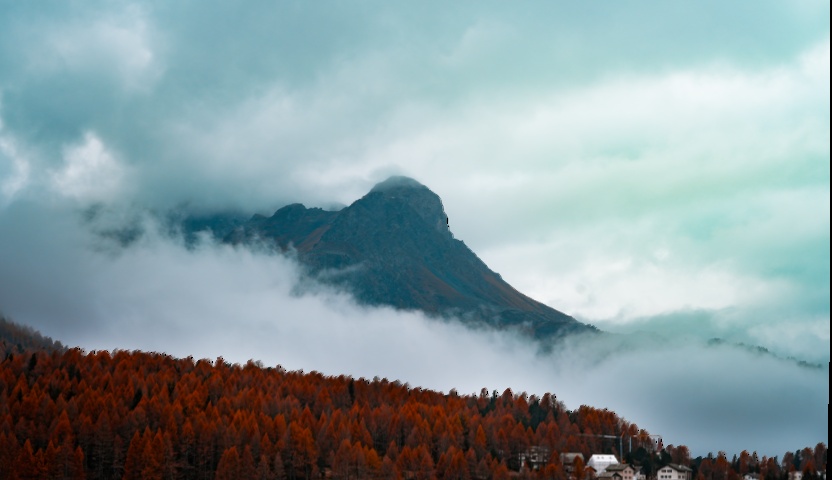}{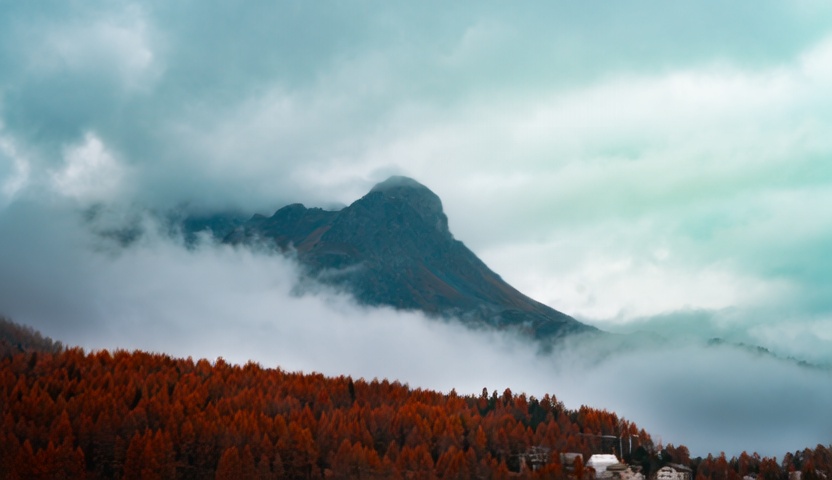}{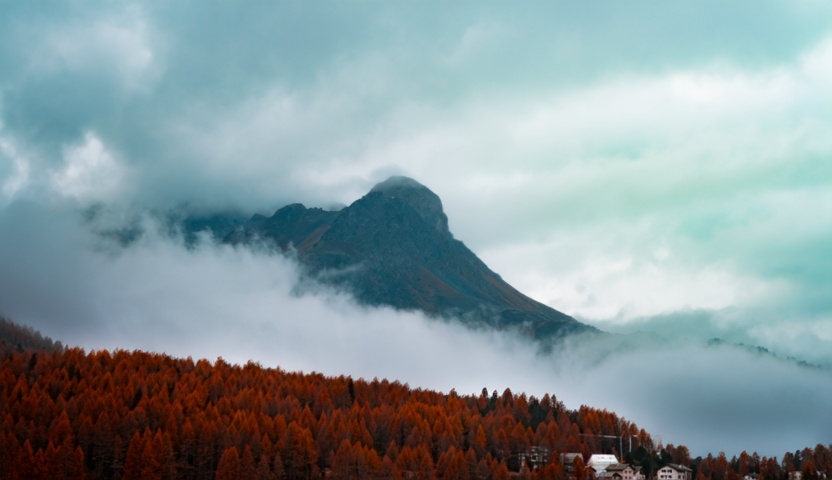}

  {Source} & {Warping} & {W/O fixer} & {\textbf{Ours}} \\
\end{tabular}
\label{fig:nvs_in-the-wild}

\caption{Visual results of in-the-wild test.}
\label{fig:in-the-wild}
\end{figure*}

\clearpage

\FloatBarrier %
\bibliographystyle{splncs04}
\bibliography{main}
\end{document}